%% file: bare_jrnl.tex
\documentclass[10pt,journal,compsoc]{IEEEtran}
\ifCLASSOPTIONcompsoc
  \usepackage[nocompress]{cite}
\else
  \usepackage{cite}
\fi

\ifCLASSINFOpdf
\else
\fi

\newcommand{\EnSi}{\texttt{En-Si}}
\newcommand{\SiEn}{\texttt{Si-En}}

\usepackage{enumitem}
\usepackage{listings}

\renewcommand{\lstlistingname}{Code Block}

\usepackage{placeins}
\usepackage{svg}

\input{headder}
\input{InsightCaptions}

\begin{document}
%
\title{Survey on Publicly Available Sinhala Natural Language Processing Tools and Research}
%
%
%

\author{Nisansa de Silva
\IEEEcompsocitemizethanks{\IEEEcompsocthanksitem Nisansa~de~Silva is with the Department of Computer Science \& Engineering, University of Moratuwa.\protect\\
E-mail: nisansa@cse.mrt.ac.lk
}
\thanks{Manuscript revised: \today.}
}

%
%

\markboth{}%
{Shell \MakeLowercase{\textit{et al.}}: Bare Demo of IEEEtran.cls for IEEE Journals}
%




\IEEEtitleabstractindextext{%
\begin{abstract}
\justify
\input{00_abstract}
\end{abstract}

}

\maketitle


%
\IEEEpeerreviewmaketitle

\input{body}

\ifCLASSOPTIONcaptionsoff
  \newpage
\fi



%
\bibliographystyle{IEEEtranN}

\footnotesize
\bibliography{sinhalaNLP,firstSources,mix}
\normalsize

\newpage

\appendices
\onecolumn

\begin{landscape}
\section{Evolutionary Eras of the Sinhala Script}
\label{app:SinhalaEras}

\begin{figure}[!hbt]	
	\centering 
\includegraphics[width=1.4\textwidth] 
{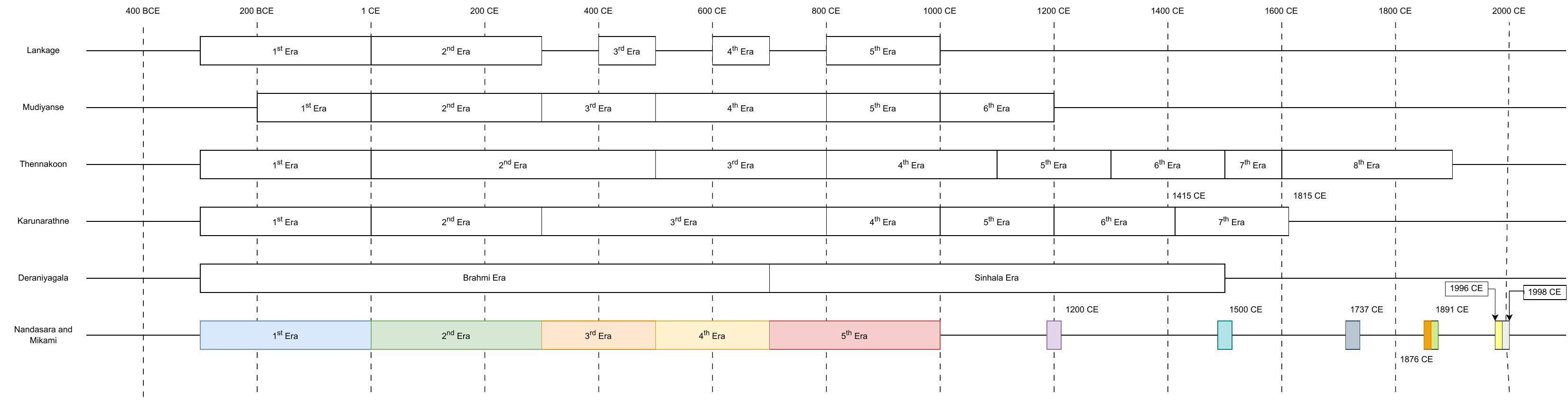}
\caption{Evolutionary eras of the Sinhala alphabet as proposed by \citet{lankage1988sinhala,lankage1996sinhala}, \citet{mudiyanse2018sinhala}, \citet{thennakoon1957parani}, \citet{karunarathne1956sinhala}, \citet{deraniyagala1992prehistory}, and \citet{nandasara2016bridging}}
	\label{fig:SinhalaEras}
\end{figure}

The evolutionary eras of the Sinhala script have raised much scholarly debate. We have shown some prominent categorizations in  Fig~\ref{fig:SinhalaEras}. Other than~\citet{mudiyanse2018sinhala}, scholars generally agree with the fact that the oldest records start at around 300 BCE. Other than \citet{deraniyagala1992prehistory}, all others agree that an era border should exist at 1AD. However, this is a peculiar observation. \citet{lankage1988sinhala,lankage1996sinhala}, \citet{mudiyanse2018sinhala}, \citet{karunarathne1956sinhala}, and \citet{nandasara2016bridging} place an era border at 300 CE while \citet{lankage1988sinhala,lankage1996sinhala}, \citet{mudiyanse2018sinhala}, \citet{thennakoon1957parani}, and \citet{nandasara2016bridging} place an era border at 500 CE. The next most common border placement is at 800 CE by \citet{lankage1988sinhala,lankage1996sinhala}, \citet{mudiyanse2018sinhala}, \citet{thennakoon1957parani}, and \citet{karunarathne1956sinhala}. The last relatively common border placement is agreed between \citet{lankage1988sinhala,lankage1996sinhala}, \citet{mudiyanse2018sinhala}, \citet{thennakoon1957parani}, and \citet{nandasara2016bridging}. 
Given that the widest coverage is provided by \citet{nandasara2016bridging}, we have used their era definitions and examples in Fig~\ref{fig:alpCaption01} and Fig~\ref{fig:alpCaption02}. The sources from which \citet{nandasara2016bridging} have extracted the ancient script are given by the relevant superscripts as follows:

\newcommand{\supe}[1]{${}^{#1}$}


\begin{multicols}{5}
\footnotesize

\textbf{300 BCE - 1 CE}

\begin{enumerate}[label=\supe{{\arabic*}}]
    \item Periyankulama (207-197 BCE)
    \item Mihintale (207-197 BCE)
    \item Situlpawwa (161-137 BCE)
    \item Korawakgala (77-63 BCE)
    \item Ritigala Weweltanne (22-7 BCE)
    \item Yatahalena Vihara (22-7 BCE)
    \item Gallena Vihara (22-7 BCE)
    \item Nuwaragala (22-7 BCE)
    \item Ritigala Andiayakanna (22-7 BCE)
    \item Boowattegala (22-7 BCE)
    \item Rajagala (44-22 BCE)
\end{enumerate}

\columnbreak

\textbf{1 CE - 300 CE}
\begin{enumerate}[label=\supe{{\arabic*}},resume]
    \item Anuradhapura (1-7 CE) 
    \item Situlpawwa (1-7 CE) 
    \item Maharatmale (7-18 CE) 
    \item Wallipuram (67-111 CE) 
    \item Viharagala (60-67 CE) 
    \item Pahala Kainattama (60-67 CE) 
\end{enumerate}

\vfill\null
\columnbreak

\textbf{300 CE - 500 CE}

\begin{enumerate}[label=\supe{{\arabic*}},resume]
    \item Tonigala (301-328 CE) 
    \item Ruwanweliseya (337-365 CE) 
    \item Thissamaharama (406-428 CE) 
    \item Anuradhapura (437-452 CE) 
\end{enumerate}

\vfill\null
\columnbreak

\textbf{500 CE - 700 CE}

\begin{enumerate}[label=\supe{{\arabic*}},resume]
    \item Kandanadu (517-518 CE) 
    \item Dhakshinathupa (639-650 CE) 
    \item Baron Paviliyan (639-650 CE) 
    \item Kuchchaweli (639-650 CE) 
    \item Murutawa (639-650 CE) 
\end{enumerate}

\vfill\null
\columnbreak

\textbf{700 CE - 1000 CE}

\begin{enumerate}[label=\supe{{\arabic*}},resume]
    \item Thiriyaya (733-771 CE) 
    \item Viyaulpotha (853-887 CE) 
    \item Dorabewila (915-923 CE) 
    \item Baddulla (946-954 CE) 
    \item Polonnaruwa (982-1029 CE) 
    \item Indikatuseya (982-1029 CE) 
\end{enumerate}

\end{multicols}

\end{landscape}

\def\alpCaption{The evolution of the Sinhala script under the era classification proposed by \citet{nandasara2016bridging}}
\def\alpCaptionNotes{1200 AD and 1500 AD from inscriptions and pillars~\cite{nandasara2016bridging}. 1737 from the first printed Sinhala book collected by~\citet{nandasara2019development}. 1876 from ``Alfabete des gesammten Erdkreises'' (\textit{Alphabets of the entire world})~\cite{austria1855alfabete} reported as \textit{CINGALESISCH}~\cite{nandasara2019development}. 1891 from the alphabet shown by~\citet{gunasekara1891comprehensive}. 1996 from the character set \textit{Sarasavi} developed by S T Nandasara~\cite{nandasara2019development}. \citet{nandasara2016bridging} report that the gray cells are not included due to being possible to be produced using consonant modifiers. 1998 from the \textit{Iskoola Pota} UNICODE font by~\citet{Microsoft1998FontList}.}

\begin{figure*}[!hbt]	
	\centering 
\includegraphics[width=1\textwidth] 
{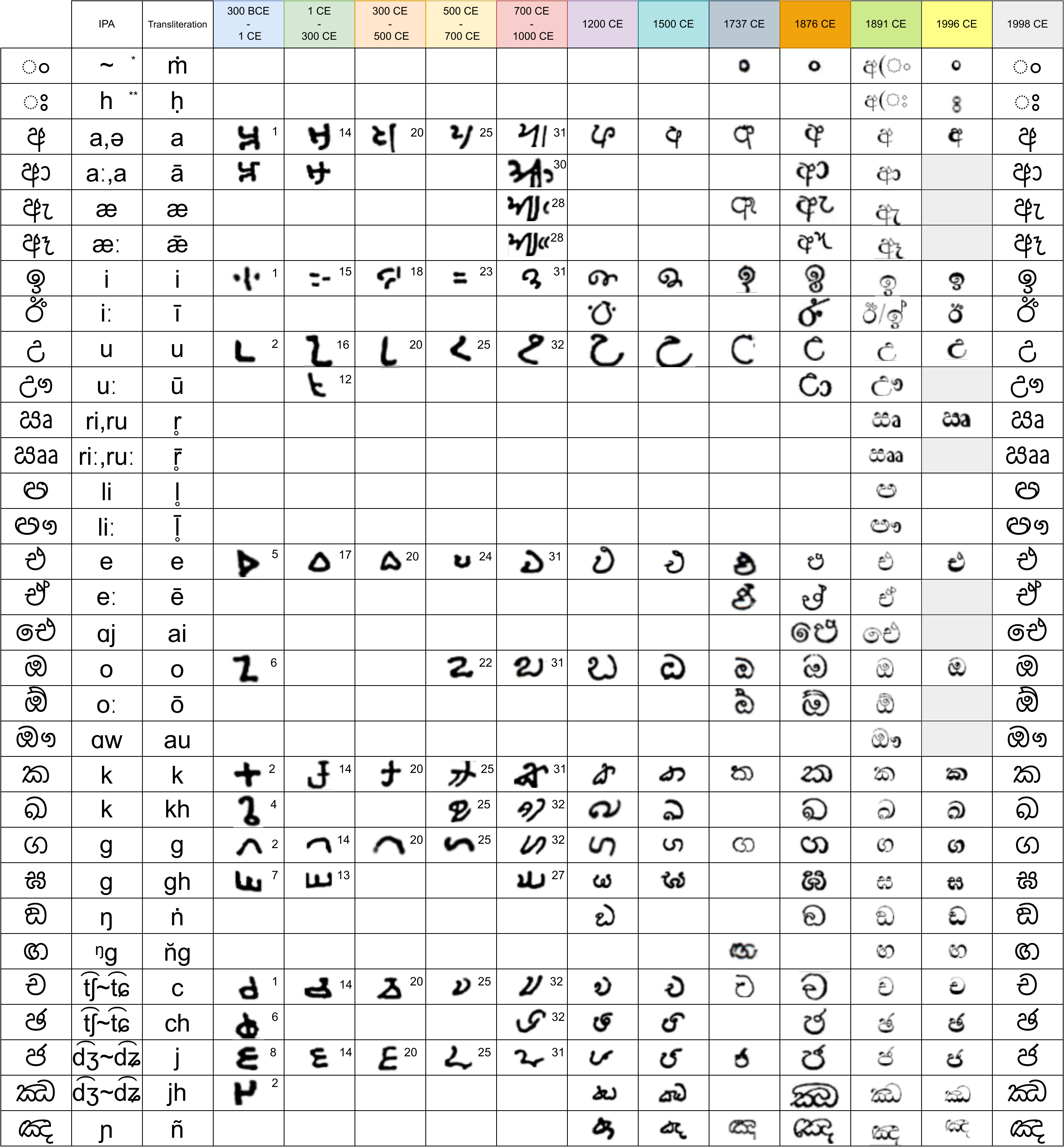}
\caption{\alpCaption{} - Part 1.\newline\textbf{Note:} The IPA for (*) anusvāraya~\cite{anusvara} and (**) visargaya~\cite{visarga} are approximations as there is no consensus on their representation. \alpCaptionNotes}
	\label{fig:alpCaption01}
\end{figure*}

\begin{figure*}[!hbt]	
	\centering 
\includegraphics[width=1\textwidth] 
{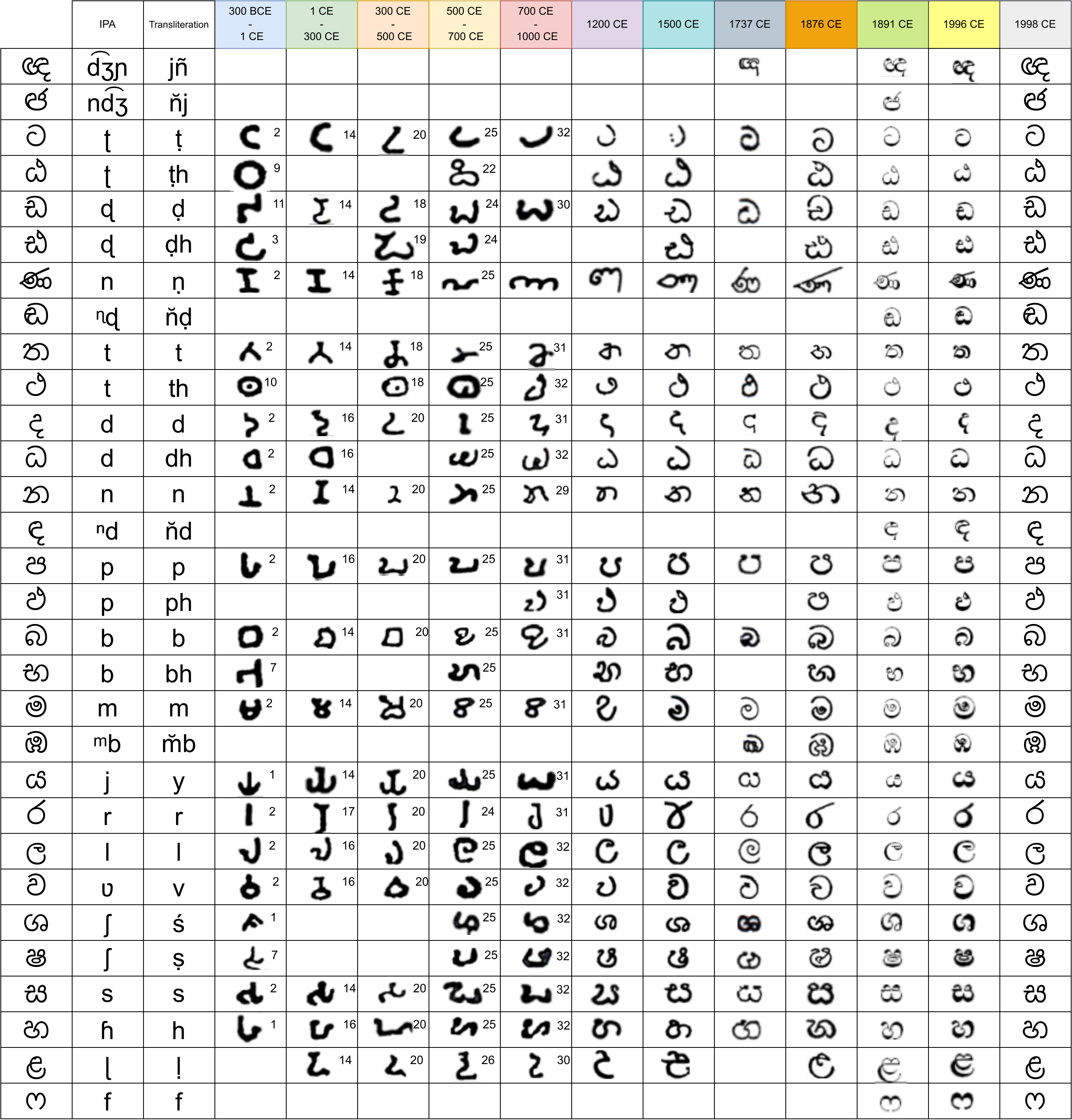}
\caption{\alpCaption{} - Part 2\newline\textbf{Note:} \alpCaptionNotes}
	\label{fig:alpCaption02}
\end{figure*}

\FloatBarrier
\clearpage

\section{Distribution of Sinhala Letters Towards the Beginning, Middle, or End of Words}
\label{app:wordStartEnd}

\begin{figure*}[!hbt]	
	\centering 
\includegraphics[width=0.84\textwidth] 
{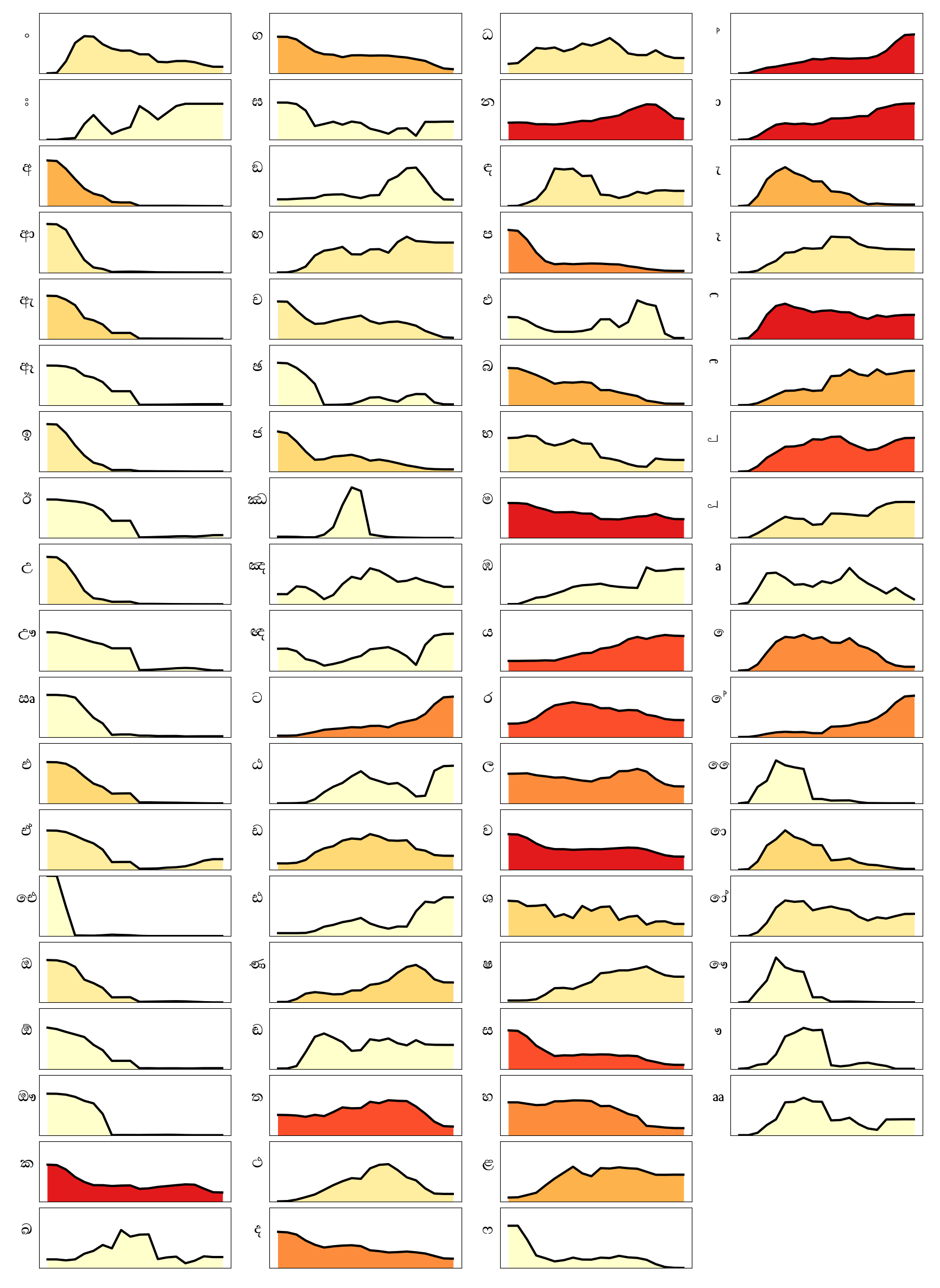}
\caption{An adaptation to Sinhala of the character position visualization proposed by~\citet{ProoffReader2014}. For this, we have used a portion of the \textit{SinMin} corpus created by~\citet{upeksha2015implementing,upeksha2015comparison}. Our sample included 3,990,838 unique words which appear more than once in the corpus. While the Spearman correlation used in this method dilutes the notability aspect such as pure Sinhala vowels only appearing as the first character, it brings to the front how diacritics, on the other hand, are generally biased towards the end.}
	\label{fig:wordStartEnd}
\end{figure*}

\begin{landscape}
\section{Author and Institution Meta Analysis}
\label{app:authors}
\begin{figure}[!hbt]	
	\centering 
\includegraphics[height=0.88\textwidth] 
{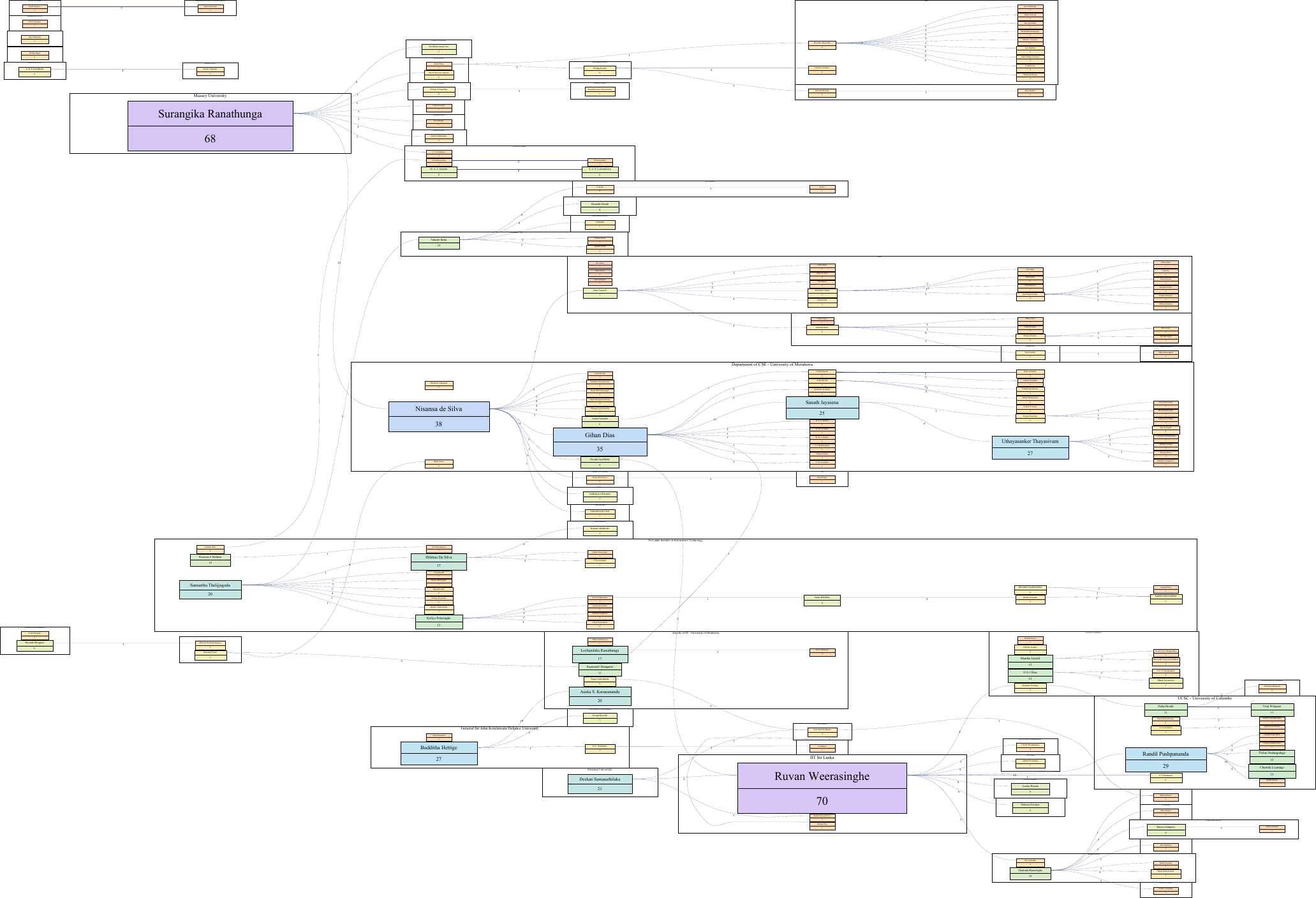}
\caption{Co-author graph of the most prolific researchers in the Sinhala NLP domain (Selected at the threshold of at least 3 publications)}
	\label{fig:authors}
\end{figure}
\end{landscape}

\begin{figure}[!hbt]	
	\centering 
\includegraphics[width=\textwidth] 
{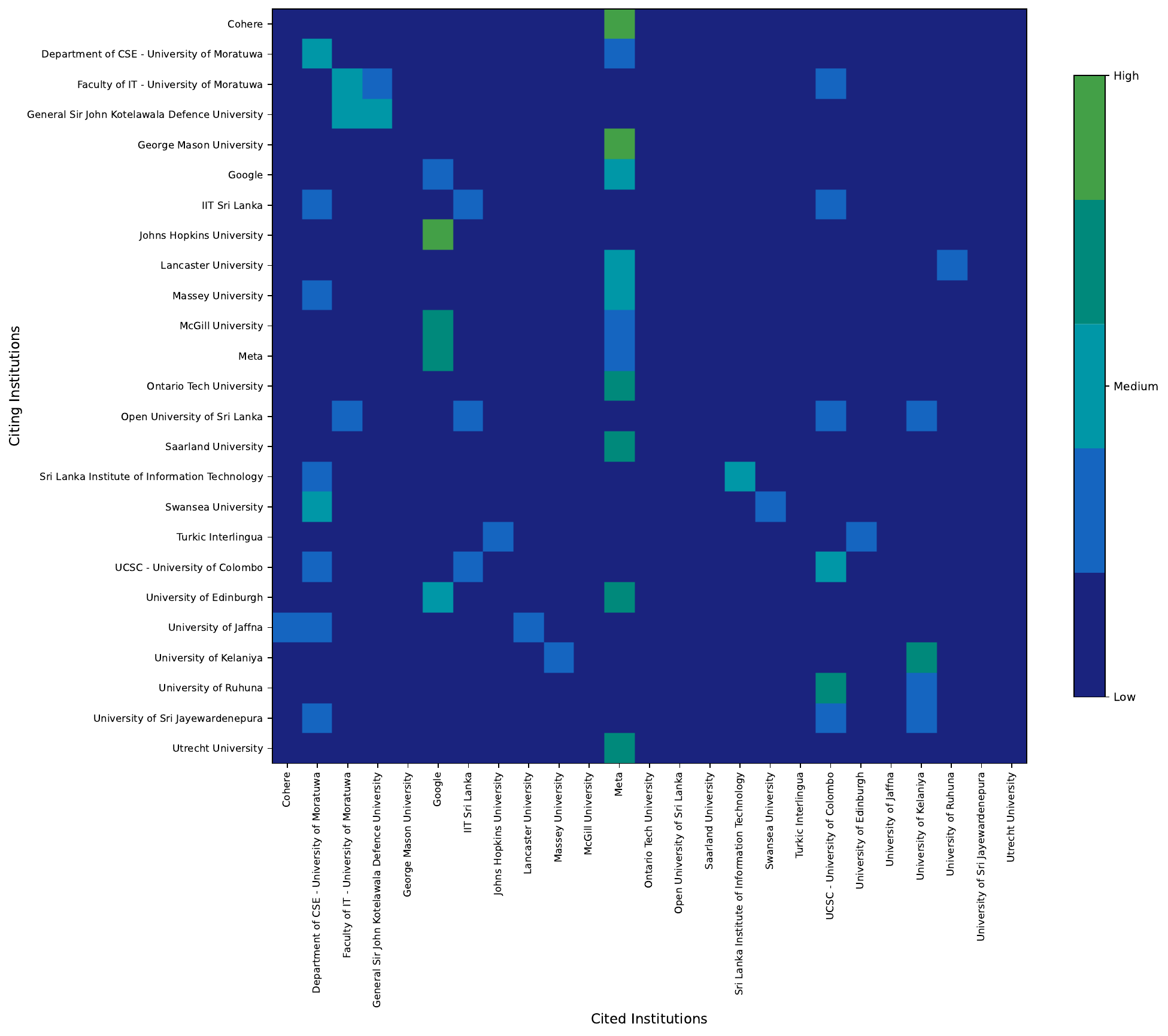}
\caption{\captionCitationProbability
}
	\label{fig:CitationProbability}
\end{figure}

\begin{landscape}
\begin{figure}[!hbt]	
	\centering 
\includegraphics[width=1.35\textwidth] 
{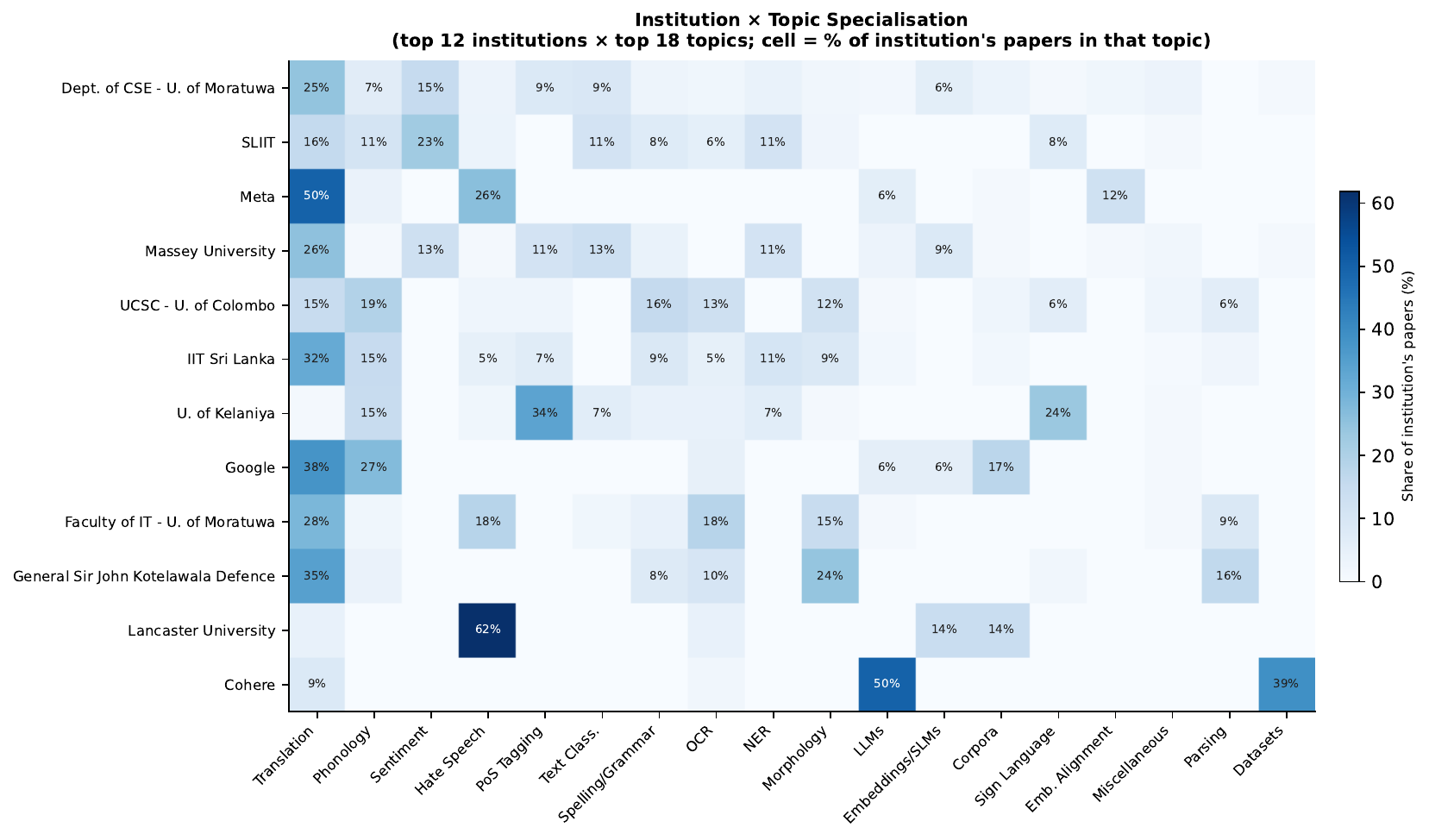}
\caption{\captionInstTopicMatrix}
	\label{fig:InstTopicMatrix}
\end{figure}
\end{landscape}

\begin{landscape}

\begin{figure}[!hbt]	
	\centering 
\includegraphics[width=1\textwidth] 
{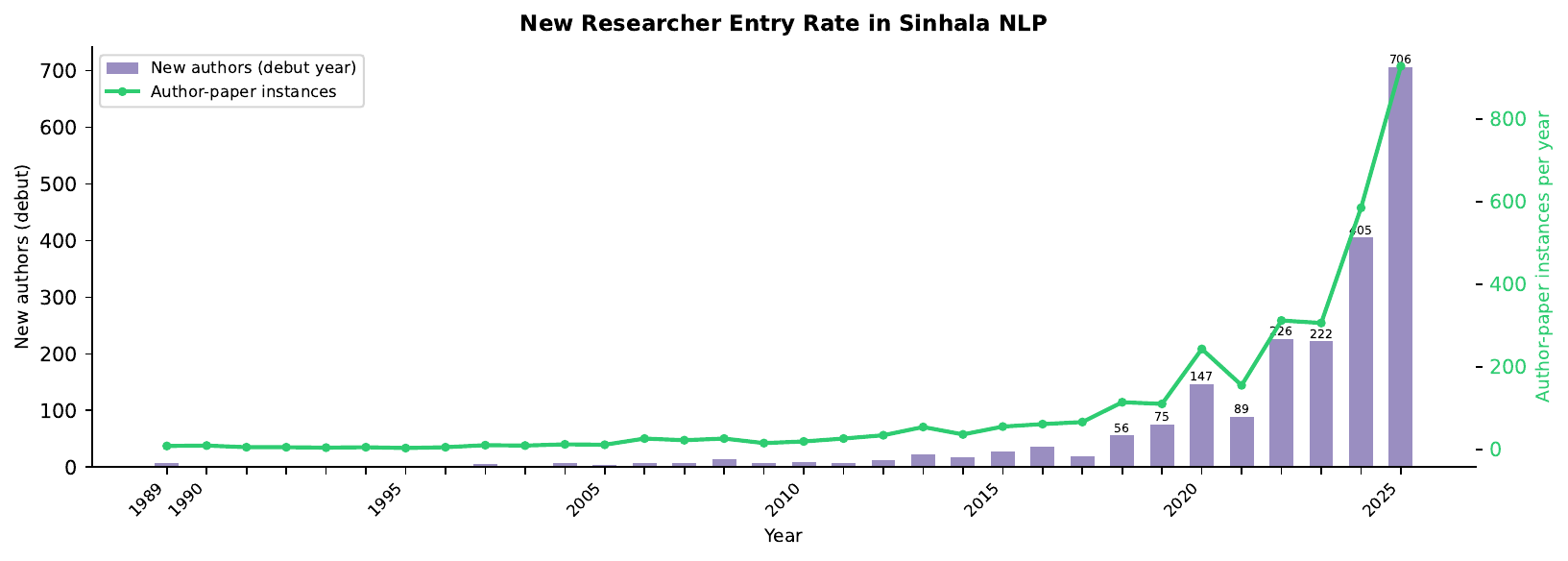}
\caption{\captionNewAuthors}
	\label{fig:newAuthors}
\end{figure}

\begin{figure}[!hbt]	
	\centering 
\includegraphics[width=1\textwidth] 
{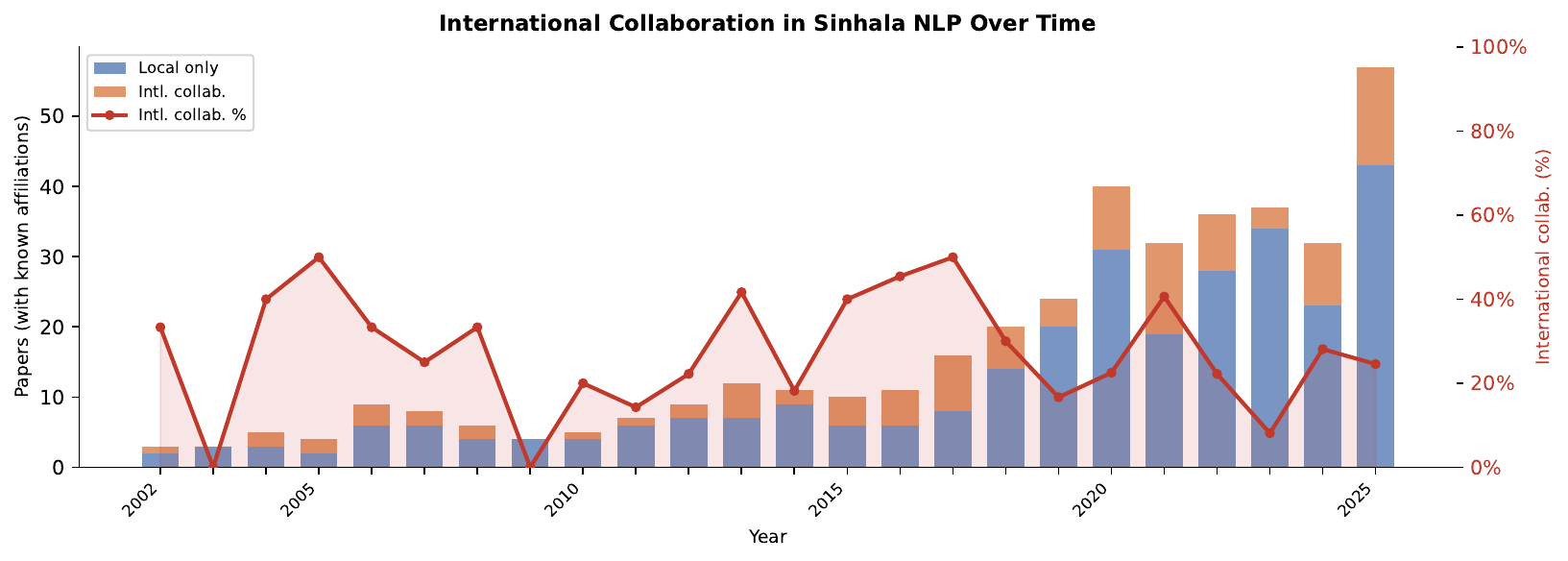}
\caption{\captionInternationalCollab}
	\label{fig:InternationalCollab}
\end{figure}

\end{landscape}

\begin{landscape}
\begin{figure}[!hbt]	
	\centering 
\includegraphics[width=1.35\textwidth] 
{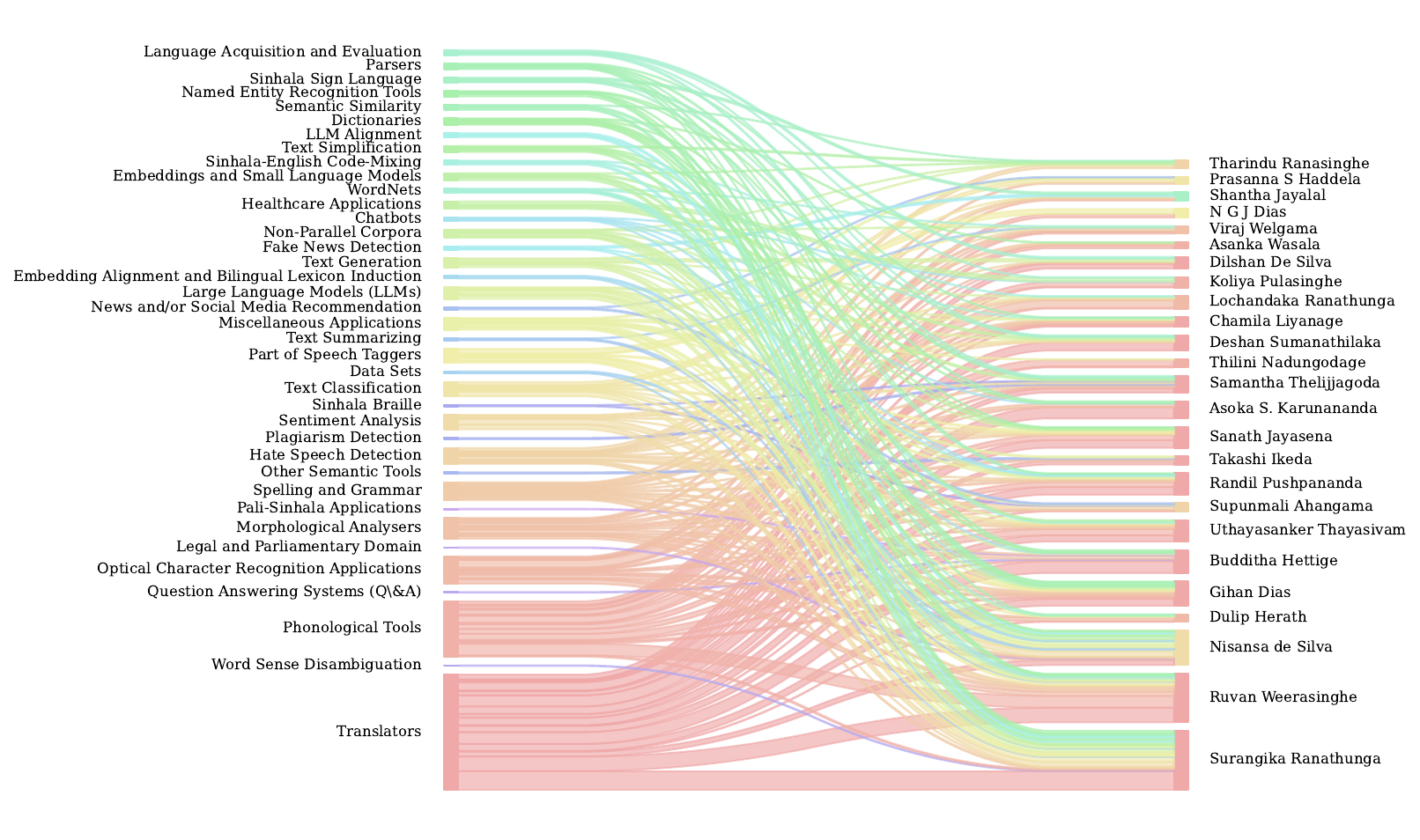}
\caption{Top~25 Sinhala NLP authors by publication count (All years), mapped to their research interests, denoted by the subsection titles of the Section~\ref{Sinhala} of this paper.}
	\label{fig:authorAreas}
\end{figure}

\end{landscape}

\begin{figure}[!hbt]	
	\centering 
\includegraphics[width=\linewidth] 
{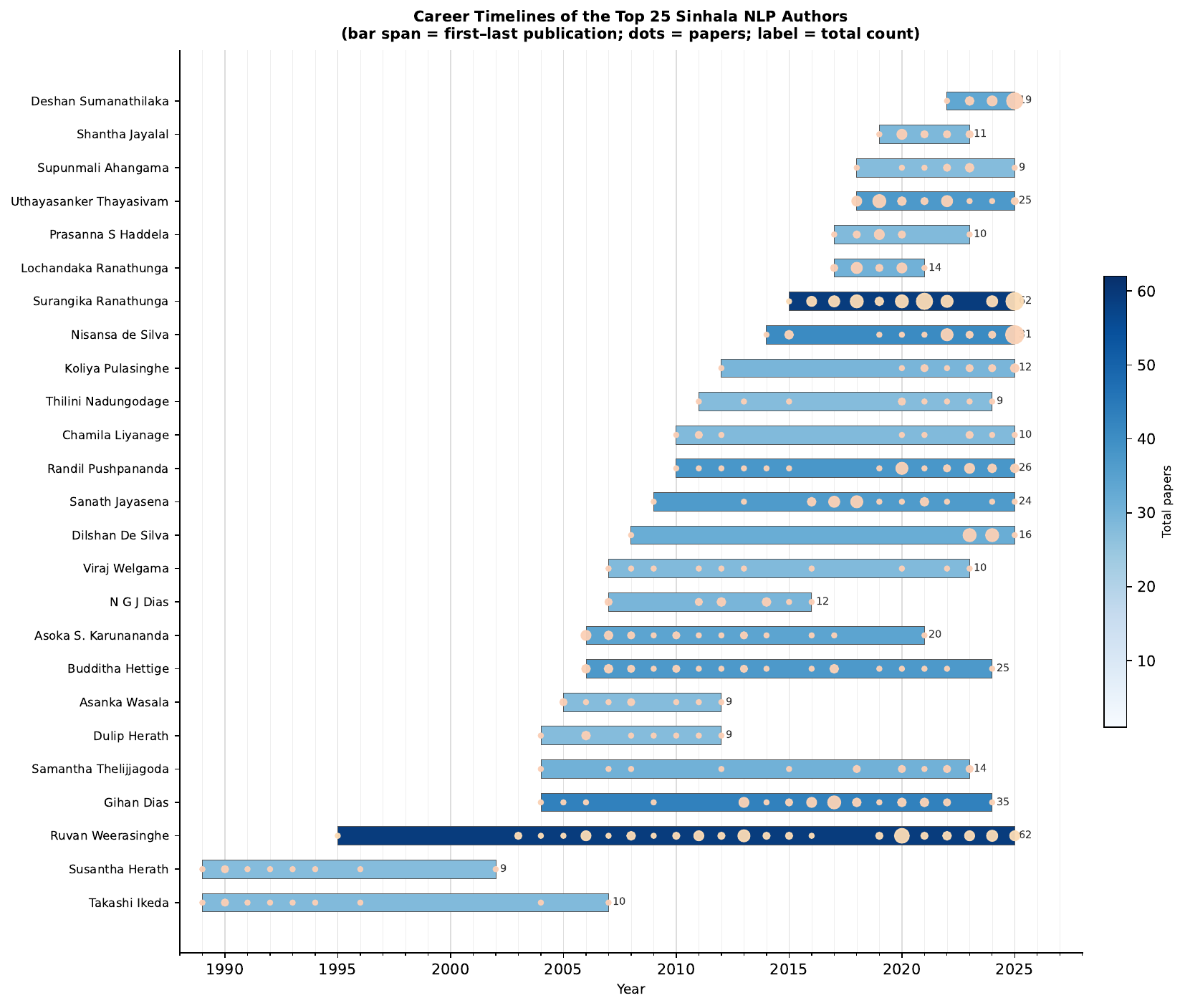}
\caption{\captionAuthorCareers~[\textbf{Notes:} (1) This analysis only contains the subset of publications of which the information was available. Therefore some of the publication counts here would be less than that shown in Fig~\ref{fig:authors}. (2) Any potential mismatch in the top~25 Sinhala NLP authors list in this figure and Fig~\ref{fig:authorAreas} is due to the year range restriction of this figure not applying to Fig~\ref{fig:authorAreas}.]}
	\label{fig:authorCareers}
\end{figure}

\end{document}

%% file: headder.tex
\usepackage[utf8]{inputenc}
\usepackage[T1]{fontenc}

\usepackage{ipa}
\usepackage{graphicx}

\usepackage[numbers,sort&compress]{natbib}

\usepackage{hyperref} 
\hypersetup{
colorlinks   = true,
citecolor    = blue,
urlcolor    =  blue
}

\usepackage{url}
\usepackage{cleveref}
\crefformat{footnote}{#2\footnotemark[#1]#3}

\newcommand{\footURL}[1]{\footnote{\url{#1}}}

\usepackage{ragged2e}

\usepackage[
singlelinecheck=false 
]{caption}
\usepackage{multirow}
\usepackage{hhline}
\usepackage{subcaption}

\usepackage{ftnxtra}
\usepackage{fnpos}

\makeatletter
\newread\pin@file
\newcounter{pinlineno}
\newcommand\pin@accu{}
\newcommand\pin@ext{pintmp}
\newcommand*\partialinput [3] {%
  \IfFileExists{#3}{%
    \openin\pin@file #3
    \setcounter{pinlineno}{1}
    \@whilenum\value{pinlineno}<#1 \do{%
      \read\pin@file to\pin@line
      \stepcounter{pinlineno}%
    }
    \addtocounter{pinlineno}{-1}
    \let\pin@accu\empty
    \begingroup
    \endlinechar\newlinechar
    \@whilenum\value{pinlineno}<#2 \do{%
      \readline\pin@file to\pin@line
      \edef\pin@accu{\pin@accu\pin@line}%
      \stepcounter{pinlineno}%
    }
    \closein\pin@file
    \expandafter\endgroup
    \scantokens\expandafter{\pin@accu}%
  }{%
    \errmessage{File `#3' doesn't exist!}%
  }%
}
\makeatother

\newcommand{\version}{\partialinput{1}{1}{ReleaseNotes.txt}}

\usepackage{pdflscape}
\usepackage{svg}

\usepackage{makecell}
\usepackage{multicol}

\newcommand{\hf}[2]{\raisebox{-2.2pt}{\includegraphics[scale=0.09]{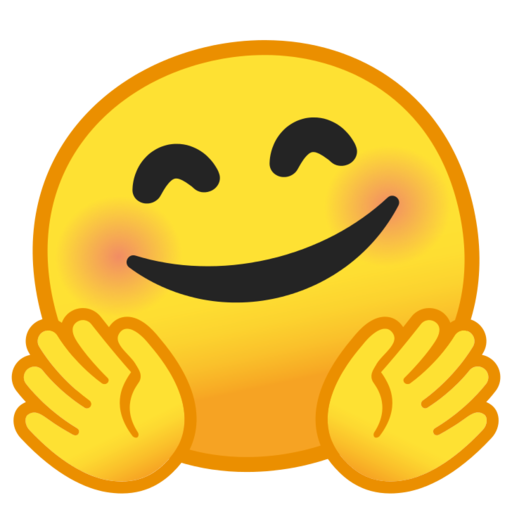}}~\href{#1}{\texttt{#2}}}

\newcommand{\gh}[2]{\raisebox{-2.2pt}{\includegraphics[scale=0.02]{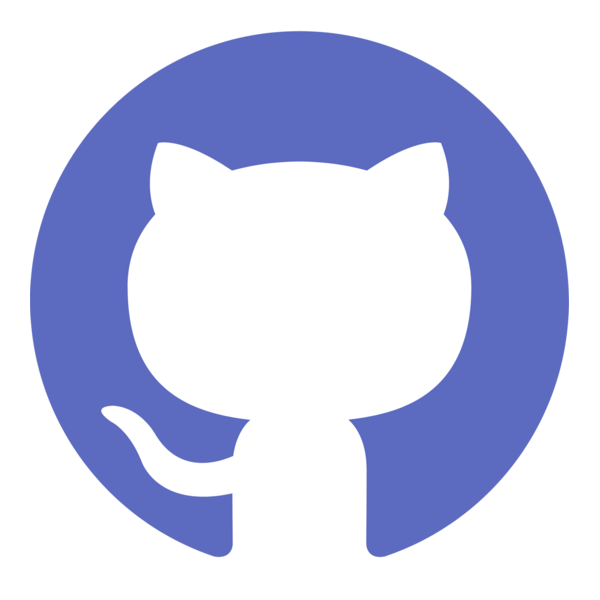}}~\href{#1}{\texttt{#2}}}

\newcommand{\gd}[2]{\raisebox{-2.2pt}{\includegraphics[scale=0.025]{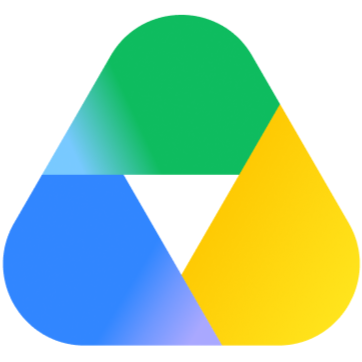}}~\href{#1}{\texttt{#2}}}

\newcommand{\osf}[2]{\raisebox{-2.2pt}{\includegraphics[scale=0.07]{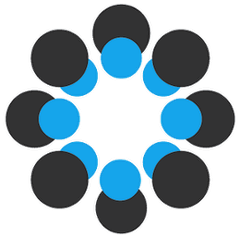}}~\href{#1}{\texttt{#2}}}

\newcommand{\opus}[2]{\raisebox{-2.2pt}{\includegraphics[scale=0.3]{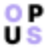}}~\href{#1}{\texttt{#2}}}

%% file: InsightCaptions.tex
\newcommand{\captionCitationProbability}{%
  The probability of research from an institution citing that of other institutions. Note that the calculations are limited by four factors: (1)~the availability of free-to-download PDFs of the paper; (2)~the aforementioned PDF containing a reference list (some extended abstracts do not include one); (3)~the text extraction capabilities of \textit{pdftotext}\footURL{https://www.xpdfreader.com/pdftotext-man.html}; (4)~the accuracy of the paper-title look-up. With those limitations in mind, we can still make a few interesting observations. The institution with the highest incoming citation probability, \textit{Meta}, appears to be getting the most citations from most sources. However, they themselves seem to almost exclusively cite their own papers (0.2657). The only notable exception is the smaller number of works they cite (0.4638) from \textit{Google}. Notably, they cite \textit{Department of CSE - University of Moratuwa} at only 0.0177, despite it having the second highest publication count. This leaves them with a 0.2528 probability of citing anyone else.  Comparatively, \textit{Department of CSE - University of Moratuwa} seems to be more egalitarian in citing. It has a lower self-citation probability (0.3276) and a higher probability of citing \textit{Meta} (0.2127). This results in a probability of citing others at 0.4596.  Among other Sri Lankan institutions, \textit{Open University of Sri Lanka} cites \textit{Department of CSE - University of Moratuwa} at only 0.0000, preferring instead \textit{University of Kelaniya} (0.1765); \textit{University of Ruhuna} cites \textit{Department of CSE - University of Moratuwa} at only 0.0323, preferring instead \textit{UCSC - University of Colombo} (0.4516); \textit{General Sir John Kotelawala Defence University} cites \textit{Department of CSE - University of Moratuwa} at only 0.0357, preferring instead \textit{Faculty of IT - University of Moratuwa} (0.4107); \textit{University of Kelaniya} cites \textit{Department of CSE - University of Moratuwa} at only 0.0833, preferring instead \textit{Massey University} (0.1458). 
}

\newcommand{\captionTopicLifecycle}{%
  Topic lifecycle heatmap for Sinhala NLP research (1989--2025). Each row represents one of the 39 surveyed research areas; colour intensity encodes the number of papers published in that year relative to the area's own peak (row-normalised), making the emergence, growth, and decline of each topic visible at a glance.  \textit{Morphological Analysers} is the earliest documented area, with work dating back to 1989.  The largest area by publication volume is \textit{Translators} (182 papers), which peaked in 2020.  Several areas are almost entirely a post-2020 phenomenon, with at least 90\% of their papers published since 2020: \textit{Large Language Models (LLMs)}, \textit{Fake News Detection}, \textit{Healthcare Applications}, \textit{Language Identification (LID) Task}.  Entirely new research directions that did not exist before 2020 include \textit{News and/or Social Media Recommendation}.  Conversely, some areas appear to have gone dormant: \textit{WordNets} (last active 2014).  Overall, 20 of the 39 areas reached their peak publication rate within the last three years (2023--2025), indicating that the field as a whole is still accelerating rather than maturing.
}

\newcommand{\textTopicLifecycle}{%
  shows the lifecycle of each Sinhala NLP research area over time, with colour encoding relative publication activity
}

\newcommand{\captionVenueMix}{%
  Publication venue mix across Sinhala NLP research areas (1989--2025). Each 100\% stacked bar shows the proportion of a research area's papers published in peer-reviewed journals (green), theses (purple), conferences (blue), preprints such as arXiv (orange), and other sources (grey). arXiv and other preprints are separated from peer-reviewed journals so the journal share reflects only fully refereed publications. Bars are sorted by journal share (descending), with ties broken by thesis share, then conference share, then preprint share. Across all 798 surveyed papers, 236 (30\%) appeared in journals, 446 (56\%) in conference proceedings, 70 (9\%) as preprints, and 37 (5\%) as theses.  \textit{Fake News Detection} has the highest journal share at 50\%, indicating a mature body of peer-reviewed work.  By contrast, \textit{Named Entity Recognition Tools} is the most conference-driven area (88\% conference papers), typical of fast-moving engineering topics.  \textit{Data Sets} has the highest preprint share (71\%), reflecting how rapidly emerging areas disseminate results before formal publication. 
}

\newcommand{\textVenueMix}{%
  shows the publication venue mix for each research area, with bars sorted by journal share
}

\newcommand{\captionAuthorCareers}{%
  Career timelines of the top~25 Sinhala NLP authors by publication count within 1989--2025. Each bar spans an author's first to last recorded publication; dots within each bar mark individual paper years (dot size scales with the number of papers in that year); bar colour encodes total paper count (darker = more prolific).  The most prolific author in the group is \textit{Ruvan Weerasinghe} with 62 papers (1995--2025).  The earliest debut in this cohort is \textit{Takashi Ikeda} (1989), who appears to have concluded their active involvement (last active~2007).  \textit{Nisansa de Silva} stands out with 31~papers across a 11-year career (2014--2025), making them one of the field's most sustained mid-career contributors.  Notably, several authors debuted as recently as 2018 or later yet still rank among the top~25: \textit{Uthayasanker Thayasivam} (debut~2018, 25~papers); \textit{Deshan Sumanathilaka} (debut~2022, 19~papers); \textit{Shantha Jayalal} (debut~2019, 11~papers). This illustrates the rapid acceleration of the field, where researchers can accumulate a substantial body of work within a short period.  By contrast, some earlier contributors appear to have concluded their active involvement: \textit{N G J Dias} (last active~2016), \textit{Susantha Herath} (last active~2002), \textit{Dulip Herath} (last active~2012), \textit{Asanka Wasala} (last active~2012). 
}

\newcommand{\textAuthorCareers}{%
  shows the career timelines of the top Sinhala NLP authors, with each bar spanning their first to last publication year
}

\newcommand{\captionInstTopicMatrix}{%
  Specialisation of the top~12 institutions (by paper count) across the top~18 research topics. Each cell shows the percentage of that institution's papers attributed to a given topic. \textit{Lancaster University} is the most specialised, devoting 62\% of its output to Hate Speech Detection. \textit{UCSC - University of Colombo} has the broadest spread, with its largest single-topic share at 19\%. The largest institution, \textit{Department of CSE - University of Moratuwa}, concentrates mainly on Translators (25\%) and Sentiment Analysis (15\%). The second largest, \textit{Sri Lanka Institute of Information Technology}, leads in Sentiment Analysis (23\%) and Translators (16\%).
}

\newcommand{\textInstTopicMatrix}{%
  shows the topic specialisation of the top institutions, with each cell giving the percentage of that institution's papers in a given area
}

\newcommand{\captionInternationalCollab}{%
  International collaboration in Sinhala NLP over time (1989--2025). Bars show the annual count of papers involving only Sri Lankan institutions (blue) versus those with at least one international co-author (orange). The red line shows the international collaboration rate. Across all years with known affiliations, 106 of 403 papers (26\%) involved international collaboration. The collaboration rate peaked in 2005 at 50\%. In the most recent five-year window (2021--2025), the rate was 24\%, suggesting that absolute output has grown faster than international partnership.
}

\newcommand{\textInternationalCollab}{%
  shows the trend of international collaboration in Sinhala NLP research over time
}

\newcommand{\captionNewAuthors}{%
  New researcher entry rate in Sinhala NLP (1989--2025). Purple bars show the number of authors making their debut publication each year; the green line shows total author-paper instances per year (a proxy for overall community activity). A total of 2,152 unique authors appear in the surveyed literature. The field reached half that total by 2024, meaning the second half of all-time contributors joined after 2024. The year with the highest intake was 2025 (706 new authors). The five years 2021--2025 account for 1,648 debuts (77\% of all time), reflecting a rapid recent expansion of the research community.
}

\newcommand{\textNewAuthors}{%
  shows the rate at which new researchers entered the Sinhala NLP field each year
}

%% file: 00_abstract.tex
Sinhala is the native language of the Sinhalese people who make up the largest ethnic group of Sri Lanka. The language belongs to the globe-spanning language tree, Indo-European. However, due to poverty in both linguistic and economic capital, Sinhala, in the perspective of Natural Language Processing tools and research, remains a resource-poor language which has neither the economic drive its cousin English has nor the sheer push of the law of numbers a language such as Chinese has. A number of research groups from Sri Lanka have noticed this dearth and the resultant dire need for proper tools and research for Sinhala natural language processing. However, due to various reasons, these attempts seem to lack coordination and awareness of each other. The objective of this paper is to fill that gap of a comprehensive literature survey of the publicly available Sinhala natural language tools and research so that the researchers working in this field can better utilize the contributions of their peers. As such, we shall be uploading this paper to arXiv and perpetually update it periodically as a living research article to reflect the advances made in the field. This manuscript is at version $\version$.   

%% file: body.tex
\input{01_Introduction}


\input{02_Properties}

\input{03_Sinhala}
\input{04_FirstSources}

\section{Conclusion}
\label{Conclusion}
At this point, a reader might think, there seems to be a significant number of implementations of NLP for Sinhala. Therefore, how can one justify listing Sinhala as a resource poor language? The important point which is missing in that assumption is that in the cases of almost all of the above listed implementations and findings, the only thing that is publicly available for a researcher is a set of research papers. The corpora, tools, algorithm, and anything else that were discovered through these research are either locked away as properties of individual research groups or worse lost to the time with crashed ancient servers, lost hard drives, and expired web hosts. This reason and probably 
academic/research rivalry has caused these separate research groups not to cite or build upon the works of each other. In many cases where similar work is done, it is a re-hashing on the same ideas adopted from resource-rich languages because of, the unavailability of (or the reluctance to), referring and building on work done by another group (Refer Fig.~\ref{fig:CitationProbability} in Appendix~\ref{app:authors}). This has resulted in multiple groups building multiple foundations behind closed doors but no one ending up with a completed end-to-end NLP workflow. 
In their analysis of low-resourced languages,~\citet{ranathunga2022some} observed that only 11.43\% of Sinhala NLP papers have released the relevant data sets. Further, according to them, code being released sits at 9.71\% while tools being released sits at 5.71\%. While the 11.43\% figure of data release may induce a feeling of availability,~\citet{ranathunga2022some} further observes that it is only 1.14\% has been released in public repositories.     
Research publications promising access to data and code only to be found lacking later is a common academic shortfall according to~\citet{gabelica2022many} however, given that Sinhala NLP is already having minimal work done as a whole compared to some other languages, we simply cannot afford to lose any of the generated data or code. 
In conclusion, what can be said is that, even though there are islands of implementations done for Sinhala NLP, they are of very small scale and/or are usually not readily accessible for further use and research by other researchers. Thus, so far, sadly, Sinhala stays a resource-poor language. 


\section*{Acknowledgments}
The authors would like to thank Romain Egele for checking the examples we have provided in French for their accuracy. Similarly, the authors would also like to thank Shravan Kale for checking the examples we have provided in Hindi for their accuracy. 


%% file: 01_Introduction.tex
\section{Introduction}

\begin{figure}[!t]	
	\centering 
\includegraphics[width=0.99\linewidth]  
{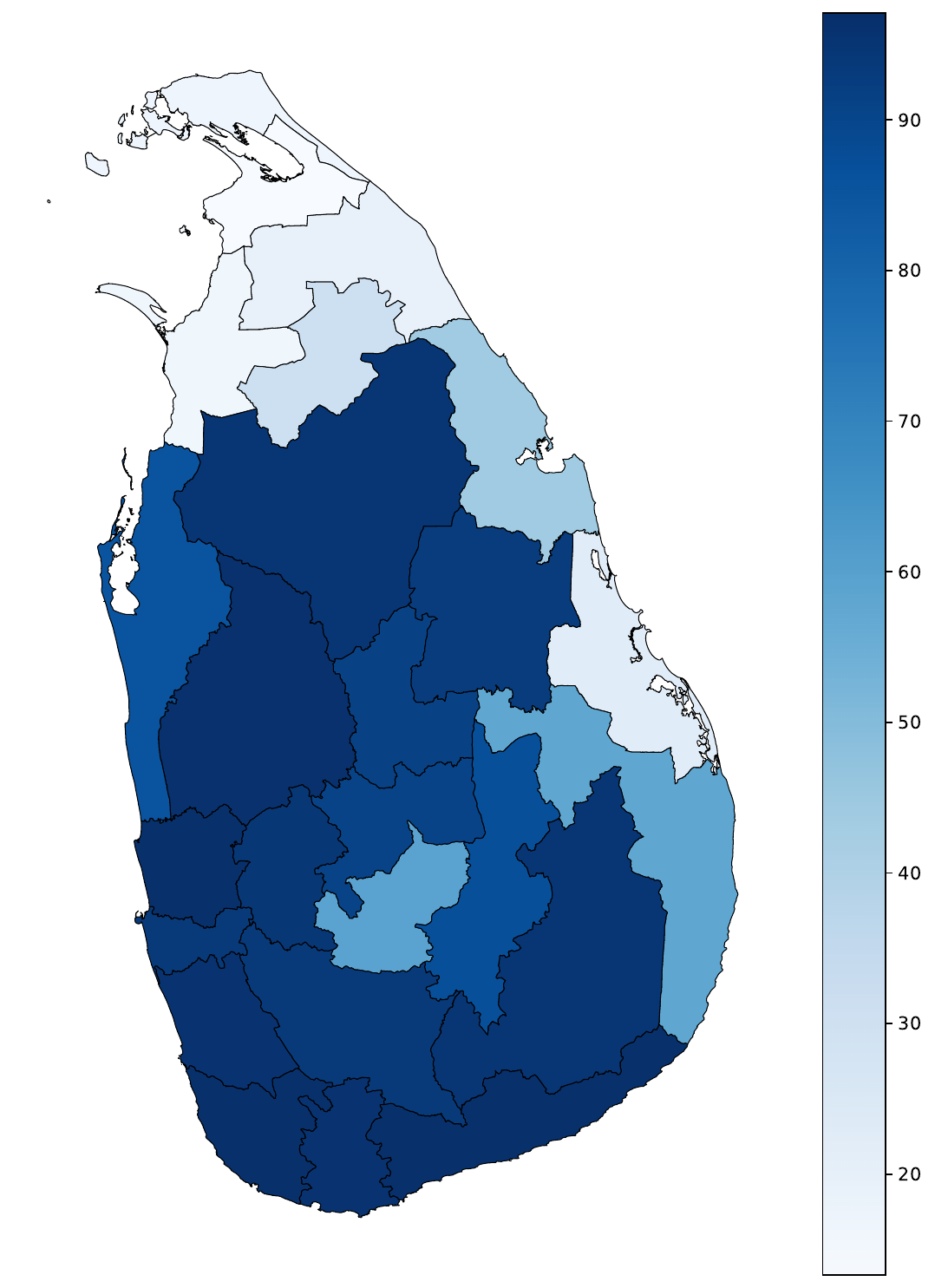}
	\caption{Sinhala language literacy as a percentage of the population aged 10 and over in Sri Lankan administrative districts as reported by the 2024 Census~\cite{department2024Census}
    }	
	\label{fig:sinhala_literacy_map}
\end{figure}

Sinhala\footnote{\citet{englebretson2005santa} observe that in some contexts the \textit{Sinhala} language is also referred as \textit{Sinhalese}, \textit{Singhala}, and \textit{Singhalese}} language, being the native language of the Sinhalese people~\cite{disanayaka1976national,perera1985sinhala,bauer2007linguistics}, who make up the largest ethnic group of the island country of Sri Lanka, enjoys being reported as the mother tongue (L1) of approximately 17 million people~\cite{2007Percentage,gair1974literary}. When both L1 and L2 speakers are counted, $85.2\%$ of the total Sri Lankan population is literate in Sinhala~\cite{department2024Census}\footnote{This is an increment from the percentage of $79.7\%$ reported in 2012~\cite{department2012Census}} (Fig~\ref{fig:sinhala_literacy_map} shows finer granular detail at the administrative district level). A strong correlation between the island of Sri Lanka and the usage of Sinhala has been observed~\cite{dunn2024geographically}\footnote{\citet{dunn2024geographically} in their work to augment the Language Identification (LID) task with geography observe that for Sinhala LID \textit{does not} benefit from the geography model given that the models already have an alignment of 99\%.}. This implies two things: 1) the majority of the Sinhala linguistic sources are located in the geographical area of Sri Lanka, and 2) the majority of the available linguistic sources in Sri Lanka are in Sinhala. Sri Lanka is the only country where Sinhala is recognised as an official language~\cite{parliament2022constitution}.

\begin{figure*}[!hbt]	
	\centering 
\includegraphics[width=\textwidth]  
{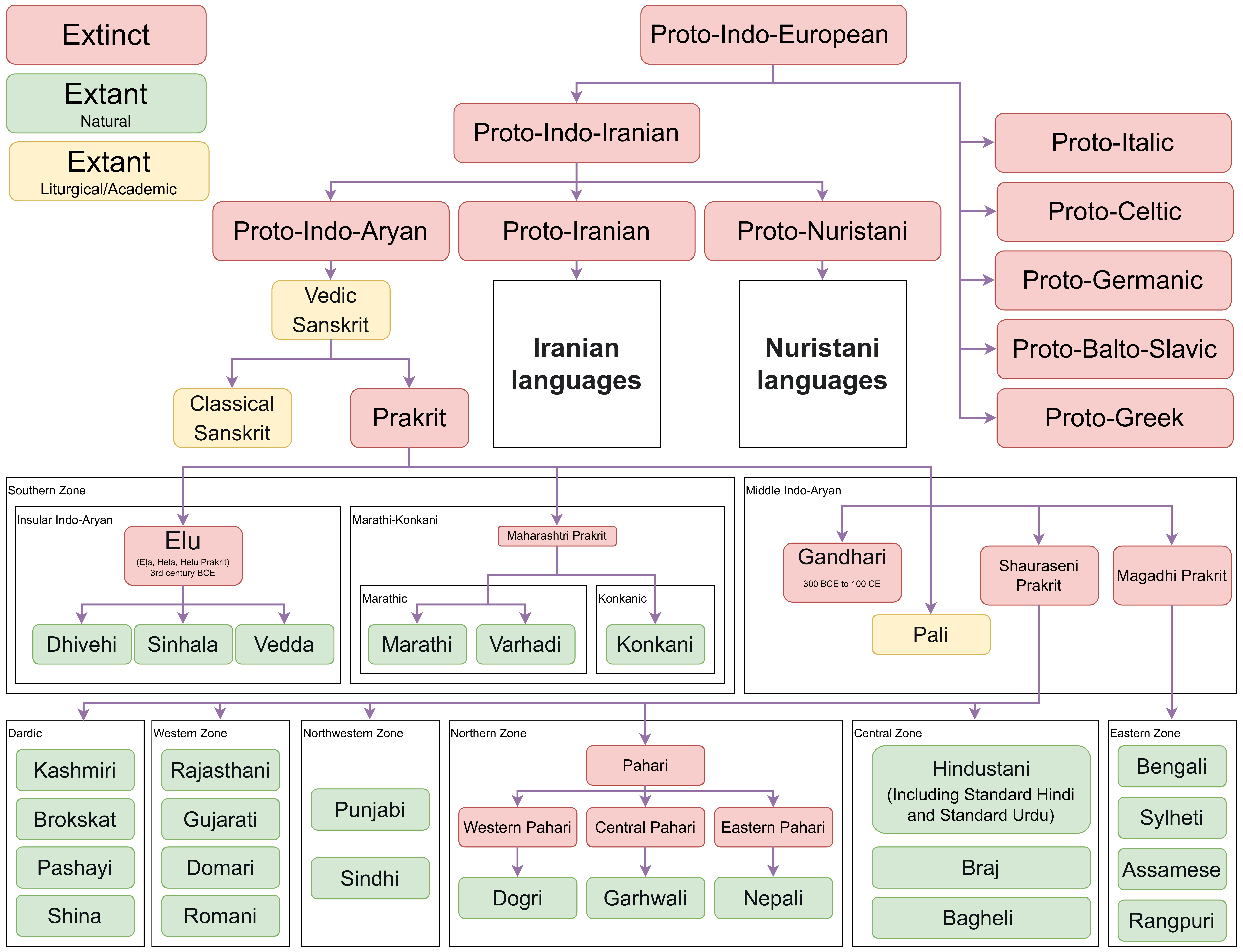}
	\caption{A portion of the Indo-European language tree with a focus on Sinhala and its close relatives. Only the \textit{Elu} sub-tree is shown exhaustively. All other sub-trees only list some languages for representation purposes, chosen randomly or otherwise. Some of the languages shown as \textit{Extinct} have archaeological proof (eg: \textit{Prakrit}) while others were retroactively reconstructed by linguists by observing the commonalities of languages that they begot (eg: \textit{Proto-Indo-Aryan}). The status of some languages have spread beyond linguistics to be a social and political question. In this depiction, we have attempted to side with academic consensus when available and the majority when not. As an example of one contentions point on this diagram, note the positioning of \textit{Pali}. While what we show here is the academically accepted relationship, there are other schools of thought: 1) Some argue that \textit{Pali} is a constructed language (ConLang) of \textit{Sinhala} and thus should be depicted as a child of \textit{Sinhala}., 2) Others claim it to be a sibling of \textit{Sinhala}, 3) Further, some claim it to be either be the link between \textit{Elu} and \textit{Sinhala} or the parent of \textit{Elu}, 4) Some claim it needs to be merged with \textit{Gandhari}, and 5) Some claim it to be the same as \textit{Magadhi Prakrit} given that in some Buddhist traditions, it is noted Buddha to have spoken \textit{Magadhi}.}	
	\label{fig:Sinhala_Language_Tree}
\end{figure*}

To give a brief linguistic background for the purpose of aligning the Sinhala language with the baseline of English, primarily it should be noted that the Sinhala language belongs to the same Indo-European language tree~\cite{Young2015language,kanduboda2011role,arangala2024location}. However, unlike English, which is part of the Germanic branch, Sinhala belongs to the Indo-Aryan branch. An extremely simplified version of the Indo-European language tree focusing mainly on Sinhala and its close relatives is shown in Fig~\ref{fig:Sinhala_Language_Tree}.
Further, Sinhala, unlike English, which borrowed the Latin alphabet, has its own writing system, which is a descendant of the Indian Brahmi script~\cite{fernando1949palaeographical,bandara2012creation,daniels1996world,sirisoma1990brahmi,dias1996lakdiwa,hettiarachchi1990investigation,paranavitana1970inscriptions}. By extension, this makes Sinhala Script a member of the Aramaic family of scripts~\cite{salomon1998indian,falk1993schrift}. Thus, by inheritance, the Sinhala writing system is \textit{abugida} (\textit{alphasyllabary}), which to say, that consonant-vowel sequences are written as single units~\cite{hettige2011computational}. An extremely simplified evolution tree of the Sinhala script is shown in Fig~\ref{fig:Sinhala_Script_Tree}. In it, the Sinhala script is shown as a singular derivation from \textit{Brahmi} directly appearing with the contemporary script. But this is a simplification. We show the actual evolutionary eras of the Sinhala script in Appendix~\ref{app:SinhalaEras}, where the gradual change from \textit{Brahmi} to the contemporary script is clearer. The phonetic and graphemic view of the contemporary Sinhala script is shown in Fig~\ref{fig:Sinhala_Script}.
It should be noted that, due to various historical reasons~\cite{gunasekara1986comprehensive}, the modern Sinhala language has loanwords from languages such as Tamil~\cite{vasuthavaninfluence,damodaram2026tamil}, English, Portuguese, and Dutch. 
An analysis of the distribution of Sinhala letters towards the beginning, middle, or end of words conducted in the style of proposed by~\citet{ProoffReader2014} using the \textit{SinMin} Sinhala corpus created by~\citet{upeksha2015implementing,upeksha2015comparison} is given in Appendix~\ref{app:wordStartEnd}.

\begin{figure*}[!hbt]	
	\centering 
\includegraphics[width=\textwidth]  
{images/Sinhala_Script_Tree.pdf}
	\caption{A portion of the writing systems (scripts) tree with a focus on Sinhala and its close relatives. It is important to note that this is distinct from the language tree shown in Fig~\ref{fig:Sinhala_Language_Tree}. The writing systems used by a language being related, not being related, or the degree to which they are related, is not directly correlated to the said languages being related, not being related, or the degree to which they are related. As an example, note that the writing systems of \textit{Sinhala} and \textit{Tamil} are reasonably closely related. But \textit{Tamil} is not shown on Fig~\ref{fig:Sinhala_Language_Tree} given that it does not belong to the \textit{Indo-European} language family. It belongs to the \textit{Dravidian} language family. A language that is shown as a single entity in Fig~\ref{fig:Sinhala_Language_Tree} may have multiple entries in this diagram. An example is the language shown as \textit{Hindustani} in Fig~\ref{fig:Sinhala_Language_Tree} splits into \textit{Hindi} and \textit{Urdu} in this diagram, where the former uses the \textit{Devanagari} script and the latter a derivation of \textit{Arabic} script. Conversely, languages that may be shown as distinct entities in Fig~\ref{fig:Sinhala_Language_Tree}, may map to a singular entity in the script tree. An example is \textit{English} which would go under \textit{Proto-Germanic} and \textit{Italian} which would go under \textit{Proto-Italic} in an extended version of Fig~\ref{fig:Sinhala_Language_Tree}, would both map to \textit{Latin} in this diagram, given that both of them use the \textit{Latin} script.}	
	\label{fig:Sinhala_Script_Tree}
\end{figure*}

\begin{figure*}[!hbt]	
	\centering 
\includegraphics[width=\textwidth]  
{images/Sinhala_Script.pdf}
	\caption{The phonetic and graphemic view of the contemporary Sinhala script. \textbf{Notes:} (1) Each unmarked consonant in Sinhala carries an inherent vowel. Thus, there is no diacritic form here. Typically, \texttt{[a]} is sounded when an unmarked consonant appears at the beginning of a word, and \texttt{[\schwa]} is sounded when an unmarked consonant appears anywhere else in a word. However, there are exceptions to this rule. These exceptions are the primary reason Sinhala cannot be said to have \textit{regular (perfect) orthography} and stops at \textit{near regular (perfect) orthography}~\cite{chandralal2010sinhala}. (2) The \textit{hal kirīma} mark is used when the aforementioned inherent vowel is removed from a consonant and thus has no sound by itself but rather the negation of one~\cite{gair1996sinhala}. (3) There is controversy in including the voiceless affricate in the \textit{śuddha} set because some purists object to the fact, pointing out that it does not appear in the main text of \textit{Sidatsangarāwa}. Nevertheless, \textit{Sidatsangarāwa} uses it in some examples; so we can be sure that the character did exist in \textit{Elu}~\cite{gunasekara1891comprehensive}. (4) The Retroflex pronunciations, which could be found in historic speech, are no longer phonemic in modern Sinhala~\cite{karunatillake1990introduction,wasala2010data}. This change was resisted for a time by forcing children to recite \textit{Sakaskadaya} intended to train them to use the \textit{correct} pronunciations. But with that practice falling into disuse and with the popular usage suppressing the academic, it is rare to encounter them with distinct pronunciations~\cite{ruberu1974educational,gunasena2017teaching}. (5) This letter is not used in ancient or contemporary use. Its usefulness is questionable, but it is included in the standard alphabet~\cite{daniels1992unicode}.}
	\label{fig:Sinhala_Script}
\end{figure*}

Regardless of the rich historical array of literature spanning several millennia (starting between $3^{rd}$ to $2^{nd}$ century BCE~\cite{ray2003archaeology,herath1994practical}), modern natural language processing tools for the Sinhala language are scarce~\cite{de2015Sinhala,ranathunga2022some}. Further, Sinhala is a \textit{diglossic} language~\cite{mallikadevi2023analysis} which has two variations: the literary form and the spoken form.
The \textit{Vedda (Veddah)} language is considered a Sinhala creole~\cite{dharmadasa1974creolization,surendra2024preservation,nawarathna2024preserving} or a language isolate~\cite{welikala2024genetic}.

\begin{figure*}[!hbt]	
	\centering 
\includegraphics[width=\textwidth]  
{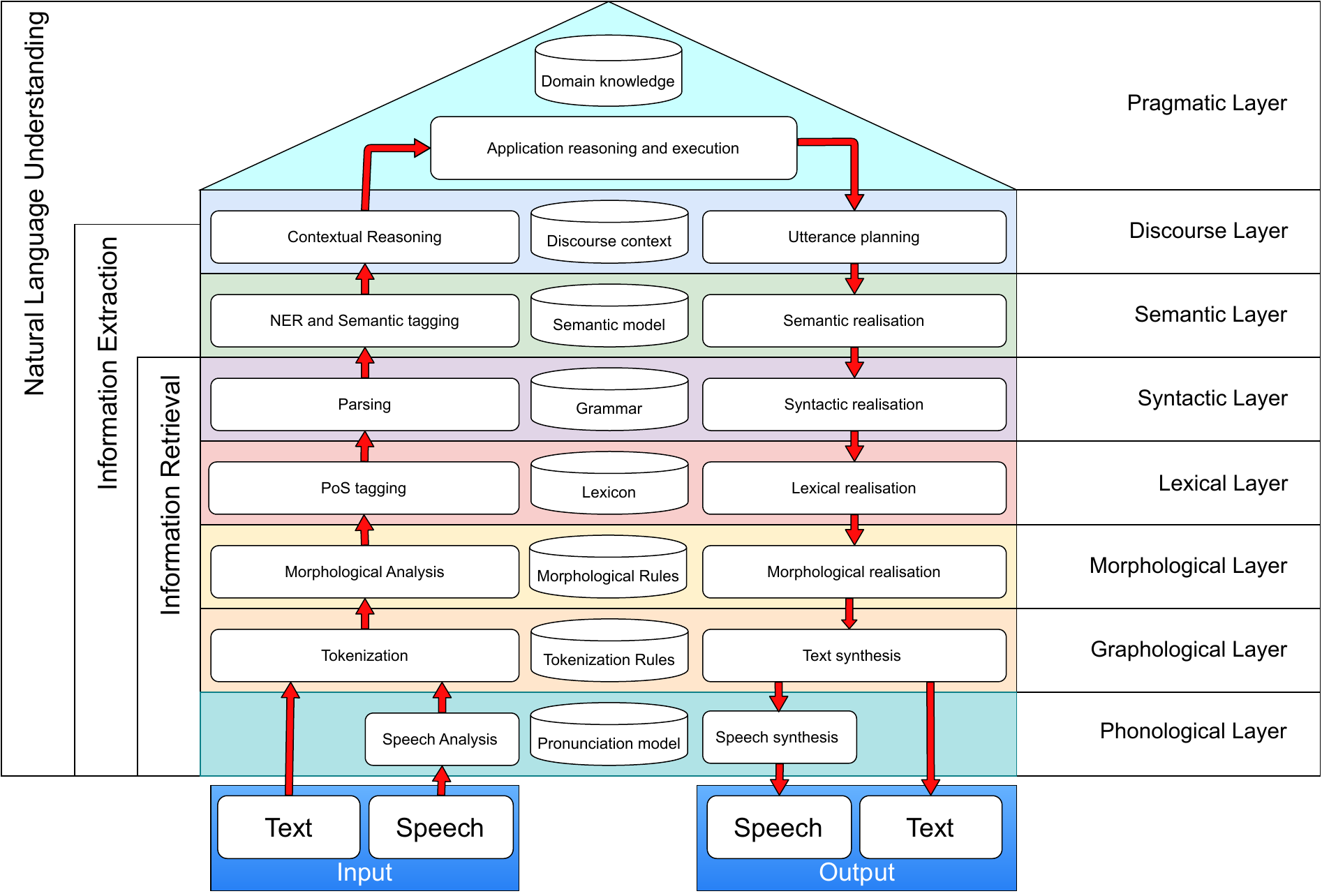}
	\caption{Traditional NLP layers and tasks (Derived from the work of~\citet{wijeratne2019natural} and extended to encapsulate a wider range by this work.)}	
	\label{fig:nlpLayers}
\end{figure*}

Natural Language Processing (NLP) is a broad area covering all computational processing and analysis of human languages. To achieve this end, NLP systems operate at different levels~\cite{wijeratne2019natural,liddy2001natural,wimalasuriya2010ontology}. A graphical representation of NLP layers and application domains are shown in Figure~\ref{fig:nlpLayers}. On one hand, according to~\citet{liddy2001natural}, these systems can be categorized into the following layers; \textit{phonological}, \textit{morphological}, \textit{lexical}, \textit{syntactic}, \textit{semantic}, \textit{discourse}, and \textit{pragmatic}. The \textit{phonological} layer deals with the interpretation of language sounds. As such, it consists of mainly speech-to-text and text-to-speech systems. In cases where one is working with the written text of the language rather than speech, it is possible to replace this layer with tools which handle Optical Character Recognition (OCR) and language rendering standards (such as Unicode~\cite{unicode1996unicode}). The \textit{morphological} layer analyses words at their smallest units of meaning. As such, analysis of word lemmas and prefix-suffix-based inflection is handled in this layer. \textit{Lexical} layer handles individual words. Therefore, tasks such as Part of Speech (PoS) tagging happens here. The next layer, \textit{syntactic}, takes place at the phrase and sentence level, where grammatical structures are utilised to obtain meaning. \textit{Semantic} layer attempts to derive the meanings from the word level to the sentence level. Starting with Named Entity Recognition (NER) at the word level and working its way up by identifying the contexts they are set in until arriving at the overall meaning. The \textit{discourse} layer handles meaning in textual units larger than a sentence. In this, the function of a particular sentence maybe contextualised within the document it is set in. Finally, the \textit{pragmatic} layer handles contexts read into contents without having to be explicitly mentioned~\cite{wijeratne2019natural,liddy2001natural}. Some forms of anaphora (coreference) resolution~\cite{van1992presupposition,lappin1994algorithm,soon2001machine,ng2002improving,mitkov2014anaphora} fall into this application. 

On the other hand,~\citet{wimalasuriya2010ontology} categorize NLP tools and research by utility. They introduce three categories with increasing complexity;~\textit{Information Retrieval} (IR), \textit{Information Extraction} (IE), and \textit{Natural Language Understanding} (NLU). \textit{Information Retrieval} covers applications, which search and retrieve information which are relevant to a given query. For pure IR, tools and methods up-to and including the \textit{syntactic} layer in the above analysis are used. \textit{Information Extraction}, on the other hand, extracts structured information. The difference between IR and IE is the fact that IR does not change the structure of the documents in question. Be them structured, semi-structured, or unstructured, all IR does is fetch them as they are. In comparison, IE, takes semi-structured or unstructured text and puts them in a machine-readable structure. For this, IE utilizes all the layers used by IR and the \textit{semantic} layer. \textit{Natural Language Understanding} is purely the idea of cognition. Most NLU tasks fall under AI-hard category and remain unsolved~\cite{wijeratne2019natural}. However, with varying accuracy, some NLU tasks such as machine translation\footnote{This is, however, not without the criticism of being nothing more than a \textit{Chinese room}~\cite{preston2002views} rather than true NLU.} are being attempted. The \textit{pragmatic} layer of the above analysis belongs to the NLU tasks while the \textit{discourse layer} straddles information extraction and natural language understanding~\cite{wijeratne2019natural}.   

The objective of this paper is to serve as a comprehensive survey on the state of natural language processing resources for the Sinhala language. The initial structure and content of this survey are heavily influenced by the preliminary surveys carried out by \citet{de2015Sinhala} and~\citet{wijeratne2019natural}. However, our hope is to host this survey at arXiv as a perpetually evolving, living research article~\cite{shanahan2015living,sopinka2020envisioning} which continuously gets updated as new research and tools for the Sinhala language are created and made publicly available\footnote{In contrast to subsequent similar studies~\cite{kumar2026bhashasutra}}. 
We also discuss how the non-compliance of policies to put data and code online~\cite{gabelica2022many}, after the research is concluded and the paper is published, has negatively impacted the growth and sustainability of Sinhala NLP. In fact, studies such as \citet{occhini2026artificial} put the \textit{AI readiness} of Sinhala at a low 2.9\%\footURL{https://nisansa.lk/?s=45m7Kd}. This puts it in second place behind Tamil when considering Sri Lanka\footURL{https://nisansa.lk/?s=ixjmZ8} and globally at 53rd\footURL{https://nisansa.lk/?s=njrVq3}.
Hence, it is our hope that this work will help future researchers who are engaged in Sinhala NLP research to conduct their literature surveys efficiently and comprehensively. For the success of this survey, we shall also consider the Sri Lankan NLP tools repository, \gh{https://github.com/lknlp/lknlp.github.io}{lknlp}. This manuscript is at version $\version$. The latest version of the manuscript can be obtained from arXiv\footURL{https://arxiv.org/abs/1906.02358} or ResearchGate\footURL{http://bit.ly/31AhvvR}. Given that this work was initially published in 2019, when citing this work, it is recommended to cite the most recent version to avoid the reviewers questioning its continued contemporary relevance. An example of an updated reference is given in~\lstlistingname~\ref{lst:ref}.

\begin{lstlisting}[caption={Example of an updated reference},captionpos=b,label={lst:ref},float,floatplacement=!htb]
@article{de2026survey,
  title={{Survey on Publicly Available Sinhala Natural Language Processing Tools and Research}},
  author={de Silva, Nisansa},
  journal={arXiv preprint arXiv:1906.02358v27},
  year={2026}
}
\end{lstlisting}

Figure~\ref{fig:authors} in Appendix~\ref{app:authors} shows the most prolific researchers in the domain of Sinhala NLP. The nodes contain the name of the researcher along with the total number of Sinhala NLP papers that the researcher has authored. The edges between the two researchers are labelled with the number of Sinhala NLP papers the relevant pair of researchers have co-authored. When selecting authors, we have applied a threshold of 3  Sinhala NLP publications. Given that the objective of the visualization is to portray the cooperation between researchers, we have also added the strongest edge that connects each researcher to the rest of the researchers in the graph. The few isolated nodes are researchers who have authored at least 3 Sinhala NLP publications but do not have any coauthored papers with anyone else on the graph. We have also added labels to clusters in cases where all or the majority of researchers in those clusters have the same affiliation. It is observable that the cluster from the \textit{Department of Computer Science \& Engineering, University of Moratuwa} is the most prolific in Sinhala NLP research.

Figure~\ref{fig:CitationProbability} in Appendix~\ref{app:authors} shows the probability of studies from the institutions to which the most prolific authors from Figure~\ref{fig:authors} are affiliated citing institutions of the same set. We observe both interesting and disturbing trends which we discuss in the figure caption. 
Figure~\ref{fig:InstTopicMatrix} in Appendix~\ref{app:authors} \textInstTopicMatrix.
Figure~\ref{fig:newAuthors} in Appendix~\ref{app:authors} \textNewAuthors.
Figure~\ref{fig:InternationalCollab} in Appendix~\ref{app:authors} \textInternationalCollab.
Figure~\ref{fig:authorAreas} in Appendix~\ref{app:authors} shows the mapping between authors with at least 10 papers in the Sinhala NLP domain and their research interests, denoted by the subsection titles of the Section~\ref{Sinhala} of this paper. If two or more authors on the diagram have co-authored a paper, it gets counted for each of the authors separately without any bias on the author order listed on the publication. For example, the paper \textit{Building a wordnet for Sinhala}~\cite{wijesiri2014building} is counted for both \textit{Nisansa de Silva} and \textit{Gihan Dias}. If a single paper contributes to more than one research area, that paper is counted for all of the research areas to which it contributes. For example, the paper \textit{Sinhala Text Classification: Observations from the Perspective of a Resource Poor Language}~\cite{de2015Sinhala} which introduces a new Sinhala text classification data set is counted for both \textit{Data Sets} (Section~\ref{DataSets}) and \textit{Text Classification} (Section~\ref{Sec:TextClassification}).
Figure~\ref{fig:authorCareers} in Appendix~\ref{app:authors} \textAuthorCareers captured by this survey (not counting the current year).

The remainder of this survey is organized as follows: Section~\ref{properties} introduces some important properties and conventions of the Sinhala language, which are important for the development and understanding of Sinhala NLP. Section~\ref{Sinhala} discusses the various tools and research available for Sinhala NLP. In this section, we will discuss both pure Sinhala NLP tools and research, as well as hybrid Sinhala-English work. We will also discuss research and tools that contribute to Sinhala NLP, either along with or with the help of Tamil, the other official language of Sri Lanka. Section~\ref{FirstSources} gives a brief introduction to the primary language sources used by the studies discussed in this work. Finally, Section~\ref{Conclusion}, concludes the survey.





%% file: 02_Properties.tex
\section{Properties of the Sinhala Language}
\label{properties}

\begin{table*}[tb]
\caption{
\textbf{Examples for grammatical cases and inflection of animate common nouns} 
}
	\label{tab:animate}
	\centering 
\includegraphics[width=1.0\textwidth]{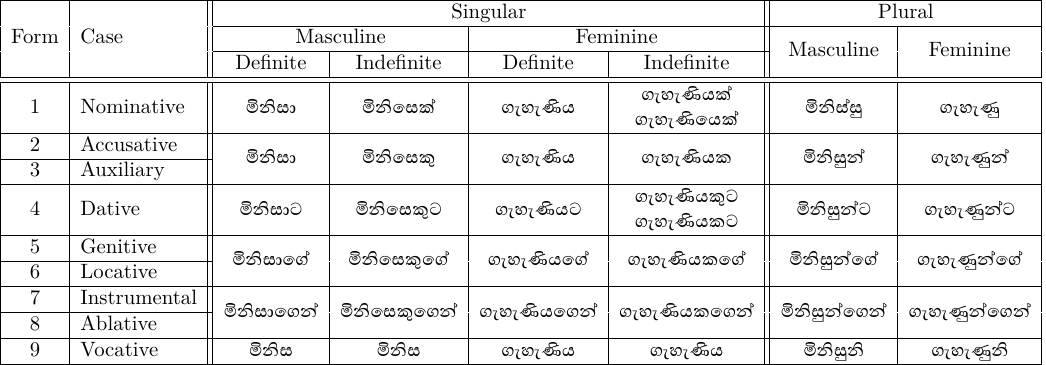}
\end{table*}

\begin{table}[!htb]
\caption{
\textbf{Examples for grammatical cases and inflection of inanimate common nouns}
\newline\footnotesize Note that the grammatical cases of \textit{Auxiliary} and \textit{Vocative} do not exist for inanimate nouns in Sinhala.
}
\label{tab:inanimate}
	\centering 
\includegraphics[width=0.48\textwidth]{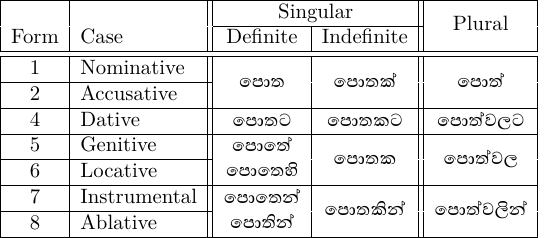}
\end{table}

\begin{figure*}[t!]
    \centering
    \begin{subfigure}[t]{0.495\textwidth}
        \centering
        \includegraphics[width=0.99\textwidth]{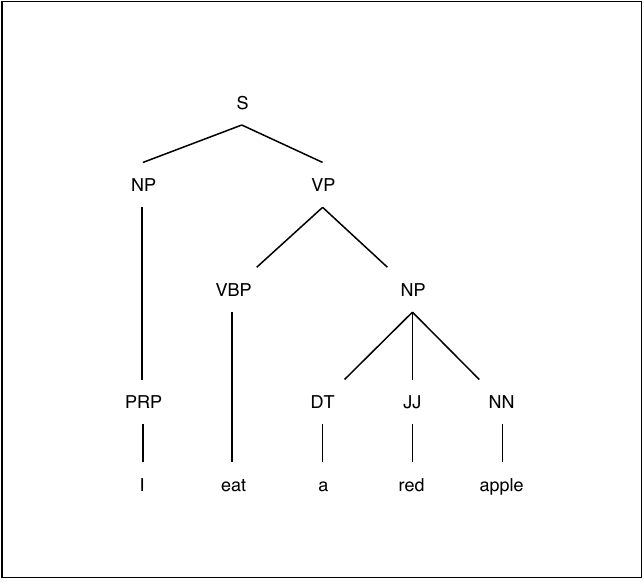}
        \caption{English}
    \end{subfigure}%
    \begin{subfigure}[t]{0.495\textwidth}
        \centering
        \includegraphics[width=0.99\textwidth]{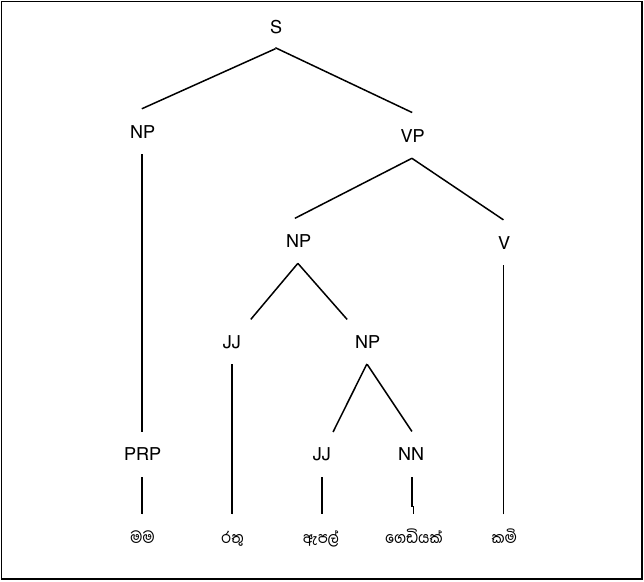}
        \caption{Sinhala}
    \end{subfigure}
    
    \begin{subfigure}[t]{0.495\textwidth}
        \centering
        \includegraphics[width=0.99\textwidth]{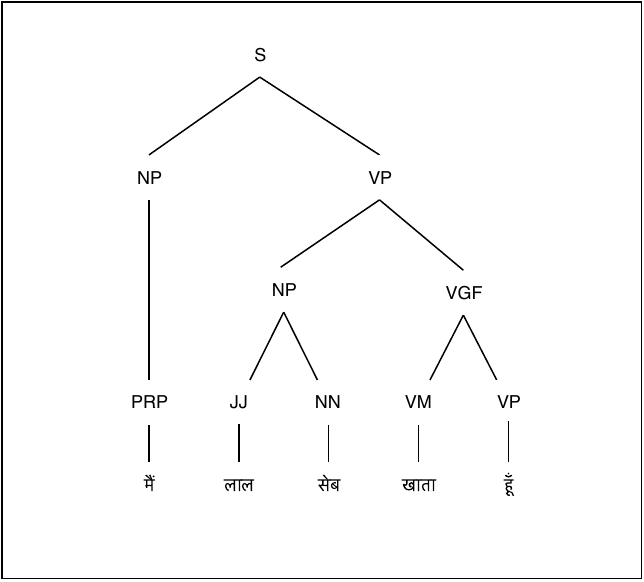}
        \caption{Hindi}
    \end{subfigure}%
    \begin{subfigure}[t]{0.495\textwidth}
        \centering
        \includegraphics[width=0.99\textwidth]{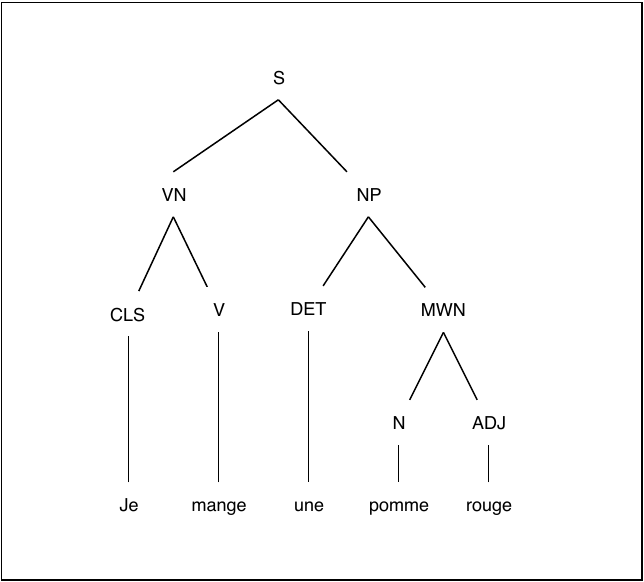}
        \caption{French}
    \end{subfigure}
    \caption[Parse trees for the sentence ``\textit{I am eating a red apple}" in four languages.]{\textbf{Parse trees for the sentence ``\textit{I eat a red apple}" in four languages.}\newline 
    }
    \label{parse}
\end{figure*}

Before moving on to discussing Sinhala NLP resources, we shall give a brief introduction to some of the important properties of Sinhala language, which impact the development of Sinhala NLP resources.
Sinhala grammar has two forms: written (literary) and spoken. These forms differ from each-other in their core grammatical structures~\cite{mallikadevi2023analysis,fairbanks1968colloquial,englebretson2005santa,miyagishi2005accusative}. The written form strictly adheres to the \textit{SOV} (Subject, Object, and Verb) configuration~\cite{disanayaka1985say,pallatthara1966sinhala}. Further, in the written form, \textit{subject-verb agreement} is enforced~\cite{kanduboda2013usage} such that, in order to be grammatically correct, the subject and the verb must agree in terms of: gender (male/female), number (singular/plural) and person (1st/2nd/3rd). However, in spoken Sinhala, the \textit{SOV} order can be neglected~\cite{liyanage2012computational} and \textit{male singular 3rd person} verb can be used for all nouns~\cite{kanduboda2013usage}. Sinhala is also a head-final language, where the complements and modifiers would appear before their heads~\cite{karunatilaka1997sinhala} this is similar to that of English and dissimilar to that of French. In total, according to~\citet{Abhayasinghe1998sinhala}, there are 25 types of simple sentence structures in Sinhala and according to~\citet{madushani2025linguistic} there are 8 types of Verb Phrases.
Similar to many Indo Aryan languages, \textit{animacy} plays a major role in Sinhala grammar in syntactic and semantic roles~\cite{jany2006relationship,garland2005morphological,henderson2005between}. Comparative studies done by~\citet{noguchi1984shinharago} and by~\citet{miyagishi2003comparison, miyagishi2005accusative} have found that \textit{animacy} extends its influence from phrase level to sentence level in Sinhala (e.g., Usage of post-positions~\cite{disanayaka1985say,chandralal2010sinhala}). On this matter, Table~\ref{tab:animate} explains grammatical cases and inflections of animate common nouns while Table~\ref{tab:inanimate} explains grammatical cases and inflections of inanimate common nouns.   
We provide a comparative analysis of parsing the very simple English sentence ``\textit{I eat a red apple}" and its Sinhala, Hindi, and French translations in Fig~\ref{parse}. English and French parsing was done using the \textit{Stanford Parser}\protect\footURL{http://nlp.stanford.edu:8080/parser/}. Hindi parsing was done using the \textit{IIIT-Hyderabad Parser}\footURL{http://ltrc.iiit.ac.in/analyzer/} and the study by~\citet{singh2016syntax}. 

\citet{herath1989sinhalese,herath1990formalization} argue that pure Sinhala words did not have suffixes and that adding suffixes was incorporated to Sinhala after 12th century BC with the influx of Sanskrit words. With this, they declare Sinhala to have to following types of words:
\begin{enumerate}
    \item Suffixes
    \item Nouns
    \item Cases
    \item Verbs
    \item Conjunctions and articles
    \item Adjectives
    \item Demonstratives, Interrogatives, and negatives
    \item Particles and prefixes
\end{enumerate}
They further divide nouns into five groups: material, agentive, common, abstract, and proper. In addition to these, they also introduce compound nouns.We show the noun categorization proposed by~\citet{herath1989sinhalese} in Table~\ref{tab:nouns}.

\begin{table}[!htb]
\caption{
\textbf{Noun categorization by~\citet{herath1989sinhalese}}
}
\label{tab:nouns}
	\centering 
\includegraphics[width=0.48\textwidth]{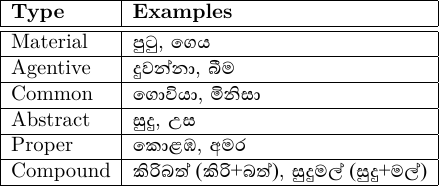}
\end{table}

\citet{herath1990formalization} categorize Sinhala suffixes along the attributes of: gender, number, definiteness, case, and conjunctive. They further claim that there are 3 types of suffixes: \textit{Suf1} adds gender, number, and definiteness; \textit{Suf2} adds case; and \textit{Suf3} adds conjunctive. Conjunctive is claimed to be equivalent to \textit{too} and \textit{and} in English. We show an extension of the suffix structure proposed by~\citet{herath1990formalization} in Table~\ref{tab:suffix}. In their analysis on register variation (vocabulary) of 60 languages,~\citet{li2022register} observes that Sinhala exhibits homogeneity between 0.5 and 1.0 in the three corpora considered: \textit{CC} (macro-web register - Common Crawl), \textit{TW} (social media register - Twitter), and \textit{WK} (Wikipedia register - March
2020). \citet{dissanayake2025meh} discusses a few Sinhala discourse markers that have seeped into Sri Lankan English.

\begin{table*}[!htb]
\caption{
\textbf{Extension of the suffix structure proposed by~\citet{herath1990formalization}}
\newline\footnotesize Number = Singular (S) / Plural (P)
\newline\footnotesize Definite = Definite (D) / Indefinite (I) / Undecided (U)
\newline\footnotesize Case = Nominative (N) / Accusative (A) / Dative (Da) / Genitive (G) / Instrumentive (In) / Auxiliary (Au) / Locative (L) / Ablative (Ab)
\newline\footnotesize Conjun = with suffix \textit{th} (Y) / without suffix \textit{th} (N)
}
\label{tab:suffix}
	\centering 
\includegraphics[width=\textwidth]{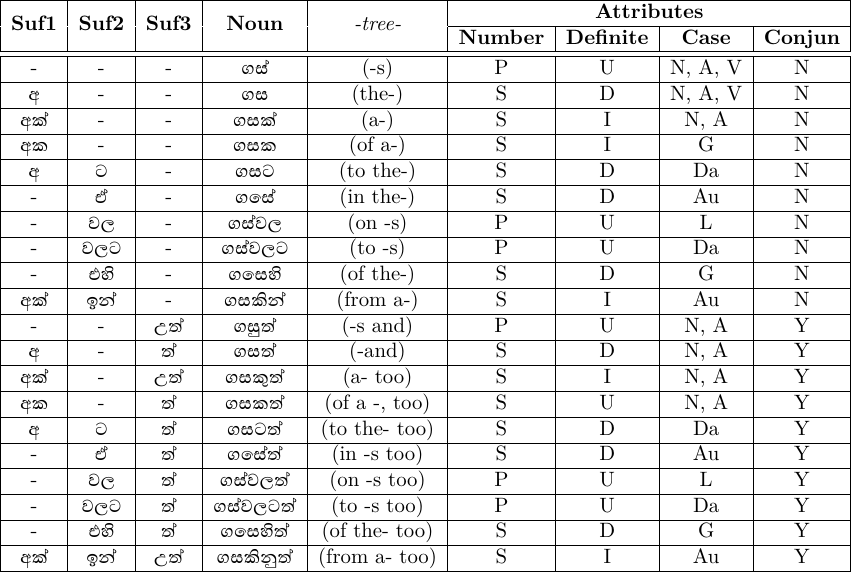}
\end{table*}



%% file: 03_Sinhala.tex
\section{Sinhala NLP resources}
\label{Sinhala}

\begin{figure*}[!hbt]	
	\centering 
\includegraphics[width=\linewidth] 
{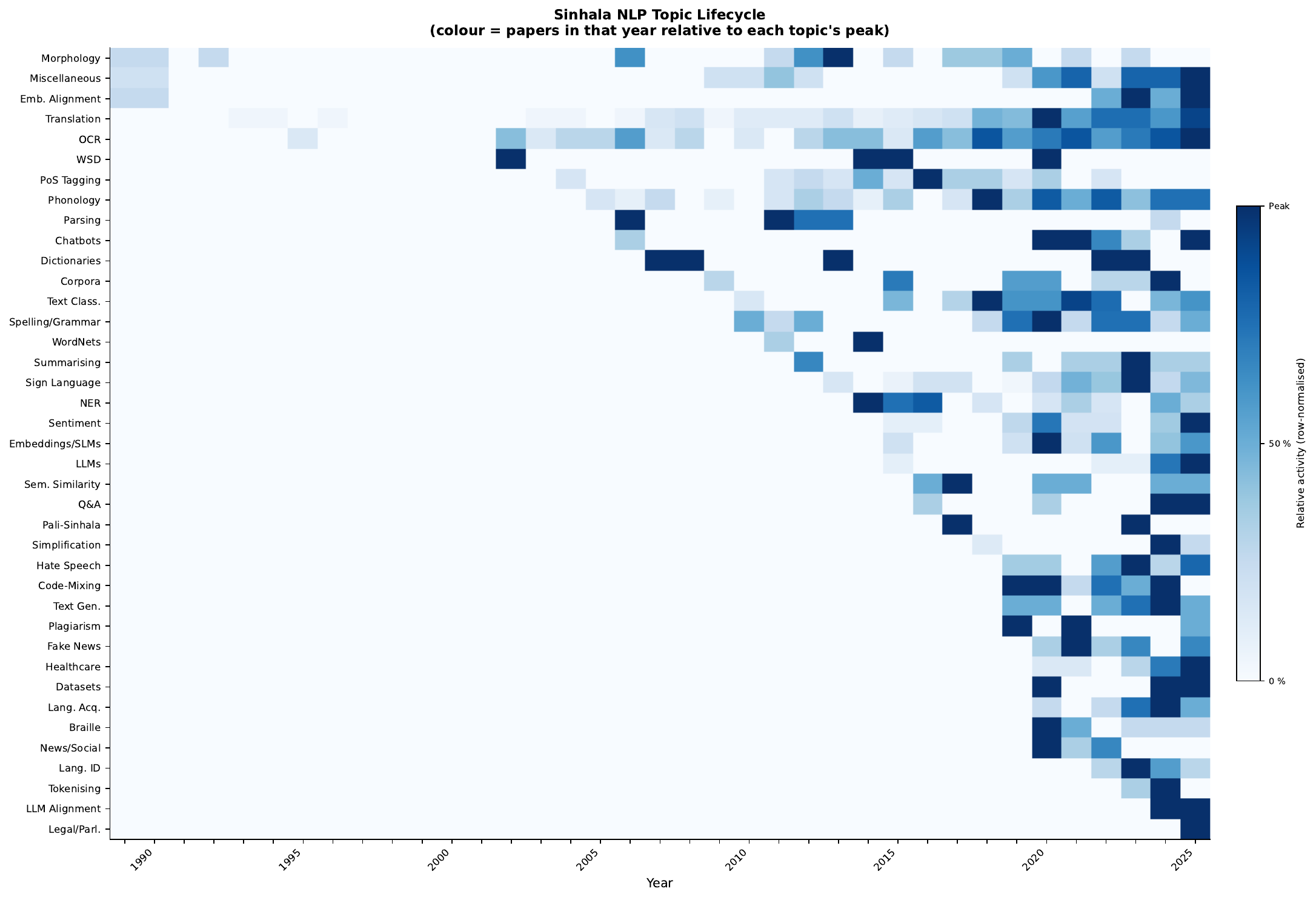}
\caption{\captionTopicLifecycle}	\label{fig:topicLifecycle}
\end{figure*}

\begin{figure*}[!hbt]	
	\centering 
\includegraphics[height=0.825\textheight ] 
{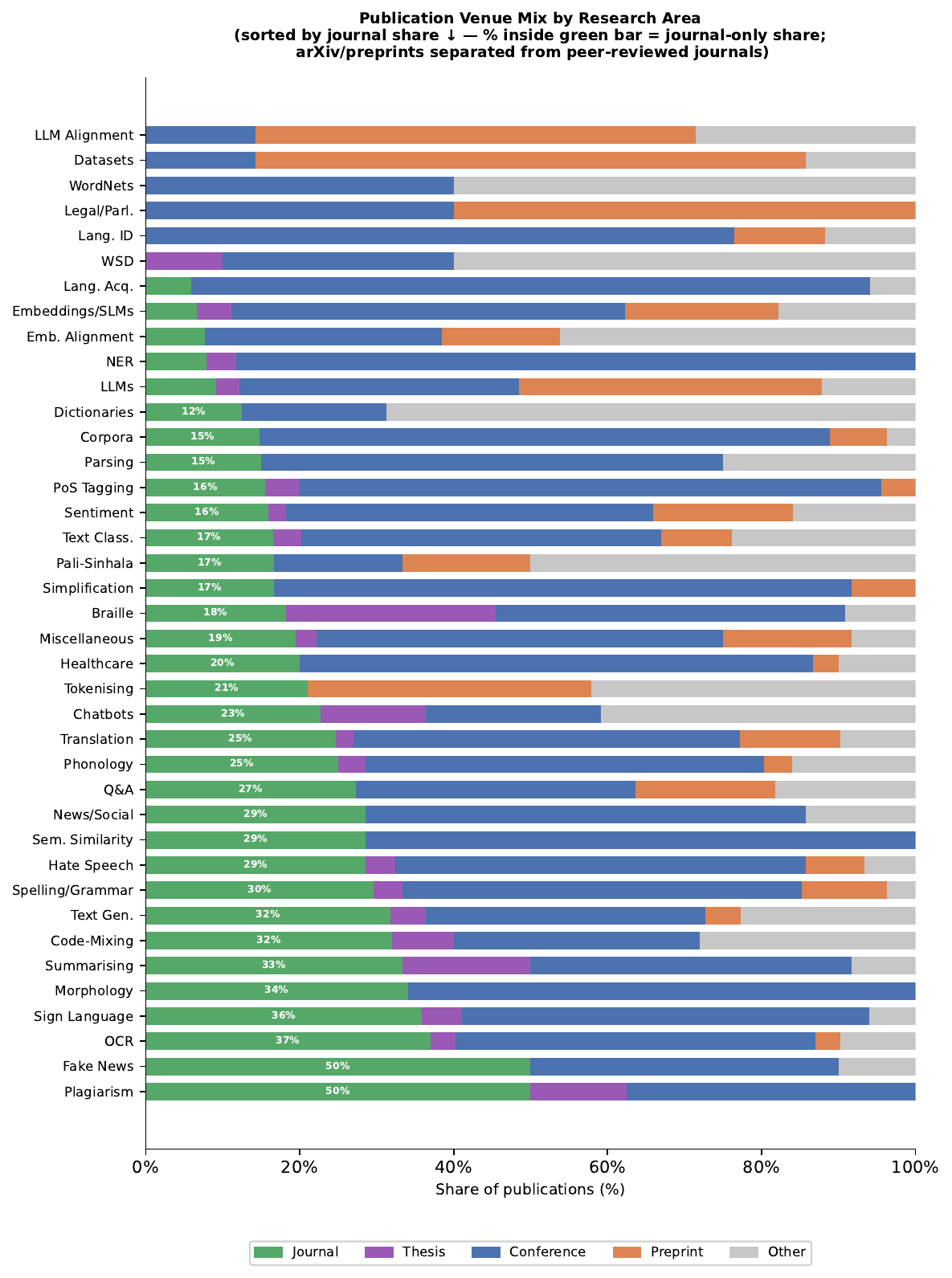}
\caption{\captionVenueMix}	\label{fig:venueMix}
\end{figure*}

This section surveys the resources, tools, and applications that constitute the Sinhala NLP ecosystem. We broadly follow the hierarchical organisation illustrated in Figure~\ref{fig:nlpLayers}, while also covering language resources that do not naturally fit within that hierarchy, including corpora, datasets, dictionaries, and WordNets. Throughout this survey, we focus on NLP resources and computational methods rather than the underlying mechanics of Sinhala script handling, which have been comprehensively discussed elsewhere~\cite{samaranayake1989standard,samaranayake2003introduction,dias2004development,dias2005challenges,weerasinghe2006sinhala,sandeva2009design}. Fig~\ref{fig:topicLifecycle} \textTopicLifecycle{} for Sinhala NLP research discussed in this work. The topic axis was populated using the subsections included in this section of the study. Fig~\ref{fig:venueMix} \textVenueMix{} followed by conference, thesis, and preprint in order.

As shown by Fig~\ref {fig:topicLifecycle}, research on Sinhala NLP spans more than three decades. One of the earliest documented efforts was the pioneering work of~\citet{herath1991machine}. Although ambitious for its time, the project was constrained by the limited computational resources and linguistic data available during the early 1990s, resulting in relatively limited subsequent development. Later, \citet{nandasara2009past} reviewed the state of Sinhala language computing. Given that it was a decade old by the time the first edition of this survey was compiled, we observe the existence of many new discoveries in Sinhala NLP which have not been taken into account by it. Similarly, \citet{nandasara2016bridging} discussed the challenges and opportunities associated with Sinhala in computer science, providing valuable context for subsequent developments in the field.

Since those surveys were published, Sinhala NLP has experienced substantial growth across nearly every layer of the NLP pipeline. Nevertheless, research activity has been unevenly distributed. Optical character recognition (OCR) has attracted the greatest volume of research. In contrast, comparatively fewer studies have addressed higher-level language understanding tasks such as semantic analysis, discourse processing, and reasoning. Another notable direction is the development of end-to-end machine translation systems. Among these, the long-running Sinhala--English translation project~\cite{hettige2006morphological,hettige2006parser,hettige2006first,hettige2007developing,hettige2007transliteration,hettige2007using,hettige2008web,hettige2008web1,hettige2009theoretical,hettige2010evaluation,hettige2010varanageema,hettige2011computational,hettige2013selected,hettige2012multi,hettige2013masmt,hettige2014sinhala,hettige2016multi,hettige2017phrase,hettige2021masmt4} represents one of the most sustained and comprehensive research efforts in Sinhala NLP.
%
The increasing maturity of the Sinhala NLP ecosystem is also reflected in the availability of reusable software. For example, \gh{https://github.com/ysenarath/sinling/}{Sinling} provides an open-source collection of NLP utilities, including a tokenizer, stemmer, part-of-speech tagger, morphological joiner, and morphological splitter. The toolkit is distributed through \textit{PyPI}.

Although this survey focuses on Sinhala, it is important to recognise its close relationship with Tamil, the other official language of Sri Lanka. Tamil is likewise considered a low-resource language; however, its substantially larger global speaker population has fostered a richer ecosystem of datasets, tools, and research~\cite{wijeratne2019natural}. Consequently, many studies have explored multilingual approaches that jointly leverage Sinhala and Tamil resources. The bilingual nature of Sri Lankan government publications, legislation, and news media naturally produces valuable parallel corpora that have supported numerous multilingual NLP studies discussed later in this survey. Beyond Tamil, several studies have also investigated Sinhala in combination with other languages, including Japanese~\cite{herath1994practical,herath1993generation,herath1996bunsetsu,thelijjagoda2004japanese,kanduboda2011role}.

Finally, Sinhala has increasingly been incorporated into broader multilingual resources. For example, it is included in the metadata collection of more than 2,800 languages compiled by~\citet{van2022writing}, enabling comparative analyses across diverse linguistic typologies. In addition, \citet{herathresearch,herathresearch2} compiled a comprehensive report on the Sinhala lexicon, providing an important linguistic foundation for subsequent NLP research.

\subsection{Non-Parallel Corpora}
\label{sec:corpora}

The availability of large, diverse, and publicly accessible text corpora is one of the primary factors determining the progress of NLP research for a language. Non-parallel corpora are particularly valuable because they support a wide range of tasks, including language modelling, pre-training, information retrieval, text classification, and the construction of lexical resources. This subsection surveys publicly reported Sinhala non-parallel corpora. Parallel corpora, which are primarily intended for machine translation and multilingual learning, are discussed separately in Section~\ref{sec:ParCorp}.

Early publicly available Sinhala corpora were relatively modest in size and were typically created through web crawling. One of the earliest substantial efforts was the \osf{https://osf.io/a5quv/}{SinMin}\footnote{\hf{https://huggingface.co/datasets/sinhala-nlp/sinmin}{Third-party mirror}} corpus introduced by~\citet{upeksha2015implementing,upeksha2015comparison}. The corpus was constructed by crawling Sinhala web pages and was subsequently extended with additional resources, including \gh{https://github.com/adeesha/SinMinTemp/tree/master/jathaka_txt}{Jathaka Stories} and a collection of \gh{https://github.com/scnakandala/sinhala_text_mining/tree/master/stm/resources/news_supervized_learning}{web-crawled news articles}. Although modest by the standards of contemporary multilingual corpora, \texttt{SinMin} represented an important milestone by providing one of the first publicly reusable corpora for Sinhala NLP research. 
%
Later, a smaller \osf{https://osf.io/tdb84/}{Sinhala news corpus} was created by~\citet{de2015Sinhala}. 
%

The emergence of large multilingual language models substantially increased the demand for web-scale corpora. Consequently, Sinhala became part of several multilingual corpus construction efforts. \citet{nguyen2023culturax} introduced \hf{https://huggingface.co/datasets/uonlp/CulturaX}{CulturaX}, a cleaned multilingual corpus intended for pre-training large language models. Their release contains 753,655 Sinhala documents comprising approximately 880 million tokens. Nevertheless, Sinhala remains severely under-represented within the overall collection, accounting for only 0.01\% of the multilingual corpus. By comparison, Tamil contributes 0.07\%, Hindi 0.27\%, and English 45.13\%. 

A similar trend is evident in \citet{kudugunta2024madlad}, who introduced the \gh{https://github.com/google-research/google-research/tree/master/madlad_400}{MADLAD-400} multilingual corpus spanning 419 languages. For \hf{https://huggingface.co/datasets/allenai/MADLAD-400/tree/main/data/si}{Sinhala}, the authors report approximately 349k cleaned documents (out of 788k noisy), 16M clean sentences (out of 22.1M noisy), 1.9B clean tokens (out of 3.4B noisy). In addition to monolingual text, the project reports more than 7.3 million sentence pairs involving Sinhala. They also released the corresponding multilingual \gh{https://github.com/google-research/google-research/tree/master/madlad_400}{trained checkpoints}
%

Several other multilingual corpora also include substantial Sinhala text. The \textit{OSCAR}\footURL{https://oscar-project.org/} corpus~\cite{ortiz-suarez-etal-2020-monolingual,OrtizSuarezSagotRomary2019} contains 301,066 Sinhala documents comprising approximately 173 million words obtained through large-scale web mining. More recently, 
%

Likewise, \citet{imani2023glot500} introduced the \gh{https://github.com/cisnlp/Glot500}{Glot500} corpus covering 500 languages, including a Sinhala component comprising over 7.2 million sentences. Beyond releasing the corpus, the authors evaluated pretrained Sinhala language models on tasks such as sentence retrieval, named entity recognition, part-of-speech tagging, text classification, round-trip alignment, and perplexity estimation.

Alongside these multilingual resources, several corpora have been developed specifically for Sinhala. \citet{dhananjaya2022bertifying} introduced the \gd{https://tinyurl.com/42un7a9y}{sin-cc-15M} corpus, which the authors reported to be the largest monolingual Sinhala corpus available at the time of publication. 
A large-scale \gh{https://github.com/Sinhala-NLP/NSINA}{Sinhala NEWS corpus} containing more than 500,000 articles collected from multiple online news outlets was later released by~\citet{hettiarachchi2024nsina}.
Similarly, a corpus collected from the Sinhala blog \textit{Kalaya}\footURL{http://www1.kalaya.org/}, together with the associated processing code, has been made publicly available on \gh{https://bit.ly/KalayaCorpus}{GitHub}. 

Social media has also emerged as an important source of Sinhala language data. \citet{wijeratne2020sinhala} publicly released a large corpus and accompanying stop-word list derived from a \gh{https://github.com/LIRNEasia/FacebookDecadeCorpora}{decade of Sinhala Facebook posts}. Although the original repository is no longer maintained, a third-party \hf{https://huggingface.co/datasets/sinhala-nlp/FacebookDecadeCorpora}{mirror} has preserved public access to the complete collection. 

Several additional corpora have contributed to the broader Sinhala NLP ecosystem. \citet{weerasinghe2009corpus} developed a word corpus containing approximately 35,000 entries, although it no longer appears to be publicly accessible. Furthermore, \citet{guzman2019flores} released two monolingual Sinhala corpora as part of the FLORES project, consisting of more than 155,000 filtered \opus{http://bit.ly/2EQZ7oM}{Sinhala Wikipedia} sentences and over 5.1 million \opus{http://bit.ly/2ZaQFZo}{Sinhala common crawl} sentences. Although originally intended to support multilingual machine translation, these corpora have subsequently been adopted in a variety of Sinhala NLP tasks beyond translation.
%

Overall, the development of Sinhala non-parallel corpora has progressed from relatively small web-crawled collections to web-scale multilingual resources containing billions of tokens. Nevertheless, the quantitative comparison with higher-resource languages demonstrates that Sinhala remains substantially underrepresented in publicly available corpora. Addressing this resource gap remains one of the most important prerequisites for further advances in Sinhala NLP, particularly for data-intensive approaches such as large language models and foundation models.

\subsection{Data Sets}
\label{DataSets}

This subsection surveys publicly available Sinhala datasets that are either task-agnostic or intended to support multiple NLP tasks. Datasets developed specifically for individual applications, such as sentiment analysis, named entity recognition, or machine translation, are discussed in their respective sections.

The growing popularity of multilingual benchmark suites has led to increasing efforts to incorporate Sinhala into standardised evaluation datasets. One such initiative is \hf{https://huggingface.co/datasets/CohereForAI/Global-MMLU}{Global-MMLU}, introduced by~\citet{singh2024global} as a multilingual extension of the original Massive Multitask Language Understanding (MMLU) benchmark~\cite{hendrycks2020measuring}. The authors report improvements over the earlier multilingual extension proposed by~\citet{ustun2024aya}. 
Nevertheless, Sinhala remains underrepresented within the benchmark, and the project actively solicits community volunteers to extend coverage for low-resource languages such as Sinhala\footURL{https://nisansa.lk/?s=Ww20rp}. At present, contributors are acknowledged on the project website but are not formally included as co-authors of the associated research publication.

Several datasets have been developed to capture naturally occurring Sinhala text from online platforms. For example, \citet{de2025linguistic} released a \gh{https://github.com/Yomald93/Linguistic-Analysis-of-Sinhala-YouTube-Comments-on-Sinhala-Music-Videos}{corpus of 63,471 Sinhala YouTube comments} collected from Sinhala music videos. Of these, 35,428 comments are written directly in Sinhala script, whereas the remaining 28,043 comments were transliterated from Singlish, making the collection useful for studying both native Sinhala text and transliterated user-generated content.

In addition to individual datasets, several curated repositories aggregate Sinhala resources from multiple domains. \citet{senaratna2025sri} introduced and continuously maintains the \texttt{Sri Lanka Document Datasets}, a collection of publicly available datasets containing Sinhala text from a variety of sources. These include \gh{https://github.com/nuuuwan/lk_news/}{news articles}, \gh{https://github.com/nuuuwan/lk_edupub/}{educational publications} issued by the Educational Publications Department, and government press releases from organizations such as the \gh{https://github.com/nuuuwan/lk_treasury/}{Treasury} and the \gh{https://github.com/nuuuwan/lk_pmd/}{Presidential Media Division}. The legal and parliamentary resources contained within this collection are discussed separately in Section~\ref{sec:Legal}.

General-purpose knowledge resources also provide valuable supplementary data for Sinhala NLP. A wiki version of the Sinhala Encyclopedia maintained by the Government of Sri Lanka is publicly available\footURL{https://t.co/k8gFT9puM0}. Although the resource appears to have received relatively few recent updates, several of its articles provide more comprehensive coverage than their counterparts in Sinhala Wikipedia\footURL{https://si.wikipedia.org/}. 

A number of lightweight lexical resources have likewise been released to facilitate downstream NLP research. While~\citet{wijeratne2020sinhala} extracted a stop-word list algorithmically from a large Sinhala corpus,~\citet{lakmal2020word} published a manually curated \gh{https://github.com/nlpcuom/Sinhala-Stopword-list/blob/master/stop words.txt}{Sinhala stop-word list}. 

Reusable software packages often bundle both linguistic resources and baseline processing tools. For example, the open-source package \gh{https://github.com/ysenarath/sinling}{sinling} provides a rudimentary Sinhala tokeniser, stemmer, and part-of-speech tagger, together with supporting lexical resources. Although these components are discussed individually in later sections, the package itself represents one of the earliest attempts to provide an integrated toolkit for Sinhala NLP.
%
%

Overall, the range of publicly available Sinhala datasets has expanded considerably in recent years, evolving from isolated resources to curated repositories and multilingual benchmark collections. Nevertheless, compared with higher-resource languages, the ecosystem remains fragmented, with relatively few standardised benchmark datasets that are widely adopted across the research community. Continued efforts toward dataset curation, long-term maintenance, and standardised evaluation benchmarks will therefore play an important role in improving the reproducibility and comparability of future Sinhala NLP research.

\subsection{Dictionaries}
\label{Sinhala:Dictionaries}

Bilingual and monolingual dictionaries form one of the most fundamental lexical resources in NLP. They support a wide range of applications, including machine translation, lexical alignment, information retrieval, word sense disambiguation, and language learning. Despite the long history of Sinhala lexicography, however, relatively few dictionaries are openly available in formats suitable for computational research.

The earliest and most comprehensive English--Sinhala dictionary remains that of~\citet{malalasekera1967english}. But this resource, together with several other widely used printed dictionaries~\cite{jayathilake1937sinhala,maitipe1988gunasena,weerasinghe1999godage,wijayathunga2003maha,ranaweera2004wasana,gunaratne2006ratna}, is protected by copyright and is therefore unavailable for unrestricted research and software development. 
Although the dictionary developed by~\citet{Madura2018Madura} is publicly accessible through an online web interface\footURL{https://www.maduraonline.com/}, it neither provides an application programming interface (API) nor allows direct access to the underlying lexical database. 

Among publicly accessible lexical resources, one of the largest English--Sinhala dictionary datasets originates from the discontinued Firefox plug-in \textit{EnSiTip}~\cite{wasala2008ensitip}. The released lexicon bears a strong resemblance to the dictionary of~\citet{Madura2018Madura}, although the relationship between the two resources has not been formally documented. 
In addition,~\citet{hettige2007developing} reported the construction of an English--Sinhala lexicon to support their English-to-Sinhala machine translation system, while~\citet{samarawickramarequirements} investigated the design requirements for intelligent English--Sinhala bilingual dictionaries.

More recently,~\citet{wickramasinghe2023sinhala} introduced one of the largest openly available Sinhala--English dictionaries. Their resource contains 1,368,416 unfiltered and 195,255 filtered \EnSi{} word pairs, and both the dictionary and the accompanying code have been publicly released on \gh{https://github.com/kasunw22/sinhala-para-dict/tree/main}{GitHub}. 
Dictionary development has also extended beyond bilingual lexical resources.~\citet{iniyage2022wahara} introduced \textit{Wahara}\footURL{http://crawler.nlpc.uom.lk/}, a publicly maintained Sinhala--Sinhala dictionary developed at the University of Moratuwa. 

For multilingual applications, the Sri Lankan government has released a trilingual Sinhala--English--Tamil dictionary~\cite{Lang2018Tri}. Although this resource is accessible through a web interface, it currently lacks an efficient API or bulk data access mechanism, limiting its practical use for computational research. Similarly,~\citet{weerasinghe2013construction} developed a multilingual gazetteer of Sri Lankan place names that serves both as a lexical resource and as supporting data for named entity recognition tasks.

In addition to conventional dictionaries, pronunciation lexicons have become increasingly important for speech technologies. Google has publicly released a comprehensive IPA pronunciation dictionary covering a \gh{https://raw.githubusercontent.com/google/language-resources/master/si/data/lexicon.tsv}{large Sinhala vocabulary}. 

Overall, Sinhala lexical resources have evolved considerably from traditional printed dictionaries to openly available bilingual lexicons, multilingual dictionaries, gazetteers, and pronunciation dictionaries. Nevertheless, the majority of comprehensive lexical resources remain either proprietary or accessible only through web interfaces, creating significant barriers to reproducible research and large-scale NLP system development. The continued release of openly licensed, machine-readable dictionaries therefore remains an important priority for advancing Sinhala NLP.

\subsection{WordNets}
WordNets~\cite{miller1995wordnet} are among the most influential lexical resources in NLP, representing words as interconnected semantic concepts rather than isolated lexical entries. They encode a variety of semantic relations, including \textit{synonymy}, \textit{hyponymy}, \textit{hypernymy}, and \textit{meronymy}, making them valuable for numerous downstream applications. These range from relatively simple ontology and gazetteer construction~\cite{wimalasuriya2010ontology} to more advanced semantic processing tasks, including semantic similarity computation~\cite{wu1994verbs,jiang1997semantic} and semantic oppositeness detection~\cite{de2017discovering}.

Despite their importance, the development of a comprehensive Sinhala WordNet has remained an unsolved problem. One of the earliest large-scale efforts was undertaken by~\citet{wijesiri2014building}, who developed a crowd-sourced Sinhala WordNet that was temporarily hosted online~\cite{Sinhala2015}. Although the project demonstrated the feasibility of collaboratively constructing a lexical semantic resource for Sinhala, the website has since become defunct, resulting in the loss of the original data and supporting software. To the best of our knowledge, the only publicly accessible remnant of the project is the subset of the resource preserved by~\citet{arukgoda2014word}, who \gh{https://github.com/jseanm1/aruthSWSD}{cloned} the available data for use in their word sense disambiguation system. Even during its active development, however, the resource remained incomplete because the crowd-sourced methodology failed to attract a sufficiently large volunteer community.

An independent effort to construct a Sinhala WordNet was also initiated by~\citet{welgama2011towards}. However, this project appears to have ceased development at an even earlier stage and did not reach the level of lexical coverage reported by~\citet{wijesiri2014building}.

Overall, unlike many higher-resource languages, Sinhala still lacks a comprehensive, actively maintained, and publicly accessible WordNet. This absence continues to limit research on lexical semantics and constrains the development of knowledge-based NLP methods that rely on structured semantic relations. The construction of an open, sustainable, and community-maintained Sinhala WordNet therefore remains an important open challenge for the language.

\subsection{Tokenising}

Tokenisation is one of the most fundamental preprocessing steps in NLP, as it determines how text is segmented into units that are subsequently processed by language models. While tokenisation has traditionally been regarded as a relatively straightforward preprocessing task, recent work has shown that its design has a substantial impact on the performance and efficiency of multilingual large language models, particularly for morphologically rich and low-resource languages such as Sinhala.

A comprehensive evaluation of multilingual tokenisers by~\citet{petrov2023language} compared 17 off-the-shelf tokenisation schemes across a large number of languages, including Sinhala. The authors observed that higher-resource languages such as English generally exhibit tokenisation lengths close to the ideal value of $1.0$, whereas some languages require substantially longer token sequences, with the worst-performing language reaching a value of approximately $15.0$. For Sinhala, most tokenisers produced relatively poor tokenisation lengths ranging from $8.83$ to $12.86$, highlighting the difficulty of efficiently representing Sinhala text using tokenisers designed primarily for higher-resource languages. 
Notable exceptions were MBart50~\cite{tang2020multilingual}, mT5~\cite{xue2020mt5,tang2020multilingual}, and CANINE~\cite{clark2022canine}, which achieved tokenization lengths of $1.35$, $1.66$, and the ideal value of $1.00$, respectively.

Subsequent work has investigated the causes of these discrepancies. Rather than focusing solely on the tokenisation algorithm itself,~\citet{velayuthan2024egalitarian} demonstrated that the pre-tokenisation stage has an even greater influence on tokenisation efficiency for languages such as Sinhala in the context of large language models. Their findings suggest that language-specific preprocessing strategies are at least as important as the choice of the underlying subword tokenisation algorithm. Later~\citet{nisfer2026grapheme} introduced a Python toolkit named \gh{https://github.com/vmenan/grapheme-kit}{grapheme-kit} building on this work.

Task-specific tokenisers have also been proposed for Sinhala. \citet{ranasinghe2024sltk} introduced the \gh{https://github.com/Buddhilive/sltk}{SLTK} tokeniser, which attempts to improve tokenisation by explicitly handling Sinhala compound words. However, the proposed approach relies heavily on a manually curated \gh{https://github.com/Buddhilive/sltk/blob/main/sltkpy/shared/dict.json}{dictionary} that currently contains only seven entries. Although this may be sufficient for the accompanying \gh{https://github.com/Buddhilive/sltk/blob/main/data/sin.csv}{news title dataset}, the limited lexical coverage raises concerns regarding its applicability to general-domain Sinhala text.

Building upon these observations,~\citet{darshana2026separate} proposed Syllable-aware Grapheme Pair Encoding (SGPE), an extension of Grapheme Pair Encoding (GPE)~\cite{velayuthan2024egalitarian} that explicitly incorporates the syllabic structure of Sinhala during tokenisation. The proposed method was evaluated using Llama 4 Scout and DeepSeek V3.

Overall, recent studies indicate that tokenisation remains a significant challenge for Sinhala despite being largely considered a solved problem for higher-resource languages. The evolution from conventional subword tokenisers to language-aware approaches that exploit graphemic and syllabic structure reflects a broader trend towards designing tokenisation methods that better accommodate the linguistic characteristics of low-resource languages.

\subsection{Morphological Analysers}

\begin{table*}[!htb]
	\centering
	\caption{\textbf{Morphological Analyzers comparison}
	\newline\footnotesize Base: Rule-based (RB) / Machine Learning (ML) 
	\newline\footnotesize Able to Handle Part of Speech (Handles): Yes (Y) / No (N)
	\newline\footnotesize Outputs: Yes (Y) / No (N) / No Information (O)
	\newline\footnotesize Abbreviations: Nouns (Nu), Verbs (Ve), Adjectives (Aj), Adverbs (Av), Function Words (Fn), Root (R), Person (P), Number (Nb), Gender (G), Article (A), Case (C)
	}
	\label{tab:Morphological}
	\small
	\begin{tabular}{|l||c|c||c|c|c|c|c||c|c|c|c|c|c|c}
		\hline
		& \multirow{2}{*}{Base} & \multirow{2}{*}{Modus Operandi} & \multicolumn{5}{c||}{Handles} & \multicolumn{6}{c|}{Output}\\ 
		\hhline{~~~------------} 
		& & & Nu & Ve & Aj & Av & Fn & R & P & Nb & G & A & C  \\ 
		\hline 
		\hline 
		~\citet{hettige2006morphological} & RB & Finite State Automata & Y & Y & N & N & N & Y & Y & Y & Y & Y & Y\\
		\hline
        ~\citet{hettige2012multi} & RB & Agent-based & Y & Y & Y & Y & Y & N & N & N & N & N & N\\
		\hline
		~\citet{nandathilaka2018rule} & RB & N/A & Y & N & N & N & N & Y & N & N & N & N & N\\
		\hline
		~\citet{welgama2013evaluating} & ML & Morfessor algorithm & Y & Y & Y & Y & Y & Y & N & N & N & N & N\\
		\hline
		~\citet{fernando2013morphological} & RB & Finite State Transducer & N & Y & N & N & N & Y & Y & Y & Y & N & Y\\
		\hline
		~\citet{dilshani2017corpus} & RB & N/A & N & Y & N & N & N & O & O & O & O & O & O\\
		\hline
	\end{tabular}
	\normalsize
\end{table*}  

As shown in Fig.~\ref{fig:nlpLayers}, morphological analysis is one of the foundational components of an NLP pipeline. Its importance is particularly pronounced for Sinhala, which exhibits rich inflectional morphology and complex word formation processes~\cite{liyanage2012computational,kanduboda2013usage,de2015Sinhala}. Accurate morphological analysis is therefore essential for numerous downstream tasks, including part-of-speech tagging, parsing, machine translation, information retrieval, and language modelling.
%

The earliest studies of Sinhala morphology were conducted by~\citet{herath1989sinhalese,herath1990formalization}. These works focused primarily on formalising the morphological structure of Sinhala rather than developing computational tools. Consequently, their linguistic observations are discussed in Section~\ref{properties}. It is also worth noting that these studies predate the adoption of the Unicode standard for Sinhala and therefore represent Sinhala words using a Latin-script transliteration.

Building upon these linguistic foundations,~\citet{herath1992analysis} proposed a modular architecture for Sinhala morphological analysis. Subsequently, as part of their broader effort to develop an English-to-Sinhala machine translation system,~\citet{hettige2006morphological} reported the development of a Sinhala morphological analyser that complemented their work on a Sinhala parser~\cite{hettige2006parser} and computational grammar~\cite{hettige2011computational}. 
Although these studies collectively describe a relatively comprehensive framework for Sinhala language processing, neither the implementation nor the underlying lexical resources were made publicly available. The same research group later proposed a multi-agent architecture for morphological analysis~\cite{hettige2012multi}, further extending this line of work.

Several independent efforts subsequently explored alternative approaches to Sinhala morphological analysis.~\citet{welgama2013evaluating} investigated the applicability of machine learning techniques, representing one of the earliest attempts to move beyond purely rule-based approaches. Independently,~\citet{fernando2013morphological} developed a morphological parser specifically for Sinhala verbs. A similar verb-centric study was later presented by~\citet{dilshani2017corpus}. However, neither work was extended to cover the full Sinhala lexicon, and no publicly accessible software or datasets were released beyond the associated publications.

Research has also addressed individual components of the morphological analysis pipeline. For example,~\citet{nandathilaka2018rule} proposed a rule-based approach for Sinhala lemmatisation, 
while~\citet{virajdefining} reported the development of a set of gold-standard definitions for Sinhala morphology. However, because the latter resource has not been made publicly available, its broader adoption and independent evaluation have not been possible.
%
The  table~\ref{tab:Morphological}  provides a comparative summary of the discussion above.

Subsequent work focused on developing more practical and reusable morphological analysis tools. A combined study introduced both a rule-based Sinhala stemmer~\cite{kariyawasam2019rule} and an enhanced tokenizer~\cite{senanayake2019enhanced}, addressing two closely related components of the morphological processing pipeline. 
Building upon these efforts,~\citet{kumarasinghe2021sinmorphy} introduced \textit{SinMorphy}\footURL{https://nlp-tools.uom.lk/sin-morphy/}, a rule-based morphological analyser that leverages a comprehensive Sinhala vocabulary to improve lexical coverage and analysis accuracy.
More recently, researchers have explored neural approaches to Sinhala morphological analysis. \citet{ekanayaka2023applying} compared recurrent neural network (RNN), Long Short-Term Memory (LSTM), and Gated Recurrent Unit (GRU) architectures using both Sinhala and Singlish datasets. Their experiments showed that a bidirectional GRU (BiGRU) model achieved the best performance, reporting an accuracy of 87.96\%. 

Beyond the development of analysers themselves, comparative linguistic studies have also sought to better understand Sinhala morphology in relation to other languages. For example,~\cite{goonatillekestudy} compared the morphological characteristics of Sinhala with those of several other languages to facilitate the development of multilingual machine translation systems. Such studies provide valuable linguistic insights that can inform the design of future multilingual morphological analysers.

We can conclude that recent research demonstrates an increasing diversification of approaches to Sinhala morphological analysis, ranging from traditional rule-based systems to vocabulary-driven analysers and neural sequence models. Nevertheless, the scarcity of standardised evaluation benchmarks and the limited public availability of implementations continue to hinder direct comparison between competing methods. Consequently, despite more than three decades of research, there remains no widely adopted, openly maintained morphological analyser that has become the de facto standard for Sinhala NLP.

\subsection{Part of Speech Taggers}
Part-of-Speech (PoS) tagging is one of the fundamental stages of syntactic analysis, assigning each word in a sentence to its corresponding grammatical category. As with other morphologically rich languages, the design of an appropriate tag set is a prerequisite for developing accurate PoS taggers for Sinhala. 
Comprehensive Sinhala PoS tag sets have been proposed independently by~\citet{fernando2016comprehensive} and~\citet{dilshani2017comprehensive}, providing a common foundation for subsequent research.

The availability of annotated datasets has enabled the development and evaluation of Sinhala PoS tagging systems. A PoS-tagged corpus containing more than 250,000 Sinhala words is publicly available through Google Groups\footURL{https://t.co/jpQz74YEdQ}. 
In addition, another publicly available \gh{https://github.com/nlpcuom/Sinhala-POS-Data}{Sinhala PoS tagged dataset} has been released~\cite{fernando2016comprehensive,dilshani2017comprehensive,fernando2018evaluation}. 

Early research focused primarily on statistical sequence models. Among the earliest studies, \citet{herath2004stochastic}, \citet{jayaweera2012evaluation}, and \citet{jayasuriya2013learning} investigated stochastic approaches for Sinhala PoS tagging. 
Building upon this line of research,~\citet{jayaweera2011part} established the initial framework that was subsequently extended through a Hidden Markov Model (HMM)-based tagger~\cite{jayaweera2014hidden}. Recognising that unseen words are a persistent challenge for morphologically rich languages, the same research group also investigated methods for handling unknown words~\cite{jayaweera2014unknown,jayaweera2014handling}. 
They later compared several Sinhala PoS tagging approaches~\cite{jayaweera2016comparison} and released a RESTful PoS tagging web service~\cite{jayaweera2015restful}, which remains accessible\footURL{http://bit.ly/2F0jKid} through standard HTTP POST and GET requests.

In parallel, other researchers explored alternative learning paradigms. Expanding upon the standardised tag sets,~\citet{fernando2016comprehensive} proposed an SVM-based Sinhala PoS tagger, while~\citet{fernando2018evaluation} compared several machine learning classifiers for the same task.
Although these studies represent one of the more sustained lines of research on Sinhala PoS tagging, much of the work appears to have remained within a single research group, with limited public release of the corresponding implementations and trained models. 
A hybrid PoS tagger combining multiple approaches was subsequently proposed by~\citet{gunasekara2016hybrid}.
More recently,~\citet{wijerathna2020svm} also employed Support Vector Machines (SVMs) using the 30-tag inventory proposed by~\citet{fernando2016comprehensive}.

More recent studies have continued to investigate both data availability and modern learning techniques. The study by~\citet{kothalawala2019online} highlighted the broader problem of limited annotated resources for Sinhala NLP through a PoS tagging case study. 
To address data scarcity in informal text,~\citet{withanage2020stochastic} proposed a stochastic PoS tagger trained on a comparatively small corpus of 10,000 words collected from Facebook and Twitter.
\citet{sathsarani2022sinhala} compared traditional rule-based and stochastic methods against deep learning models, reflecting the broader transition within NLP toward neural approaches for sequence labelling.

Overall, research on Sinhala PoS tagging has evolved from the establishment of standardised tag sets and annotated corpora to stochastic, hybrid, machine learning, and deep learning approaches. Nevertheless, compared with higher-resource languages, the field still suffers from a shortage of large, standardised benchmark corpora and widely adopted open-source implementations. Greater emphasis on publicly available datasets, reproducible evaluation protocols, and community-maintained tagging systems would substantially strengthen future research in Sinhala PoS tagging.

\subsection{Parsers}

The PoS tagged data then needs to be handed over to a parser. This is an area which is not completely solved even in English due to various inherent ambiguities in natural languages. However, in the case of English, there are systems which provide adequate results~\cite{manning2014the} even if not perfect yet. 
The Sinhala state of affairs is that the first parser for the Sinhala language was proposed by~\citet{hettige2006parser} with a model for grammar~\cite{hettige2011computational}. The study by~\citet{liyanage2012computational} is concentrated on the same, given that they have worked on formalising a computational grammar for Sinhala. While they do report reasonable results, yet again, do not provide any means for the public to access the data or the tools that they have developed.~\citet{kanduboda2013usage} have worked on Sinhala differential object markers relevant for parsing. A meta-study was conducted by~\citet{stephen2024light} on the Sinhala data from \textit{Universal Dependencies (UD)} for identifying light verb constructions in Sinhala.

The first attempt at a Sinhala parser, as mentioned above, was by~\citet{hettige2006parser} where they created a prototype Sinhala morphological analyser and a parser as part of their larger project to build an end-to-end translator system. The function of the parser is based on three dictionaries: \textit{Base Dictionary}, \textit{Rule Dictionary}, and \textit{Concept Dictionary}. They are built as follows: 
\begin{itemize}
    \item \textbf{The Base Dictionary:} \textit{prakurthi} (base words), \textit{nipatha} (prepositions), \textit{upasarga} (prefixes), and \textit{vibakthi} (Irregular Verbs).
    \item \textbf{The Rule Dictionary:} inflection rules used to generate various forms of verbs and nouns from the base words.
    \item \textbf{The Concept Dictionary:} synonyms and antonyms for the words found in the base dictionary.
\end{itemize}

Parsers are, in essence, a computational representation of the grammar of a natural language. As such, in building Sinhala parsers, it is crucial to create a computational model for Sinhala grammar. The first such attempt was taken by~\citet{hettige2011computational} with special consideration given to Morphology and the Syntax of the Sinhala language as an extension to their earlier work~\cite{hettige2006parser}. Here, it is worthy to note that, unlike in their earlier attempt~\cite{hettige2006parser}, where they explicitly mentioned that they are building a parser, in this study~\cite{hettige2011computational}, they use the much conservative claim of building a computational grammar. Under Morphology, they again handled Sinhala inflection. Their system is based on a Finite State Transducer (FST) and Context-Free Grammar (CFG) where they they modeled 85 rules for nouns and 18 rules for verbs. The specific implementation is  more partial to a rule-based composer rather than parser. It is also worthy to note that this system could only handle simple sentences which only contained the following 8 constituents: \textit{Attributive Adjunct of Subject}, \textit{Subject}, \textit{Attributive Adjunct of Object}, \textit{Object}, \textit{Attributive Adjunct of Predicate}, \textit{Attributive Adjunct of the Complement of Predicate}, \textit{Complement of Predicate}, and \textit{Predicate}. With these, they propose the following grammar rules for Sinhala: 

\begin{verbatim}
S = Subject Akkyanaya
Subject = SimpleSubject | ComplexSubject
ComplexSubject = SimpleSubject ConSub
SimpleSubject = Noun | Adjective Noun
ConSub = Conjunction SimpleSubject
Akkyanaya = VerbP | Object VerbP
Object = SimpleObject | ComplexObject
ComplexObject = Conjunction SimpleObject
SimpleObject = Noun | Adjective Noun
VerbP = Verb | Adverb Verb
\end{verbatim}

The later work by~\citet{liyanage2012computational} also involves formalizing a computational grammar for Sinhala. They claim that Sinhala can have any order of words in practice. However, they do not note that this is happening because practices of the spoken language, which does not share the strong SOV conventions of the written language, are slowly seeping into written text. However, they do make note of how Sinhala grammar is modeled as a head-final language~\cite{karunatilaka1997sinhala}. They propose the Sinhala Noun Phrase ($NP$) to be defined as shown in equation~\ref{NP} where $NN$ is a noun which can be of types: common noun ($N$), pronoun ($PrN$) or proper noun ($PropN$). The adjectival phrase ($ADJP$) is then defined as as shown in equation~\ref{ADJP} where: $Det$ is a Determiner, $Adj$ is the adjective, and $Deg$ is an optional operator \textit{Degrees} which can be used to intensify the meaning of the adjective in cases where the adjective is qualitative. While they note that according to~\citet{gunasekara1891comprehensive}, there has to three classes of adjectives (\textit{qualitative}, \textit{quantitative}, and \textit{demonstrative}), they do not implement this distinction in their system. Similarly, they propose Sinhala Verb Phrase ($VP$) to be defined as shown in equation~\ref{VP} where $V$ is a \textit{single verb}. They here note that they are ignoring \textit{compound verbs} and \textit{auxiliary verbs} in their grammar. The adverbial phrases ($ADVP$) are then recursively defined as as shown in equation~\ref{ADVP}. 

\begin{equation}
    NP= [ADJP] [NN]
\label{NP}
\end{equation}

\begin{equation}
ADJP = \bigg[ [Det] \Big[[Deg][Adj]\Big]\bigg]
\label{ADJP}
\end{equation}

\begin{equation}
    VP= [ADVP] [V]
\label{VP}
\end{equation}

\begin{equation}
ADVP = \Bigg[[NP] \bigg[[ADVP] \Big[[Deg][ADV]\Big]\bigg]\Bigg]
\label{ADVP}
\end{equation}

Similar to~\citet{hettige2011computational}, the work by~\citet{liyanage2012computational} also builds a CFG for Sinhala covering 10 out of the 25 types of simple sentence structures in Sinhala reported by~\citet{Abhayasinghe1998sinhala}. This parser is unable to parse sentences where inanimate subjects do not consider the number. Further, sentences which contain, \textit{compound verbs}, \textit{auxiliary verbs}, \textit{present participles}, or\textit{ past participles} cannot be handled by this parser. If the verbs have \textit{imperative mood} or \textit{negation} those too cannot be handled by this. Non-verbal sentences which end with \textit{adjectives}, \textit{oblique nominals}, \textit{locative predicates}, \textit{adverbials}, or any other language entity which is not a verb cannot be handled by this parser. 

The study by~\citet{kanduboda2013usage} covers not the whole of Sinhala parsing but analyzes a very specific property of Sinhala observed by~\citet{aissen2003differential} which states that it is possible to notice Differential Object Marking (DOM) in Sinhala active sentences.~\citet{kanduboda2013usage} define this as the choice of \textit{/wa/} and \textit{/ta/} object markers. They further observe three unique aspects of DOM in Sinhala: (a) it is only observed in active sentences which contain transitive verbs, (b) it can occur with accusative marked nouns but not with any other cases, (c) it exists only if the sentence has placed an animate noun in the accusative position. They do a statistical analysis and provide a number of short gazetteer lists as appendixes. However, they observe that further work has to be done for this particular language rule in Sinhala given that they found some examples which proved to be exceptions to the general model which they proposed. 

\subsection{Spelling and Grammar}
Reliable benchmark datasets are essential for evaluating and comparing spelling correction systems. To this end,~\citet{sonnadarasinhala} introduced a \gh{https://github.com/chason94/SinNeuSpellCorrector}{benchmark dataset} for Sinhala spelling correction together with a neural baseline model. This resource was subsequently \hf{https://huggingface.co/datasets/lm-spell/sinhala-spell-correction-dataset}{extended} by~\citet{gunathilakea2025lmspell}. Although the newer dataset is hosted on Hugging Face, access currently requires submitting a request that must be approved by the dataset authors. 

\subsubsection{Sinhala Spelling}

Early research on Sinhala spelling correction focused primarily on data-driven and rule-based approaches. The open-source systems proposed by~\citet{wasala2010data,wasala2011open} introduced one of the earliest publicly described data-driven spelling correction frameworks for Sinhala. 
Subsequently,~\citet{jayalatharachchi2012data} proposed combining two complementary correction algorithms to improve spelling correction performance, and this line of research~\cite{wasala2010data,jayalatharachchi2012data} was further extended by~\citet{subhagya2018data}.

Rule-based spell checking has also received considerable attention. \citet{liyanapathirana2021sinspell} introduced \textit{SinSpell}, a Sinhala spell checker built on top of \textit{Hunspell}\footURL{http://hunspell.github.io/}. In addition to describing the system, the authors provide an online demonstration\footURL{http://nlp-tools.uom.lk/sinspell/}, making it one of the few publicly accessible Sinhala spell-checking services. 
Similarly,~\citet{samarawickrama2019lasi} proposed the multi-agent spelling correction system \textit{LaSi Spell}, while~\citet{sithamparanathan2019sinhala} extended the \textit{Generic Environment for Context-Aware Spell Correction} to support both Sinhala and Tamil.

Several studies have addressed specialized spelling correction problems. For example,~\citet{udagedara2022language} focused specifically on correcting Sri Lankan names and postal addresses, while~\citet{sudesh2022erroff} introduced \textit{Erroff}, a system designed to detect and correct real-word spelling errors in Sinhala text. More recently,~\citet{praveen2024machine} reported a Random Forest-based spell checker achieving an accuracy of 82\%.

The recent success of large language models has also influenced Sinhala spelling correction. \citet{gunathilakea2025lmspell} introduced \gh{https://github.com/Nadil-K/lm-spell}{LMSpell}, an LLM-based toolkit for Sinhala spelling correction. In addition to releasing the toolkit, the authors provide public access to the corresponding \hf{https://huggingface.co/lm-spell}{trained models} and an online demonstration\footURL{https://nisansa.lk/?s=eDbXYk}. This work represents the latest transition from traditional rule-based and statistical approaches toward foundation-model-based spelling correction.

\subsubsection{Sinhala Grammar}
Compared with spelling correction, grammatical error detection and correction has received considerably less attention in Sinhala NLP. One of the earliest efforts was presented by~\citet{pabasara2020computational}, who developed a model for detecting grammatical errors. This work was subsequently extended into a grammatical error detection and correction system~\cite{pabasara2020grammatical}. Around the same period,~\citet{fernando2020sinhala} proposed a Hidden Markov Model (HMM)-based Sinhala grammar checker.

Alternative approaches have also been explored. Rather than treating grammar correction as a classification task,~\citet{widyaratna2019sinhala} formulated it as a sequence-to-sequence translation problem, using an attention-based encoder--decoder model to transform grammatically incorrect sentences into their corrected forms.
Meanwhile,~\citet{jayasuriya2023grammar} combined rule-based methods with Google Translate to perform grammatical correction, illustrating a hybrid approach that leverages existing machine translation systems.

Research has also addressed more specialized grammatical transformations. For example,~\citet{ilukkumbura2023sinhala} proposed a rule-based system for converting Sinhala sentences from active voice to passive voice while simultaneously correcting grammatical errors. Although designed for a specific transformation, this work demonstrates how grammar correction techniques can be integrated with broader text rewriting tasks.

While not directly addressing grammar correction,~\citet{gunasekara2020using} investigated annotation projection for semantic role labeling in Sinhala. Although this work falls outside the traditional scope of grammar checking, it contributes to the broader understanding of Sinhala syntax and grammatical structure.

\subsubsection{Sinhala Spelling and Grammar Both}
Several systems attempt to jointly address spelling and grammatical errors within a unified framework. \citet{goonawardena2022automated} proposed a rule-based system capable of both spell checking and grammatical error detection and correction. Similarly, \citet{navoda2023automated} reported the development of an automated tool that simultaneously performs Sinhala spelling and grammar checking. 

\subsubsection{Other Relevant Applications}

Research on error correction has also influenced several related text restoration tasks. 
\citet{disara2026restoring} fine-tuned a RoBERTa model using a physics-domain Sinhala corpus derived from G.C.E. Advanced Level educational materials and evaluated its ability to restore missing or corrupted words in Sinhala sentences.
%
More recently,~\citet{fernando2026reconstructing} employed ByT5~\cite{xue2022byt5} to reconstruct modern Sinhala sentences from direct Brahmi--Sinhala translations. Their approach used one fine-tuned ByT5 model to recover the correct word order and a second model to perform morphological correction. 

\subsection{Named Entity Recognition Tools}
The development of Sinhala Named Entity Recognition (NER) has been supported by a small but growing collection of annotated datasets. The \gh{https://github.com/suralk/multiNER}{multiNER} dataset introduced by~\citet{ranathunga2024multi} provides parallel NER annotations for Sinhala, Tamil, and English and contains approximately 99,000 tagged Sinhala words. It follows a conventional annotation scheme comprising the entity classes \texttt{PER}, \texttt{ORG}, \texttt{LOC}, and \texttt{MISC}, together with the non-entity label \textit{O}. The \gh{https://github.com/suralk/AnanyaSinhalaNERDataset}{Ananya} dataset, created as part of the work by~\citet{manamini2016ananya}, is also publicly available.

\begin{table*}[!htb]
	\centering
	\caption{\textbf{NER system comparison} \newline\small
	* Denotes a baseline.\newline
	$F_1$ to $F_{11}$ denotes \textit{Context Words}, \textit{Word Prefixes and Suffixes}, \textit{Length of the Word}, \textit{Frequency of the Word}, \textit{First Word/ Last Word of a Sentence}, \textit{(POS) Tags}, \textit{Gazetteer Lists}, \textit{Clue Words}, \textit{Outcome Prior}, \textit{Previous Map}, and \textit{Cutoff Value}    
	}
	\label{tab:NER}
	\small
	\begin{tabular}{|l||c|c||c|c|c|c|c|c|c|c|c|c|c|c|c|c}
		\hline
		& \multirow{2}{*}{CRF} & \multirow{2}{*}{ME} & \multicolumn{12}{c|}{Features}\\ 
		\hhline{~~~------------} 
		& & & $F_1$ & $F_2$ & $F_3$ & $F_4$ & $F_5$ & $F_6$& $F_7$ & $F_8$ & $F_9$ & $F_{10}$ & $F_{11}$  \\ 
		\hline 
		\hline 
		~\citet{dahanayaka2014named} & Yes & Yes* & Yes & Yes & No & No & No & No & No & No & No & No & No\\
		\hline
        ~\citet{senevirathne2015conditional} & Yes & No & Yes & Yes & Yes & No & Yes & No & No & Yes & No & Yes & No\\
		\hline
		~\citet{manamini2016ananya} & Yes & Yes & Yes & Yes & Yes & Yes & Yes & Yes & Yes & Yes & Yes & Yes & Yes\\
		\hline
	\end{tabular}
	\normalsize
\end{table*}

As shown in Fig.~\ref{fig:nlpLayers}, NER forms an important stage in the extraction of structured information from processed text. The earliest identified attempt to develop an NER system for Sinhala was presented by~\citet{dahanayaka2014named}. As no established Sinhala NER framework was available at the time, the study adapted methods previously developed for other languages, with particular attention to research on Indic languages. An important observation made in this work was that NER can be more difficult for Sinhala and other Indic languages than for English because Sinhala does not use capitalization as a cue for identifying proper nouns and named entities.
Following the dominant sequence-labelling approaches of the period, they employed Conditional Random Fields (CRFs) and compared their performance against a Maximum Entropy (ME) baseline. The feature representation was relatively limited, consisting primarily of the \textit{candidate word}, its surrounding \textit{context words}, and basic information derived from Sinhala \textit{suffixes}. 

The follow-up study by~\citet{senevirathne2015conditional} retained the CRF architecture and the original feature set, although it did not report a corresponding comparison with an ME model. Its principal contribution was the introduction of a richer collection of language-sensitive features. These included the \textit{length of the word}, represented using a threshold-based feature, and a \textit{first word} feature motivated by regularities in Sinhala sentence structure. The study also refined the use of contextual information by introducing \textit{clue words}, a selected subset of context words expected to provide stronger evidence for entity identification. In addition, the \textit{previous map} feature incorporated the predicted named-entity label of the preceding word. Several of these features were supported by a rule-based post-processing component that relied on context-sensitive word lists.

The third major Sinhala NER system, \textit{Ananya}, was introduced by~\citet{manamini2016ananya}. This work retained both the CRF model and the ME baseline used by~\citet{dahanayaka2014named}, while substantially extending the feature inventory developed by~\citet{senevirathne2015conditional}. A \textit{frequency of the word} feature was introduced based on the assumption that highly frequent words are less likely to be named entities. Word frequency was therefore converted into a Boolean feature through the application of a predefined threshold.
The positional feature introduced in the earlier work~\cite{senevirathne2015conditional} was also expanded from \textit{first word} to \textit{first word/last word of a sentence}, with the latter motivated by the predominantly subject--object--verb (\textit{SOV}) structure of Sinhala. In line with NER research for other languages, \textit{Ananya} additionally incorporated a \textit{Part-of-Speech (PoS) tag} feature and features derived from \textit{gazetteer lists}. The \textit{clue words} feature, which had previously been treated as part of the broader contextual representation by~\citet{dahanayaka2014named}, was formalized as an independent feature.
Because \textit{Ananya} retained the ME component unlike~\citet{dahanayaka2014named}, \citet{manamini2016ananya} were also able to introduce an \textit{outcome prior} feature complementary to the \textit{previous map} feature. This feature exploited the underlying probability distribution produced by the ME model. Finally, a \textit{cutoff value} feature was incorporated to reduce overfitting by excluding features whose frequencies fell below a specified threshold.

Table~\ref{tab:NER} provides a comparative summary of the three early Sinhala NER systems discussed above. All three systems were restricted to three principal entity categories: \textit{person names}, \textit{location names}, and \textit{organisation names}. The dataset associated with the \gh{https://github.com/suralk/AnanyaSinhalaNERDataset}{Ananya} system by~\citet{manamini2016ananya} is publicly available. In contrast, the data and implementations associated with~\citet{dahanayaka2014named} and~\citet{senevirathne2015conditional} are not publicly accessible.

Subsequent research expanded Sinhala NER beyond the relatively coarse entity inventories and feature-engineered models used in the earlier systems. Building upon their previous work on Sinhala NER~\cite{manamini2016ananya} and PoS tagging~\cite{fernando2018evaluation},~\citet{azeez2020fine} proposed a fine-grained NER model for Sinhala. This work represents an important shift from recognizing only broad categories such as persons, locations, and organizations toward distinguishing more specific entity types.

Alternative learning paradigms were also explored. \citet{anuruddha2021reinforcement} proposed a reinforcement-learning-based method for Sinhala NER, extending the task beyond conventional supervised sequence-labelling approaches. 
In parallel,~\citet{mallikarachchi2021support} employed Support Vector Machines to identify Sinhala named entities belonging to the \textit{person}, \textit{location}, and \textit{organisation} categories. Although the entity inventory remained similar to that of the earlier systems, this work provided an additional classical machine learning formulation of the task.

Domain-specific NER has received comparatively limited attention. One notable exception is the sports-domain Sinhala NER system introduced by~\citet{wijesinghe2022sinhala}, who evaluated classical machine learning models for identifying entities in sports-related text. The restricted domain can reduce some of the ambiguity associated with general-domain NER, but it also limits the direct applicability of the resulting models to other genres.
%

Sinhala NER has additionally been incorporated as an intermediate component in broader NLP applications. For example, although the principal task investigated by~\cite{peiris2024sinhala} was the clustering of Sinhala news articles, named-entity information was used to support the clustering process. 

More recent work has examined the use of pretrained and large language models for Sinhala NER. \citet{gurgurov2024adapting} compared several adaptation strategies for applying language models to the task. Their strongest results were obtained using a masked language modelling objective for Sinhala, with combinations of ConceptNet (\textit{CN}) and Wikipedia (\textit{wiki}) language adapters together with task adapters (\textit{TA}). 

In contrast, \citet{rathnayake2025intelligent} applied LLaMA directly to Sinhala NER. 
Although the effectiveness of such approaches depends on factors such as prompt design, model coverage of Sinhala, and the availability of suitable evaluation data.

The development of Sinhala NER should also be viewed within the wider context of NER research for South Asian languages. Sinhala is discussed alongside other Indic languages in the survey by~\citet{deshmukh2024named}, which situates the challenges faced by Sinhala, including limited annotated data, weak capitalization cues, and restricted tool availability, within a broader multilingual perspective.

\subsection{Word Sense Disambiguation}

Word Sense Disambiguation (WSD) is the task of determining the intended meaning of an ambiguous word from its surrounding context~\cite{yarowsky1992word,ide1998introduction,yarowsky1995unsupervised,banerjee2002adapted,navigli2009word}. As one of the core semantic analysis tasks in NLP, WSD underpins numerous downstream applications, including information retrieval, question answering, machine translation, and semantic search. Despite its importance, relatively few studies have addressed Sinhala WSD compared with higher-resource languages.

One of the earliest dedicated Sinhala WSD systems was \textit{Aruth}, proposed by~\citet{arukgoda2014word}. The system adopts the Lesk algorithm~\cite{lesk1986Automatic}, a knowledge-based approach that identifies the most appropriate word sense by measuring the overlap between dictionary definitions and the surrounding context. Unlike many early Sinhala NLP systems, \textit{Aruth} was accompanied by several publicly accessible resources, including an online demonstration\footURL{http://aruth.herokuapp.com/}, a publicly available \gd{https://drive.google.com/file/d/0B1ND7M3X2GhueXRKaDREeEMzdEU/view?resourcekey=0-wfrEg-hG7RtoS-45G0yl9A}{API}, and both the source code and lexical resources hosted on \gh{https://github.com/jseanm1/aruthSWSD}{GitHub}. These releases have made \textit{Aruth} one of the most reproducible Sinhala WSD systems reported in the literature.

Alternative approaches have also been investigated. \citet{marasinghe2002word} proposed a probabilistic model for Sinhala WSD, representing an early attempt to move beyond purely knowledge-based methods toward statistical inference. Although the published literature provides relatively limited details regarding the long-term development of this approach, it illustrates the exploration of probabilistic techniques for lexical ambiguity resolution.
Research related to lexical ambiguity has also appeared within broader semantic processing tasks. For example,~\citet{palihakkara2015dialogue} developed a dialogue act recognition system that employed comparatively simple classification algorithms. While dialogue act recognition extends beyond conventional WSD, both tasks rely on contextual semantic interpretation and therefore share many underlying challenges.

More recently,~\citet{subasingha2020sinsense} introduced the WSD system \textit{SinSense}. Rather than performing disambiguation directly in Sinhala, the proposed method adopts a cross-lingual strategy in which English word sense disambiguation is first performed and the resulting sense assignments are then transferred to Sinhala. This approach enables Sinhala WSD to benefit from the substantially richer lexical resources available for English. However, neither the implementation nor the associated datasets have been made publicly available.

Overall, Sinhala WSD research has evolved from knowledge-based algorithms to probabilistic and cross-lingual approaches. Nevertheless, progress has been constrained by the absence of a comprehensive Sinhala WordNet, the scarcity of large annotated sense corpora, and the limited availability of openly released implementations. Consequently, reproducible evaluation remains challenging, and WSD continues to be one of the least explored areas of Sinhala semantic processing despite its importance to many downstream NLP applications.

\subsection{Semantic Similarity}

Measuring semantic similarity between words, sentences, or documents is a fundamental task in NLP, underpinning applications such as information retrieval, duplicate detection, question answering, automatic grading, and semantic search. Compared with other core NLP tasks, research on semantic similarity for Sinhala remains relatively limited, although the field has gradually evolved from rule-based sentence similarity measures to neural representation learning and multilingual evaluation benchmarks.

One of the earliest studies on Sinhala semantic similarity was conducted by~\citet{kadupitiya2016sinhala}, who proposed a semantic similarity measure for short Sinhala sentences. This work was subsequently extended by~\citet{kadupitiya2017sinhala} to support the practical application of automatic \textit{short-answer grading}. Although these studies established an important foundation for semantic similarity research in Sinhala, neither the corresponding datasets nor the software implementations have been made publicly available. 

Later research shifted towards neural representation learning. \citet{nilaxan2021monolingual} developed a sentence similarity model based on Siamese neural networks and demonstrated its applicability to both Sinhala and Tamil. Building upon this direction, \citet{deepal2024siamese} proposed a hybrid Siamese architecture combining Long Short-Term Memory (LSTM) and Convolutional Neural Networks (CNNs) for Sinhala sentence similarity estimation. Their model achieved an F1-score of 0.9041 using cosine similarity as the evaluation metric. Although the authors describe how several publicly available datasets were refined to construct their training corpus, the resulting processed dataset has not been released. 

Semantic similarity has also been investigated in multilingual settings. \citet{isuranga2020improved} proposed a cross-lingual document similarity framework for Sinhala and English, extending semantic similarity beyond monolingual sentence comparison to multilingual document retrieval and comparison. Such approaches are particularly valuable for low-resource languages, where multilingual transfer can compensate for the limited availability of large annotated corpora.

More recently, emphasis has shifted toward standardized multilingual evaluation. \citet{ranasinghe2025musts} introduced the \gh{https://github.com/TharinduDR/MUSTS}{MUSTS} benchmark, a multilingual semantic textual similarity dataset that includes Sinhala. The Sinhala portion of the benchmark is derived from the short-sentence similarity dataset originally developed by~\citet{kadupitiya2017sinhala}, providing one of the first standardized resources for evaluating multilingual semantic similarity systems involving Sinhala.

Overall, research on semantic similarity for Sinhala has progressed from handcrafted sentence-level similarity measures to neural representation learning and multilingual benchmarking. Nevertheless, the field continues to be constrained by the limited public availability of annotated datasets and reproducible implementations. 
Larger and more diverse semantic similarity datasets remain necessary to support the development and comparison of modern embedding-based and large language model approaches.

\subsection{Embeddings and Small Language Models}

Distributed representations have become the dominant paradigm for representing linguistic information in modern NLP. Unlike manually engineered lexical features, embeddings encode semantic and syntactic properties in continuous vector spaces, enabling models to generalize beyond individual words. The evolution of Sinhala embeddings mirrors the broader development of NLP, progressing from static word embeddings to contextual transformer models and, more recently, task-specific Small Language Models (SLMs).

The earliest publicly available pretrained embeddings for Sinhala were released by Facebook as part of the FastText project~\cite{joulin2016fasttext,bojanowski2017enriching,joulin2017bag}. These models, trained on the Sinhala Wikipedia corpus, are available both as text models\footURL{http://bit.ly/2JXAyL8} and binary models\footURL{http://bit.ly/2JY5J9c}. Building upon these embeddings,~\citet{lakmal2020word} trained an extended FastText model using a substantially more diverse corpus consisting of Wikipedia articles, news publications, and official government documents. The resulting binary model\footURL{http://bit.ly/2WowH0h} has also been released publicly.
\citet{ekanayake2026helabert} have released \hf{https://huggingface.co/ThisenEkanayake/HelaBERT}{HelaBERT-Small} and \hf{https://huggingface.co/ThisenEkanayake/HelaBERT_Large}{HelaBERT-Large} a 23.3 million parameter BERT model trained on approximately 900 million tokens of Sinhala text sourced from MADLAD-400~\cite{kudugunta2024madlad}, CulturaX~\cite{nguyen2023culturax} , and a custom private corpus.

The availability of pretrained embeddings naturally led to comparative studies of their effectiveness. \citet{lakmal2020word} conducted one of the earliest evaluations of Sinhala word embeddings, while~\citet{silva2022generating} revisited the problem several years later to assess the progress achieved by subsequent developments in Sinhala NLP. Cross-lingual representation learning has likewise received attention. For example,~\citet{liyanage2021bilingual} analysed Sinhala FastText and Word2Vec embeddings within bilingual embedding spaces, investigating their suitability for multilingual representation learning.

Research has also extended beyond word-level representations to sentence embeddings. \citet{weeraprameshwara2022sinhala} proposed a two-tier architecture for generating Sinhala sentence embeddings and reported that a \textit{Seq2Seq} Gated Recurrent Unit (GRU) model with attention operating on FastText embeddings achieved the strongest performance for Sinhala sentence embedding. 
More recently,~\citet{ranaweera2025custom} introduced \texttt{AshenBERTo}, a RoBERTa model specifically pretrained for Sinhala sentence similarity estimation. 

The emergence of transformer-based language models significantly changed the landscape of Sinhala representation learning. \citet{dhananjaya2022bertifying} released two pretrained Sinhala BERT models, \hf{https://huggingface.co/NLPC-UOM/SinBERT-large}{SinBERT-large} and \hf{https://huggingface.co/NLPC-UOM/SinBERT-small}{SinBERT-small}, providing dedicated transformer encoders for Sinhala. A further independently developed model, \hf{https://huggingface.co/keshan/SinhalaBERTo}{SinhalaBERTo}, was trained by adapting the RoBERTa architecture~\cite{liu2019roberta} using the \textit{OSCAR} corpus~\cite{ortiz-suarez-etal-2020-monolingual,OrtizSuarezSagotRomary2019}. Although this model has no accompanying research publication, it demonstrates the growing community interest in developing publicly available pretrained Sinhala language models.

Several studies have subsequently explored multilingual embedding resources and parameter-efficient adaptation techniques. \citet{demotte2020sentiment} publicly released both the \gd{https://drive.google.com/drive/folders/1CgqPIloW5DKLVj3WViMitIhtrswOv_qR}{word embeddings} and the accompanying \gh{https://github.com/theisuru/sentiment-tagger}{implementation} used in their embedding-based Sinhala sentiment analysis system. 
These embeddings were later incorporated into the multilingual \textit{LowREm} repository~\cite{gurgurov2024lowrem}, which provides embedding resources for 87 low-resource languages and explicitly credits~\citet{demotte2020sentiment} as the Sinhala source. 
Building upon multilingual pretrained models,~\citet{gurgurov2025small} argued that multilingual language models (mLMs) such as mBERT and XLM-R are better suited than contemporary large language models for many low-resource languages, including Sinhala, and proposed parameter-efficient adapter-based methods for their adaptation. Their implementation has been made publicly available on \gh{https://github.com/d-gurgurov/Knowledge-Driven-Adaptation-LLMs}{GitHub}.

As pretrained representations became increasingly widespread, standardized evaluation benchmarks also began to emerge. \citet{enevoldsen2025mmteb} introduced \gh{https://github.com/embeddings-benchmark/mteb}{MMTEB}, together with a public \hf{https://huggingface.co/spaces/mteb/leaderboard}{leaderboard} for multilingual embedding evaluation. The benchmark includes Sinhala through the news classification dataset developed by~\citet{de2015Sinhala}. 
More recently,~\citet{ranasinghe2025sinhala} introduced \gh{https://github.com/Sinhala-NLP/Sinhala-GLUE}{SINHALA-GLUE}, a comprehensive language understanding benchmark accompanied by three new Sinhala encoder models: \texttt{Sinhala-BERT} (Raja), \texttt{Sinhala-RoBERTa} (Koliya), and \texttt{Sinhala-Electra} (Mahasen). These models are not publicly available.

The recent emergence of Small Language Models (SLMs) has further expanded the scope of Sinhala representation learning. \citet{devine2026kakugo} introduced \texttt{Kakugo}, a general-purpose SLM training pipeline evaluated across multiple languages, including Sinhala. While the absolute performance obtained for Sinhala is moderate, the authors report consistently favourable pairwise win rates for the \texttt{Kakugo} models, suggesting that carefully trained SLMs can provide competitive performance even for relatively low-resource languages.

Overall, Sinhala representation learning has evolved rapidly from static FastText embeddings to contextual transformer encoders, multilingual adaptation methods, benchmark suites, and dedicated Small Language Models. Compared with many other areas of Sinhala NLP, this is one of the fastest-moving research directions. Nevertheless, challenges remain in terms of publicly releasing pretrained models, establishing standardized evaluation protocols, and developing sufficiently large pretraining corpora. Continued progress in these areas is likely to benefit virtually every downstream Sinhala NLP task discussed throughout this survey.

\subsection{Embedding Alignment and Bilingual Lexicon Induction}

Embedding alignment and Bilingual Lexicon Induction (BLI) seek to map the vector spaces of different languages into a common semantic space, thereby enabling cross-lingual transfer, multilingual information retrieval, and machine translation without requiring large parallel corpora. For low-resource languages such as Sinhala, these techniques are particularly valuable because they leverage the abundant lexical resources available for higher-resource languages while minimizing the need for expensive manual annotation.

One of the first comprehensive studies of Sinhala--English embedding alignment was conducted by~\citet{wickramasinghe2023sinhala2}. Using the Sinhala--English dictionary introduced by~\citet{wickramasinghe2023sinhala}, the authors established one of the first benchmark datasets for Sinhala--English embedding alignment and compared its performance with several widely studied language pairs, including English--Spanish, English--French, English--German, English--Russian, and English--Chinese. To facilitate reproducible research, both the benchmark data and the corresponding implementation have been released on \gh{https://github.com/kasunw22/sinhala-word-embedding-alignment}{GitHub}.

Subsequent work shifted attention toward improving alignment algorithms. \citet{rathnayake2024unsupervised} investigated unsupervised BLI for the Sinhala--English language pair using Unsupervised VecMap (UVecMap)~\cite{artetxe2017learning} as the baseline alignment method. Their experiments demonstrated that combining UVecMap with linear transformation achieved the strongest performance for English--Sinhala Word2Vec embeddings, reporting a score of 33.18. For English--Sinhala FastText embeddings, the highest score of 29.49 was obtained by combining UVecMap, linear transformation, and CSCBLI~\cite{zhang2021combining}. 

More recently,~\citet{wickramasinghe2025good} conducted a broader comparative study of multiple BLI techniques across several language pairs. Their experiments showed that Language-agnostic BERT Sentence Embeddings (LaBSE)~\cite{feng2022language} combined with Cross-domain Similarity Local Scaling (CSLS)~\cite{joulin2018loss} consistently achieved the strongest overall performance. In addition to comparing existing methods, the authors proposed vocabulary pruning and stemming as effective preprocessing techniques when one or both languages exhibit rich inflectional morphology, as is the case for Sinhala. 

Embedding alignment has also been explored in broader multilingual semantic representation settings. \citet{yu2025research} employed the \texttt{FLORES-200}~\cite{costa2022no} dataset to evaluate a cross-model semantic alignment framework and demonstrated improvements over both CLIP~\cite{radford2021learning} and ALIGN~\cite{jia2021scaling}. Notably, Sinhala was identified as one of the languages exhibiting the largest improvements in Recall@1 and Mean Rank.

Overall, research on Sinhala embedding alignment has progressed from establishing benchmark datasets to developing increasingly sophisticated alignment algorithms and multilingual evaluation frameworks. A recurring theme across these studies is that the linguistic characteristics of Sinhala, particularly its rich morphology, necessitate language-aware preprocessing strategies in addition to generic multilingual alignment methods. As multilingual language models continue to improve, embedding alignment and bilingual lexicon induction are likely to remain key enabling technologies for cross-lingual transfer involving Sinhala.


\subsection{Text Classification}
\label{Sec:TextClassification}

Text classification is one of the most widely studied applications at the semantic layer of the NLP stack. Sinhala research in this area spans news categorization, source and writing-style identification, authorship attribution, fake-news detection, code-mixed text analysis, textual entailment, and topic discovery. The field has progressed from rule-based and conventional machine learning methods to neural architectures, multilingual transformers, parameter-efficient adaptation, and active learning. However, the availability and quality of datasets remain uneven across these applications.

Several publicly accessible datasets support Sinhala text classification research. Although the primary objective of~\citet{ranasinghe2024sltk} was to develop a Sinhala tokenizer rather than a classification benchmark, the associated repository contains a \gh{https://github.com/Buddhilive/sltk/blob/main/data/sin.csv}{dataset} of approximately 26,000 labelled Sinhala news titles. 
An earlier and considerably smaller implementation of Sinhala news classification was presented by~\citet{de2015Sinhala}. As discussed in Section~\ref{DataSets}, the corresponding \osf{https://osf.io/tdb84/}{news corpus} is publicly available.
A further set of classification resources was released by~\citet{dhananjaya2022bertifying}. The authors derived three datasets from previously published corpora: a \hf{https://huggingface.co/datasets/NLPC-UOM/Sinhala-News-Source-classification}{Sinhala news source classification} dataset based on the corpus introduced by~\citet{sachintha2021exploiting}; a second \hf{https://huggingface.co/datasets/NLPC-UOM/Sinhala-News-Source-classification}{Sinhala news source classification} dataset derived from the corpus of~\citet{de2015Sinhala}; and a \hf{https://huggingface.co/datasets/NLPC-UOM/Writing-style-classification}{Sinhala writing style classification} dataset constructed from the corpus introduced by~\citet{upeksha2015implementing}.
For authorship analysis,~\citet{faumi2025stylomech} released \gh{https://github.com/cipherdragon/SimpleAA}{SimpleAA}, a writing-style identification dataset that also includes Romanised (transliterated) Sinhala text. 
Although the research paper associated with~\citet{ekanayaka2018sinhala} appears to be unavailable now, the corresponding news corpus has been released through \gh{https://github.com/rksk/sinhala-news-analysis/tree/master/sinhala-news-corpus}{GitHub}. It contains 1,000 labelled and 50,000 unlabelled Sinhala news reports collected from multiple sources, while the wider repository also provides some of the associated \gh{https://github.com/rksk/sinhala-news-analysis/}{code}.

Early Sinhala text classification research predominantly relied on conventional statistical and rule-based methods. \citet{gallege2010analysis} investigated basic classification using Naïve Bayes, Support Vector Machines (SVMs), and features motivated by Zipf's law. 
\citet{lakmali2017effectiveness} subsequently compared six popular rule-based algorithms and reported the construction of a corpus named \textit{SinNG5}, although no mechanism was provided for obtaining the corpus. 
\citet{kumari2019use} later used \textit{SinNG5} to investigate the application of LIME~\cite{ribeiro2016should} to interpretable Sinhala document classification, but likewise did not release the data. 
In a related direction,~\citet{nanayakkara2018clustering} developed a classification system based on corpus-derived similarity measures.
%

Feature selection and document representation were also central themes in this early research. \citet{gunasekara2018context,gunasekara2018effective} proposed a context-aware stop-word extraction method for Sinhala text classification based on TF--IDF, while~\citet{koralage2019sinclassify} presented a simpler TF--IDF-based classification system. 
\citet{haddela2020use} adopted a different approach by applying a Genetic Algorithm to Lucene\footURL{https://lucene.apache.org/} search queries in order to obtain interpretable classification models for Sinhala documents. 
Later,~\citet{weerasiri2022word} compared Word2Vec, FastText, and Doc2vec~\cite{le2014distributed} representations against TF--IDF for Sinhala news classification. 

Topic-based methods have similarly been used for organizing Sinhala news. \citet{kirindage2020automatic} employed Latent Dirichlet Allocation (LDA)~\cite{blei2003latent} to construct Sinhala news topic hierarchies and classify documents according to the resulting structure. 
\citet{sameemdeen2021topic} considered Naïve Bayes, SVM, and KNN while also reviewing earlier work by~\citet{nanayakkara2018clustering},~\citet{gunasekara2018context},~\citet{lakmali2017effectiveness}, and~\citet{buddhika2018domain}. The authors proposed active learning~\cite{novak2006text,yang2009effective} as an alternative for reducing annotation requirements, although no experiments were provided to demonstrate its effectiveness for Sinhala text classification. 
More recent work by~\citet{kariyakaranage2026cal} used Sinhala text classification to evaluate \texttt{CAL-Log}, a logarithmic model of annotation cost for active learning, thereby providing a more explicit investigation of annotation efficiency.

Several studies proposed classification frameworks intended to generalize across domains. \citet{kodithuwakku2021adapttext} introduced \textit{AdaptText}, a domain-independent and domain-adaptive framework for Sinhala text classification. 
\citet{bandara2021ontology} proposed an ontology-based approach to Sinhala fake-news detection. However, its literature review omitted foundational work in ontology-based information extraction, including~\citet{wimalasuriya2010ontology}, which weakened the methodological grounding of the proposed system.

The adoption of neural methods broadened both the models and tasks explored for Sinhala. \citet{jayasinghe2019deep} proposed an LSTM-based system for Sinhala textual entailment. \citet{demotte2021dual} introduced a dual-state capsule network architecture and evaluated it using the Sinhala dataset established by~\citet{senevirathne2020sentiment}. Capsule-based methods were also used by~\citet{chathuranga2021classification} to classify Sinhala--English code-mixed text, following the approach recommended by~\citet{senevirathne2020sentiment}. For social-media content,~\citet{caldera2022long} employed stacked LSTMs to classify Sinhala and Singlish discussions of COVID-19 collected from YouTube and Twitter.

Transformer-based models subsequently became an important focus. \citet{dhananjaya2022bertifying} conducted a comparative evaluation of BERT-based architectures for Sinhala text classification and publicly released the corresponding \gh{https://github.com/nlpcuom/Sinhala-text-classification}{code}.
For code-mixed and code-switched classification, \citet{rathnayake2022adapter} applied adapter-based fine-tuning~\cite{houlsby2019parameter,pfeiffer2020adapterhub,pfeiffer2020mad,pfeiffer2020adapterfusion,wang2021efficient,friedman2021single} to XLM-R~\cite{conneau2019unsupervised}, demonstrating how parameter-efficient adaptation can be used when fully fine-tuning large multilingual models is impractical. 
More recently,~\citet{ilangeshwaran2024saliency} introduced Saliency-based token Swap (SSwap), a data augmentation method designed to improve Sinhala text classification, while~\citet{wickramasinghe2025context} proposed a SinBERT-based model for Sinhala news personalization.

Text classification has also been applied to several specialized tasks. \citet{hettigoda2019english} collected English--Sinhala code-mixed comments from Facebook pages operated by online businesses in Sri Lanka's clothing industry and classified them as \textit{Inquiries}, \textit{Maybe Inquiries}, or \textit{Not Inquiries}. 
\citet{wijayarathnahybrid} used classical machine learning methods, including Random Forest, to classify Sinhala fake news on Twitter. 
\citet{ranathunga2025identifying} investigated writing-style identification specifically for distinguishing AI-generated from human-written Sinhala answers and reported an accuracy of 86\% using an LSTM operating on TF--IDF features. 
In literary authorship attribution,~\citet{fernando12026authorship} applied a CNN to TF--IDF representations to identify the authors of Sinhala poems.

Other work has extended classification beyond assigning conventional document labels. \citet{gunawardana2024popularity} attempted to predict the future popularity of Sinhala-language YouTube videos using multiple associated factors. 
\citet{rathnayake2025intelligent} used LLaMA to classify named entities in Sinhala documents before applying Fuzzy C-Means clustering and subsequently using LLaMA again to generate names for the resulting clusters. 

Overall, Sinhala text classification has developed from small-scale rule-based and conventional machine learning experiments into a broad research area encompassing neural models, multilingual transformers, data augmentation, parameter-efficient adaptation, and active learning. News remains the dominant domain, while writing-style analysis, fake-news detection, code-mixed classification, and social-media applications have received increasing attention. Despite the growing number of publicly released corpora, several influential datasets remain unavailable. Progress would therefore benefit from consolidated benchmark suites, clearer dataset provenance, standardized train--test divisions, and evaluations that compare classical, transformer-based, and generative approaches under consistent experimental conditions.

\subsection{Sentiment Analysis}
\label{Sec:SentimentAnalysis}

Sentiment analysis is among the most extensively studied semantic processing tasks in Sinhala NLP. Research has covered binary and multiclass sentiment classification, emotion recognition, reaction prediction, aspect-based sentiment analysis, code-mixed social-media content, and multimodal emotion analysis. The methodological progression broadly follows developments in NLP more generally, moving from lexicon-based and conventional machine learning approaches to recurrent neural networks, multilingual transformers, language adapters, explainable models, and fine-tuned large language models.

Several datasets have supported this research. \citet{jenarthanan2019actsea} introduced the \gh{https://github.com/Jenarthanan14/Tamil-Sinhala-Emotion-Analysis}{ACTSEA} dataset, which contains 318,308 Sinhala tweets annotated with emotion labels. More recently, \citet{galwatta2026salangabhava} released \texttt{SalAngaBhava}, an aspect-tagged Sinhala sentiment analysis dataset available through both \gh{https://github.com/lakshani-98/sinhala-absa-dataset}{GitHub} and \hf{https://huggingface.co/datasets/lakshani005/SalAngaBhava}{HuggingFace}. The dataset contains 10,989 customer-feedback records from online marketplaces, of which 1,858 were manually annotated. Using \texttt{InstructABSA}~\cite{scaria2024instructabsa}, the authors established baseline Macro F1-scores of 0.72 for Aspect Term Extraction (ATE) and 0.28 for Aspect-Level Sentiment Classification (ALSC).
Most Sinhala sentiment datasets employ discrete sentiment or emotion categories. In contrast,~\citet{de2025geesanbhava} introduced \gh{https://github.com/YomalCSE/Analysis-of-Emotional-Expressions-in-Sinhala-Music-Video-Comments-on-YouTube/blob/main/annotated_YouTube _Comment Data.xlsx}{GeeSanBhava} by manually annotating the raw Sinhala YouTube comment dataset released by~\citet{de2025linguistic}. This is the first reported Sinhala sentiment resource annotated according to Russell's Valence--Arousal model~\cite{russell1980circumplex}, providing a continuous dimensional representation of emotion rather than relying solely on predefined sentiment classes.

Early computational work on Sinhala sentiment analysis employed relatively simple neural and lexical methods. Building upon their earlier study~\cite{medagoda2015sentiment}, \citet{medagoda2016sentiment} proposed a multilayer perceptron-based method for classifying sentiment in Sinhala text. A publicly available \gh{https://github.com/theisuru/sentiment-tagger}{Word2Vec-based tool} was also developed for analysing Sinhala news comments. To address the scarcity of manually constructed lexical resources, \citet{chathuranga2019sinhala} proposed a semi-automated methodology for deriving a Sinhala sentiment lexicon from a given corpus.

Subsequent research increasingly adopted recurrent neural architectures and distributed representations. \citet{demotte2020sentiment} introduced a \gh{https://github.com/theisuru/sentiment-tagger}{sentiment analysis} system based on sentence-state LSTM networks for Sinhala news comments. In follow-up work,~\citet{karunanayake2020sinhala} used word similarity to construct a Sinhala semantic lexicon, while~\citet{senevirathne2020sentiment} evaluated several deep learning architectures, including recurrent neural networks and bidirectional LSTMs, for Sinhala sentiment analysis. \citet{ranathunga2021sentiment} further argued that word embeddings can serve as effective semantic features and partially compensate for the limited availability of Sinhala-specific linguistic resources when analysing sentiment in news comments.

Several studies compared alternative lexical representations and classification strategies. \citet{jayasuriya2021sentiment1} compared word-level and character-level $n$-grams for the sentiment classification of Sinhala social-media content and subsequently extended the work using an ensemble approach~\cite{jayasuriya2021sentiment2}. 
\citet{liyanaarachchi2024sentiment} investigated character-level embeddings for binary sentiment classification of Sinhala news comments and reported that the 1-of-$m$ representation outperformed the Log-$m$ representation. More recently, the short comparative study by~\citet{mohamed2025sinhala} found that XLM-R and SinBERT achieved the strongest overall performance on Sinhala comment sentiment analysis, although BiLSTM and Logistic Regression remained competitive for certain tasks.

Social-media sentiment has been studied across several platforms and domains. \citet{karunarathne2020sentiment} used word embeddings to analyse manually annotated Sinhala tweets, while~\citet{jayasuriya2020Sentiment} classified Sinhala sports-related posts into positive and negative sentiment categories. \citet{jayawickrama2021seeking,jayawickrama2022facebook} used the Facebook corpus released by~\citet{wijeratne2020sinhala} to predict the reactions elicited by Sinhala posts. This work was subsequently extended by~\citet{weeraprameshwara2022sentiment}, who compared results obtained from the Facebook reaction dataset against those obtained using the sentiment dataset of~\citet{senevirathne2020sentiment}. Sentiment-related classification has also been applied to music:~\citet{abeyratne2019classification} conducted a multimodel analysis of emotion classification for Sinhala songs.

Code-mixed Sinhala--English content has become another prominent research direction. \citet{aththanayaka2020sentimental} applied Random Forest, Support Vector Machines, and Multinomial Naïve Bayes to sentiment analysis of Sinhala--English code-mixed social-media text. \citet{uthpala2024sinhala} created a sentiment-annotated dataset of Sinhala--English code-mixed YouTube comments. However, both the publication and the dataset are behind access restrictions, and the resource is therefore not freely available for external research.

More recent work has applied multilingual pretrained language models to compensate for the scarcity of Sinhala-labelled data. \citet{dhananjaya2024lexicon} used the sentiment lexicon of a high-resource language to fine-tune pretrained multilingual language models, including mBERT and XLM-R, on an intermediate task before adapting them to Sinhala sentiment classification. \citet{gurgurov2024adapting} compared language-model adaptation strategies for Sinhala sentiment analysis and obtained the strongest results using a masked language modelling objective. Their configurations combined ConceptNet (\textit{CN}) and/or Wikipedia (\textit{wiki}) language adapters with task adapters (\textit{TA}).

Transformer models have also been evaluated directly on established Sinhala sentiment datasets. \citet{bandaranayake2025sentiment} compared BERT, DistilBERT, RoBERTa, and XLM-R using the Sinhala news-comment dataset introduced by~\citet{senevirathne2020sentiment}, reporting that XLM-R-large achieved the strongest performance. In their work on improving the cross-lingual representations of multilingual language models through linguistic entity masking,~\citet{fernando2025linguistic} used Sinhala code-mixed sentiment analysis as one of the evaluation tasks. More recently, \texttt{SinLlama}~\cite{aravinda2025sinllama} was fine-tuned by~\citet{abeynayake2026fine} for sentiment analysis of Romanised Sinhala--English code-mixed social-media text.

Aspect-based sentiment analysis and explainability have received increasing attention. \citet{rizvi2025keyword} used XLM-RoBERTa to extract keywords and classify aspects in Sinhala code-mixed data. In subsequent work~\cite{rizvi2025enhancing}, the authors again employed XLM-RoBERTa for sentiment classification and incorporated SHAP~\cite{lundberg2017unified} and LIME~\cite{ribeiro2016should} to explain the resulting predictions. Similarly,~\citet{senevirathna2025explainable} reported the creation of a dataset containing 13,000 aspect-tagged banking reviews in Sinhala, English, and code-mixed language. Their hybrid XLM-RoBERTa model incorporated domain-specific lexicon correction and achieved a reported accuracy of 88.4\% and an F1-score of 0.84 for the Sinhala and code-mixed data. SHAP~\cite{lundberg2017unified} and LIME~\cite{ribeiro2016should} were again used to provide model-level explanations.

Sentiment and emotion analysis have also been extended beyond text-only classification. \citet{peiris2025hybrid} analysed emotion in Sinhala songs using both lyrical and acoustic information. For the textual component, the authors used RoBERTa to represent Sinhala words before combining the resulting features with audio-based evidence. 

Several domain-specific sentiment datasets and models have additionally been reported. \citet{dissanayake2025sentiment} collected a dataset of \textit{Sinhala patient feedback in a hospital setting} and reported that a BiLSTM model achieved the highest accuracy of 98.08\%. \citet{perera2026sinhala} reported the construction of a dataset containing 13,000 manually annotated YouTube comments for Sinhala code-mixed sentiment analysis. Their evaluation covered conventional machine learning methods---including Logistic Regression, Support Vector Machines, Multinomial Naïve Bayes, K-Nearest Neighbour, Decision Trees, and Random Forests---as well as deep learning and transformer models, including BERT, DistilBERT, ALBERT, and XLM-R.

Overall, Sinhala sentiment analysis has developed from small-scale binary classification and manually constructed lexicons into a diverse research area encompassing large social-media datasets, recurrent neural networks, multilingual transformers, adapter-based transfer, aspect-level analysis, explainable AI, multimodal emotion recognition, and fine-tuned LLMs. News comments and social-media text remain the dominant domains, although recent work has expanded into online marketplaces, banking, healthcare, music, and Romanised code-mixed communication. The field would benefit from clearer dataset licensing, standardised train--validation--test partitions, consistent sentiment and emotion taxonomies, and broader comparisons across classical, encoder-based, and generative approaches.

\subsection{Hate Speech Detection}
In the Sinhala NLP literature, the terms \textit{hate speech detection}, \textit{offensive speech detection}, and \textit{inappropriate speech detection} are frequently used interchangeably. This contrasts with research on higher-resource languages, where these tasks are typically defined as related but distinct phenomena~\cite{mathew2021hatexplain}. Hate speech generally targets individuals or groups on the basis of protected characteristics, whereas offensive or inappropriate language may include insults, profanity, abuse, or other harmful content without necessarily expressing identity-directed hatred. Because the existing Sinhala literature rarely maintains this distinction consistently, this subsection jointly reviews work described under one or more of these task labels.

A growing collection of annotated resources supports research in this area. \citet{costa2022no} introduced a large \gh{https://github.com/facebookresearch/flores/tree/main/toxicity}{Sinhala toxicity detection dataset} by extending material from \textit{FLORES}~\cite{guzman2019flores}. The resource can support direct toxicity detection as well as translation-based approaches to hate-speech analysis. \citet{gammanpila2026hate} reported the creation of \texttt{SOCOLD}, an obfuscated and code-mixed Sinhala offensive-language dataset designed to capture deliberately altered harmful expressions.

Platform-specific datasets have also been released. \citet{de2020self} published a \gh{https://github.com/eranga-de-saa/Sinhala_Youtube_Hate_Video_Detector_Analysis/tree/4ca2d90930c53a6c9e75fc96064d2f8789443ca8/dataset}{YouTube comment dataset} together with a feature model for Sinhala hate-speech detection. A separate collection of Sinhala hate-speech comments obtained from Facebook is available through Kaggle\footURL{https://bit.ly/3j9r3Za}, although it has no accompanying research publication. \citet{perera2023comparative} released a \gh{https://github.com/Isurie/Text-Classification-Module/tree/master/Dataset}{hate-annotated dataset} containing more than 1,600 Sinhala tweets. In addition, a lexical resource of Sinhala swear and obscene words is publicly available in both \gh{https://github.com/thambaru/sinhala-bad-word-moderation-list/blob/master/unicode-word-list.txt}{Unicode} and \gh{https://github.com/thambaru/sinhala-bad-word-moderation-list/blob/master/singlish-word-list.txt}{Singlish} formats.

More recently,~\citet{chavinda2025dual} introduced \hf{https://huggingface.co/datasets/krishan-CSE/Facebook_Sinhala_Hate_Speech}{Facebook} and \hf{https://huggingface.co/datasets/krishan-CSE/Twitter_Sinhala_Hate_Speech}{Twitter} datasets for Sinhala hate-speech detection. They evaluated multilingual large language models using Dual Contrastive Learning (DCL) and reported that \texttt{TwHIN-BERT-base}~\cite{zhang2023twhin} achieved the strongest performance.

The most influential benchmark in this area is the \hf{https://huggingface.co/datasets/sinhala-nlp/SOLD}{Sinhala Offensive Language Dataset (SOLD)} introduced by~\citet{ranasinghe2022sold}. SOLD contains 10,000 Sinhala Twitter posts annotated at both sentence and token levels using the binary classes \textit{offensive} and \textit{not offensive}. The same study introduced \hf{https://huggingface.co/datasets/sinhala-nlp/SemiSOLD}{SemiSOLD}, a substantially larger collection of 145,000 Sinhala tweets annotated using a semi-supervised procedure. The accompanying code is publicly available through \gh{https://github.com/Sinhala-NLP/SOLD}{GitHub}. SOLD subsequently became the principal shared benchmark for comparing Sinhala offensive-language detection systems.

Early Sinhala hate-speech detection research predominantly relied on conventional machine learning and manually engineered features. \citet{de2019approach} proposed a machine learning approach for Sinhala hate-speech detection, while~\citet{kariyawasam2019machine} investigated the identification of toxic Sinhala content on social media. \citet{sandaruwan2019sinhala} collected a corpus of 3,000 Sinhala comments and applied Multinomial Naïve Bayes, whereas~\citet{sheran2019detection} reported a machine learning system for detecting hate speech written in either Sinhala or Singlish.

Further studies expanded the range of harmful-content categories and platforms considered. \citet{sandaruwan2020identification} used text mining and machine learning methods to identify abusive Sinhala comments on social media. \citet{amali2020classification} examined Sinhala cyberbullying classification using classical machine learning algorithms, while~\citet{hettiarachchi2020detecting} applied similar methods to Romanised Sinhala social-media posts. Although the latter work did not explicitly describe the text as \textit{transliterated} Sinhala, its treatment of Romanised content corresponds to a transliteration-based setting.

The transition toward neural methods began with studies such as~\citet{samarasinghe2020machine}, who proposed a Convolutional Neural Network for Sinhala hate-speech detection. \citet{munasinghe2022deep} later explored deep learning for detecting hate speech in Sinhala tweets, while~\citet{fernando2022sinhala} reported the use of both machine learning and deep learning methods for the task. \citet{gamage2022improving} compared several embedding representations with classical frequency-based features, providing a broader evaluation of alternative text representations for Sinhala hate-speech detection.

Classical models nevertheless remained competitive in several comparative studies. \citet{guruge2022analyze} evaluated an ensemble of Naïve Bayes, Support Vector Machines, XGBoost, multilayer perceptrons, and AdaBoost using 49,019 tweets collected between February and April 2021. \citet{shalinda2022hate} compared a Linear Support Vector Classifier, Random Forest, SGD, Logistic Regression, XGBoost, and Multinomial Naïve Bayes on Sinhala and Singlish content to identify  \textit{hate words}. Their strongest result was obtained using an SGD classifier with TF--IDF representations constructed from unigrams and bigrams.

Beyond classifying individual messages, researchers have also investigated the users and temporal dynamics associated with harmful content. \citet{perera2023predicting} attempted to predict Sinhala hate speech using Twitter user behaviour and ensemble machine learning, with users represented according to the Big Five personality traits. Their follow-up study~\cite{perera2023comparative} analysed hate-speech propagation on Twitter and released the accompanying \gh{https://github.com/Isurie/Text-Classification-Module/tree/master/Dataset}{dataset} of more than 1,600 annotated Sinhala tweets. \citet{rajapaksha2023analyzing} used deep learning to identify periods during which particular hate-related topics became prominent on Sinhala Twitter.

Several studies have addressed inappropriate or harmful content through indirect or specialised formulations. \citet{arachchi2023inappropriate} used a web-based Sinhala--English translation service as part of a system for detecting inappropriate Sinhala word usage. \citet{ehelepola2023hate} analysed hate speech in Sinhala text obtained from both social-media and e-commerce platforms. \citet{hewage2025detecting} concentrated specifically on \textit{body shaming}, constructing a corpus containing both Sinhala-script and Romanised Sinhala instances.

Obfuscated profanity represents another important challenge because users frequently alter spellings to evade automated moderation. \citet{premachandra2025monolingual} reported a manually annotated dataset of 4,000 Facebook comments containing deliberately modified profanity. Using SinBERT~\citet{dhananjaya2022bertifying}, they reported F1-scores of 0.671 for profanity detection and 0.387 for altered-speech detection. The comparatively lower performance on altered expressions attests to the difficulty of detecting adversarial orthographic variation.

Code-mixed and Romanised Sinhala have likewise received substantial attention. \citet{muthuthanthri2024hate} evaluated TF--IDF and \textit{fastText} representations with SVM, CNN, RNN, LSTM, BERT, and GPT-2 models using an annotated Facebook dataset. They concluded that BERT achieved the strongest performance for detecting hate speech in English--Sinhala code-mixed text. \citet{tanjim2025multilingual} studied cyberbullying detection across several low-resource languages and, when treating Sinhala as the target language, recommended Assamese, Bengali, Nepali, Urdu, Hindi, or Bodo as transfer-learning source languages.

More recent systems increasingly rely on pretrained multilingual transformers. \citet{jahnavi2025hate} combined BERT with CNN and LSTM components for Sinhala hate-speech detection. \citet{perera2025identification} used XLM-RoBERTa to assign Sinhala YouTube titles and comments to different hate-severity levels, thereby formulating the task as multi-class rather than binary classification.

Hate-speech analysis has also expanded beyond text-only inputs. \citet{dikwatta2024exploring} applied supervised algorithms to Sinhala text extracted from Internet memes. In subsequent work~\cite{dikwatta2025can}, the authors introduced a Sinhala meme dataset annotated using both textual and visual components and reported an accuracy of 95\% for their SVM-based approach. Although the reported model is not inherently multimodal, the dataset enables future work that jointly models the two modalities.

A related multimodal formulation was explored by~\citet{wickramaarachchi2023identifying}, who used an LSTM over BART-derived representations to compare the titles and descriptions of Sinhala YouTube videos with their audio content in order to determine whether the videos contained hate speech. This extends the task from message-level classification to consistency analysis across textual and acoustic evidence.

The release of SOLD~\cite{ranasinghe2022sold} led to a more standardised phase of Sinhala offensive-language research. \citet{fernando2023enhancing} evaluated traditional machine learning algorithms on the benchmark. \citet{ranasinghe2023text} compared pretrained mT5 performance on Sinhala SOLD tweets with corresponding results for German, Spanish, Hindi, Korean, and Marathi. \citet{bestgen2023using} compared classical machine learning models using simple $n$-gram features on SOLD against equivalent experiments for Assamese and Bengali.
The benchmark has also supported research on translation, transliteration, and cross-lingual transfer. \citet{dmonte2025does} used SOLD to investigate whether machine translation affects offensive-language detection performance. \citet{chamikara2026hate} first transliterated SOLD into Romanised Sinhala and then compared XLM-RoBERTa, IndicBERT, and distil-mBERT on the transformed dataset. These studies demonstrate that script representation and translation quality can materially affect downstream classification.
SOLD was adopted as the Sinhala dataset for the Hate Speech and Offensive Content (\textit{HASOC}) identification task introduced by~\citet{ranasinghe2023overview}\footURL{https://hasocfire.github.io/hasoc/2023/task1.html}. For this shared task,~\citet{narayan2023hate} compared several pretrained deep learning models and reported that XLM-RoBERTa-base achieved a Macro F1-score of 83.49\%, compared with 75.30\% for an LSTM baseline. \citet{ojo2023hate} used mBERT for the same Sinhala hate- and offensive-content identification task.
Generative models have also been evaluated on SOLD. \citet{rostamkhani2023hate} compared the zero-shot performance of ChatGPT against several Sinhala models and concluded that zero-shot ChatGPT outperformed systems that had not themselves been fine-tuned on SOLD. More recently, \citet{haturusinghe2025subasa} introduced the \textit{Subasa} family of models and evaluated them on SOLD. Their trained \gh{https://github.com/haturusinghe/subasa-plm}{PMLs} and \gh{https://github.com/haturusinghe/subasa-llm}{LLMs} are publicly available in separate repositories. The authors report that \texttt{Subasa-XLM-R} outperforms even state-of-the-art commercial LLMs on the SOLD benchmark.
Machine-translated offensive-language resources have provided another route for expanding Sinhala data. \citet{dmonte2024effects} used SOLD to evaluate the \gh{https://github.com/LanguageTechnologyLab/MT-Offense}{MT-Offense} dataset, which was created by translating the English Offensive Language Identification Dataset (OLID)~\cite{zampieri2019predicting} using neural machine translation. In their detailed evaluation,~\citet{dmonte2024effects} reported a best F1-score of 0.550 for translations generated by nllb-200-3.3B~\cite{costa2022no}. This comparatively modest result illustrates the limitations of constructing offensive-language datasets through automatic translation alone, particularly when culturally and linguistically specific expressions do not transfer directly.

Overall, Sinhala hate- and offensive-speech detection has developed from small, independently collected social-media corpora and conventional classifiers into a diverse field encompassing code-mixed and Romanised text, obfuscated profanity, user behaviour, temporal propagation, multimodal memes, multilingual transfer, adapter and transformer models, and generative LLMs. SOLD has played a central role in enabling more reproducible comparison and shared-task evaluation. Nevertheless, the field continues to face inconsistent task definitions, overlapping annotation schemes, limited coverage of targeted identities and hate categories, and insufficient evaluation of model robustness to transliteration, spelling variation, cultural context, and domain shift. Future benchmarks should clearly distinguish hate speech, offensiveness, profanity, cyberbullying, and other forms of harmful language while supporting multiclass, target-aware, token-level, and multimodal evaluation.

\subsection{Fake News Detection}

Fake news and misinformation detection has received growing attention in Sinhala NLP, driven largely by the widespread use of social media and online news platforms. Compared with sentiment analysis or hate-speech detection, however, research in this area remains relatively limited. Existing work spans misinformation corpora, hybrid credibility-assessment frameworks, semantic similarity approaches, multimodal verification, and, more recently, evaluations using large language models.

The principal publicly available benchmark for Sinhala misinformation detection was introduced by~\citet{jayawickrama2021corpus}. The authors released a \gh{https://github.com/LIRNEasia/MisinformationCorpusSinhala}{dataset} containing 3,576 Sinhala news articles collected from Sri Lankan news websites and annotated with four credibility labels: \textit{CREDIBLE}, \textit{FALSE}, \textit{PARTIAL}, and \textit{UNCERTAIN}. Alongside the dataset, they also established baseline misinformation classification models, providing one of the first standardised benchmarks for Sinhala fake-news detection.

Early research primarily relied on hybrid systems that combined textual analysis with external credibility indicators. \citet{wijayarathna2021text} proposed a framework that verified the textual content of social-media posts against trusted news sources while simultaneously evaluating the credibility of the user account through a rule-based scoring scheme. 

Several studies have instead focused on semantic representations of the textual content itself. \citet{wijayarathna2020sinhala} collected a dataset containing 120 fake-news tweets and 250 genuine tweets. Each tweet was represented by averaging the FastText embeddings of its constituent words, after which vector similarity was used to predict whether the associated news item was genuine or fabricated. 

Subsequent work explored increasingly sophisticated hybrid scoring frameworks. \citet{udurawana2022hybrid} proposed combining an accuracy score derived from analysing the news content with a credibility score obtained through a rule-based scoring mechanism. Their framework additionally incorporated passive-aggressiveness classification. 
The adoption of deep learning methods has been comparatively limited but continues to increase. \citet{adihetti2023sinhala} investigated the use of autoencoders for detecting fake news in Sinhala social-media posts.
Researchers have also explored multimodal formulations of the problem. \citet{wickramaarachchi2023identifying} detected fake content in Sinhala YouTube videos by comparing the titles and descriptions of videos against automatically derived representations of the spoken audio using an LSTM operating on BART representations. 

Recent work has shifted toward benchmarking and evaluating large language models. \citet{atthanayake2025comprehensive} conducted a comparative study between the English \texttt{Fakeddit} benchmark~\cite{nakamura2020fakeddit} and the Sinhala misinformation corpus introduced by~\citet{jayawickrama2021corpus}, highlighting similarities and differences between misinformation datasets in high-resource and low-resource languages. Building upon the rapid progress of generative language models,~\citet{premadasa2025exploring} evaluated GPT-4o, DeepSeek-Chat, and Gemini-2.5-Flash-Lite for misinformation detection in Sinhala online news articles under zero-shot, few-shot, Chain-of-Thought (CoT), and role-based prompting settings. This represents a shift away from task-specific supervised classifiers toward prompt-based reasoning with foundation models.

Overall, Sinhala fake-news detection has evolved from small-scale embedding-based classifiers and hybrid credibility-scoring systems to multimodal verification frameworks and prompt-driven large language model approaches. Compared with related areas such as sentiment analysis and hate-speech detection, however, the availability of publicly accessible benchmark datasets remains relatively limited, with the LIRNEasia misinformation corpus serving as the primary evaluation resource. Future research would benefit from larger multi-domain datasets, multimodal benchmarks that jointly model text, images, videos, and user metadata, and standardised evaluation protocols that enable robust comparisons between conventional machine learning, neural architectures, and modern generative language models.

\subsection{Language Identification (LID) Task}

Language Identification (LID), also referred to as language identification (LangID), is the task of automatically determining the language in which a given piece of text is written. For Sinhala, this task is generally regarded as comparatively straightforward because the language is predominantly written using the unique Sinhala script. Consequently, most existing benchmarks report near-perfect performance for Sinhala. However, this apparent success masks several practical challenges that are largely absent from current evaluations, including Romanised (Singlish) text, code-mixed content, closely related languages such as Pali and Sanskrit written in the Sinhala script, and deliberately obfuscated spelling.

One of the largest publicly available multilingual resources for this task was introduced by~\citet{burchell2023open}, who released a \gh{https://github.com/laurieburchell/open-lid-dataset}{dataset} together with multilingual language identification models covering Sinhala and many other languages. Their models generally outperform the language identification component of NLLB~\cite{costa2022no} across most languages. For Sinhala, however, they reported performance comparable to that of NLLB, ostensibly suggesting that conventional Sinhala language identification is already approaching saturation under existing benchmark conditions.

Several studies have investigated factors beyond purely textual representations. \citet{dunn2024geographically} explored whether incorporating geographical priors could improve language identification. While geographic information benefited certain languages, the authors found virtually no improvement for Sinhala because the language identification model and the geographical model already agreed on approximately 99\% of the instances. 

Research has also focused on developing lightweight multilingual language identification tools. \citet{kargaran2023glotscript} introduced \textit{GlotScript}, a written language identification system that achieved over 80\% accuracy for Sinhala. 
More recently,~\citet{van2025identifying} surveyed the remaining challenges of language identification across more than 2,000 languages. Their experiments reported F1-scores exceeding 99\% for Sinhala, with both \texttt{Textcat}~\cite{cavnar1994n}\footURL{https://www.let.rug.nl/vannoord/TextCat/} and \texttt{GLOT500}~\cite{imani2023glot500} achieving 99.9\%. However, the authors' evaluation considered only Sinhala text written in the Sinhala script. Consequently, the reported performance reflects script identification as much as language identification. 
This is further illustrated by the original \gh{https://github.com/cisnlp/Glot500}{GLOT500} study, where~\citet{imani2023glot500} reported only 45.0\% accuracy for Sinhala under substantially more challenging multilingual evaluation settings.

These studies imply that conventional Sinhala language identification is largely a solved problem when the language is written using the Sinhala script. Nevertheless, this conclusion should be interpreted with caution because current benchmarks do not adequately capture many real-world scenarios encountered in Sri Lankan digital communication. Existing evaluations largely ignore Romanised Sinhala, Sinhala--English code-mixed text, closely related languages such as Pali and Sanskrit written using the Sinhala script, dialectal variation, and adversarial spellings. Future research should therefore move beyond script-based discrimination and develop benchmarks that evaluate language identification under these more challenging and practically relevant conditions.


\subsection{Plagiarism Detection}
An extremely simple plagiarism detection tool which only uses n-grams of simply tokenized text was proposed by~\citet{basnayakeplagiarism}. Another simple plagiarism detection tool that uses synonymy and Hyponymy-Hypernymy (which they call \textit{Generalization} in the paper) was attempted by~\citet{rajamanthrisinhala}. They later extended this work~\cite{rajamanthri2021plagiarism} to propose a more advanced plagiarism detection tool which uses Internet resources.~\citet{kasthuriarachchi2019deep} proposed using Word2Vec vector cosign similarity to detect plagiarism. A multi-document Sinhala similarity detection tool to detect plagiarism was proposed by\citet{piyarathna2019sinhala}. \citet{punchihewa2021language} developed a character-level model which can identify the author for Sinhala text in student answers. The work by~\citet{jayasundara2025bilingual} used cosine similarity on  TF-IDF vectors to identify plagiarism in Sinhala English mixed student essays and reports. \citet{primal2026sinhala} have proposed a plagiarism checking system which uses LaBSE embeddings. 

\subsection{Sinhala-English Code-Mixing}
The problem of recognizing Sinhala and English code-mixed data where the Sinhala text is written in Singlish was explored by~\citet{smith2019language} and later by~\citet{smith2020sinhala} using an XGB classifier and a CRF model building on their previous work~\cite{smith2019sinhala}, which analysed such data.~\citet{shanmugalingam2019language} also attempted to identify the language in Sinhala-English code-mixed text using Support Vector Machines (SVM), Naive Bayes, Logistic Regression, Random Forest, and Decision Trees. A dictionary based approach to standardize Sinhala Code-Mixed text was proposed by~\citet{kugathasan2020standardizing}. They later used it for NMT in Sinhala-English Code-Mixed text~\cite{kugathasan2022neural}.
\citet{shakir2023compiling} claims to have developed a corpus of South Asian languages in the context of code-mixing which includes Sinhala. However, this is not publicly available. It should further be noted that they report Sinhala data using the heading \textit{Sri Lanka} rather than \textit{Sinhala}. In their follow-up work~\citet{shakir2024code} claim to extend this code-mixed data set to include meme data.

As discussed in Section~\ref{Sec:TextClassification},~\citet{rathnayake2022adapter} used adapter-based~\cite{houlsby2019parameter,pfeiffer2020adapterhub,pfeiffer2020mad,pfeiffer2020adapterfusion,wang2021efficient,friedman2021single} fine-tuning on XLM-R~\cite{conneau2019unsupervised}, for classifying code-mixed and code-switched Sinhala text.~\citet{hettigoda2019english} classified English-Sinhala code-mixed comments from Facebook. As discussed in Section~\ref{Sec:SentimentAnalysis},~\citet{aththanayaka2020sentimental} utilized traditional machine learning methods for sentiment analysis on Sinhala-English code-mixed text from social media.~\citet{chathuranga2021classification} used capsule-based methods to classify Sinhala-English code-mixed data.
~\citet{dissanayake2022enhancing} claim that the web-based code-less chatbot development platform for Sinhala proposed by them is capable of handling Sinhala-English code-switching.
The feature set derived by~\citet{walawage2020devising} for text on social media images (memes) attempts to separately identify Sinhala and English text. 
The study by~\citet{fazal2023depression} used classical machine learning algorithms on TF-IDF features to predict depression in Sinhala-English code-mixed data from Twitter and Facebook.
The study by~\citet{udawatta2024use} shows that prompt-based learning of pre-trained language models (PLMs) outperforms full fine-tuning of PLMs on Code-mixing and code-switching (CMCS) English-Sinhala data across various NLP tasks such as sentiment classification, hate-speech detection, and humour detection.
\citet{uthpala2024sinhala} claims to have created a sentiment-annotated Sinhala-English code mixed dataset using comments from YouTube videos. However, given that even their paper is behind a paywall, there is no free and public way to access this dataset. 
\citet{senanayaka2024singrag} proposes \textit{SingRAG}, a RAG and translation framework built on LLaMA that converts code-mixed Sinhala to English for processing and then converts whatever the English results is of the attached backend to code-mixed Sinhala before sending it back to the user.

\subsection{Question Answering Systems (Q\&A)}
\citet{romero2024cvqa} introduced \textit{CVQA}, a multilingual visual Question Answering benchmark which includes Sinhala. They have publicly released both their \hf{https://huggingface.co/datasets/afaji/cvqa}{data} and leader board\footURL{https://nisansa.lk/?s=sOrx7r}. \citet{vayani2024all} created a dataset \hf{https://huggingface.co/datasets/MBZUAI/ALM-Bench}{ALM-Bench} by machine translating English QA pairs from \textit{LLaVA-Bench} (In-the-Wild) dataset~\cite{liu2024visual} and extended it by searching for images under \textit{country name}, \textit{language name}, and \textit{cultural category} on the web so that they can include culturally relevant questions and answer for each language. They then reported that for Sinhala Q\&A (especially visual Q\&A), GPT-4o~\cite{achiam2023gpt} is superior to other tested LLMs.

\citet{dissanayake2020thematic} implemented a question-and-answer generator for Sinhala with the limited PoS: pronouns, adjectives, verbs, and adverbs.~\citet{jayakody2016mahoshadha} uses simple KNN and SVM methods on a PoS tagged Sinhala corpus to create a question-answering system which they name \textit{Mahoshadha}. \citet{shah2026answer} introduced a \gh{https://github.com/globalsouthml/public_service_triage_public_repo}{small benchmark dataset} of 90 items in the public-service question domain where systems are evaluated on whether they should answer the question, ask for further clarifications, or escalate the user to an official channel.

\citet{wansekara2024intelligent} developed a visual Sinhala question and answer system using YOLO for object identification and a BERT model trained on a Sinhala corpus for Sinhala text. \citet{dhanawardhana2025enhancing} frame their work just as a multi-model deep learning approach. But in actuality, their lost-and-found application is utilising techniques that are relevant to Sinhala VQA. 
 The system created by~\citet{dasanayaka2025multimodal} is claimed to be capable of first, generating radiology reports for orthopantomography (related to dentistry) using Llama on captions generated by a Blip-2-based caption generator on dental panoramic tomography (DPT) images and then serve a Sinhala chatbot where the patients may ask questions on the said report and get answers. 
\citet{shafique2025culturally} introduced \texttt{ViMUL-Bench} a multimodal video benchmark for LLMs that includes Sinhala. With 49.6, they show that GPT4o has the best results for Sinhala; however, the Sinhala accuracy is the lowest of all the tested languages across almost all the LLMs.

\subsection{Text Summarizing}
\citet{hasan2021xl} introduced the \gh{https://github.com/csebuetnlp/xl-sum/tree/master}{XL-sum} dataset, which includes 3,414 Sinhala documents and their summaries collected from the Sinhala BBC website\footURL{https://www.bbc.com/sinhala}. Their \hf{https://huggingface.co/csebuetnlp/mT5_multilingual_XLSum}{trained model} is also available.
The \gh{https://github.com/anubhav-jangra/M3LS}{M3LS} multi-modal dataset introduced by~\citet{verma2023large} contains 10,148 Sinhala documents, relevant images (making this a multi-modal dataset), and their summaries collected from the Sinhala BBC website\footURL{https://www.bbc.com/sinhala}. Further, they claim that MT5~\cite{xue2020mt5} obtains the best ROUGE-1 and ROUGE-L f scores for their Sinhala dataset.
The first Sinhala multi-document summarizing dataset, \hf{https://huggingface.co/datasets/KushanH/m2ds}{M2DS}, was created by~\citet{hewapathirana2024m2ds}. The \textit{M2DS} dataset consists of 23.5k Sinhala documents in 5.5k clusters, with each cluster having a golden summary. Their code and pre-trained models are also available on \gh{https://github.com/KushanMH/m2ds}{GitHub}.

A deterministic process flow for automatic Sinhala text summarizing was proposed by~\citet{welgama2012automatic}. The study by~\citet{wimalasuriya2019automatic}, which has the same name as the above work by~\citet{welgama2012automatic}, uses the graph-based TextRank algorithm for automatic Sinhala text summarizing. The use case of Sinhala Text summarization for government gazettes was explored by~\citet{jayawardane2022automatic}. The study by~\citet{rathnayake2023talking} compared the results of extractive and abstractive summarization on Sinhala textbooks. The study by~\citet{jahan2023automated} compared the abstractive summarizing methods of TF-IDF and Text-Rank for Sinhala using ROUGE as the evaluation score. They concluded TF-IDF to be the superior choice.
\citet{patabandi2025whatsapp} used BART and DistilBART to summarise Romanised Sinhala (Singlish) messages collected from WhatsApp group chats. \citet{seneviratne2026sinhala} claims to have created a dataset of human-validated 2,493 Sinhala article-summary pairs and established a benchmark of ROUGE-1 0.286 with a fine-tuned Gemma-3-4B model.


\subsection{Text Simplification}
The \hf{https://huggingface.co/datasets/MLSP2024/MLSP2024}{MultiLS} data set introduced by~\citet{shardlow2024extensible} for \gh{https://github.com/MLSP2024/MLSP_Data}{MLSP 2024 Shared Task} includes a Sinhala text simplification data set with data from News and Religious domains. \citet{ranathunga2025sitse} introduced \gh{https://github.com/brainsharks-fyp17/Sinhala-Text-Simplification-Dataset-and-Evaluation}{SiTSE}, a Sinhala text simplification dataset. It has 1000 complex Sinhala texts along with two separate simplifications for each of them. Further, they provide the evaluation metrics that can be used to evaluate any subsequent research conducted using the same data. \citet{liu2026oasissimp} created \gh{https://oasissimpdataset.github.io/OasisSimp.zip}{OasisSimp}, a text simplification dataset that includes  Sinhala entries that were sourced from the SiTa trilingual parallel corpus~\cite{ranathunga2018si}.

The text simplification shared task of Building Educational Applications (BEA) at MLSP 2024~\cite{shardlow2024bea} included the \textit{NSINA}~\cite{hettiarachchi2024nsina} Sinhala corpus. The aggregated data from the shared task is available online as \gh{https://github.com/MLSP2024/MLSP_Data/}{Train-Test data} and \hf{https://huggingface.co/datasets/MLSP2024/MLSP2024}{cleaned gold data}. \citet{enomoto2024tmu} experimented with GPT-4 and noted that re-ranking systems outperformed the base system for Sinhala while the opposite was true for all the other nine languages. \citet{goswami2024gmu} also tested GPT-4 (Terbo) for this task against the top 10 suggestion model and reported better results for GPT-4 for Sinhala than the top 10 suggestion model. \citet{seneviratne2024anu} compared multiple models for the English-Sinhala language pair simplification for the lexical complexity prediction task and lexical substitution task. In the submission by~\citet{cristea2024machine} word frequency and other meta-knowledge are not used for Sinhala text simplification while they are used for the other languages in the task. Their Sihala models are trained only using Sinhala text Google translated to English. As such the results they report for Sinhala for this task is quite weak compared to their results for the other languages. 
Even though the main focus on the work by~\citet{nohejl2025japanese} is introducing a Japanese text simplification data set and a model, to demonstrate the generalizability of the system, they use the \texttt{MultiLS} data set~\cite{shardlow2024extensible} which includes Sinhala. 

\subsection{Text Generation}
The study by~\citet{kao2020automated} used LSTM to generate Sinhala lyrics.~\citet{bandarasibil} introduced \textit{Sibil AI}, a GTP-2 based Children's story generator for Sinhala. However, they did not train a GTP-2 model in Sinhala. Instead, they trained an English GTP-2 model and generated stories in English, which were then translated to Sinhala using Google Translate API.
\citet{amarasinghe2019evolutionary} used an ontology-based approach to generate Sinhala essay questions. \citet{warnasooriya2026adaptive} used a Sinhala-adapted LLaMA model to generate Sinhala stories and quizzes.
\citet{ranasinghe2023image} trained a Sinhala image caption generator model on Google translated \textit{MS COCO}~\cite{lin2014microsoft} dataset and tested it on Google translated \textit{Flickr8K}~\cite{rashtchian2010collecting} and \textit{MS COCO} datasets. \citet{chen2026munichus} introduced \hf{https://huggingface.co/datasets/tharindu/MUNIChus}{MUNIChus}, a multilingual news image captioning benchmark dataset which includes Sinhala. However, with 2,418 training examples and 998 testing examples, Sinhala is the least represented language in this dataset, and the authors specifically observed that Sinhala has the worst results among the included languages. Nevertheless, they also release two models fine-tuned on the dataset: \hf{https://huggingface.co/alita9/xl-munichus-CohereLabs-aya-vision-8b}{Aya-vision-8b} and \hf{https://huggingface.co/alita9/xl-munichus-meta-llama-Llama-3.2-11B-Vision-Instruct}{Llama-3.2-11B}.
The work by~\citet{liyanage2019multi,liyanage2020multi} attempt to use LSTMs for mathematical word problem generation in Sinhala and other languages. The follow-up work by~\citet{niyarepola2022math} used pre-trained mBART~\cite{liu2020multilingual} and mT5~\cite{xue2020mt5,tang2020multilingual} models for the same task and showed better results. \citet{gamage2025multilingual} created a dataset named \gh{https://github.com/OmegaGamage/multiMWP}{MultiMWP} which consists of Math Word Problems (MWPs) in nine languages, including Sinhala.

A Sinhala programming language named \textit{Helaa} based on Java was proposed by~\citet{yasasri2023helaa}. \citet{senarathne2023translate} introduced a model to \textit{translate} Sinhala pseudocode to computer code written in C\#. The works by~\citet{de2024sinhala,de2024transformer} claim to be able to provide Sinhala language explanations to JAVA code as well as generate JAVA code snippets from Sinhala descriptions. They later extend this work to create a JAVA programming assistance tool~\cite{de2024sinhala3,athukorala2024sinhala,athukorala2025bridging} for Sinhala native speakers.

\subsection{Chatbots}
A simple Sinhala chatbot which utilizes a small knowledge base has been proposed by~\citet{hettige2006first}. A study on the effect of word embeddings on a Sinhala chatbot was conducted by~\citet{gamage2020impact} where they used, the fasttext model trained by Facebook~\cite{joulin2016fasttext,bojanowski2017enriching,joulin2017bag}, on a RASA\footURL{https://rasa.com/}~\cite{bocklisch2017rasa} chat bot. A Sinhala chat bot for train information was proposed by~\citet{harshani2021sinhala}. Similarly, the tool proposed by~\citet{chandrasena2021sinhala} serves as a chat bot-based recommendation system for Sri Lankan traditional dancers. The chat bot discussed by~\citet{kumanayake2021sinhala} has the very specific purpose of answering user inquiries about the degree programs at University of Ruhuna. 
\citet{avishka2022novel,avishka2023towards} used off-the-shelf RASA NLU Engine~\cite{bocklisch2017rasa} and Microsoft Bot Network~\cite{biswas2018microsoft} to set up a generic Sinhala chat bot architecture. They demonstrated the effectiveness of their architecture by creating a food ordering chat bot. A web-based code-less chat bot development platform for Sinhala was proposed by~\citet{dissanayake2022enhancing}. Further, they claimed that their system can handle OOV tokens as well as Sinhala-English code-switching.
The work by~\citet{dasanayaka2020contextual} used a deep learning Intent Mapping (IM) model to map consumer responses in their Sinhala chat bot framework.~\citet{rajapakshe2020sinhala} proposed a Sinhala conversational interface for the purpose of scheduling medical appointments and giving medical advice. The chat bot component was in the back end while the front end contained speech recognition and voice synthesizer components.
The end product of work by~\citet{dasanayaka2025multimodal} is a Sinhala chatbot that can answer patient questions on orthopantomography radiology reports which were previously generated by the system itself by analysing dental panoramic tomography (DPT) images.
\citet{kandamby2025cloud} introduced \texttt{Cloud Era}, a multilingual chatbot utilising \texttt{LightRAG}~\cite{guo2024lightrag} for agentic behaviour in multi-cloud setups. They claim to have achieved 81\% recall for Sinhala.
The work by~\citet{weerakoon2025sinhala} set out to build a Sinhala speech-to-speech chatbot. They select \texttt{Wav2Vec2-BERT}~\cite{barrault2023seamless} for ASR and utilise a RAG approach for the chatbot. The TTS component is handled with the \gh{https://github.com/jaywalnut310/vits}{VITS}~\cite{kim2021conditional} model. 

\subsection{Translators}
A meta-study on the viability of machine translators replacing English to Sinhala human translators was conducted by~\citet{dilshani2019study}. However, this study only involves 100 combined and complex English sentences translated to Sinhala by human translators as well as MT software. Given that reason and the fact that they seem to only used Google translate and \textit{Akura} Sinhala dictionary app for comparison, the conclusions of this study may not be generalized.
Another meta-study on the impact of pre-trained multilingual sequence-to-sequence models on low-resource language translation has been conducted by~\citet{lee2022pre}; while they consider Sinhala as one of the examples, they do not go much into the specific impact due to the general nature of the paper.
The meta-analysis by~\citet{ramadasa2022analysis} attempts to evaluate the goodness of the Google Sinhala-to-English translation by using the Google cloud API to translate Sinhala to English and then analysing the accuracy of the \textit{Sentiment Analysis} task and the \textit{Named Entity Recognition} task on the translated text. 
The meta-analysis by~\citet{das2023statistical} compared the results of translating English to 15 Indic languages including Sinhala using statistical translation methods. They used datasets from OPUS~\cite{tiedemann2012parallel} for model building and utilized Flores-200 for fine-tuning.
The NMT for Indic languages study conducted by~\citet{sheshadri2023voyage} discusses the Sinhala language translation in the abstract and conclusion but the paper itself focuses more on the languages spoken in India. Nevertheless, it puts Sinhala into a regional linguistic perspective.
The meta-study by~\citet{bapna2022building} discusses the task of building clean, web-mined datasets for a number of languages including Sinhala for the task of machine translation. This discussion was continued by~\citet{jones2023bilex} who discussed the bilingual lexica (BILEXs) in the context of a number of languages including Sinhala.
Yet another meta-analysis by~\citet{halpege2024google} compared the commercially available English-Sinhala translation systems provided by Google and Bing through a comprehensive error analysis. This study used academic articles as its domain and ultimately concluded that both translation systems need significant improvements before they can be recommended for widespread usage.
The survey by~\citet{tafa2025machine} compared the performance of a number of MT models on Low-Resource languages, including Sinhala.
\citet{de2024comparative} have also conducted another shallow study on rule-based, statistical, neural, and hybrid MT models for Low-Resource languages.
The short survey by~\citet{dias2025systematic} discusses translation and transliteration of few code-mixed Indo-Aryan languages which includes a major focus on Sinhala.

\subsubsection{Parallel Corpora}
\label{sec:ParCorp}
The open parallel corpus (\textit{OPUS})\footURL{https://opus.nlpl.eu/} by~\citet{tiedemann2012parallel} curates the largest collection of parallel data between Sinhala and other languages.~\citet{zhang2020improving} sampled data from \textit{OPUS} for 100 languages (including Sinhala) and created the \gh{https://github.com/EdinburghNLP/opus-100-corpus}{OPUS100} data set as a benchmark for neural machine translation (NMT). Their \gh{https://github.com/bzhangGo/zero}{code} is also available. 
\citet{ranathungaen2026sita} introduced \texttt{EnSiTa}, with over 200,000 human-annotated parallel sentence pairs involving Sinhala, making it the largest human-annotated parallel corpus for Sinhala. The data spans 7 domains and supports English–Sinhala and Sinhala–Tamil, alongside English–Tamil, with additional manually translated test data from an 8th domain. It benchmarks Sinhala MT across in-domain, cross-domain, and multi-domain settings, providing quantitative baselines for future Sinhala MT research.
Further, \citet{tiedemann2021development} created an NMT benchmark named \gh{https://github.com/Helsinki-NLP/Tatoeba-Challenge}{Tatoeba MT Challenge} using OPUS data based on Tatoeba\footURL{https://tatoeba.org/}. The English-Sinhala data set can also be directly downloaded from a link on \gh{https://github.com/Helsinki-NLP/Tatoeba-Challenge/tree/master/data}{GitHub}.
A benchmark Sinhala-English translation data set named \textit{FLORES}\footURL{https://tinyurl.com/flores200dataset} was created by~\citet{guzman2019flores}. This includes a 600k+ Sinhala-English \opus{http://bit.ly/2KsFQxm}{subtitle pairs} initially collected by~\cite{lison2016opensubtitles2016}, 45k+ Sinhala-English sentence pairs from \opus{http://bit.ly/2Z8q0fo}{GNOME}, \opus{http://bit.ly/2WLY6bI}{KDE}, and \opus{http://bit.ly/2wLVZGt}{Ubuntu}. They further provided two monolingual corpora for Sinhala. Those were a 155k+ sentences of filtered \opus{http://bit.ly/2EQZ7oM}{Sinhala Wikipedia} and 5178k+ sentences of Sinhala \opus{http://bit.ly/2ZaQFZo}{common crawl}. In addition to the data set, they have also made their \gh{https://github.com/facebookresearch/flores}{code} publicly available. This work was then extended by~\citet{costa2022no}, of which the data and code are available under the \gh{https://github.com/facebookresearch/fairseq/tree/nllb}{No Language Left Behind (NLLB)} project. 
\citet{sofalas2026sinfos} introduced \hf{https://huggingface.co/datasets/SloppyCalculator/SinFoS}{SinFoS}, a parallel data set of 2,344 Sinhala figures of
speech including proverbs, idioms, adages, idiosyncratics, and other sayings.
\citet{pushpananda2024tamsipara} claims to have created a manually annotated parallel corpus named \texttt{TamSiPara} with 25k parallel sentences in the Si-Ta direction and 22k parallel sentences in the Ta-Si direction. But it is not freely and openly accessible. 
\citet{burchell2025expanded} introduced \hf{https://huggingface.co/datasets/HPLT/HPLT2.0_cleaned}{HPLT v2} a multilingual monolingual and parallel corpora that includes 273k Sinhala Sentences. These sentences were extracted from the Internet Archive(IA) and the Common Crawl (CC). They have also publicly released their \hf{https://huggingface.co/collections/HPLT/hplt-20-uni-direction-translation-models-67f2fc7ae54845f9b182957a}{translation models}. \citet{ranathunga2024quality} have \gh{https://github.com/nlpcuom/quality-matters}{publicly released} their data, code, and models for Sinhala-English translation. Very importantly, this includes a human-cleaned portion of the NLLB dataset~\cite{costa2022no} in \hf{https://huggingface.co/datasets/NLPC-UOM/nllb-top25k-ensi-cleaned}{En-Si}.
\citet{fernando2022exploiting} has created aligned corpora for Sinhala–English, Tamil–English, and Sinhala–Tamil language pairs. The \gh{https://github.com/kdissa/comparable-corpus} corpora as well as \gh{https://github.com/nlpcuom/parallel_corpus_mining}{code} for their document and sentence alignment task, are available. A parallel corpus of Sinhala and English was collected by~\citet{banon2020paracrawl} containing 217,407 sentences and available to download from their website\footURL{https://www.paracrawl.eu/}. However, the later audit by~\citet{caswell2021quality} raised issues on the quality of that data set. A \gh{https://github.com/kdissa/comparable-corpus}{parallel corpus} of aligned Sinhala-English documents and sentences obtained from crawling the web was released by~\citet{sachintha2021exploiting}.
As for Sinhala-Tamil corpora,~\citet{hameed2016automatic} claim to have built a sentence-aligned Sinhala-Tamil parallel corpus and~\citet{mohamed2017automatic} claim to have built a word-aligned Sinhala-Tamil parallel corpus. However, at the time of writing this paper, neither of them was publicly available. A very small \gh{https://github.com/nlpc-uom/Sinhala-Tamil-Aligned-Parallel-Corpus}{Sinhala-Tamil aligned parallel corpus} created by~\citet{farhath2018improving} using order papers of government of Sri Lanka is available to download. \citet{vasantharajan2021tamizhi} used Printed Character Recognition (PCR) to create a \gh{https://github.com/Chaarangan/tamizhinet-corpus}{Tamil-Sinhala-English parallel corpus}. Their follow-up work~\cite{vasantharajan2022adapting} on adapting the \textit{Tesseract} engine to handle non-Unicode (legacy fonts) in pdf documents resulted in an improved version of the corpus, which also can be found on \gh{https://github.com/aaivu/Tamizhi-Net-OCR/tree/master/corpus/parallel corpus}{GitHub}.  

\subsubsection{Translation Quality studies}
As mentioned in Section~\ref{sec:corpora}, the study by~\cite{caswell2021quality} raised questions on the quality of the existing Sinhala-English corpora. In a follow-up study by~\citet{ranathunga2024quality}, it was pointed out that using an automated ranking based on a similarity measure on web-mined corpora and using the resultant top samples can yield better translation models for English-Sinhala and Sinhala-Tamil. These models sometimes were better performing than a model trained on the full dataset. They further showed that using human labour to clean web-mined parallel corpora only gives marginal benefits over automated ranking and filtering. Thus they concluded that using expensive human labour for this task might not be efficient.  The study by~\citet{nllb2024scaling} conducted a study on the quality of the translator models trained on their NLLB data set~\cite{costa2022no}. Naturally, this includes Sinhala. 
\citet{sen2019parallel} on the other hand attempted to improve the quality of the Sinhala-English parallel corpora using fuzzy string matching where they tried to match the English translation of the given Sinhala sentence to the English sentence in the dataset pair. 
\citet{alastruey2026omnilingual} introduce \hf{https://huggingface.co/datasets/facebook/bouquet}{BOUQuET}, a translation benchmark. However, Sinhala is not included in the 8 handcrafted languages; it is only included under the 260+ machine-translated languages. They do welcome annotators to donate their labour\footURL{https://bouquet.metademolab.com/} to change this.
\citet{abubacar2026parameter} claimed to have created a \hf{https://huggingface.co/surrey-nlp/Low-resource-QE-DA-dataset}{quality estimation dataset} that includes 7k Sinhala-English pairs and \gh{https://github.com/surrey-nlp/TRMQE}{associated code}, but neither of the links currently work.

A study on the viability of using  Google Translate for the legal domain English-Sinhala and Sinhala-English translation was conducted by~\cite {ubhayawardhana2023study}. They used human experts to extensively analyse the end result of the translation with many concrete examples of legal phrases that needed to be translated. 
The multilingual lexicon for low-resource machine translation dataset \textit{GATITOS} introduced by~\citet{jones2023gatitos} does not include Sinhala among their 170 languages set. However, they use Sinhala data from other sources and include several observations for Sinhala in this task. They observe that \textit{PanLex}~\cite{kamholz2014panlex} and \textit{GATITOS} improve results in the \EnSi{} direction for smaller models (Transformer
Big, 475M) but weaken the results in the \SiEn{} direction. For bigger models (Transformer
1.6B), they do not observe an improvement or degradation of \EnSi{} or \SiEn{} direction. 

\subsubsection{Sinhala-English Non-NMT}
A series of work has been done by a group towards English to Sinhala translation as mentioned in some of the above subsections. This work includes; building a morphological analyzer~\cite{hettige2006morphological}, lexicon databases~\cite{hettige2007developing}, a transliteration system~\cite{hettige2007transliteration}, an evaluation model~\cite{hettige2010evaluation}, a computational model of grammar~\cite{hettige2011computational}, and a multi-agent solution~\cite{hettige2016multi}. After working on human-assisted machine translation~\cite{hettige2007using}, ~\citet{hettige2009theoretical,hettige2010varanageema} have attempted to establish a theoretical basics for English to Sinhala machine translation. A very simplistic web based translator was proposed~\cite{hettige2008web,hettige2008web1}. The same group have worked on a Sinhala ontology generator for the purpose of machine translation~\cite{hettige2014sinhala} and a phrase level translator~\cite{hettige2017phrase} based on the previous work on a multi-agent system for translation~\cite{hettige2013masmt}. Further, an application of the English to Sinhala translator on the use case of selected text for reading was implemented~\cite{hettige2013selected}. They later continued their work on multi-agent English to Sinhala translation with the AGR organizational model~\cite{hettige2021masmt4}.

Another group independently attempted English-to-Sinhala machine translation~\cite{liyanapathirana2011english} with a statistical approach~\cite{liyanapathirana2013statistical}.~\citet{wijerathna2012translator} and~\citet{de2008sinhala} have proposed simple rule based translators. An example-based method applied on the English-Sinhala sentence aligned government domain corpus was proposed by~\citet{silva2008example}. A translator based on a look-up system was proposed by~\citet{vidanaralage2018sinhala}. In a preprint,~\citet{joseph2019evolutionary} proposes an evolutionary algorithm for Sinhala to English translation with a basis of Point-wise Mutual Information (PMI) and claims that the code will be shared once the paper is accepted. However, they do not report any quantitative results to be compared and the reported qualitative results are also superficial.~\citet{pushpananda2015statistical} utilized statistical machine translation to translate between Sinhala and Tamil.~\citet{fernando2020data} tries to solve the Out of vocabulary (OOV) problem for Sinhala in the context of Sinhala-English-Tamil statistical machine translation. The approach proposed by~\citet{rajitha2020sinhala} uses statistical machine translation and transliteration to align Sinhala and English documents. 

\subsubsection{Sinhala-English NMT}
~\citet{fonseka2020english} used Byte Pair Encoding (BPE) for English to Sinhala neural machine translation. As another solution to the OOV problem, an analysis of subword techniques to improve English to Sinhala Neural Machine Translation (NMT) was conducted by~\citet{naranpanawa2020analyzing}. A data augmentation method to expand bilingual lexicon terms based on case markers for the purpose of solving the OOV problem in the domain of NMT was proposed by~\citet{fernando2021data} which they later extended further~\cite{fernando2022data}.~\citet{epaliyana2021improving} proposed iterative filtering and data selection be used to improve Sinhala-English NMT.~\citet{perera2022improving} used English Part-of-Speech (PoS) tags to improve English to Sinhala NMT.~\citet{linimprovement} used a model based on \textit{fairseq}~\cite{ott2019fairseq} to improve machine translation between English and Sinhala. The second half of the study by~\citet{arafath2020polylingo} dealt with translating Sinhala speech to other languages. \citet{kugathasan2021neural} proposed an NMT system for Sinhala-English Code-Mixed text using the standardized Sinhala Code-Mixed text they proposed earlier~\cite{kugathasan2020standardizing}. Later, they published a slightly different version of the work at a different venue but under the same title~\cite{kugathasan2022neural}.
\citet{nguyen2021contrastive} introduced a new fine-tuning objective \textit{LAgSwAV} (Language-Agnostic Constraint for SwAV loss), using which they obtained 5.4 BLEU for English-Sinhala. Later, this method was further discussed by~\citet{nguyen2023improving}. 
\citet{attigala2023effectiveness} conducted an analysis on the effectiveness of ChatGPT~\cite{brown2020language} in translating Sinhala songs to English. 
\citet{utsa2024neural} used transfer learning and back translation as well as focal loss of the Sinhala-English dataset from FLoRes~\cite{guzman2019flores} and reported better results than mBART~\cite{liu2020multilingual}.
The study by~\citet{ranathungaa2024exploiting} presents an evaluation of using parallel data from auxiliary domains to enhance EnSI NMT by fine-tuning or further pre-training the models. 
In their study on improving cross-lingual representation of multilingual
language models for low-resource languages using linguistic entity masking, \citet{fernando2025linguistic} used \texttt{SiTa} and \texttt{EnSi} NMT on NLLB and \textit{CCAligned} as one of the evaluations. 
\citet{muscat2026contrastmt}, when introducing \texttt{ContrastMT}, their pivot language augmentation method that uses Hindi as the pivot, tested English to Sinhala as one of the translation pairs and reported a BLEU score of 24.61 on 48K pairs from Flores.
\citet{thillainathan2025beyond} proposed to use continual pre-training and intermediate task transfer learning for fine-tuning what they call \textit{multilingual sequence-to-sequence large language models (msLLMs)}. However, in actuality, what their experiments are on is mBART. 
The work by~\citet{zebaze2025compositional} introduces a new NMT pardigm that they call \texttt{CompTra} (compositional translation) which is based on LLMs. They report the results for Sinhala using the FLORES-200~\cite{costa2022no} data set, where \texttt{CompTra} applied on NLLB~\cite{costa2022no} gives the best result. When introducing \texttt{ALOPE}, a translation quality optimisation study for LLMS, \citet{sindhujan2025alope} used Sinhala-to-English translation for their results. However, they achieved a better  Spearman correlation score when they tested the same with \texttt{CometKiwi}~\cite{rei2023scaling}. Specialised Dictionary-supported LLMs are used to translate agricultural Packages of Practices (PoPs) into Sinhala in the work of~\citet{wilson2025enhancing}. The work by~\citet{okadini2026resource} tested Llama 3 and Llama 4 against NLLB-200 600M on the Sinhala-Tamil and Tamil-Sinhala translation task under multiple prompting strategies.

\citet{fernando2025improving} in their meta-study on low-resource language NMT, introduced debiasing heuristics for web mined parallel corpora that improve \texttt{EnSi} translation quality. They showed the improvements on \textit{CCMatrix} and \textit{CCAligned} datasets. The meta study by \citet{velayuthan2025encoder} proposes an improved sequence level Knowledge Distillation method that improves encoder knowledge transfer. They used Sinhala-English NMT to discuss the $\alpha$ value ablation by using \textit{CCAligned}~\cite{el2019ccaligned}, \textit{OpenSubtitles}~\cite{lison2016opensubtitles2016}, and \textit{SITA (Gov)}~\cite{fernando2020data} corpora.    
Sinhala is one of the languages considered in the short survey by~\citet{perera2025machine} on machine translation and transliteration for Indo-Aryan Languages. They mention a sub-set of works~\cite{joseph2019evolutionary,fonseka2020english,ranathunga2018si,rajitha2020sinhala,hisan2020cross,sandaruwan2021neural} discussed in this survey for EnSi translation.

\subsubsection{Singlish to Sinhala (Transliteration)}
The \gh{https://github.com/google-research/xtreme-up}{XTREME-UP} data set created by~\citet{ruder2023xtremeup} contains a Sinhala data set for the transliteration task. \citet{perera2025evaluating} introduced a \gh{https://github.com/Sameera2001Perera/Romanized-Sinhala-Transliteration-Disambiguation-Dataset}{data set} to disambiguate Sinhala transliteration. The data set contains 600 sentences across 22 ambiguous Romanised Sinhala words. The \gh{https://github.com/google-research-datasets/dakshina}{Dakshina} dataset by~\cite{roark2020processing} contains Native Sinhala script data (200k train and 28k validation sentences) along with Romanised (Singlish) data (10k sentences) from Wikipedia. It also contains a Romanisation Lexicon (25k train and 5k test).

The early work by~\citet{goonetilleke2008srishell} attempted Sinhala transliteration through the Latin alphabet. However, their work does not use the word \textit{transliteration} and instead focuses on the predictive aspect.~\citet{priyadarshani2019statistical} used statistical machine learning for transliteration of names between Sinhala, Tamil, and English. A rule-based system on trigrams was proposed by~\citet{liwera2020combination} for Singlish to Sinhala transliteration of social media text. A Singlish to Sinhala converter which uses an LSTM was proposed by~\citet{de2020singlish}. A rule-based approach for the same was proposed by~\citet{de2021singlish}. The study by~\citet{nanayakkara2022context} introduced an English-to-Sinhala transliteration system.

\citet{athukorala2022swa} proposed a fuzzy logic-based Sinhala transliteration system named \textit{Swa-Bhasha}. \citet{sumanathilaka2023sinhala,sumanathilaka2023swa} then proposed a Trie~\cite{bodon2003trie} data structure-based algorithm for word suggestion for the \textit{Swa-Bhasha} system. An extended analysis of the same work was presented in a later work~\cite{sumanathilaka2023romanized}. A further extension, \textit{Swa-Bhasha 2.0}, by~\citet{dharmasiri2024swa} used NMT to eliminate word selection ambiguity in the transliteration process. Subsequently, \citet{sumanathilaka2024swa} have released  \hf{https://huggingface.co/datasets/deshanksuman/SwaBhasha_Transliteration_Sinhala}{Swa-Bhasha Sentence (En-Si)}, \hf{https://huggingface.co/datasets/deshanksuman/Augmented_SinhalatoRomanizedSinhala_Dataset}{Swa-Bhasha Sentence (Si-En)}, and \hf{https://huggingface.co/datasets/deshanksuman/Swabhasha_RomanizedSinhala_Dataset}{Swa-Bhasha Word} datasets. \texttt{Swa-bhasha Resource Hub}~\cite{sumanathilaka2025swa} was then created by putting the previous work together for easy access, which also contains \hf{https://huggingface.co/deshanksuman/swabhashambart50SinhalaTransliteration}{a trained BART model}. Continuing this work, \citet{dias2026sinmix2mono} claim to introduce a dataset named \texttt{SinMix2Mono}; however, none of the supposed dataset links in the paper is clickable.

\citet{amarasekara2023developing} proposed a rule-based method supported by N-gram analysis and a corpus dataset to transliterate Singlish tweets to Sinhala.
The study by~\citet{rajapaksha2023sinhala} claims to have created a translation system, however, what they have created is a transliteration system between Singlish and Sinhala. While they also claim to have trained ASR and TTS systems, the paper does not identify them by name or citation. The same is true for the data sets they have used to train the various modules in their pipeline.  
The studies by~\citet{kumaravithana2023sinhala} and~\citet{jayawardhana2023bridgetalk} claim to have built Sinhala-English translators but in actuality are just invoking Google API for their translation tasks.

The work by~\citet{kirov2024context} used the \textit{Dakshina}~\cite{roark2020processing} dataset and compared the accuracy of Singlish to Sinhala transliteration with vanilla LSTM, vanilla Transformers, mT5~\cite{xue2020mt5}, ByT5~\cite{xue2022byt5}, and a non-neural finite-state transducer (FST) based on work by~\citet{bisani2008joint}. They discuss in detail how the zero-width joiner (\textit{ZWJ}, \verb|U+200D|) character cause issues in this task and Sinhala rendering as a whole. Their code is available on \gh{https://github.com/google-research/google-research/tree/master/context_aware_transliteration}{GitHub}.
\citet{khiu2024predicting} fine-tuned mBART on Sinhala Government Corpus and Bible corpus to predict the NMT performance dependence on domain similarity.    
\citet{de2024sinhala2} conducted a comparative analysis for Sinhala transliteration between an extension of the rule-based approach proposed by~\citet{tennage2018transliteration} and the  transliteration as a translation model proposed by~\citet{deselaers2009deep}. They reported across-the-board better performance with the latter. All their code as well as the test-sets are \gh{https://github.com/kasunw22/Sinhala-Transliterator/}{publicly available}. \citet{perera2025indonlp} attempted the same task with BERT. Their \gh{https://github.com/Sameera2001Perera/Singlish-Transliterator}{code} is also publicly available. \citet{rajapakse2026script} evaluated multiple LMs on the task of Sinhala transliteration on a dataset of their own, which contains 1500 sentences.

In their short survey discussing Indo-Aryan Languages machine translation and transliteration,~\citet{perera2025machine} covers a number of works~\cite{vidanaralage2018sinhala,tennage2018transliteration,priyadarshani2019statistical,liwera2020combination,de2021singlish,athukorala2022swa,nanayakkara2022context,sumanathilaka2023swa,dharmasiri2024swa,sumanathilaka2024swa} on Sinhala transliteration which we have already discussed above. The survey by~\citet{deshpande2025bridging} covers the works by~\citet{perera2025indonlp} and \citet{de2024sinhala2} for Sinhala transliteration, which we have discussed above.

\subsubsection{Singlish and English (Translation via Transliteration)}
An LSTM-based sequence-to-sequence model was used by~\citet{sandaruwan2021neural} for Singlish to English NMT task. 
\citet{nalinka2023shattering} used simple transformers with positional embedding to translate Singlish into English.
The study by~\citet{de2023art} claims to have achieved Singlish to English translation by simple stemming. 

\subsubsection{Sinhala to Singlish (Reverse-Transliteration)}

For their data set development, \citet{de2026sinhala} created a Sinhala to Singlish transliteration tool and have made it publicly available\footURL{https://nisansa.lk/?s=GUUANj}.

\subsubsection{Between Sinhala and Non-English Languages}
Most of the cross-Sinhala and Tamil work has been done in the domain of machine translation. A neural machine translation for Sinhala and Tamil languages was initiated by~\citet{tennage2017neural,tennage2017neural1}. Then they further enhanced it with transliteration and byte pair encoding~\cite{tennage2018transliteration} and used synthetic training data to handle the rare word problem~\cite{tennage2018handling}. This project produced \textit{Si-Ta}~\cite{ranathunga2018si} a machine translation system of Sinhala and Tamil official documents. In the statistical machine translation front,~\citet{farhath2018integration} worked on integrating bilingual lists. The attempts by~\citet{weerasinghe2003statistical} and~\citet{sripirakas2010statistical} were also focused on statistical machine translation while~\citet{jeyakaran2013novel} attempted a kernel regression method. A yet another attempt was made by~\citet{pushpananda2013towards} which they later extended with some quality improvements~\cite{pushpananda2014sinhala}. An attempt at real-time direct translation between Sinhala and Tamil was done by~\citet{rajpirathap2015real}.~\citet{dilshani2018linguistic} have done a study on the linguistic divergence of Sinhala and Tamil languages with respect to machine translation.~\citet{mokanarangan2019translation} claims to have built a named entity translator between Sinhala and Tamil for official government documents. But this work is locked behind an institutional repository wall and thus is not accessible by other researchers.~\citet{arukgoda2019improving} studied the possibility of using deep learning techniques to improve Sinhala-Tamil translation which they further improved later~\cite{arukgoda2021improving}.~\citet{pramodya2020comparison} compared Transformers, Recurrent Neural
Networks, and Statistical Machine Translation (SMT) in the context of Tamil to Sinhala machine translation. The work by~\citet{nissanka2020exploring} used monolingual word embedding to improve NMT between Sinhala and Tamil.~\citet{thillainathan2021fine} uses pre-trained mBART~\cite{liu2020multilingual} models for six directional translations between Sinhala, Tamil, and English. \citet{yashothara2023utility} discussed the use of the Hierarchical Phrase-Based Model for Tamil to Sinhala and Sinhala to Tamil translations. \citet{pramodya2023exploring} presented a comparison of SMT~\cite{pushpananda2015statistical} and NMT models for Sinhala-Tamil translation. For NMT they have used base transformer, turned transformer, and mT5~\cite{xue2020mt5}. 
The work by~\citet{su2024unlocking} which compared 8 Parameter-efficient fine-tuning (PEFT) methods on the Si-Ta language pair using the Government corpus~\cite{fernando2020data} and NLLB~\cite{costa2022no} reports that Houlsby adapter~\cite{houlsby2019parameter}, with a 33.34 SacreBLEU~\cite{post2018call} score, to be the best PEFT method. However, they also note that the Pfeiffer adapter~\cite{pfeiffer2020mad} runs the fastest at 52.59 hours with a reasonable score of 31.24. (Comparatively, the Houlsby adapter runs for 78.65 hours).
The work by~\citet{pramodya2024enhancing} conducted a comparative analysis of the available models for Sinhala-Tamil translation and reaffirmed the observations of~\citet{duh2020benchmarking} that, in low-resource scenarios, SMT and NMT both may work similarly, however, NMT needs more careful tuning to fit performance.
The short survey~\cite{perera2025machine} on Indo-Aryan Languages machine translation and transliteration, discusses a few works~\cite{ranathunga2018si,nissanka2020exploring,fonseka2020english,pramodya2020comparison,arukgoda2019improving,thillainathan2021fine} under SiTa translation which have already been discussed in this survey.

There have been attempts to link Sinhala NLP with Japanese by Herath et al.~\cite{herath1994practical,herath1996bunsetsu,herath1993generation},~\citet{thelijjagoda2004japanese},~\citet{thelijjagoda2007japanese}, and~\citet{kanduboda2011role}.
The translation component of the work of~\citet{rathnayaka2026brahmi} translates Brahmi inscriptions to Sinhala with an LSTM with Byte Pair Encoding (BPE).
\citet{jayasinghe2023analytical} discusses the importance of translation between Sinhala and the liturgical language of Buddhism, Pali~\cite{childers1875dictionary,salaville1938introduction,liddicoat1993choosing}. \citet{shalini2017dictionary} attempted to use a dictionary-based machine-translation method for this task. 
The study by~\citet{anuradha2020machine} uses machine translation on the unique application of converting Sinhala and English Braille documents, which they have run OCR on, into voice.
Natural Language Interfaces to Databases (NLIDB) where Sinhala natural language queries of a user is translated to SQL has been proposed~\cite{mahdi2025hybrid,refaideen2025sinhala}.


\subsection{Phonological Tools}
\label{Sinhala:Phonological}

On the case of phonological layer, a report on Sinhala phonetics and phonology was published by~\citet{wasala2005research}.~\citet{wickramasinghe2007practical} discussed the practical issues in developing Sinhala Text-to-Speech and Speech Recognition systems. 
\textit{The Massively Multilingual Speech} (\gh{https://github.com/facebookresearch/fairseq/tree/main/examples/mms}{MMS}) data set created by~\citet{pratap2023mms} has Sinhala data for the Spoken Language Identification (LID) task. The meta-study by~\citet{al2022speech} compared the status of Sinhala speech recognition research against 17 other languages. 
A \gh{https://github.com/pnfo/sinhala-tts-dataset}{text-to-speech data set} with 3300 Sinhala sentences with 7.5 hours of recordings was released by the \textit{Path Nirvana Foundation}.
\citet{buddhika2018voicer} introduce a  crowd sourcing tool that they named \textit{Voicer} to collect speech data. They claim that the tool is open source and that they have created a Sinhala Speech corpus of 10 hours with 39 different sentences in the banking domain. Neither the code nor the data is publicly linked in the research paper. However, a subsequent work by~\citet{hellarawa2022domain} uses this data set. Thus, it can be assumed that this data set may be available if the authors are contacted. 
The study by~\citet{warusawithana2022systematic} created a \gh{https://github.com/SinSpeech-Development/Refined-OpenSLR-52-Corpus}{
refined version} of the OpenSLR-52 speech corpus for Sinhala\footURL{https://openslr.org/52} by~\citet{kjartansson-etal-sltu2018}\footnote{A later further overview on the corpus is also available~\cite{butryna2020google}.}.

\subsubsection{Text-to-Speech (TTS)}
Based on the earlier work by~\citet{weerasinghe2005rule},~\citet{wasala2006sinhala} have developed methods for Sinhala grapheme-to-phoneme conversion along with a set of rules for schwa epenthesis. This work was then extended by~\citet{nadungodagesinhala}.~\citet{weerasinghe2007festival} developed a Sinhala text-to-speech system. However, it is not publicly accessible.
They internally extended it to create a system capable of helping a mute person achieve synthesized real-time interactive voice communication in Sinhala~\cite{amarasekara2013real}. A rule based approach for automatic segmentation of a small set of Sinhala text into syllables was proposed by~\citet{kumara2007automatic}. An \textit{ew prosodic phrasing} method to help with Sinhala Text-to-Speech process was proposed by~\citet{bandara2017ew,dias2009sinhala,bandara2013new}.~\citet{sodimana2018text} proposed a text normalization methodology for Sinhala text-to-speech systems. Further,~\citet{sodimana2018step} formalized a step-by-step process for building text-to-speech voices for Sinhala. Both~\citet{jayamanna2014android} and~\citet{mishangi2018android} have created Sinhala document readers for visually impaired persons to be used on Android devices. An OCR and  Text-to-Speech system for Sinhala named Bhashitha was proposed by~\citet{de2018project}. The works by~\citet{lakmal2021adapting} and~\citet{senarathna2022step} adapted MaryTTS~\cite{schroder2003german} to synthesize Sinhala speech. The study by~\citet{jayawardhana2019intelligent} used \textit{Deep Voice}~\cite{ping2017deep} for Sinhala and English TTS.~\citet{gamage2020dnn} included a Sinhala text-to-speech module as one of the three modules present in their currency recognition system. The study by~\citet{madhusha2023mobile} claims to have created a mobile app with Sinhala Text-to-Speech and OCR to read books for visually impaired students. \citet{praveen2024machine} claim to have created a Sinhala TTS system with a custom neural network that archives 98\% accuracy. The work by~\citet{senarath2024enhancing} discusses using existing TTS systems for Sinhala. They especially highlight \texttt{VAENAR-TTS}~\cite{lu2021vaenar} as the best candidate for this task. \citet{nayanathara2025enhancing}, on the other hand, propose to use the \gh{https://github.com/jaywalnut310/vits}{VITS} architecture~\cite{kim2021conditional} for the same task. \citet{botheju2025multilingual} used \texttt{SpeechT5}~\cite{ao2022speecht5} model with \texttt{X-vector} speaker embeddings~\cite{snyder2018x} for Sinhala Personalized voice synthesis and accent adaptation.
~\citet{anuradha2020machine} proposed a machine translation system to convert Sinhala and English Braille documents into voice.
A separate group has done work on  Sinhala text-to-speech systems independent to above~\cite{nanayakkarahuman}. 

\subsubsection{Speech-to-Text (ASR)}
On the converse,~\citet{nadungodagespeech} have done a series of work on Sinhala speech recognition with special notice given to Sinhala being a resource-poor language. This project divides its focus on: continuity~\cite{nadungodage2011continuous}, active learning~\cite{nadungodage2013efficient}, and speaker adaptation~\cite{nadungodage2015speaker}. A Sinhala speech recognition for voice dialling, which is speaker independent, was proposed by~\citet{amarasingha2012speaker}, and on the other end, a Sinhala speech recognition methodology for interactive voice response systems, which are accessed through mobile phones, was proposed by~\citet{manamperi2018sinhala}. A Sinhala speech to Unicode text converter for the disaster relief domain was proposed by~\citet{prasangini2018sinhala}.~\citet{priyadarshani2012speaker} proposes a method for speaker-dependent speech recognition based on their previous work on: dynamic time warping for recognising isolated Sinhala words~\cite{priyadarshani2012dynamic}, genetic algorithms~\cite{priyadarshani2012genetic}, and a syllable segmentation method utilising acoustic envelopes~\cite{priyadarshani2011automatic}. The method proposed by~\citet{gunasekara2015real} utilises an HMM model for Sinhala speech-to-text. A Sinhala speech recogniser supporting bi-directional conversion between Unicode Sinhala and phonetic English was proposed by~\citet{punchimudiyanse2015unicode}. The work by~\citet{karunanayake2019transfer} transfer-learns CNNs for transcribing free-form Sinhala and Tamil speech data sets for the purpose of classification.~\citet{dilshan2018transcribing} conducted a study for the specific use case of transcribing number sequences in continuous Sinhala speech.
~\citet{gamage2020usage} explored the use of combinational acoustic models such as Deep Neural Network - Hidden Markov Model (DNN-HMM)~\cite{giuliani2015large} and Subspace Gaussian Mixture Model (SGMM)~\cite{giuliani2015large} in Sinhala speech recognition. In the follow-up work,~\citet{gamage2021improve} extended that work with an end-to-end Lattice-Free Maximum Mutual Information  (e2e LF-MMI) model~\cite{manohar2018semi}, which is claimed to be a viable solution for low-resource language speech recognition by~\citet{carmantini2019untranscribed}. However, it was shown that the new model slightly underperforms compared to the state-of-the-art result. Later, they conducted further development on their model in a follow-up work~\cite{gamage2024applicability}. 
~\citet{karunathilaka2020low} explore Sinhala speech recognition using deep learning models such as pre-trained DNN, DNN, TDNN, TDNN+LSTM. The first half of the study by~\citet{arafath2020polylingo} dealt with recognising Sinhala speech using LSTMs.~\citet{gamage2020dnn} included a Sinhala speech recognition module as one of the three modules present in their currency recognition system. Time-delay neural network architectures (including multistream CNN architectures) were used for acoustic modelling of Sinhala Automatic Speech Recognition (ASR) by~\citet{warusawithana2022enhanced}. They have used the \textit{Kaldi speech recognition toolkit}~\citep{povey2011kaldi} for training the ASR models. As part of their child [sic] cognitive ability assessment model,~\citet{kahawanugoda2022development}, proposed a Sinhala speech recognition system. The study by~\citet{azir2021sinhala} attempts to identify number sequences spoken in Sinhala.
\textit{TacoSi} introduced by~\citet{arachchige2023tacosi}, is based on \textit{Tacotron}~\cite{wang2017tacotron} and has been evaluated with 10 human evaluators to determine its text-to-speech quality.
\citet{nanayakkara2023exploring,nanayakkara2024exploring} used \gh{https://github.com/mozilla/DeepSpeech}{DeepSpeech} by Mozilla for Sinhala speech recognition. 
\citet{gunarathne2017sinhala} used an earlier version of the CMUSphinx toolkit\footURL{https://cmusphinx.github.io/} to transcribe Sinhala speech to text. The later work by~\citet{akesh2023real} also used CMUSphinx toolkit on features extracted from Mel-frequency cepstral coefficients (MFCC)~\cite{bridle1974experimental,mermelstein1976distance} to automatically generate Sinhala subtitles from Speech. 
The work by~\citet{wickramaarachchi2024automatic} also used MFCC features but focused on Sinhala speech intonation recognition for the purpose of detecting speech impediments in young children. 
\citet{dissanayaka2024voice} have introduced \textit{Word Sri} an application that is voice-activated and capable of Sinhala grammar checking and plagiarism checking.
\citet{thayasivam2025sita} have introduced a \gh{https://github.com/SiTa-SpeakerDiarization/SiTa}{Sinhala text-to-speech} dataset transcribed from 60 videos with a total of 602 minutes. As the baseline, they report the lowest  Diarization Error Rate (DER) of 6.0 for Powerset Cross-Entropy Diarization~\cite{plaquet2023powerset}. While the data set link points to a GitHub repository, it does not contain the data set; it points to a project website\footURL{https://sita-speakerdiarization.github.io/} which then points to a download page\footURL{https://rtuthaya.staff.uom.lk/resources/dataset/44} where you need to send a request to the authors to obtain the data set.      
\citet{omnilingual2025omnilingual} includes \hf{https://huggingface.co/datasets/facebook/omnilingual-asr-corpus/tree/main/data/sin_Sinh}{Sinhala} among the 1600+ languages for which they have created a \gh{https://github.com/facebookresearch/omnilingual-asr}{multilingual ASR system}. \citet{thameera2026developing} also fine-tuned Whisper for Sinhala ASR. They have used multiple augmentation techniques to increase the size of their dataset from 3078 clips (5.83 hours) to 12312 clips (23.32 hours). \citet{harischandra2025pre} tested Whisper, Wav2Vec2-XLSR, and Wav2Vec2-BERT for the Sinhala ASR task and reported that Wav2Vec2-BERT obtains the best results with a Word Error Rate (WER) of 1.79\% and a Character Error Rate (CER) of 0.33\%. \citet{prabhashwara2026real} implemented an ASR system with Wav2Vec2-XLSR-300M and a 5-gram KenLM~\cite{heafield2011kenlm} for phonetic error correction for the specific task of transcribing Sinhala lectures in the political science domain. \citet{javed2022towards} tested the Sinhala dataset created by~\citet{kjartansson-etal-sltu2018} on their model and reported a WER of 18.6.

\subsubsection{Speech-to-Speech}
\citet{layansan2015android} created a speech-to-speech translation system for Sinhala on the Android platform. The system developed by~\citet{rajapakshe2020sinhala} is also speech-to-speech in the sense that, it is a chatbot for scheduling medical appointments and giving medical advice where the front end contains speech recognition and voice synthesizer components that interfaced with a chatbot component in the back end. The work by~\citet{athas2024calltran} uses Google APIs (speech-to-text API, text-to-text-translator API, and text-to-speech synthesizer API) to achieve voice translation for Sinhala and Tamil. \citet{dilshani2025bridging} have conducted a survey on Speech-to-Speech systems available for Sinhala and Tamil translation.
In their speech-to-speech chatbot, \citet{weerakoon2025sinhala} uses \texttt{Wav2Vec2-BERT}~\cite{barrault2023seamless} for ASR and \gh{https://github.com/jaywalnut310/vits}{VITS}~\cite{kim2021conditional} model for TTS. The Text-to-Text chatbot component is the middle is handled by an LLM with RAG.

\subsubsection{Speech-to-Intent}
The work by~\citet{karunanayake2019sinhala} used English phoneme-Based Automatic Speech
Recognition (ASR) for intent identification in Sinhala and Tamil.~\citet{ignatius2021speaker} proposed a speaker-invariant speech-to-intent classification model with i-vector based speaker normalization, which was then evaluated on Sinhala, and Tamil speech intent data sets. The later work by~\citet{yadav2021intent} used pre-trained embeddings for Sinhala speech intent classification.
\citet{hellarawa2022domain} proposes a BiLSTM-based ASR system for intent classification which they have tested on the banking domain Sinhala speech dataset created by~\citet{buddhika2018voicer}. For this, they report an accuracy of 98.53\%. 

\subsubsection{Speech classification}
\citet{luger2025building} introduced a data set named \hf{https://huggingface.co/datasets/MLCommons/unsupervised_peoples_speech}{unsupervised\_peoples\_speech} that conatains 981 hours of Sinhala speech data.
The Sinhala speech classification system proposed by~\citet{buddhika2018domain} does so without converting the speech-to-text. However, they report that this approach only works for specific domains with well-defined limited vocabularies. The work by~\citet{dinushika2019speech} uses automatic speech recognition of Sinhala for speech command classification. Extending that,~\citet{kavmini2020improved} presented a Sinhala speech command classification system which can be used for downstream applications. The voice assistant system created by~\citet{senarathne2022automated} is capable of handling Sinhala voice commands. \citet{kathriarachchi2025child} developed a system to classify and predict Sinhala speech clips of children into age groups.   
The work by~\citet{welarathna2021automated} used CNNs to classify emotions (sad, disgust, surprise, neutral, happy, calm, fear, and angry) of Sinhala speech by Autistic children. \citet{sundarapperuma2023automatic} created a speech emotion detection system for Sinhala.

\subsubsection{Lip Synchronization}
The study by~\citet{weerathunga2020lip} worked on lip synchronization for Sinhala speech where videos of people speaking Sinhala were mapped to a visemes alphabet created by them. Further of this line of study,~\citet{wakkumbura2022phoneme} came up with Phoneme-Viseme mapping for Sinhala speech that they intended to be used for future applications of robotics. 

\subsubsection{Music to Notation}
The work by~\citet{dulmi20241d} has created a pipeline which converts audio recordings to western music notation, which is then converted to Sinhala musical notations using an API call to a Large Language Model with a custom-tailored prompt.

\subsubsection{Sinhala ASR as a Source}
A rather unique study by~\citet{ilyas2026sinhala} uses Sinhala ASR as the source for transfer learning to improve Dhivehi ASR. While, as shown in Fig~\ref{fig:Sinhala_Language_Tree}, Sinhala is very closely related to Dhivehi and thus transfer between the two languages should be possible, the lack of resources for both these languages so far discouraged any attempt at transfer learning between the two. But the success of this study is promising proof that the status of Sinhala NLP is improving. All the code for this study is \gh{https://github.com/lukmalilyas/From-Sinhala-to-Dhivehi-ASR}{publicly available}. 

\subsection{Optical Character Recognition Applications}
While it is not necessarily a component of the NLP stack shown in Fig~\ref{fig:nlpLayers}, which follows the definition by~\citet{liddy2001natural}, it is possible to swap out the bottom-most phonological layer of the stack in favour of an Optical Character Recognition (OCR) and text rendering layer. 
\citet{gunathilaka2025sinfund} introduced \gh{https://github.com/SriDoc/datasets}{SinOCR}, consisting of 100,000 images which include 1,135 \gd{https://drive.google.com/file/d/16X6lAzinZTIgiy6iw8Z0mLEyPyEwZJkL/view?usp=drive_link}{handwritten Sinhala texts} and texts printed in 200 \gd{https://drive.google.com/file/d/1uuPNN8M4jHLuZLF94CApkiG3kysDALia/view?usp=drive_link}{different Sinhala fonts}. They also released \gd{https://drive.google.com/file/d/1hfAfCJcZciEqMgO_wCuvTQ4pWI8t0L5D/view?usp=drive_link}{SinFUND}, a fully annotated dataset of 100 manually filled Sinhala forms.

\subsubsection{Printed Text}
The \gh{https://github.com/google-research/xtreme-up}{XTREME-UP} data set created by~\citet{ruder2023xtremeup} contains a Sinhala data set for the OCR task. The data was obtained from book transcriptions. 
\citet{dilhara2026cross} introduced \hf{https://huggingface.co/datasets/avishadilhara/sinhala-ocr-lk-acts-1010}{sinhala-ocr-lk-acts-1010} an annotated dataset of 1,010 page images and their transcriptions collected from Sri Lankan Legislative Acts published between 1981-1989 and 2000-2019.

An attempt for a Sinhala OCR system has been taken by~\citet{rajapakse1995neural} before any other work has been done on the topic. 
Much later, a linear symmetry-based approach was proposed by~\citet{premaratne2002recognition,premaratne2004segmentation}. They then used hidden Markov model-based optimization on the recognized Sinhala script~\cite{premaratne2006lexicon}. 
Similarly,~\citet{hewavitharana2002off} used hidden Markov models for off-line Sinhala character recognition.
\citet{herathresearch,herathresearch2} developed a prepossessing engine based on a template matching for printed Sinhala OCR.
Statistical approaches with histogram projections for Sinhala character recognition is proposed by~\citet{hewavitharana2002statistical}, by~\citet{ajward2010converting}, and by~\citet{madushanka2017sinhala}.
~\citet{karunanayaka2004off} also did off-line Sinhala character recognition with a use case for postal city name recognition. 
A separate group had attempted Sinhala OCR~\cite{weerasinghe2008nlp} mainly involving the nearest-neighbor method~\cite{weerasinghe2006nearest,weerasinghe2006knn}. 
A study by~\citet{ediriweera2012improviing} uses dictionaries to correct errors in Sinhala OCR. 
An early attempt for Sinhala OCR by~\citet{dias2013sinhala} has been extended to be online and made available to use via desktops~\cite{dias2013online} and hand-held devices~\cite{ranmuthugala2006online} with the ability to recognize handwriting. 
A simple neural network based approach for Sinhala OCR was utilized by~\citet{rimas2013optical}. A fuzzy-based model for identifying printed Sinhala characters was proposed by~\citet{gunarathna2014fuzzy}.~\citet{premachandra2016artificial} proposes a simple back-propagation artificial neural network with hand crafted features for Sinhala character recognition. Another neural network with specialized feature extraction for Sinhala character recognition was proposed by~\citet{naleer2016technique}. On the matter of neural networks and feature extraction, a feature selection process for Sinhala OCR was proposed by~\citet{kumara2016systematic}.
~\citet{jayawickrama2018letter} worked on Sinhala printed characters with special focus on handling diacritic vowels. However, they opted to refer to diacritic vowels as \textit{modifiers} in their work.~\citet{gunawardhana2018segmentation} proposed a limited approach to recognize Sinhala letters on Facebook images.
A CNN-based methodology to improve printed Sinhala character OCR was proposed by~\citet{liyanage2018improving}.
Printed Character Recognition (PCR) was used by~\citet{vasantharajan2021tamizhi} to create a large-scale Tamil-Sinhala-English parallel corpus.
A meta-study on the effects of text genre, image resolution, and algorithmic complexity needed for Sinhala OCR from books and newspapers was conducted by~\citet{anuradha2021estimating}. 
~\citet{anuradha2020deep} used \textit{Tesseract 3}\footURL{https://tesseract-ocr.github.io/}~\cite{smith2007overview} for Sinhala OCR. The later study by~\cite{balasooriya2021improving} improved the accuracy of \textit{Tesseract} OCR engine on Sinhala from 53.22\% to 86.16\% for the data set they tested on.
~\citet{maduranga2022multi} used an Artificial Neural Network (ANN) based on Universe of Discourse and Self-Organisation Map methods to recognise multi-style printed Sinhala characters. The study by~\citet{de2023ceylon}, even though ostensibly presented as translation research, is just invoking Google Cloud service for translation. Their novelty in Sinhala NLP lies in the facility provided to OCR the text from photos and documents. 
The work by~\citet{thamarasee2024sinhala} used a histogram-oriented gradient descriptor (HOG descriptor)~\cite{mcconnell1986method} and SVMs to recognise 400+ variations of Sinhala characters.
After conducting a survey of existing methods~\cite{hulathdoowage2025exploring}, \citet{hulathdoowage2025enhanced} proposed a U-Net~\cite{ronneberger2015u} based system for Sinhala printed document layout analysis. For this task they have created a data set using Sinhala textbooks. However, they do not provide a public link to this data set. 
\citet{jayatilleke2025zero} conducted a comparative analysis on zero-shot Sinhala OCR using a \hf{https://huggingface.co/datasets/Ransaka/sinhala_synthetic_ocr-large}{synthetic Sinhala OCR data set}. They concluded that \gh{https://github.com/VikParuchuri/surya}{Surya} to be the best Sinhala OCR system in this configuration. This aligns with the previous short study conducted by~\citet{velayuthan2024benchmarking}. However, later in their subsequent work, \citet{jayatilleke2025sidiac}, they showed that for real-world applications \texttt{Google Document AI}\footURL{https://cloud.google.com/document-ai/} obtains better results for Sinhala.

An OCR and  Text-to-Speech system for Sinhala named Bhashitha was proposed by~\citet{de2018project}.
A study on Sinhala text extraction from social media images (memes) was conducted by~\citet{samarajeewa2020approach}. They specifically handled the character-touching issue. The study by~\citet{walawage2020devising} attempted to devise a feature set to separately identify Sinhala and English text on social media images (memes).
The study by~\citet{de2021sinhala2} used Convolutional Spiking Neural Networks to extract Sinhala text from YouTube thumbnails.
~\citet{chanda2008word} proposed a Gaussian kernel SVM based method for word-wise Sinhala, Tamil, and English script identification. The work by~\citet{vasantharajan2022adapting} adapted the \textit{Tesseract} engine to handle non-Unicode (legacy fonts) in pdf documents to create a Tamil-Sinhala-English parallel corpus.

\subsubsection{Handwritten Text}
\citet{fernando2003database} claim to have created a database for handwriting recognition research in the Sinhala language and further claims that the data set is available at the National Science Foundation (NSF) of Sri Lanka. However, the paper provides no URLs, and we were not able to find the dataset on the NSF website either. \citet{weraduwa2024developing} claims to have created a dataset of  373 Sinhala handwriting images from 84 participants where 73 are dysgraphic and 300 are non-dysgraphic. Their objective is that this dataset may be used to detect dysgraphia in children. 

The work by~\citet{karunanayaka2005thresholding} is focused on noise reduction and skew correction of Sinhala handwritten words. A genetic algorithm-based approach for non-cursive Sinhala handwritten script recognition was proposed by~\citet{jayasekara2005non}.~\citet{nilaweera2007comparison} compare projection and wavelet-based techniques for recognizing handwritten Sinhala script. ~\citet{silva2014segmenting} worked on segmenting Sinhala handwritten characters with special focus on handling diacritic vowels. A comparative study of few available Sinhala handwriting recognition methods was done by~\citet{silva2014state}.~\citet{silva2015contour} uses contour tracing for isolated characters in handwritten Sinhala text. A Sinhala handwriting OCR system which utilizes zone-based feature extraction has been proposed by~\citet{dharmapala2017sinhala}. The study by~\citet{walawage2018segmentation} and its follow up study by~\citet{walawage2019segmentation} specifically focus on segmentation of overlapping and touching Sinhala handwritten characters.~\citet{silva2020character} focused on recognizing character modifiers in Sinhala handwriting. The similarly named studies by~\citet{mariyathas2020sinhala} and~\citet{wasalthilake2020sinhala}, both utilize CNN to recognize Sinhala handwriting; as does the study by~\citet{weerasinghe2019sinhala}.~\citet{ifhaam2019sinhala} used genetic algorithms to recognize Sinhala handwritten postal addresses for postal sorting. A segmentation-based approach that utilizes an n-gram model to recognize and validate Sinhala words written on touch screens was proposed by~\citet{mahesh2022segmentation}. They used a CNN classifier and were able to classify 19 different Sinhala characters.
The study by~\citet{rowel2021learning} frames their work as an E-Learning platform for hearing-impaired children. However, their research does not contain any work done towards Sinhala sign language to be included in Section~\ref{Sec:signLang}. What they do have is an OCR system that they claim to recognise letters and digits. Even there, we are only given an example of a recognised digit. Whether or not their system can recognise Sinhala letters is not explicitly discussed.
As part of their child [sic] cognitive ability assessment model,~\citet{kahawanugoda2022development}, proposed a Sinhala handwriting letter recognition system.  
The study by~\citet{withana2023detecting} used Sinhala handwritten text classification to detect Dyslexia and Dysgraphia. The study by~\citet{ekanayake2023design} has used CNN to recognise Sinhala handwritten text.
\citet{karunarathne2024efficiency} used a Gabor-initialised CNN (GCNN) on a dataset of 6000 handwritten Sinhala character images.
\texttt{Akura}, introduced by~\citet{mihiran2026akura}, is a mobile application that includes handwriting recognition aimed at Sinhala language acquisition.
\citet{gunathilaka2026quantitative} proposed using a feature-driven model incorporating stroke thickness, size consistency, spacing uniformity, slant angle, and baseline alignment for Sinhala handwriting detection. Their system obtained a Pearson correlation of 0.509 against human ratings when evaluated on a Likert scale. The study by~\citet{ranathunga2026reconstruction} suggested a flow tangent-guided endpoint pairing algorithm and Bézier curve-based~\cite{bezier1974mathematical} stroke restoration for Sinhala handwritten text restoration prior to recognition. 
Even though the work by~\citet{ramanayake2025deep} applies OCR on Sinhala handwritten text, its ultimate objective is not recognising the text itself, but the gender of the author. For this, they use several classical ML models as well as GoogleNet and  ResNet.

\subsubsection{Ancient Text}
Summarising optically recognised old Sinhala text for the purpose of archival search and preservation was explored by~\citet{rathnasena2018summarization}. The work of~\citet{peiris2012recognition} also focused on OCR for ancient Sinhala inscriptions. Building upon the architecture proposed by~\citet{ruwanmini2016architecture}, a neural network-based method for recognising ancient Sinhala inscriptions was proposed by~\citet{karunarathne2017recognizing}. The study by~\citet{wickramarathna2019data} created a system to recognise Brahmi characters, correct errors, and generate Sinhala meanings. 
\citet{heenkenda2023computational} used Inception-v3~\cite{szegedy2016rethinking}, VGG-19~\cite{simonyan2014very}, and ResNet-50~\cite{he2016deep} to classify Sinhala inscriptions to historical time periods. They later extended their work~\cite{heenkenda2025automated} by proposing to use \gh{https://github.com/ultralytics/yolov5}{YOLOv5} to detect ancient Sinhala inscriptions.
The work by~\citet{gunasekara2024deep} claims to have created a mobile app that is capable of recognising and translating early Brahmi characters. The work by~\citet{pabasara2021period} used CNN models to predict the historical period of ancient Sinhala text. In their subsequent work~\cite{surasinghe2024automated}, they extended this work by adding attention to the CNN models. The work by~\citet{rathnayaka2026brahmi} also used YOLO for recognising Brahmi characters.


\subsection{Sinhala Sign Language}
\label{Sec:signLang}

\begin{figure*}[!hbt]	
	\centering 
\includegraphics[width=0.95\textwidth]  
{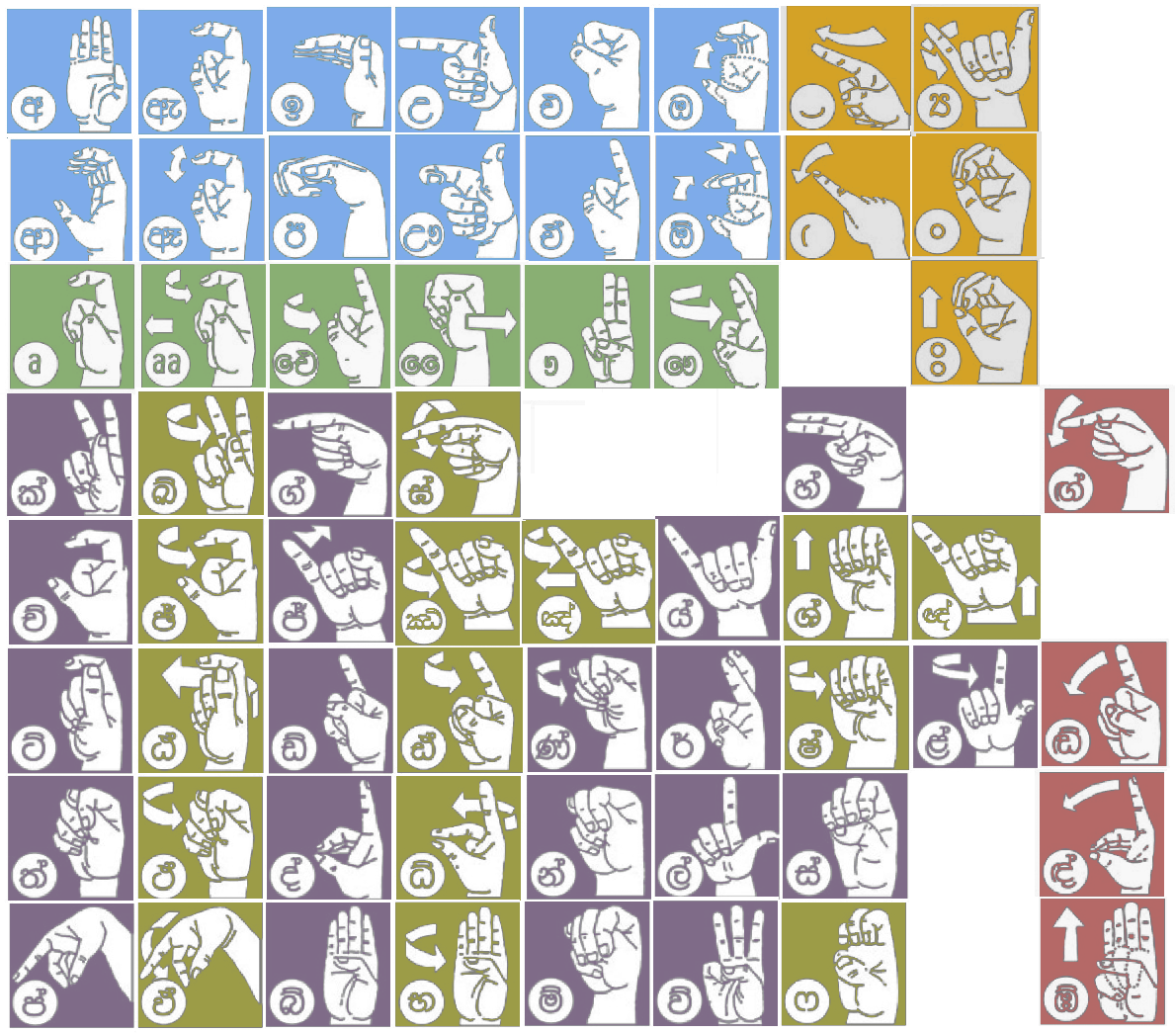} 
	\caption{Sinhala Sign Language Fingerspelling Alphabet as reported by~\citet{punchimudiyanse2017animation} and \citet{munasinghe2023interpretation}. We have attempted to preserve the colour coding and alignment that we introduced in Fig~\ref{fig:Sinhala_Script} as much as possible.}	\label{fig:SinhalaFingerSpellingSignLanguageAlphabet}
\end{figure*}

\citet{Learning2021meyler} claim that there is no such thing as a \textit{Sinhala Sign Language} or a \textit{Tamil Sign Language} and there is only \textit{Sri Lankan Sign Language}. Further, \citet{rishan2022translation} claim that SSL was derived from British Sign
Language (BSL). However, the literature uses \textit{Sri Lankan Sign Language} and \textit{Sinhala Sign Language} interchangeably with the same acronym, \textit{SSL}. In this section, we will discuss works that have used either term. But when relevant we will point out the works that claimed to work on \textit{Sri Lankan Sign Language} as opposed to \textit{Sinhala Sign Language}. As far as we can observe, there does not seem to be a technical or functional difference between the two. We show the Sinhala Sign Language Fingerspelling Alphabet as reported by~\citet{punchimudiyanse2017animation} and \citet{munasinghe2023interpretation} in Fig~\ref{fig:SinhalaFingerSpellingSignLanguageAlphabet}.

In the domain of Sinhala sign language,~\citet{liyanaarachchi2020signing} claims to have created a signing dataset for the Sinhala sign language however only the abstract of their work can be publicly accessible.~\citet{dissanayake3image} also claims to have created a database of gestures that are included in Sinhala sign language. \citet{charuka2023sign} has published a dynamic word-level Sinhala Sign Language video dataset of 50 classes. The dataset contains 1110 videos. \citet{abhishek2025ssl400} claims to introduce \textit{SSL400}, a word-level Sinhala sign language dataset consisting of 384 commonly used words. However, given that their paper is behind a paywall, it is not possible to verify whether the dataset is publicly available or not.

A study for the limited use case of translating 15 Sinhala signs to text was conducted by~\citet{fernando2016sign}. The work by~\citet{de2019sign} used a CNN model to convert Sinhala sign language to Sinhala text. Along with a Sinhala sign language recognition system.~\citet{wijegoonaratna2020realtime} has created a neural network-based approach for real-time Sinhala sign language gesture recognition. 
The approach by~\citet{hettiarachchimachine}, the approach by~\citet{dilakshan2020convolutional}, as well as the approach by~\citet{peiris2021sinhala} have used CNN to recognise the fingerspelling alphabet of Sinhala sign language. The work by~\citet{sewwantha2021mask} used Mask Region-Based CNN (Mask R-CNN) to identify some Sinhala signs. \citet{perera2021sri} also used CNN to translate Sinhala sign Language to Sinhala text. However, they have also explicitly used Scale-Invariant Feature Transform (SIFT)~\cite{lowe1999object} features. The work by~\citet{dhanawansa2021sinhala} used a CNN to classify features of static Sinhala signs, while dynamic signs were modelled as a tree structure of static signs. \citet{dahanayaka2021multi} also used CNN in their multi-model approach to recognise Sinhala sign language.  \citet{weerasooriya2022sinhala} used Random Forest (RF), K-Nearest Neighbour (KNN), Logistic Regression (LR), and a rule-based classifier to recognise Sinhala fingerspelling. \citet{kumarawadu2022sinhala} used a \textit{Leap Motion Controller}\footURL{https://leap2.ultraleap.com/leap-motion-controller-2/} and Deep Neural Networks (which most probably is, but not explicitly stated, an MLP) to recognise Sinhala sign language.  The study by~\citet{charuka2023sign} introduces a novel skeleton-based sign language recognition method named \textit{ProtoSign} built upon prototypical networks~\cite{snell2017prototypical} to do few-shot learning for Sinhala sign language recognition. The study by~\citet{haputhanthri2023multi} used \textit{ResNet} followed by a transformer encoder with multi-head attention to recognise sentence-level Sinhala sign language. The study by~\citet{fernando2023empowering} introduced a mobile app capable of translating Sinhala Sign Language to text. They further enhance the results of their system with a facial emotion detection system created with CNNs. The study by~\citet{gamage2023sinhala} also combines Sinhala Sign language recognition and emotion recognition. \citet{krishnananthan2023handtalk} introduced \textit{HANDTALK} web-based application designed to teach sign-language. As Such it has a component that translates the sign language gestures of the learner into text. In their study~\citet{gedaragoda2023hand} used traditional machine learning approaches to recognize Sinhala Sign Language. A pose-based method for Sinhala Sign Language recognition was introduced by~\citet{indatissa2023dynamic}. The work by~\citet{perera2023gesture} uses CNN for Sri Lanka Sign Language recognition.  The work by~\citet{nadeesha2024sinhala} used classical machine learning techniques to recognize 40 Sinhala signs from 100 pose videos. They reported a success rate of 86.7\%. The work by~\citet{navodya2024real} reports accuracies between 87\% and 92\% for recognising Sinhala sign language. The work by~\citet{kavinda2024multimodal} proposed a multi-modal system that converts both sign language and miming language to Sinhala text. Further, they have a component to convert Sinhala audio to miming language and Sinhala text as well.  \citet{rajapakshe2025robust} used a hybrid CNN-LSTM model and reports 94\% accuracy on the 20 Sinhala signs they tested on. \citet{thennakon2025real} on the other hand reports 96\% accuracy in recognising real-time Sinhala Sign Language signs. \citet{heshanee2025bridging} attempted Sinhala sign language recognition with 3D CNN, LSTM, and Temporal Segment Networks (TSN). The study by~\citet{herath2025deep} created a dataset of 23 SSL gestures representing major Sri Lankan cities and used  CNN-LSTM and transformer-based models to recognise the said signs. \cite{kumari2026explainability} analysed commonly used  Sinhala Sign Language detection models with SHAP~\cite{lundberg2017unified} to report the efficacy of transfer learning in this matter. \citet{rajamanthri2025sri} tested CNN, LSTM, and YOLOv8 for Sinhala sign language recognition. \citet{silva2026ai} claim to have built a system that, apart from recognising Sinhala sign language, is also capable of other relevant tasks such as providing sign language training, an environmental sound alert module, and a two-way communication module for visual and sign-based communication.
Instead of using a vision-based approach, \citet{madushanka2016framework} utilises a wearable armband for this task. The work by~\citet{godage2021sign} uses \textit{Myo} armbands for gesture-capturing in trying to recognise Sinhala signs at a sentence level. The work by~\citet{munasinghe2023interpretation} used a sensor-based approach that requires a special gesture recognition glove.

\citet{punchimudiyanse20153d} used a 3D avatar to display 200 known signs and 40 fingerprinting signs of SSL. Strides have been made in the domains of computer interpreting for written Sinhala~\cite{punchimudiyanse2017computer} and animation of finger-spelt words and number signs~\cite{punchimudiyanse2017animation}. \citet{thrimahavithana2025text} used 3D avatars to show Sinhala sign language on screen. Their system was evaluated with a usability score, error rate in text-to-sign language conversion, and overall user satisfaction.

The work by~\citet{idushan2023sinhala} claims to be able to convert Sinhala audio to SSL and also recognize SSL and interpret dynamic signs. The system proposed by~\citet{gunaratne2023multifunctional} attempts to build a full communication system where Sinhala text, Sinhala voice, Sinhala sign, and Sinhala Braille may be translated from one-another.

\citet{wickramasuriya2013computer} used template matching to recognise Sri Lankan sign language. \citet{dissanayake2020utalk} have introduced \textit{Utalk}, a mobile app to interpret Sri Lankan Sign Language. \citet{kumar2020easytalk} introduced \textit{EasyTalk}, a tool that translates Sri Lankan sign language to text as well as converts English text into Sri Lankan Sign Language. Both the approaches by~\citet{herath2022approach} and \citet{perera2023real} integrated \textit{MediaPipe}~\cite{lugaresi2019mediapipe} with Deep Neural Networks to interpret Sri Lankan sign language. \citet{perera2023real} only used LSTM while \citet{herath2022approach} used LSTM, CNN, and CNN-LSTM. \citet{rishan2022translation} uses \textit{Leap Motion Controller} (Similar to \citet{kumarawadu2022sinhala}) to translate Sri Lankan sign language to Sinhala text. \citet{jayasekara2025real} uses \texttt{EfficientNet-B0} model~\cite{hussain2023efficient} to detect static Sri Lankan Sign Language.

\citet{priyankara2023sign} have conducted a short survey on the available Sinhala sign language recognition systems. They look at 12 studies and conclude that image processing with CNN is the most used technique for Sign Language translation.
The work by~\citet{chithrani2024advancements} compares a few Sinhala sign language recognition works~\cite{fernando2016sign,de2019sign,
dissanayake2020utalk,dahanayaka2021multi,peiris2021sinhala,dhanawansa2021sinhala,rishan2022translation,hettiarachchimachine} against work conducted for other languages. 
The review by~\citet{abhishek2024deep} on deep learning in SSL recognition also covers a subset~\cite{dilakshan2020convolutional,fernando2016sign,gedaragoda2023hand,gamage2023sinhala,godage2021sign,gunaratne2023multifunctional,dhanawardhana2025enhancing,dissanayake3image,herath2022approach,dhanawansa2021sinhala,idushan2023sinhala,charuka2023sign,kumar2020easytalk,kumarawadu2022sinhala,madushanka2016framework,munasinghe2023interpretation,nadeesha2024sinhala,peiris2021sinhala,perera2023real,fernando2023empowering,sewwantha2021mask,rishan2022translation,weerasooriya2022sinhala,wickramasuriya2013computer} of the works discussed in this section.
The survey by~\citet{ahinsa2025bridging} discusses text-to-Sinhala sign language in the contexts of: finger-spelling and translation. It should be noted here that the \textit{translation} they describe here is text-to-sign language translation and not translation between any two or more natural languages. 
\citet{ahinsa2025comprehensive} in their survey of Sri Lankan sign language recognition and Sinhala text/speech to sign language translation cover a subset~\cite{Learning2021meyler,fernando2016sign,sewwantha2021mask,rishan2022translation,munasinghe2023interpretation,haputhanthri2023multi,punchimudiyanse2017animation,punchimudiyanse20153d,idushan2023sinhala,abhishek2025ssl400} of the works discussed in this section. 
Similarly, the survey by~\citet{maleesha2025review} on Sri Lankan Sign Language Interfaces for LLM integration covers a subset~\cite{abhishek2024deep,perera2021sri,fernando2016sign,punchimudiyanse2017computer,nadeesha2024sinhala,hettiarachchimachine,dissanayake3image,sewwantha2021mask,charuka2023sign,gedaragoda2023hand,haputhanthri2023multi,abhishek2025ssl400,liyanaarachchi2020signing} of the works discussed in this section.

\subsection{Sinhala Braille}

\begin{figure*}[!hbt]	
	\centering 
\includegraphics[width=\textwidth]  
{images/Sinhala_Braille.pdf}
	\caption{Sinhala Braille Alphabet. We have attempted to preserve the colour coding and alignment that we introduced in Fig~\ref{fig:Sinhala_Script} as much as possible. Given that Sinhala Braille is an extension of Bharati Braille and as such uses the same punctuation schema~\cite{mackenzie1954world}, we did not include it in this diagram.}	\label{fig:Sinhala_Braille}
\end{figure*}

Figure~\ref{fig:Sinhala_Braille} shows the Sinhala Braille Alphabet.
The work by~\citet{de2021sinhala} uses KNN, SVM, and a simple ANN system to recognize Sinhala Braille text.~\citet{vithanage2021braille} also claims to have created a conversion engine to easily convert the Braille text into the corresponding Sinhala text. The study by~\citet{madubashana2020automated}, the study by~\citet{ariyarathna2020projection} as well as the study by~\citet{weerasinghe2020system} also focused on creating an automated Braille to Sinhala recognition system. \citet{dampalessa2025identification} used a dataset of 230 characters to train a model to transliterate Braille letters into Sinhala text.
~\citet{anuradha2020machine} uses OCR on Sinhala and English Braille documents on which they then run a machine translation system in order to convert them into voice. \citet{dananjanee2026ai} used \texttt{YOLOv8} for Sinhala Braille character detection and classification, which were then post-processed through an LLM to correct grammar and refine context. The fixed text was then sent to a TTS system to obtain Sinhala speech.  
\citet{udana2024tactile} introduces a prototype tactile Braille display for Sinhala. They claim to be able to convert digital Sinhala text into Braille using an electromechanical process that also utilises a CNN. 
\citet{gunaratne2023multifunctional} proposed a system that may translate between Sinhala text, Sinhala voice, Sinhala sign, and Sinhala Braille.

\subsection{Large Language Models (LLMs)}
\citet{pramodya2025sinhalammlu} partially released, \hf{https://huggingface.co/datasets/naist-nlp/SinhalaMMLU}{SinhalaMMLU}, a data set of more than 7000 Sinhala questions and answers covering secondary to collegiate education levels in Sri Lanaka. Only around 1900 pairs are publicly available. The authors request that they be contacted to receive the rest of the data set. Further, it is released under \textit{CC BY-NC-ND 4.0} license; therefore, no derivatives of this data set can be published. They used this dataset to evaluate 26 LLMs and reported that Claude 3.5 achieved the highest average accuracy of 67\%.
\citet{singh2024aya} introduced the \hf{https://huggingface.co/datasets/CohereForAI/aya_dataset}{Aya Data set}, a human-curated multilingual instruction fine-tuning (IFT) data set for large language models (LLM). They also introduced \hf{https://huggingface.co/datasets/CohereForAI/aya_collection}{Aya Collection}, 44 instruction-style datasets that were created by transforming existing NLP datasets into pairs of \textit{prompt} and \textit{completion} and then using NMT on them to obtain data in multiple languages. Both of these data sets include Sinhala. Further, they also released \hf{https://huggingface.co/datasets/CohereForAI/aya_evaluation_suite}{Aya Evaluation suite} to measure multilingual open-ended generation quality.

The short survey by~\citet{unjum2025llms} discusses the status of Sinhala in contemporary LLMs. They discuss the importance of cultural alignment of LLMs for Sinhala, but do not discuss \texttt{GlobalPIQA}~\cite{chang2025global,de2026sinhala}, which is the first dataset with Sinhala for LLM cultural alignment. 
\citet{ustun2024aya} introduced \hf{https://huggingface.co/CohereForAI/aya-101}{Aya}, a massively multilingual generative language model trained on their \textit{Aya data set}~\cite{singh2024aya}. \citet{amarakoon2025leveraging} further fine-tuned the \textit{Aya} model using Sri Lankan business and regulatory datasets to automate multilingual customer interactions and generate context-sensitive business insights in Sinhala and Tamil for Sri Lankan Small and Medium Enterprises (SMEs). \cite{salamanca2026tiny} introduced \hf{https://huggingface.co/CohereLabs/tiny-aya-base}{Tiny Aya}, a scaled-down version of the Aya model. Further, they have a model named \hf{https://huggingface.co/CohereLabs/tiny-aya-global}{Tiny Aya: Global} balanced for multilingual performance and specifically a model named \hf{https://huggingface.co/CohereLabs/tiny-aya-fire}{Tiny Aya: Fire} specialised for South Asian languages. However, Sinhala is not listed in the HuggingFace description of the latter.

\citet{aravinda2025sinllama} introduced \hf{https://huggingface.co/polyglots/SinLlama_v01}{SinLlama}, the first pre-trained Sinhala LLM-base model. They used continual pre-training with a 10-million-sentence Sinhala corpus on Llama-3-8B to obtain this model. They show the utility of this model with three text classification tasks. Further, for this task, they created a \hf{https://huggingface.co/polyglots/Extended-Sinhala-LLaMA}{Sinhala tokeniser}. \citet{abeynayake2026fine} fine-tuned \texttt{SinLlama} for Sentiment analysis in Romanised Sinhala-English Code-Mixed social media text. 
\citet{sandaruwan2026end} also claims to have trained a Sinhala LLM by continual pre-training on LLaMA  3.1. However, they do not build upon or compare with any other work in the Sinhala LLM domain other than their own prior work~\cite{sandaruwan2025approach}. Further, neither their data nor their models are publicly available for any comparative study.

\citet{jayakody2024performance} conducted a comparative analysis on LLMs for Sinhala. In it, they observed that the small (\~7B) models of Claude\footURL{https://docs.anthropic.com/en/docs/about-claude/models} and
GPT-4o\footURL{https://openai.com/index/hello-gpt-4o/} handled Sinhala better than Llama 3 (8B)\footURL{https://llama.meta.com/docs/get-started/} and Mistral 7B\footURL{https://docs.mistral.ai/}. Further, they reported that they had to exclude Falcon 2\footURL{https://falconllm.tii.ae/} and Gemini\footURL{https://deepmind.google/technologies/gemini/} from the study due to poor or non-existent support for Sinhala. \citet{hansaka2025systematic} also claims to have conducted a review on fine-tuning approaches for LLMs such as GPT-3, GPT-4, and LlaMA in the context of Sinhala. 
\citet{chang2024goldfish} introduced \hf{https://huggingface.co/goldfish-models}{Goldfish}, a collection of monolingual language models which includes Sinhala trained on data from MADLAD~\cite{kudugunta2024madlad}, NLLB~\cite{costa2022no}, and Glot500~\cite{imani2023glot500}. The work by~\citet{ayesha2024multilingual} used a fined-tuned Llama 3.1 (8B) model~\cite{dubey2024llama} to predict student performance by analyzing textual entries created by students in Sinhala and English. \citet{tang2025framework} evaluated the susceptibility of LLMs being tricked into answering illegal questions in low-resource languages when the same questions would have been successfully rejected had they been asked in a high-resourced language. They provide the framework for testing with the results on eight low-resourced languages, which include Sinhala. 
In their work to overcome data sparsity for LLMs, \citet{lai2024llms}\footnote{Later elaborated as \citet{lai2025adaptation}.} tested the Sinhala text generation and understanding results of Google Translate against LLaMA, BLOOM, and several other variations. Even though they point to a \gh{https://github.com/boschresearch/ACL24-MLLM}{repository} of the instruction data set they created, currently, the data set cannot be found at the given URL.
\citet{puranegedara2025utilizing} proposed an intermediate-layer fusing method to LLMs so that they would perform better for low-resource languages such as Sinhala. They show their improvement with Sinhala News Results on the data set created by~\citet{de2015Sinhala}. \citet{senanayaka2026well} claim to have created a dataset named \texttt{SinMet-5K}, which is a multi-annotator Sinhala metaphor dataset aimed to be a benchmark to test the detection capabilities of LLMs.
\citet{devine2026kakugo} introduced \texttt{Kakugo}, a pipeline designed to train general-purpose Small Language Models (SLMs)

\citet{vayani2024all} analyzed the performance of a number of LLMs over 100 languages that included Sinhala. The basic dataset was created by machine translating English QA pairs from \textit{LLaVA-Bench} (In-the-Wild) dataset~\cite{liu2024visual}. This was further extended with culturally specific data points by searching for images under \textit{country name}, \textit{language name}, and \textit{cultural category}. This dataset, which they named \hf{https://huggingface.co/datasets/MBZUAI/ALM-Bench}{ALM-Bench}, is publicly available. They made a number of observations in this aspect. First, they observed that for the Q\&A task, the performance of GPT-4o~\cite{achiam2023gpt} for Sinhala is better than Qwen2-VL~\cite{wang2024qwen2}. Next, they report that including images in the question further improved Sinhala results of GPT-4o by 38.7\%. In the error analysis of the same, they report that 36\% of the errors are language errors (fluency), 31\% of the errors are due to lack of knowledge, and 18\% of the errors are due to a lack of cultural understanding. 
\citet{shafique2025culturally} when introducing \texttt{ViMUL-Bench}, a multimodal video Q\&A benchmark for LLMs that includes Sinhala, show that GPT4o has the best results for Sinhala.

\subsection{LLM Alignment}
As LLMs grow in usage, their bias towards \textit{WEIRD}~\cite{henrich2010weirdest} (Western, Educated, Industrialised, Rich, and Democratic) cultures has been observed and attempts to align them (also) with other cultures have materialised. 
\citet{chang2025global} created \hf{https://huggingface.co/datasets/mrlbenchmarks/global-piqa-nonparallel}{Global PIQA}\footURL{https://mrlbenchmarks.github.io/}, a multilingual extension to \texttt{PIQA}~\cite{bisk2020piqa} in which they include physical commonsense reasoning triples covering 100+ languages, such that the examples are culturally aligned to the relevant languages. The objective of this dataset is to function as a benchmark to measure the cultural alignment of LLMs. Their analysis results, which include Sinhala, are also available on \gh{https://github.com/mrlbenchmarks}{GitHub}. Specifically, the Sinhala portion, created by~\citet{de2026sinhala}, can be accessed in multiple forms: \hf{https://huggingface.co/datasets/mrlbenchmarks/global-piqa-nonparallel/viewer/sin_sinh}{Nonparallel}, \hf{https://huggingface.co/datasets/mrlbenchmarks/global-piqa-nonparallel/viewer/sin_latn}{Nonparallel-Romanised}, \hf{https://huggingface.co/datasets/mrlbenchmarks/global-piqa-parallel/viewer/sin_sinh}{Parallel}, and \hf{https://huggingface.co/datasets/mrlbenchmarks/global-piqa-parallel/viewer/sin_latn}{Parallel-Romanised}
The \hf{https://huggingface.co/datasets/CohereLabs/CultureMarkers/}{CultureMarkers} dataset released by~\citet{sahu2026culture} contains 9705 Sinhala entries, all sourced from \hf{https://huggingface.co/datasets/uonlp/CulturaX}{Cultura-X}~\cite{nguyen2024culturax}.
\citet{muthugala2026lkvalues} introduced \gh{https://github.com/NextME14/LKValues}{LKValues}, the first survey-grounded resource suite for Sri Lankan value alignment. It includes \texttt{LKvaluesIT}, an instruction dataset to align LLMs to Sri Lankan values and \texttt{LKvaluesBench}, a benchmark dataset to test the aforementioned alignment. The datasets are in Sinhala, Tamil, and English.

\subsection{Healthcare Applications}
\label{sec:ment}

In discussing the management of certain mental conditions (eg: Autism, ADHD), this section covers systems targeted for teaching and evaluating individuals with such special needs. Information on systems handling Sinhala language acquisition and its evaluation in the general setting is discussed in Section~\ref{sec:acq} instead.

Using classical machine learning algorithms on TF-IDF features,~\citet{fazal2023depression} predicted depression in Sinhala-English code-mixed data from Twitter and Facebook. \citet{hewapathirana2024emoscan} introduced \textit{EmoScan}, a method to detect depression in Romanized Sinhala (Singlish) tweets. The study by~\citet{herath2024social} used classical machine learning models to predict suicidal ideation from Sinhala Facebook posts. \citet{fazal2024machine} compared multiple algorithms to detect depression in Sinhala-English code-mixed social media posts and reported that \texttt{ExtraTreesClassifier}~\cite{geurts2006extremely} obtains the best results. \citet{vitharana2026benchmarking} also attempted to detect depression, suicidal ideation, or irrelevance from Sinhala Facebook posts. They report the highest F1 of 0.88 with RoBERTa. The work by~\citet{sewwandi2026ai} claims to have built a set of models to monitor depression, where the voice-based model achieves 86.69\% test accuracy, social media depression detection achieves 97.15\% accuracy, and typing behaviour analysis achieves 80\% accuracy. 

\citet{welarathna2021automated} used CNNs to classify emotions of Sinhala speech by Autistic children. They handle the following emotions: sad, disgust, surprise, neutral, happy, calm, fear, and angry. \citet{jayawardena2024multimodal} used a \textit{NAO Robot}~\cite{gouaillier2009mechatronic} to create an interactive system with Sinhala language content detection (Speech to Text), dialogue management, and voice synthesis (Text-to-Speech) for Autism Intervention in children. Continuing their work, \citet{jayawardena2025advancing} used the RASA framework~\cite{bocklisch2017rasa} to build a dialogue system that provides speech therapy to Sinhala-speaking Children with Autism Spectrum Disorder (ASD). Further, \citet{manukalpa2025designing} extended this work with cultural alignment. A Sinhala speech-based Autism intervention extension was then proposed by~\citet{bopage2026sinhala}.

The system by~\citet{kulasooriya2024psychological} used SVM and a simple ANN to detect psychological stress in Sinhala speech. 
\citet{rathnayake2025sinhala} used voice classification by biomarkers to detect  Parkinson's disease with classical machine learning algorithms.
The work by~\citet{alahakoon2026ai} used Random Forest on Sinhala handwriting of children as a pre-screening for Attention-Deficit Hyperactivity Disorder (ADHD) diagnosis.
\cite{iddamalgoda2026huruwa} claim to have created an AI-IoT robotic system named \texttt{Huruwa} aimed at supporting Sinhala-speaking children with speech sound disorders.

\citet{perera2025recent} discusses the lack of annotated datasets and NLP resources to cater for Sinhala-speaking adults with dyslexia.  \textit{LexiLearn}, a gamified system intended to help Sinhala learners with Dyslexia was introduced by~\citet{weerasinghe2025lexilearn}. Dyslexia and dysgraphia were detected using Sinhala handwritten text classification by~\citet{withana2023detecting}. A system named  \textit{Arunalu} that they claimed to use Voice recognition, Natural Language Processing, Machine Learning, and Deep Learning concepts to help individuals with dyslexia overcome problems of reading Sinhala was proposed by~\citet{sandathara2020arunalu}. \citet{lokubalasuriya2026development} proposed \texttt{REaDDS}, a tool for early detection and screening of dyslexia in Sinhala. \citet{weraduwa2025early} used VGG16 to detect Dysgraphia among Sinhala-speaking primary school children. The work by~\citet{perera2025low} created a pipeline with: Whisper for speech-to-text, SinBERT for dyslexic error detection, mT5 and Mistral to generate corrected text, and finally GoogleTTS (gTTS) for text-to-speech. An Automated Essay Scoring (AES) system with integrated dyslexic writing-pattern classification named \texttt{Akura} was created by~\citet{fonseka2026akura}. 
The \texttt{Nena Sawiya} system proposed by~\citet{thisera2026machine} aims to detect and intervene in multiple Learning Disabilities (LD) such as: Receptive Language Disorder (RLD), Expressive Language Disorder (ELD), Visual Discrimination and Memory difficulties, and Visual Closure deficits. They employ speech-to-text as well as text classification with classical models such as Random Forest and regression models. \texttt{Shilpa} introduced by~\citet{ishara2026shilpa} is a primary education mobile learning application for students with visual, auditory, physical, or cognitive impairments. The system developed by~\citet{pathiraja2026ai} uses multimodal emotional analysis with therapeutic story generation in Sinhala as well as a Sinhala voice-first mood prediction system using fine-tuned XLM-RoBERTa to help early detection of mental health issues in Children.
\cite{tharana2026unified} proposed a system they name \texttt{IPITPH} (Integrated Predictive Intelligence Tool for Pediatric Health). One of its features is voice-first caregiver interaction in Sinhala using a RAG-based multilingual conversational AI.

\subsection{Legal and Parliamentary Domain}
\label{sec:Legal}
Perpetually maintained \texttt{Sri Lanka Document Datasets} by~\citet{senaratna2025sri} includes a number of data sets in the Legal and Parliamentary domain: \gh{https://github.com/nuuuwan/lk_hansard/}{Hansards}, \gh{https://github.com/nuuuwan/lk_legal_docs/}{Legal Documents}, \gh{https://github.com/nuuuwan/lk_cabinet_decisions/}{Cabinet Decisions}, and \gh{https://github.com/nuuuwan/lk_police_press_releases/}{Police Press Releases}. From this, \citet{lasandi2026sinhalegal} derived \hf{https://huggingface.co/datasets/Minduli-Lasandi/SinhaLegal}{SinhaLegal}, a curated corpus of Sinhala legal documents including 1,065 acts and 141 bills. They transformed the original PDF-only dataset by running OCR with data cleaning and structuring so that a dedicated machine-readable corpus is obtained. The OCR dataset \hf{https://huggingface.co/datasets/avishadilhara/sinhala-ocr-lk-acts-1010}{sinhala-ocr-lk-acts-1010} created by~\citet{dilhara2026cross} uses Sri Lankan Legislative Acts published between 1981-1989 and 2000-2019 as their domain. This contains an annotated set of 1,010 page images and their transcriptions. \citet{dhanapala2026trilingual} created a \hf{https://huggingface.co/datasets/sl-parliamentary-nlp/sl-parliamentary-hansard-17-26}{dataset} of 19,553 speeches spanning 2017-2026 from the Sri Lankan parliament. 15,455 (78.8\%) of these Speeches are in Sinhala while another 2,084 (10.6\%) are in Sinhala-English and 202 (1\%) in Sinhala-Tamil. A further 56 (0.3\%) speeches contain all three languages. They have also applied \gh{https://github.com/HimathX/lk-hansard-topic-modeling}{topic modelling} on this dataset.

The work by~\cite {ubhayawardhana2023study} attempted to study the viability of using  Google Translate for the legal domain English-Sinhala and Sinhala-English translation. Many concrete examples of legal phrases that need to be translated were extensively analysed by human experts in this work.
\citet{jayawardana2025topic} conducted TF-IDF-based topic modelling for Sinhala parliamentary discussion categorisation into policy domains such as Education, Agriculture, and Health.

\subsection{News and/or Social Media Recommendation}
A trending topic detection model for Sinhala tweets using simple clustering and ranking algorithms was proposed by~\citet{jayasekara2020trend}. \citet{sandamini2022singlish} proposed a post recommendation system, which supports Singlish, for social media.
\citet{tennakoon2020hybrid} proposes a hybrid system which uses skip-gram and collaborative Filtering on a Multi-Layer Perceptron for recommending categorised Sinhala news articles.~\citet{tennakoon2020hybrid2} then extends the system to also be able to identify grey sheep users while performing the aforementioned hybrid recommendation using LDA~\cite{blei2003latent} and SVM. 
Following the above work, a news aggregator with news categorisation, comment filtering, and two types of recommendation systems was proposed by~\citet{malsha2021automated}.
~\citet{madhushika2022analyzing} analysed Twitter trending topics to understand how Sinhala Twitter data affects news dissemination on mass media. They proposed calculating a \textit{news value} to a tweet which can be utilised to sort tweets by their news-worthiness in order to give better recommendations.

\subsection{Pali-Sinhala Applications}

Pali is the liturgical language of Buddhism~\cite{childers1875dictionary,salaville1938introduction,liddicoat1993choosing}. Given its cultural relation to the Sinhalese people, the Sinhala script is one of the commonly used scripts for Pali. In addition to that, many Sinhala-Pali bridging NLP datasets and projects exist. 

A collection of Sinhala-Sinhala, Sinhala-English, English-Sinhala, and Pali-Sinhala dictionaries is available on \gh{https://github.com/pnfo/arutha.lk/tree/main/dev/dict-input}{GitHub} courtesy of \texttt{aruth.lk}. Another Pali-Sinhala dictionary has been made \gh{https://github.com/pnfo/pali-sinhala-dictionary}{available} by the \texttt{Path Nirvana} project. The same project provides a large repository of Buddhist books with Sinhala and Pali text named as \gh{https://github.com/pathnirvana/pitaka.lk/tree/master/books}{books} and \gh{https://github.com/pathnirvana/sinhala-tools/tree/master/dev/buddhist-books/text}{buddhist-books}. Further, they have some News data collected from \gh{https://github.com/pathnirvana/sinhala-tools/tree/master/dev/silumina}{Silumina}, \gh{https://github.com/pathnirvana/sinhala-tools/tree/master/dev/mawbima}{Mawbima}, and \gh{https://github.com/pathnirvana/sinhala-tools/tree/master/dev/adaderana}{AdaDerana}. An associated profile contains a large \gh{https://github.com/pnfo/sinhala-tts-dataset}{Sinhala TTS data set} consisting of 6248 sentences with 13.8 hours of recordings over two speakers: one male and one female. \citet{gurusinghe2026sipakosa} introduced \hf{https://huggingface.co/datasets/RaniduG/SiPaKosa}{SiPaKosa-Books} and \hf{https://huggingface.co/datasets/RaniduG/SiPaKosa-Sent}{SiPaKosa-Sent}, a large-scale Sinhala-Pali corpus combining classical Sinhala Buddhist texts from historical archives with complete canonical scriptures from web sources. Their \gh{https://github.com/RanxduG/SiPaKosa-Dataset}{code} is also publicly available.
The importance of translation between Sinhala and Pali was discussed by~\citet{jayasinghe2023analytical}. \citet{shalini2017dictionary} attempted to use a dictionary-based machine-translation method for this task.

\subsection{Language Acquisition and Evaluation}
\label{sec:acq}
This section only discusses Sinhala language acquisition and its evaluation in the general setting. Information on systems targeted for teaching and evaluating individuals with special needs (eg: Autism, ADHD) is discussed in Section~\ref{sec:ment} instead.

\citet{sandathara2020arunalu} proposed a system which they named \textit{Arunalu} that they claimed to use Voice recognition, Natural Language Processing, Machine Learning, and Deep Learning concepts to help individuals with dyslexia overcome problems of reading Sinhala. The learning bot \textit{MiMi} proposed by~\citet{vithana2022mimi} assists children to learn to speak without stuttering. \citet{nethmi2023narrataa} introduced \textit{Narrataa}, a mobile application targeted at developing the Sinhala vocabulary of children by using computer vision to caption images in Sinhala. The mobile application developed by~\citet{saranga2023enhancing} helps individuals learn Sinhala letters with the help of images, colours, and shapes. The application also attempts to generate stories in Sinhala. The \textit{Katha App} developed by~\citet{wijesooriya2023katha} uses SVM and LSTm to identify phonological disorders from Sinhala speech. \citet{walpola2026neural} used Quadratic SVM and Subspace KNN to study Sinhala speed reading and regular reading-induced Functional Magnetic Resonance Imaging (fMRI). Fuzzy C-Means (FCM) clustering was used for data augmentation, and SHAP~\cite{lundberg2017unified} was used for model interpretability.

\citet{francis2024sinhalearn} claims to use Sinhala OCR and PoS tagging in their effort to build the \textit{SinhaLearn}, a system for improving the Sinhala Proficiency of grade 5 scholarship students. \citet{praveen2024machine} also claims to have created a web-based education system that helps grade 5 scholarship students who are learning the Sinhala language. They claim that they have created a Sinhala Spell checker with  82\% accuracy and a Sinhala TTS system with 98\% accuracy within this system. \citet{weerasinghe2025lexilearn} introduced \textit{LexiLearn}, a gamified system intended to help Sinhala learners with dyslexia. The survey by~\citet{perera2025recent} discusses the lack of annotated datasets and NLP resources to cater for Sinhala-speaking adults with dyslexia. An interactive platform capable of handwritten Sinhala character recognition and prediction was introduced by~\citet{de2024enhancing} for the purpose of teaching preschool children the Sinhala alphabet. \citet{mihiran2026akura} introduced \texttt{Akura}, a mobile application that includes handwriting recognition, gesture-based numeracy, emotion-aware lesson adaptation, and speech pronunciation feedback. Yet another system named \texttt{Akura} was created by~\citet{fonseka2026akura}. However, their system is an Automated Essay Scoring (AES) system with integrated dyslexic writing-pattern classification. \citet{lokuhewage2026sinhala} have proposed, \texttt{Sinhala Learn}, an end-to-end educational assistant for Sinhala medium education with OCR on both printed and handwritten text, question answering and summarisation, voice recognision which includes Southern dialect normalisation. \citet{de2024interactive} introduced a digital platform to help students learn how to write Sinhala letters by using handwriting recognition. \citet{jayasekara2026developing} tested models such as \texttt{BERT-base}, \texttt{RoBERTa-large}, \texttt{DeBERTa-v3} for Sinhala automated essay scoring (AES) using a Sri Lankan GCE O/L examination dataset.

\subsection{Miscellaneous Applications}
In this section, we discuss NLP tools and research which are either hard to categorize under the above sections or reasonably involve multiples of them.
~\citet{fernando2011inexact} proposed a method for inexact matching of Sinhala proper names.
A study on determining canonical word order of colloquial Sinhala sentences using priority information was conducted by~\citet{kanduboda2009priority} which they later extended~\cite{kanduboda2010priority,tamaoka2011effects,kanduboda2012priority}.
A cross-language information retrieval system where Sinhala search queries are converted to English search queries by mapping Sinhala word embeddings to English word embeddings was proposed by~\citet{hisan2020cross}.

~\citet{rajitha2020sinhala} has proposed a statistical machine translation and transliteration method to align Sinhala and English documents.~\citet{rajitha2021metric} used the data set that they introduced in their previous work~\cite{sachintha2021exploiting} to prove that task-specific supervised distance learning metrics outperform their unsupervised counterparts, for document alignment.~\citet{fernando2022exploiting} have used pre-trained multilingual language models (PMLMs) to improve document and sentence alignment between Sinhala–English, Tamil–English, and Sinhala–Tamil language pairs.

\citet{kumari2021sintm} proposed an algorithm for Sinhala topic modelling based on LDA~\cite{blei2003latent} and RAKE~\cite{rose2010automatic}.~\citet{batawalaarachchi2021automated} proposed two methods using statistical features to select words to be included in Sinhala document titles. The study by~\citet{pallawala2023comparison} conducted a comparative analysis of LDA, LSA, and NMF for Sinhala and concluded that NMF archives best results. 
\citet{arambewela2021real} proposed a Sinhala writing assistant tool utilising CNNS.
The study by~\citet{bandara2024deep} has introduced a Sinhala virtual assistant with rule-based decision-making for a pre-set collection of user queries and a general Transformer-based model for general user questions. 
~\citet{jayaweera2019dynamic} used classical machine learning methods to propose dynamic stop word removal. They claim to have released a corpus of 100,000+ Sinhala documents in their paper. But they provide no information on where to obtain this corpus.
\citet{liyanage2023sinhala} have created a small Sinhala treebank of 100 sentences.
\citet{minixhofer2023s} have used Sinhala from \textit{OPUS100}~\cite{zhang2020improving} dataset to pre-train their  multilingual punctuation-agnostic
sentence segmentation model. Given that Sinhala is not one of their main target languages, they do not report sentence segmentation results for Sinhala in much detail. However, they do report that their method improves Sinhala sentence splitting F1 score to 86.0 above their SpaCy baseline of 75.8.
A meta-study on how language data from social media are used in research was conducted by~\citet{hewapathirana2023review}.

The study by~\cite{peiris2024sinhala} performs best in this regard.  used NER to cluster Sinhala news articles. 
\citet{subasinghe2024legibility} analysed the \textit{Noto Sans Sinhala}\footURL{https://fonts.google.com/noto/specimen/Noto+Sans+Sinhala} font in the context of legibility on small-scale digital device screens. 
\citet{zhao2024collaborative,zhao2025enhancing} discussed a Sinhala LLM-based system they utilized to conduct participatory research in Sri Lanka. While they discussed how the LLM system helped their goals and briefly touched on how they utilised a hybrid translation approach of NMT and rule-based methods to circumvent the lack of training data, the specifics of their LLM system are not given.
\citet{wijerathne2025scheduleme} introduced \texttt{ScheduleMe} an LLM multi-agent system to  manage a Google calendar. They reported a success rate of 70\% for Sinhala. It should be noted that among the languages tested by them, only Chinese achieved a worse result (65\%) than Sinhala. 
\citet{de2025sinhala} observed that Sinhala lacks a diachronic corpus. To alleviate this, \citet{jayatilleke2025sidiac} introduced \gh{https://github.com/NeviduJ/SiDiaC}{SiDiaC}, the first Sinhala diachronic corpus. It covers 58k tokens across 46 Sinhala literary works from the 5th to the 20th century CE under Religious, History, Poetry, Language, and Medical genres. They followed this up with \gh{https://github.com/NeviduJ/SiDiaC-v.2.0}{SiDiaC-v.2.0}~\cite{jayatilleke2026sidiac}, which covers 241k words across 185 literary works from the same time period and categorisation. In this extended version, they further introduced a number of advanced preprocessing steps that are not utilised in the earlier version. 
\cite{kishanthan2026beyond} evaluated the mathematical reasoning capabilities of LLMs in Sinhala compared to English and Tamil. 

The work by~\citet{jayasiri2025sinhala} attempted to identify Sinhala song genres by classifying their Mel-frequency Cepstral Coefficients (MFCCs)~\cite{logan2000mel} spectrogram images using CNN models such as VGG-19, InceptionV3, EfficientNet, and DenseNet.
\citet{karunathilake2026genre} used Support Vector Machine, k-Nearest Neighbours, Random Forest, and a Neural Network to classify a Sinhala song dataset of their own into Pop, Rock, Hip Hop, Classical, and Folk genres using the audio features in the songs.
The earliest attempt at semantic analysis was done by~\citet{herath1990syntactic} using their earlier work, which dealt with Sinhala morphological analysis~\cite{herath1989sinhalese}.


%% file: 04_FirstSources.tex
\section{Primary Sources}
\label{FirstSources}

Even though the main objective of this survey is to cover NLP tools and research, we noticed that much of these NLP tools and research depend on primary sources of Sinhala language such as printed books in the role of knowledge sources and ground truth. Therefore, for the benefit of other researchers who venture into Sinhala NLP, we decided to add a short introduction to the available primary sources of Sinhala language used by their peers. We note that the body of work by a single scholar, Disanayake~\cite{disanayaka1976national,disanayaka2000basaka,disanayaka2004basaka,disanayake2014sinhala,disanayaka2000basaka1,disanayaka2008basaka,disanayake2001basaka,disanayaka1991structure,disanayaka1985say,disanayaka2006sinhala,disanayaka2007usage,disanayaka1969Bashavaka,disanayaka1995grammar}, is quite prominent in the case of being used for NLP applications. For formal introduction of the language, the books by~\citet{disanayaka1976national} and~\citet{perera1985sinhala} are commonly used. In cases which deal with the Sinhala alphabet, the introduction by~\citet{indrasena2001sinhala} and by Disanayake~\cite{disanayaka2000basaka,disanayaka2004basaka} have been used. An analysis of modern Sinhala linguistics has been done by~\citet{jayathilake1991modern} and by~\citet{pallatthara1966sinhala}. The early study by~\citet{henadeerage2002topics} covers a number of topics on the Sinhala language such as grammatical relations, argument
structure, phrase structure and focus constructions.

As we discussed in Section~\ref{Sinhala:Dictionaries}, a number of printed Sinhala dictionaries exist,~\citet{malalasekera1967english} being the most prominent English-Sinahala dictionary among them. In addition to that seminal work, previous researchers of Sinhala NLP have utilized a number of other dictionaries of various configurations such as: English-Sinhala~\cite{gunaratne2006ratna,maitipe1988gunasena,ranaweera2004wasana}, Sinhala-Sinhala~\cite{jayathilake1937sinhala,wijayathunga2003maha}, and English-Sinhala-Tamil~\cite{weerasinghe1999godage}.   

A number of NLP applications have utilized first sources intended to teach children~\cite{dasanayaka1990kumara,dasanayaka2005kumara,fernando1994wara,fernando1994kriya,fernando1994sinhala} or foreigners~\cite{ranawake1986spoken,gunasekara1891comprehensive,gunasekara1999comprehensive}\footnote{Note that~\cite{gunasekara1999comprehensive} is an extension of~\cite{gunasekara1891comprehensive}.}. This makes sense given that an introduction written for children would start with basic principles and thus be ideal for crafting rule based NLP systems and an introduction written for foreigners would have Sinhala language described in terms of English, making easy the process of rule based translation of English NLP tools to Sinhala. 

For applications where a rule based approach for Sinhala spelling correction is utilized, the books by~\citet{disanayaka2006sinhala,disanayaka2007usage}, by~\citet{koparahewa2006dictionary}, and by~\citet{gair2006sinhala} are used to provide a basis. A number of NLP applications which handle spoken Sinhala in the capacity of phonological layer (Section~\ref{Sinhala:Phonological}) applications or otherwise, make note of the fact that spoken Sinhala is considerably distinguishable from written Sinhala, as such, they refer primary sources which explicitly deal with spoken Sinhala~\cite{ranawake1986spoken,disanayaka1991structure,disanayaka1985say,karunatillake1990introduction,inman1986duration,fernando1994wara,disanayaka1995grammar}. 

Primary sources used in NLP application for Sinhala grammar are varied. A number of them provide overviews of the entirety of Sinhala grammar~\cite{munidasa1938vyakarana,pallatthara1966sinhala,gunasekara1986comprehensive,nie1989sinhala,jayathilake1991nuthana,sannasgala1995viyakaranavimansawa,balagalle1995bashaadauanayasaha,karunatilaka1997sinhala,karunatilaka2004sinhala,karunarathna2004sinahala,alwis2006niwaeradi,alwis2007niwaeradi,pereraPrayogika,disanayaka1969Bashavaka}. There are specific primary sources focusing on verbs~\cite{munidasa1993kriya,fernando1994kriya,disanayake2001basaka}, nouns~\cite{fernando1994sinhala,disanayaka2008basaka}, prepositions~\cite{fernando1994wara}, compounds~\cite{disanayaka2000basaka1}, derivation~\cite{disanayake2014sinhala}, case system~\cite{jayawardana1989surface}, and sentence structure~\cite{Abhayasinghe1998sinhala} of the Sinhala language. The book by~\citet{rajapaksha2008sinhala} is commonly used in NLP applications as a guide for word tagging and  punctuation mark handling. NLP studies that tackle the hard problem of handling questions expressed in Sinhala often refer to the book by~\citet{kariyakarawana1998syntax}.~\citet{kekulawala1972future} has aptly discussed the much controversial topic of the situation of \textit{future tense} of Sinhala.

%% file: firstSources.bib
@book{jayathilake1991modern,
  title={Modern Sinhalese lingustics},
  author={Jayathilake, K},
  publisher={Pradeepa Publications},
  year={1991}
}

@book{indrasena2001sinhala,
  title={Sinhala Akshara Malava},
  author={Indrasena, D A},
  journal={Sridevi Printers (pvt) Ltd},
  year={2001}
}

@phdthesis{henadeerage2002topics,
  title={Topics in Sinhala syntax},
  author={Henadeerage, Deepthi Kumara},
  year={2002},
  school={The Australian National University}
}

@misc{daniels1992unicode,
author = {Andy Daniels and Joe Becker and others},
  title = {Unicode Technical Report Number 2},
  url = "https://unicode.org/reports/tr2.html",
year = {1992},
  note = "[Online; accessed 2026-07-13]"
}

@article{ruberu1974educational,
  title={Educational tradition indigenous to Ceylon},
  author={Ruberu, T Ranjit},
  journal={Paedagogica Historica},
  volume={14},
  number={1},
  pages={106--117},
  year={1974},
  publisher={Taylor \& Francis}
}

@inproceedings{gunasena2017teaching,
  title={Teaching--Learning Methodologies in Buddhism with Special reference to the Theravada Buddhist Tradition:(Its impact on Sri Lankan Education)},
  author={Gunasena, M H Pradeep Kumar},
  booktitle={15th Wu Yue Buddhist Academic Symposium, Hangzhou Buddhist Association and Hangzhou Religious Research Association},
  year={2017}  
}

@book{chandralal2010sinhala,
  title={Sinhala},
  author={Chandralal, Dileep},
  volume={15},
  year={2010},
  publisher={John Benjamins Publishing}
}

@article{gair1996sinhala,
  title={Sinhala writing},
  author={Gair, James W},
  journal={The world’s writing systems},
  pages={408--412},
  year={1996},
  publisher={Oxford University Press New York, NY/Oxford, UK}
}

@article{malalasekera1967english,
  title={English-Sinhalese Dictionary.},
  author={Malalasekera, George Peiris},
  year={1967},
  publisher={ERIC}
}

@book{jayathilake1937sinhala,
  title={Sinhala Shabdakoshaya (Sinhala Dictionary), Prathama Bhagaya (Vol 1)},
  author={Jayathilake, D B},
  year={1937},
  publisher={Sri Lankan branch of Royal Asian Society}
}

@article{weerasinghe1999godage,
  title={Godage English-Sinhala-Tamil Dictionary},
  author={Weerasinghe, A and Weerasinghe, C P},
  journal={Sri Lanka: S. Godage and brothers, Godage book shop},
  volume={661},
  year={1999}
}

@book{gunaratne2006ratna,
  title={Ratna English-Sinhalese Dictionary},
  author={Gunaratne, K},
  year={2006},
  publisher={Ratna Poth Prakasakayo, 513, Maradana road, Colombo 10, Sri Lanka}
}

@book{maitipe1988gunasena,
  title={Gunasena English-Sinhalese Concise Dictionary},
  author={Maitipe, Sirisena},
  year={1988},
  publisher={MD Gunasena \& Co.}
}

@book{ranaweera2004wasana,
  title={Wasana English-Sinhala Dictionary},
  author={Ranaweera, S},
  year={2004},
  publisher={Wasana Prakashakayo, Dankotuwa, Sri Lanka}
}

@book{wijayathunga2003maha,
  title={Maha Sinhala Sabdakoshaya},
  author={Wijayathunga, Harischandra},
  year={2003},
  publisher={M. D. Gunasena \& Co. Ltd., Colombo}
}

@book{dasanayaka1990kumara,
  title={kumara rachanaya; Grade 4},
  author={Dasanayaka, A. E. S.},
  year={1990},
  publisher={M D Gunasena Publishers, Colombo}
}

@book{dasanayaka2005kumara,
  title={kumara rachanaya; Grade 5},
  author={Dasanayaka, A. E. S.},
  year={2005},
  publisher={M D Gunasena Publishers, Colombo}
}

@book{fernando1994wara,
  title={Wara nonamena Nipatha},
  author={Fernando, S. O.},
  month={January},
  year={1994},
  publisher={Sammana}
}

@book{fernando1994kriya,
  title={Kriya pada igena ganimu},
  author={Fernando, S. O.},
  month={January},
  year={1994},
  publisher={Sammana}
}

@book{fernando1994sinhala,
  title={Sinhala Nouns year 11},
  author={Fernando, S. O.},
  month={March},
  year={1994},
  publisher={Sammana}
}

@book{ranawake1986spoken,
  title={Spoken Sinhalese for Foreigners},
  author={Ranawake, Edwin},
  year={1986},
  publisher={MD Gunasena \& Co. Ltd.}
}

@book{gunasekara1891comprehensive,
  title={A comprehensive grammar of the Sinhalese language: adapted for the use of English readers and prescribed for the Civil Service examinations},
  author={Gunasekara, Abraham Mendis},
  year={1891},
  publisher={GJA Skeen}
}

@book{gunasekara1999comprehensive,
  title={A comprehensive grammar of the Sinhalese language},
  author={Gunasekara, Abraham Mendis},
  year={1999},
  publisher={Asian Educational Services}
}

@book{disanayaka2000basaka,
  title={Basaka Mahima: 2 - Akuru ha Pili},
  author={Disanayake, J B},
  publisher={Godage \& brothers},
  year={2000}
}

@book{disanayaka2004basaka,
  title={Basaka Mahima: 6 - Prakurthi},
  author={Disanayake, J B},
  publisher={Godage \& brothers},
  year={2004}
}

@book{disanayaka2000basaka1,
  title={Basaka Mahima: 8 - Tadditha},
  author={Disanayake, J B},
  publisher={Godage \& brothers},
  year={2000}
}

@book{disanayaka2008basaka,
  title={Basaka Mahima: 10 - Nama Padaya},
  author={Disanayake, J B},
  publisher={Godage \& brothers},
  year={2008}
}

@book{disanayake2001basaka,
  title={Basaka Mahima: 11 - Kriya Padaya},
  author={Disanayake, J B},
  publisher={Godage \& brothers},
  year={2001}
}

@book{disanayake2014sinhala,
  title={Sinhala Reethiya: 7 - Pada Nirmanaya},
  author={Disanayake, J B},
  publisher={Sumitha Books},
  year={2014}
}

@book{disanayaka1991structure,
  title={The structure of spoken Sinhala: Sounds and their patterns},
  author={Disanayake, J B},
  year={1991},
  publisher={National Institute of Education}
}

@book{disanayaka1985say,
  title={Say it in Sinhala},
  author={Disanayake, Jayaratna Banda},
  year={1985},
  publisher={Lake House Investments Limited}
}

@book{disanayaka2006sinhala,
  title={Sinhala Akshara Vicharaya (Sinhala Graphology)},
  author={Disanayake, J B},
  year={2006},
  publisher={Sumitha Publishers},
  ISBN={955-1146-44-1}
}

@book{disanayaka2007usage,
  title={The Usage of Dental and Cerebral Nasals},
  author={Disanayake, J B},
  year={2007},
  publisher={Sumitha Publishers},
  ISBN={978-955-1146-66-5}
}

@book{disanayaka1969Bashavaka,
  title={Bashavaka rata samudaya},
  author={Disanayake, J B},
  year={1969},
  publisher={Lake house investment Co. Ltd. Colombo 2}
}

@book{disanayaka1995grammar,
  title={Grammar of Contemporary Literary Sinhala-Introduction to Grammar, Structure of Spoken Sinhala},
  author={Disanayake, J B},
  publisher={Godage \& Bros},
  volume={661},
  year={1995}
}

@book{koparahewa2006dictionary,
  title={Dictionary of Sinhala Spelling},
  author={Koparahewa, S},
  publisher={S. Godage and Brothers, Colombo},
  ISBN={955-20-8266-8},
  year={2006}
}

@book{gair2006sinhala,
  title={The Sinhala Writing System, A Guide to Transliteration},
  author={Gair, James W and Karunatillake, W S},
  year={2006},
  publisher={Sinhamedia, PO Box 1027, Trumansburg, NY 14886}
}

@book{karunatillake1990introduction,
  title={An Introduction to Spoken Sinhala},
  author={Karunatillake, W S},
  year={1990},
  publisher={M D Gunasena \& Company}
}

@book{inman1986duration,
  title={Duration and Stress in Sinhala},
  author={Inman, M.},
  year={1986},
  publisher={Stanford University}
}

@book{gunasekara1986comprehensive,
  title={A Comprehensive Grammar of the Sinhalese Language},
  author={Gunasekara, Abraham Mendis},
  publisher={Asian Educational Services, New Delhi, Madras, India},
  year={1986}
}

@book{jayathilake1991nuthana,
  title={Nuthana Sinhala Vyakaranaye Mul Potha},
  author={Jayathilake, K},
  publisher={Pradeepa Publications},
  year={1991}
}

@book{munidasa1938vyakarana,
  title={Vyakarana Vivaranaya},
  author={Munidasa, Kumaratunga},
  year={1938},
  publisher={MD Gunasena Publishers, Colombo}
}

@book{karunatilaka1997sinhala,
  title={Sinhala Bhasa Vyakaranaya},
  author={Karunatilaka, W S},
  year={1997},
  publisher={M D Gunasena Publishers, Colombo}
}

@book{karunatilaka2004sinhala,
  title={Sinhala Basha Viharanaya},
  author={Karunatilaka, W S},
  year={2004},
  publisher={M D Gunasena Publishers, Colombo}
}

@book{karunarathna2004sinahala,
  title={Sinahala Viharanaya},
  author={Karunarathna, S},
  publisher={Washana prakasakayo, Dankotuwa, Sri Lanka},
  year={2004}
}

@book{alwis2006niwaeradi,
  title={Niwaeradi Wahara},
  author={Alwis, P},
  publisher={Suriya Prakashakayo},
  volume={109},
  year={2006}
}

@book{alwis2007niwaeradi,
  title={Niwaeradi Wahara -2},
  author={Alwis, P},
  publisher={Suriya Prakashakayo},
  year={2007}
}

@book{sannasgala1995viyakaranavimansawa,
  title={Viyakarana Vimansawa},
  author={Sannasgala, U S and Perera, A},
  publisher={Sanhida Mudranasaha Prakashana, Pannipitiya, Sri Lanka},
  year={1995}
}

@book{balagalle1995bashaadauanayasaha,
  title={Basha Adauanayasaha Sinhala Vivaharaya},
  author={Balagalle, V G},
  publisher={S. Godage and Brothers, Colombo 10},
  year={1995}
}

@book{pallatthara1966sinhala,
  title={Sinhala Grammar in Linguistic Perspective},
  author={Pallatthara, S and Weihene, P},
  publisher={Colombo, Sri Lanka: S Godage \& Brothers},
  year={1966}
}

@book{nie1989sinhala,
  title={Sinhala Laekhana Reethiya},
  author={{National Institute of Education}},
  publisher={NIE Press},
  year={1989}
}

@book{pereraPrayogika,
  title={Prayogika Sinhla Viyakaranaya},
  author={Perera, K. C.},
  publisher={Ratna Publishers Sri Lanka}
}

@book{munidasa1993kriya,
  title={Kriya Vivaranaya},
  author={Munidasa, Kumaratunga},
  year={1993},
  publisher={MD Gunasena Publishers, Colombo}
}

@book{jayawardana1989surface,
  title={The surface case system in Sinhala},
  author={Jayawardana, T},
  year={1989},
  pages={264-277},
  publisher={University of Kelaniya}
}

@book{Abhayasinghe1998sinhala,
  title={Sinhala bhashave sarala vakya vibagaya},
  author={Abhayasinghe, A. A.},
  year={1998}
}

@book{rajapaksha2008sinhala,
  title={Sinhala bhashave pada bedima saha virama lakshana bhavithaya},
  author={Rajapaksha, D},
  journal={Dharma Rajapaksha},
  year={2008}
}

@book{kariyakarawana1998syntax,
  title={The syntax of focus and wh-questions in Sinhala},
  author={Kariyakarawana, Sunil M},
  year={1998},
  publisher={Karunaratne \& Sons Limited}
}

@book{kekulawala1972future,
  title={The future tense in Sinhalese – an ‘unorthodox’ point of view},
  author={Kekulawala, S. L.},
  year={1972},
  journal={Journal of the Vidyalankara University of Ceylon},
  publisher={Vidyalankara University of Ceylon}
}

@inproceedings{madushani2025linguistic,
  title={{A Linguistic Study of the Verb Phrase in the Sinhala Language}},
  author={Madushani, B L},
  year={2025},
  booktitle={Proceedings of the International Research Conference of the Open University of Sri Lanka},
  publisher={The Open university of Sri Lanka}
}

@phdthesis{lankage1988sinhala,
  title={Sinhala Warna Malawe Vikashanaya},
  author={Lankage, Jayasiri},
  year={1988}
}

@book{lankage1996sinhala,
  title={Sinhala Warna Malawe Vikashanaya},
  author={Lankage, Jayasiri},
  year={1996},
  isbn={9559605100},
  publisher={Godage}
}

@book{mudiyanse2018sinhala,
  title={Sinhala Akuruwala Ithihasaya},
  author={Mudiyanse, Nandasena},
  year={2018},
  publisher={Godage International Publishers}
}

@book{thennakoon1957parani,
  title={Parani Lankawa ha Shilalipi},
  author={Thennakoon, Wimalananda},
  year={1957},
  publisher={M D Gunasena}
}

@book{karunarathne1956sinhala,
  title={Sinhala Shila Lekhana},
  author={Karunarathne, W S},
  year={1956},
  publisher={S Godage \& Brothers}
}

@book{deraniyagala1992prehistory,
  title={The Prehistory of Sri Lanka},
  author={Deraniyagala, Siran Upendra},
  year={1992},
  publisher={Department of Archaeological Survey Sri Lanka}
}

@misc{anusvara,
author = {Wiktionary},
title = {anusvara},
howpublished = {\url{https://en.wiktionary.org/wiki/anusvara}},
month = {},
year = {},
note = {(Accessed on 02/05/2023)}
}

@misc{visarga,
author = {Wiktionary},
title = {visarga},
howpublished = {\url{https://en.wiktionary.org/wiki/visarga}},
month = {},
year = {},
note = {(Accessed on 02/05/2023)}
}

@phdthesis{nandasara2019development,
  title={Development and Standardization of Sinhala Script Code for Digital Inclusion of Native Computer Users},
  author={Nandasara, Shakrange Turrance},
  year={2019}
}

@book{austria1855alfabete,
  title={{Alfabete des gesammten Erdkreises}},
  author={Austria. Staatsdruckerei},
  url={https://books.google.lk/books?id=dswhyAEACAAJ},
  year={1855},
  publisher={Druck und Verlag der Kaiserlich-K{\"o}niglichen Hof- und Staatsdruckerei}
}

@misc{Microsoft1998FontList,
author = {{Microsoft}},
title = {Font List Windows 10 - Typography - Microsoft Learn},
howpublished = {\url{https://learn.microsoft.com/en-us/typography/fonts/windows_10_font_list}},
month = {},
year = {1998},
note = {(Accessed on 02/05/2023)}
}


%% file: mix.bib
@book{bauer2007linguistics,
  title={Linguistics Student's Handbook},
  author={Bauer, Laurie},
  year={2007},
  publisher={Edinburgh University Press}
}

@Online {2007Percentage,
  author= {{Department of Census and Statistics Sri Lanka}},
  title = {Percentage of population aged 10 years and over in major ethnic groups by district and ability to speak Sinhala, Tamil and English Languages},
  date = {2007-07-02},
  url = {https://goo.gl/nnVZSd},
  urldate = {2018-12-17}
}

@Online {Young2015language,
  title = {A language family tree - in pictures | Education | The Guardian},
  date = {2018-12-16},
  author = {Holly Young},
  url = {https://www.theguardian.com/education/gallery/2015/jan/23/a-language-family-tree-in-pictures},
  urldate = {2018-12-16}
}

@book{daniels1996world,
  title={The world's writing systems},
  author={Daniels, Peter T and Bright, William},
  year={1996},
  publisher={Oxford University Press on Demand}
}

@book{salomon1998indian,
  title={Indian epigraphy: a guide to the study of inscriptions in Sanskrit, Prakrit, and the other Indo-Aryan languages},
  author={Salomon, Richard},
  year={1998},
  publisher={Oxford University Press}
}

@book{falk1993schrift,
  title={Schrift im alten Indien: ein Forschungsbericht mit Anmerkungen},
  author={Falk, Harry},
  volume={56},
  year={1993},
  publisher={Gunter Narr Verlag}
}

@book{ray2003archaeology,
  title={The archaeology of seafaring in ancient South Asia},
  author={Ray, Himanshu Prabha},
  year={2003},
  publisher={Cambridge University Press}
}

@article{miller1995wordnet,
  title={WordNet: a lexical database for English},
  author={Miller, George A},
  journal={Communications of the ACM},
  volume={38},
  number={11},
  pages={39--41},
  year={1995},
  publisher={ACM}
}

@article{wimalasuriya2010ontology,
  title={Ontology-based information extraction: An introduction and a survey of current approaches},
  author={Wimalasuriya, Daya C and Dou, Dejing},
  journal={Journal of Information Science},
  volume={36},
  number={3},
  pages={306--323},
  year={2010},
  publisher={Sage Publications Sage UK: London, England}
}

@inproceedings{wu1994verbs,
  title={Verbs semantics and lexical selection},
  author={Wu, Zhibiao and Palmer, Martha},
  booktitle={Proceedings of the 32nd annual meeting on Association for Computational Linguistics},
  pages={133--138},
  year={1994},
  organization={Association for Computational Linguistics}
}

@inproceedings{jiang1997semantic,
  title={Semantic similarity based on corpus statistics and lexical taxonomy},
  author={Jiang, Jay J and Conrath, David W},
  booktitle={Proc of 10th International Conference on Research in Computational Linguistics, ROCLING’97},
  year={1997},
  organization={Citeseer}
}

@inproceedings{de2017discovering,
  title={Discovering inconsistencies in pubmed abstracts through ontology-based information extraction},
  author={de Silva, Nisansa and Dou, Dejing and Huang, Jingshan},
  booktitle={Proceedings of the 8th ACM International Conference on Bioinformatics, Computational Biology, and Health Informatics},
  pages={362--371},
  year={2017},
  organization={ACM}
}

@InProceedings{manning2014the,
  author    = {Manning, Christopher D. and  Surdeanu, Mihai  and  Bauer, John  and  Finkel, Jenny  and  Bethard, Steven J. and  McClosky, David},
  title     = {The {Stanford} {CoreNLP} Natural Language Processing Toolkit},
  booktitle = {Association for Computational Linguistics (ACL) System Demonstrations},
  year      = {2014},
  pages     = {55--60},
  url       = {http://www.aclweb.org/anthology/P/P14/P14-5010}
}

@article{liddy2001natural,
  title={Natural language processing},
  author={Liddy, Elizabeth D},
  year={2001}
}

@article{bandara2012creation,
  title={Creation of precise alphabet fonts of early Brahmi script from photographic data of ancient Sri Lankan inscriptions},
  author={Bandara, Dammi and Warnajith, Nalin and Minato, Atsushi and Ozawa, Satoru},
  journal={Canadian Journal on Artificial Intelligence, Machine Learning and Pattern Recognition},
  volume={3},
  number={3},
  pages={33--39},
  year={2012}
}

@misc{sirisoma1990brahmi,
  title={Brahmi inscriptions of Sri Lanka from 3rd century BC to 65 AD},
  author={Sirisoma, M. H.},
  journal={Inscriptions: Volume Two, Archaeological Department Centenary (1890--1990), Commemorative Series},
  pages={3--54},
  year={1990},
  publisher={Department of Archaeology Colombo}
}

@article{dias1996lakdiwa,
  title={Lakdiwa Sellipiwalin heliwana Sinhala Bhashawe Prathyartha namayange vikashanaya},
  author={Dias, M},
  journal={Department of Archaeology, Colombo Sri Lanka},
  pages={1},
  year={1996}
}

@article{hettiarachchi1990investigation,
  title={Investigation of 2nd, 3rd and 4th century inscriptions},
  author={Hettiarachchi, A. S.},
  journal={Inscriptions: Volume Two, Archaeological Department Centenary (1890--1990), Commemorative Series. Colombo: Department of Archaeology},
  pages={57--104},
  year={1990}
}

@book{paranavitana1970inscriptions,
  title={Inscriptions of Ceylon},
  author={Paranavitana, Senarat and Dep{\=a}rtam{\=e}ntuva, Sri Lanka Pur{\=a}vidy{\=a}},
  year={1970},
  publisher={Dept. of Archaeology}
}

@article{unicode1996unicode,
  title={The Unicode Standard: A Technical Introduction},
  author={Unicode Consortium and others},
  journal={online document, http://www. unicode. org/unicode/standards/principles. html},
  year={1996}
}

@article{bojanowski2017enriching,
  title={Enriching word vectors with subword information},
  author={Bojanowski, Piotr and Grave, Edouard and Joulin, Armand and Mikolov, Tomas},
  journal={Transactions of the Association for Computational Linguistics},
  volume={5},
  pages={135--146},
  year={2017},
  publisher={MIT Press}
}

@inproceedings{joulin2017bag,
  title={Bag of Tricks for Efficient Text Classification},
  author={Joulin, Armand and Grave, Edouard and Bojanowski, Piotr and Mikolov, Tomas},
  booktitle={Proceedings of the 15th Conference of the European Chapter of the Association for Computational Linguistics: Volume 2, Short Papers},
  pages={427--431},
  year={2017}
}

@article{joulin2016fasttext,
  title={Fasttext. zip: Compressing text classification models},
  author={Joulin, Armand and Grave, Edouard and Bojanowski, Piotr and Douze, Matthijs and J{\'e}gou, H{\'e}rve and Mikolov, Tomas},
  journal={arXiv preprint arXiv:1612.03651},
  year={2016}
}

@book{preston2002views,
  title={Views into the Chinese room: New essays on Searle and artificial intelligence},
  author={Preston, John and Bishop, Mark J. M.},
  year={2002},
  publisher={OUP}
}

@inproceedings{lesk1986automatic,
  title={Automatic sense disambiguation using machine readable dictionaries: how to tell a pine cone from an ice cream cone},
  author={Lesk, Michael},
  booktitle={Proceedings of the 5th annual international conference on Systems documentation},
  pages={24--26},
  year={1986},
  organization={Citeseer}
}

@article{fernando1949palaeographical,
  title={{Palaeographical Development of the Brahmi Script in Ceylon from 3rd Century BC to 7th Century AD}},
  author={Fernando, P E E},
journal={University of Ceylon Review},
volume={7},
number={4},
pages={282-301},
  year={1949},
  publisher={University of Ceylon, Peradeniya}
}

@book{disanayaka1976national,
  title={National Languages of Sri-Lanka: Sinhala},
  author={Disanayake, Jayaratna Banda},
  year={1976},
  publisher={Department of cultural affairs}
}

@book{perera1985sinhala,
  title={Sinhala Bhashawa},
  author={Perera, Theodore G},
  publisher={M D Gunasena (Sri Lanka)},
  year={1985}
}

@article{soon2001machine,
  title={A machine learning approach to coreference resolution of noun phrases},
  author={Soon, Wee Meng and Ng, Hwee Tou and Lim, Daniel Chung Yong},
  journal={Computational linguistics},
  volume={27},
  number={4},
  pages={521--544},
  year={2001},
  publisher={MIT Press}
}

@inproceedings{ng2002improving,
  title={Improving machine learning approaches to coreference resolution},
  author={Ng, Vincent and Cardie, Claire},
  booktitle={Proceedings of the 40th annual meeting on association for computational linguistics},
  pages={104--111},
  year={2002},
  organization={Association for Computational Linguistics}
}

@article{van1992presupposition,
  title={Presupposition projection as anaphora resolution},
  author={Van der Sandt, Rob A},
  journal={Journal of semantics},
  volume={9},
  number={4},
  pages={333--377},
  year={1992},
  publisher={Oxford University Press}
}

@article{lappin1994algorithm,
  title={An algorithm for pronominal anaphora resolution},
  author={Lappin, Shalom and Leass, Herbert J},
  journal={Computational linguistics},
  volume={20},
  number={4},
  pages={535--561},
  year={1994},
  publisher={MIT Press}
}

@book{mitkov2014anaphora,
  title={Anaphora resolution},
  author={Mitkov, Ruslan},
  year={2014},
  publisher={Routledge}
}

@inproceedings{yarowsky1995unsupervised,
  title={Unsupervised word sense disambiguation rivaling supervised methods},
  author={Yarowsky, David},
  booktitle={33rd annual meeting of the association for computational linguistics},
  pages={189--196},
  year={1995}
}

@article{navigli2009word,
  title={Word sense disambiguation: A survey},
  author={Navigli, Roberto},
  journal={ACM computing surveys (CSUR)},
  volume={41},
  number={2},
  pages={10},
  year={2009},
  publisher={ACM}
}

@inproceedings{yarowsky1992word,
  title={Word-sense disambiguation using statistical models of Roget's categories trained on large corpora},
  author={Yarowsky, David},
  booktitle={Proceedings of the 14th conference on Computational linguistics-Volume 2},
  pages={454--460},
  year={1992},
  organization={Association for Computational Linguistics}
}

@inproceedings{banerjee2002adapted,
  title={An adapted Lesk algorithm for word sense disambiguation using WordNet},
  author={Banerjee, Satanjeev and Pedersen, Ted},
  booktitle={International conference on intelligent text processing and computational linguistics},
  pages={136--145},
  year={2002},
  organization={Springer}
}

@article{ide1998introduction,
  title={Introduction to the special issue on word sense disambiguation: the state of the art},
  author={Ide, Nancy and V{\'e}ronis, Jean},
  journal={Computational linguistics},
  volume={24},
  number={1},
  pages={2--40},
  year={1998},
  publisher={MIT Press}
}

@article{liddicoat1993choosing,
  title={Choosing a liturgical language},
  author={Liddicoat, Anthony J},
  journal={Australian Review of Applied Linguistics},
  volume={16},
  number={2},
  pages={123--141},
  year={1993},
  publisher={John Benjamins}
}

@book{salaville1938introduction,
  title={An introduction to the study of eastern liturgies},
  author={Salaville, S{\'e}v{\'e}rien},
  year={1938},
  publisher={Sands \& Company}
}

@book{childers1875dictionary,
  title={A dictionary of the Pali language},
  author={Childers, Robert Caesar},
  year={1875},
  publisher={Tr{\"u}bner}
}

@inproceedings{singh2016syntax,
  title={Syntax based machine translation using blended methodology},
  author={Singh, Shashi Pal and Kumar, Ajai and Sahu, Pragya and Verma, Preeti},
  booktitle={2016 2nd International Conference on Next Generation Computing Technologies (NGCT)},
  pages={242--247},
  year={2016},
  organization={IEEE}
}

@article{englebretson2005santa,
  title={Santa Barbara papers in linguistics: Proceeding from the workshop on Sinhala linguistics},
  author={Englebretson, R and Genetti, C},
  journal={Santa Barbara, CA: Department of Linguistics at the University of California, Santa Barbara},
  year={2005}
}

@book{gair1974literary,
  title={Literary Sinhala.},
  author={Gair, James W and Karunatilaka, WS},
  year={1974},
  publisher={ERIC, Cornell University. New York}
}

@book{fairbanks1968colloquial,
  title={Colloquial Sinhalese.},
  author={Fairbanks, Gordon H and Gair, J. W. and Silva, M. W. S. D.},
  year={1968},
  publisher={ERIC, Cornell University. New York}
}

@article{miyagishi2005accusative,
  title={Accusative Subject of Subordinate Clause in Literary Sinhala},
  author={Miyagishi, T},
  journal={Journal of Yasuda Women's University},
  volume={33},
  year={2005}
}

@article{jany2006relationship,
  title={The relationship between case marking and S, A, and O in spoken Sinhala},
  author={Jany, Carmen},
  journal={Santa Barbara Papers in Linguistics},
  number={17},
  pages={68--84},
  year={2006}
}

@article{garland2005morphological,
  title={Morphological typology and the complexity of nominal morphology in Sinhala},
  author={Garland, Jennifer},
  journal={Santa Barbara Papers in Linguistics},
  number={17},
  pages={1--19},
  year={2005}
}

@article{henderson2005between,
  title={Between lexical and lexico-grammatical classification: nominal classification in Sinhala},
  author={Henderson, Mara},
  journal={Santa Barbara Papers in Linguistics},
  pages={29},
  year={2005}
}

@article{miyagishi2003comparison,
  title={A Comparison of Word Order between Japanese and Sinhalese},
  author={Miyagishi, T},
  journal={Bulletin of Japanese Language and Literature},
  pages={101--107},
  year={2003}
}

@article{noguchi1984shinharago,
  title={Shinharago nyuumon [Introductory to the Sinhalese language]},
  author={Noguchi, Tadashi},
  journal={Tokyo: Daigaku Shorin},
  year={1984}
}

@article{aissen2003differential,
  title={Differential object marking: Iconicity vs. economy},
  author={Aissen, Judith},
  journal={Natural Language \& Linguistic Theory},
  volume={21},
  number={3},
  pages={435--483},
  year={2003},
  publisher={Springer}
}

@inproceedings{giuliani2015large,
  title={Large Vocabulary Children's Speech Recognition with DNN-HMM and SGMM Acoustic Modeling},
  author={Giuliani, Diego and BabaAli, Bagher},
  booktitle={Sixteenth Annual Conference of the International Speech Communication Association},
  year={2015}
}

@incollection{novak2006text,
  title={Text classification with active learning},
  author={Novak, Bla{\v{z}} and Mladeni{\v{c}}, Dunja and Grobelnik, Marko},
  booktitle={From Data and Information Analysis to Knowledge Engineering},
  pages={398--405},
  year={2006},
  publisher={Springer}
}

@inproceedings{yang2009effective,
  title={Effective multi-label active learning for text classification},
  author={Yang, Bishan and Sun, Jian-Tao and Wang, Tengjiao and Chen, Zheng},
  booktitle={Proceedings of the 15th ACM SIGKDD international conference on Knowledge discovery and data mining},
  pages={917--926},
  year={2009}
}

@article{blei2003latent,
  title={{Latent Dirichlet Allocation}},
  author={Blei, David M and Ng, Andrew Y and Jordan, Michael I},
  journal={the Journal of machine Learning research},
  volume={3},
  pages={993--1022},
  year={2003},
  publisher={JMLR. org}
}

@article{rose2010automatic,
  title={{Automatic Keyword Extraction from Individual Documents}},
  author={Rose, Stuart and Engel, Dave and Cramer, Nick and Cowley, Wendy},
  journal={Text mining: applications and theory},
  volume={1},
  pages={1--20},
  year={2010},
  publisher={Citeseer}
}

@article{shanahan2015living,
  title={A living document: reincarnating the research article},
  author={Shanahan, Daniel R},
  journal={Trials},
  volume={16},
  number={1},
  pages={1--5},
  year={2015},
  publisher={Springer}
}

@article{sopinka2020envisioning,
  title={Envisioning the scientific paper of the future},
  author={Sopinka, Natalie M and Coristine, Laura E and DeRosa, Maria C and Rochman, Chelsea M and Owens, Brian L and Cooke, Steven J},
  journal={Facets},
  volume={5},
  number={1},
  pages={1--16},
  year={2020},
  publisher={Canadian Science Publishing 65 Auriga Drive, Suite 203, Ottawa, ON K2E 7W6}
}

@article{schroder2003german,
  title={{The German text-to-speech synthesis system MARY: A tool for research, development and teaching}},
  author={Schr{\"o}der, Marc and Trouvain, J{\"u}rgen},
  journal={International Journal of Speech Technology},
  volume={6},
  number={4},
  pages={365--377},
  year={2003},
  publisher={Springer}
}

@inproceedings{smith2007overview,
  title={{An overview of the Tesseract OCR engine}},
  author={Smith, Ray},
  booktitle={Ninth international conference on document analysis and recognition (ICDAR 2007)},
  volume={2},
  pages={629--633},
  year={2007},
  organization={IEEE}
}

@article{ping2017deep,
  title={Deep voice 3: Scaling text-to-speech with convolutional sequence learning},
  author={Ping, Wei and Peng, Kainan and Gibiansky, Andrew and Arik, Sercan O and Kannan, Ajay and Narang, Sharan and Raiman, Jonathan and Miller, John},
  journal={arXiv preprint arXiv:1710.07654},
  year={2017}
}

@inproceedings{lowe1999object,
  title={Object recognition from local scale-invariant features},
  author={Lowe, David G},
  booktitle={Proceedings of the seventh IEEE international conference on computer vision},
  volume={2},
  pages={1150--1157},
  year={1999},
  organization={Ieee}
}

@incollection{biswas2018microsoft,
  title={Microsoft Bot Framework},
  author={Biswas, Manisha},
  booktitle={Beginning AI Bot Frameworks},
  pages={25--66},
  year={2018},
  publisher={Springer}
}

@inproceedings{houlsby2019parameter,
  title={{Parameter-efficient transfer learning for NLP}},
  author={Houlsby, Neil and Giurgiu, Andrei and Jastrzebski, Stanislaw and Morrone, Bruna and De Laroussilhe, Quentin and Gesmundo, Andrea and Attariyan, Mona and Gelly, Sylvain},
  booktitle={International conference on machine learning},
  pages={2790--2799},
  year={2019},
  organization={PMLR}
}

@article{pfeiffer2020adapterhub,
  title={Adapterhub: A framework for adapting transformers},
  author={Pfeiffer, Jonas and R{\"u}ckl{\'e}, Andreas and Poth, Clifton and Kamath, Aishwarya and Vuli{\'c}, Ivan and Ruder, Sebastian and Cho, Kyunghyun and Gurevych, Iryna},
  journal={arXiv preprint arXiv:2007.07779},
  year={2020}
}

@article{pfeiffer2020mad,
  title={{MAD-X: An Adapter-Based Framework for Multi-Task Cross-Lingual Transfer}},
  author={Pfeiffer, Jonas and Vuli{\'c}, Ivan and Gurevych, Iryna and Ruder, Sebastian},
  journal={arXiv preprint arXiv:2005.00052},
  year={2020}
}

@article{pfeiffer2020adapterfusion,
  title={AdapterFusion: Non-destructive task composition for transfer learning},
  author={Pfeiffer, Jonas and Kamath, Aishwarya and R{\"u}ckl{\'e}, Andreas and Cho, Kyunghyun and Gurevych, Iryna},
  journal={arXiv preprint arXiv:2005.00247},
  year={2020}
}

@article{wang2021efficient,
  title={Efficient Test Time Adapter Ensembling for Low-resource Language Varieties},
  author={Wang, Xinyi and Tsvetkov, Yulia and Ruder, Sebastian and Neubig, Graham},
  journal={arXiv preprint arXiv:2109.04877},
  year={2021}
}

@article{friedman2021single,
  title={Single-dataset experts for multi-dataset question answering},
  author={Friedman, Dan and Dodge, Ben and Chen, Danqi},
  journal={arXiv preprint arXiv:2109.13880},
  year={2021}
}

@article{conneau2019unsupervised,
  title={Unsupervised cross-lingual representation learning at scale},
  author={Conneau, Alexis and Khandelwal, Kartikay and Goyal, Naman and Chaudhary, Vishrav and Wenzek, Guillaume and Guzm{\'a}n, Francisco and Grave, Edouard and Ott, Myle and Zettlemoyer, Luke and Stoyanov, Veselin},
  journal={arXiv preprint arXiv:1911.02116},
  year={2019}
}

@article{gabelica2022many,
  title={Many researchers were not compliant with their published data sharing statement: mixed-methods study},
  author={Gabelica, Mirko and Boj{\v{c}}i{\'c}, Ru{\v{z}}ica and Puljak, Livia},
  journal={Journal of Clinical Epidemiology},
  year={2022},
  publisher={Elsevier}
}

@article{ott2019fairseq,
  title={fairseq: A fast, extensible toolkit for sequence modeling},
  author={Ott, Myle and Edunov, Sergey and Baevski, Alexei and Fan, Angela and Gross, Sam and Ng, Nathan and Grangier, David and Auli, Michael},
  journal={arXiv preprint arXiv:1904.01038},
  year={2019}
}

@online{department2012Census,
    title ={{Census of Population and Housing of Sri Lanka}},
    author ={{Department of Census and Statistics, Sri Lanka}},
    year ={2012},
    url= {https://bit.ly/3bAgcXE},
    lastaccessed={05-Aug-2022}
}

@online{department2024Census,
    title ={{Census of Population and Housing of Sri Lanka}},
    author ={{Department of Census and Statistics, Sri Lanka}},
    year ={2024},
    url= {https://nisansa.lk/?s=AoRbxv},
    lastaccessed={24-July-2026}
}

@article{li2022register,
  title={Register variation remains stable across 60 languages},
  author={Li, Haipeng and Dunn, Jonathan and Nini, Andrea},
  journal={Corpus Linguistics and Linguistic Theory},
  year={2022},
  publisher={De Gruyter Mouton}
}

@inproceedings{povey2011kaldi,
  title={The Kaldi speech recognition toolkit},
  author={Povey, Daniel and Ghoshal, Arnab and Boulianne, Gilles and Burget, Lukas and Glembek, Ondrej and Goel, Nagendra and Hannemann, Mirko and Motlicek, Petr and Qian, Yanmin and Schwarz, Petr and others},
  booktitle={IEEE 2011 workshop on automatic speech recognition and understanding},
  number={CONF},
  year={2011},
  organization={IEEE Signal Processing Society}
}

@inproceedings{carmantini2019untranscribed,
  title={Untranscribed Web Audio for Low Resource Speech Recognition.},
  author={Carmantini, Andrea and Bell, Peter and Renals, Steve},
  booktitle={INTERSPEECH},
  pages={226--230},
  year={2019}
}

@inproceedings{manohar2018semi,
  title={Semi-supervised training of acoustic models using lattice-free MMI},
  author={Manohar, Vimal and Hadian, Hossein and Povey, Daniel and Khudanpur, Sanjeev},
  booktitle={2018 IEEE International Conference on Acoustics, Speech and Signal Processing (ICASSP)},
  pages={4844--4848},
  year={2018},
  organization={IEEE}
}

@article{bodon2003trie,
  title={Trie: an alternative data structure for data mining algorithms},
  author={Bodon, Ferenc and R{\'o}nyai, Lajos},
  journal={Mathematical and Computer Modelling},
  volume={38},
  number={7-9},
  pages={739--751},
  year={2003},
  publisher={Elsevier}
}

@article{wang2017tacotron,
  title={Tacotron: Towards end-to-end speech synthesis},
  author={Wang, Yuxuan and Skerry-Ryan, RJ and Stanton, Daisy and Wu, Yonghui and Weiss, Ron J and Jaitly, Navdeep and Yang, Zongheng and Xiao, Ying and Chen, Zhifeng and Bengio, Samy and others},
  journal={arXiv preprint arXiv:1703.10135},
  year={2017}
}

@article{clark2022canine,
  title={Canine: Pre-training an efficient tokenization-free encoder for language representation},
  author={Clark, Jonathan H and Garrette, Dan and Turc, Iulia and Wieting, John},
  journal={Transactions of the Association for Computational Linguistics},
  volume={10},
  pages={73--91},
  year={2022},
  publisher={MIT Press}
}

@article{xue2020mt5,
  title={mT5: A massively multilingual pre-trained text-to-text transformer},
  author={Xue, Linting and Constant, Noah and Roberts, Adam and Kale, Mihir and Al-Rfou, Rami and Siddhant, Aditya and Barua, Aditya and Raffel, Colin},
  journal={arXiv preprint arXiv:2010.11934},
  year={2020}
}

@article{liu2020multilingual,
  title={Multilingual denoising pre-training for neural machine translation},
  author={Liu, Yinhan and Gu, Jiatao and Goyal, Naman and Li, Xian and Edunov, Sergey and Ghazvininejad, Marjan and Lewis, Mike and Zettlemoyer, Luke},
  journal={Transactions of the Association for Computational Linguistics},
  volume={8},
  pages={726--742},
  year={2020},
  publisher={MIT Press}
}

@article{tang2020multilingual,
  title={Multilingual translation with extensible multilingual pretraining and finetuning},
  author={Tang, Yuqing and Tran, Chau and Li, Xian and Chen, Peng-Jen and Goyal, Naman and Chaudhary, Vishrav and Gu, Jiatao and Fan, Angela},
  journal={arXiv preprint arXiv:2008.00401},
  year={2020}
}

@article{brown2020language,
  title={Language models are few-shot learners},
  author={Brown, Tom and Mann, Benjamin and Ryder, Nick and Subbiah, Melanie and Kaplan, Jared D and Dhariwal, Prafulla and Neelakantan, Arvind and Shyam, Pranav and Sastry, Girish and Askell, Amanda and others},
  journal={Advances in neural information processing systems},
  volume={33},
  pages={1877--1901},
  year={2020}
}

@inproceedings{szegedy2016rethinking,
  title={Rethinking the inception architecture for computer vision},
  author={Szegedy, Christian and Vanhoucke, Vincent and Ioffe, Sergey and Shlens, Jon and Wojna, Zbigniew},
  booktitle={Proceedings of the IEEE conference on computer vision and pattern recognition},
  pages={2818--2826},
  year={2016}
}

@article{simonyan2014very,
  title={Very deep convolutional networks for large-scale image recognition},
  author={Simonyan, Karen and Zisserman, Andrew},
  journal={arXiv preprint arXiv:1409.1556},
  year={2014}
}

@inproceedings{he2016deep,
  title={Deep residual learning for image recognition},
  author={He, Kaiming and Zhang, Xiangyu and Ren, Shaoqing and Sun, Jian},
  booktitle={Proceedings of the IEEE conference on computer vision and pattern recognition},
  pages={770--778},
  year={2016}
}

@inproceedings{le2014distributed,
  title={Distributed representations of sentences and documents},
  author={Le, Quoc and Mikolov, Tomas},
  booktitle={International conference on machine learning},
  pages={1188--1196},
  year={2014},
  organization={PMLR}
}

@article{lugaresi2019mediapipe,
  title={Mediapipe: A framework for building perception pipelines},
  author={Lugaresi, Camillo and Tang, Jiuqiang and Nash, Hadon and McClanahan, Chris and Uboweja, Esha and Hays, Michael and Zhang, Fan and Chang, Chuo-Ling and Yong, Ming Guang and Lee, Juhyun and others},
  journal={arXiv preprint arXiv:1906.08172},
  year={2019}
}

@article{snell2017prototypical,
  title={Prototypical networks for few-shot learning},
  author={Snell, Jake and Swersky, Kevin and Zemel, Richard},
  journal={Advances in neural information processing systems},
  volume={30},
  year={2017}
}

@inproceedings{lin2014microsoft,
  title={Microsoft coco: Common objects in context},
  author={Lin, Tsung-Yi and Maire, Michael and Belongie, Serge and Hays, James and Perona, Pietro and Ramanan, Deva and Doll{\'a}r, Piotr and Zitnick, C Lawrence},
  booktitle={Computer Vision--ECCV 2014: 13th European Conference, Zurich, Switzerland, September 6-12, 2014, Proceedings, Part V 13},
  pages={740--755},
  year={2014},
  organization={Springer}
}

@inproceedings{rashtchian2010collecting,
  title={Collecting image annotations using amazon’s mechanical turk},
  author={Rashtchian, Cyrus and Young, Peter and Hodosh, Micah and Hockenmaier, Julia},
  booktitle={Proceedings of the NAACL HLT 2010 workshop on creating speech and language data with Amazon’s Mechanical Turk},
  pages={139--147},
  year={2010}
}

@inproceedings{mathew2021hatexplain,
  title={{HateXplain: A Benchmark Dataset for Explainable Hate Speech Detection}},
  author={Mathew, Binny and Saha, Punyajoy and Yimam, Seid Muhie and Biemann, Chris and Goyal, Pawan and Mukherjee, Animesh},
  booktitle={Proceedings of the AAAI conference on artificial intelligence},
  volume={35},
  number={17},
  pages={14867--14875},
  year={2021}
}

@inproceedings{kamholz2014panlex,
  title={{PanLex: Building a Resource for Panlingual Lexical Translation.}},
  author={Kamholz, David and Pool, Jonathan and Colowick, Susan M},
  booktitle={LREC},
  pages={3145--3150},
  year={2014}
}

@article{liu2019roberta,
  title={{RoBERTa: A Robustly Optimized BERT Pretraining Approach}},
  author={Liu, Yinhan and Ott, Myle and Goyal, Naman and Du, Jingfei and Joshi, Mandar and Chen, Danqi and Levy, Omer and Lewis, Mike and Zettlemoyer, Luke and Stoyanov, Veselin},
  journal={arXiv preprint arXiv:1907.11692},
  year={2019}
}

@article{bridle1974experimental,
  title={An experimental automatic word recognition system},
  author={Bridle, John S and Brown, Michael D},
  journal={JSRU report},
  volume={1003},
  number={5},
  pages={33},
  year={1974},
  publisher={Joint Speech Research Unit Ruislip, England}
}

@article{mermelstein1976distance,
  title={Distance measures for speech recognition, psychological and instrumental},
  author={Mermelstein, Paul},
  journal={Pattern recognition and artificial intelligence},
  volume={116},
  pages={374--388},
  year={1976}
}

@article{mallikadevi2023analysis,
  title={An analysis of the production of plural nouns in Sinhala},
  author={Mallikadevi, Narayanan},
   journal={Strad},
    pages={544 – 551},
    volume={10},
    issue={8},
  year ={2023}
}

@article{xue2022byt5,
  title={{Byt5: Towards a token-free future with pre-trained byte-to-byte models}},
  author={Xue, Linting and Barua, Aditya and Constant, Noah and Al-Rfou, Rami and Narang, Sharan and Kale, Mihir and Roberts, Adam and Raffel, Colin},
  journal={Transactions of the Association for Computational Linguistics},
  volume={10},
  pages={291--306},
  year={2022},
  publisher={MIT Press One Broadway, 12th Floor, Cambridge, Massachusetts 02142, USA~…}
}

@article{bisani2008joint,
  title={Joint-sequence models for grapheme-to-phoneme conversion},
  author={Bisani, Maximilian and Ney, Hermann},
  journal={Speech communication},
  volume={50},
  number={5},
  pages={434--451},
  year={2008},
  publisher={Elsevier}
}

@article{post2018call,
  title={{A call for clarity in reporting BLEU scores}},
  author={Post, Matt},
  journal={arXiv preprint arXiv:1804.08771},
  year={2018}
}

@misc{mcconnell1986method,
  title={Method of and apparatus for pattern recognition},
  author={McConnell, Robert K},
  year={1986},
  month=jan # "~28",
  publisher={Google Patents},
  note={US Patent 4,567,610}
}

@incollection{surendra2024preservation,
  title={{Preservation of Vedda’s Language in Sri Lanka}},
  author={Surendra, Suvee Vathma and Priyadarshini, Swati and Parida, Shantipriya},
  booktitle={Applying AI-Based Tools and Technologies Towards Revitalization of Indigenous and Endangered Languages},
  pages={35--44},
  year={2024},
  publisher={Springer}
}

@article{nawarathna2024preserving,
  title={Preserving Language Heritage for Socio-economic Survival? The Case of Indigenous Vedda Community in Sri Lanka},
  author={Nawarathna, Sandya and Jayawickrama, Ananda},
 year={2024}
}

@article{dharmadasa1974creolization,
  title={{The creolization of an aboriginal language: The case of Vedda in Sri Lanka (Ceylon)}},
  author={Dharmadasa, KNO},
  journal={Anthropological Linguistics},
  pages={79--106},
  year={1974},
  publisher={JSTOR}
}

@article{welikala2024genetic,
  title={{The genetic identity of the Vedda: A language isolate of South Asia}},
  author={Welikala, Anjana and Desai, Shailesh and Singh, Prajjval Pratap and Fernando, Amali and Thangaraj, Kumaraswamy and van Driem, George and Adikari, Gamini and Tennekoon, Kamani and Chaubey, Gyaneshwer and Ranasinghe, Ruwandi},
  journal={Mitochondrion},
  pages={101884},
  year={2024},
  publisher={Elsevier}
}

@inproceedings{duh2020benchmarking,
  title={{Benchmarking neural and statistical machine translation on low-resource African languages}},
  author={Duh, Kevin and McNamee, Paul and Post, Matt and Thompson, Brian},
  booktitle={Proceedings of the Twelfth Language Resources and Evaluation Conference},
  pages={2667--2675},
  year={2020}
}

@article{liu2024visual,
  title={Visual instruction tuning},
  author={Liu, Haotian and Li, Chunyuan and Wu, Qingyang and Lee, Yong Jae},
  journal={Advances in neural information processing systems},
  volume={36},
  year={2024}
}

@article{wang2024qwen2,
  title={{Qwen2-VL: Enhancing Vision-Language Model's Perception of the World at Any Resolution}},
  author={Wang, Peng and Bai, Shuai and Tan, Sinan and Wang, Shijie and Fan, Zhihao and Bai, Jinze and Chen, Keqin and Liu, Xuejing and Wang, Jialin and Ge, Wenbin and others},
  journal={arXiv preprint arXiv:2409.12191},
  year={2024}
}

@article{achiam2023gpt,
  title={{GPT-4 Technical Report}},
  author={OpenAI and Josh Achiam and Steven Adler and Sandhini Agarwal and Lama Ahmad and Ilge Akkaya and Florencia Leoni Aleman and Diogo Almeida and Janko Altenschmidt and Sam Altman and Shyamal Anadkat and Red Avila and Igor Babuschkin and Suchir Balaji and Valerie Balcom and Paul Baltescu and Haiming Bao and Mohammad Bavarian and Jeff Belgum and Irwan Bello and Jake Berdine and Gabriel Bernadett-Shapiro and Christopher Berner and Lenny Bogdonoff and Oleg Boiko and Madelaine Boyd and Anna-Luisa Brakman and Greg Brockman and Tim Brooks and Miles Brundage and Kevin Button and Trevor Cai and Rosie Campbell and Andrew Cann and Brittany Carey and Chelsea Carlson and Rory Carmichael and Brooke Chan and Che Chang and Fotis Chantzis and Derek Chen and Sully Chen and Ruby Chen and Jason Chen and Mark Chen and Ben Chess and Chester Cho and Casey Chu and Hyung Won Chung and Dave Cummings and Jeremiah Currier and Yunxing Dai and Cory Decareaux and Thomas Degry and Noah Deutsch and Damien Deville and Arka Dhar and David Dohan and Steve Dowling and Sheila Dunning and Adrien Ecoffet and Atty Eleti and Tyna Eloundou and David Farhi and Liam Fedus and Niko Felix and Simón Posada Fishman and Juston Forte and Isabella Fulford and Leo Gao and Elie Georges and Christian Gibson and Vik Goel and Tarun Gogineni and Gabriel Goh and Rapha Gontijo-Lopes and Jonathan Gordon and Morgan Grafstein and Scott Gray and Ryan Greene and Joshua Gross and Shixiang Shane Gu and Yufei Guo and Chris Hallacy and Jesse Han and Jeff Harris and Yuchen He and Mike Heaton and Johannes Heidecke and Chris Hesse and Alan Hickey and Wade Hickey and Peter Hoeschele and Brandon Houghton and Kenny Hsu and Shengli Hu and Xin Hu and Joost Huizinga and Shantanu Jain and Shawn Jain and Joanne Jang and Angela Jiang and Roger Jiang and Haozhun Jin and Denny Jin and Shino Jomoto and Billie Jonn and Heewoo Jun and Tomer Kaftan and Łukasz Kaiser and Ali Kamali and Ingmar Kanitscheider and Nitish Shirish Keskar and Tabarak Khan and Logan Kilpatrick and Jong Wook Kim and Christina Kim and Yongjik Kim and Jan Hendrik Kirchner and Jamie Kiros and Matt Knight and Daniel Kokotajlo and Łukasz Kondraciuk and Andrew Kondrich and Aris Konstantinidis and Kyle Kosic and Gretchen Krueger and Vishal Kuo and Michael Lampe and Ikai Lan and Teddy Lee and Jan Leike and Jade Leung and Daniel Levy and Chak Ming Li and Rachel Lim and Molly Lin and Stephanie Lin and Mateusz Litwin and Theresa Lopez and Ryan Lowe and Patricia Lue and Anna Makanju and Kim Malfacini and Sam Manning and Todor Markov and Yaniv Markovski and Bianca Martin and Katie Mayer and Andrew Mayne and Bob McGrew and Scott Mayer McKinney and Christine McLeavey and Paul McMillan and Jake McNeil and David Medina and Aalok Mehta and Jacob Menick and Luke Metz and Andrey Mishchenko and Pamela Mishkin and Vinnie Monaco and Evan Morikawa and Daniel Mossing and Tong Mu and Mira Murati and Oleg Murk and David Mély and Ashvin Nair and Reiichiro Nakano and Rajeev Nayak and Arvind Neelakantan and Richard Ngo and Hyeonwoo Noh and Long Ouyang and Cullen O'Keefe and Jakub Pachocki and Alex Paino and Joe Palermo and Ashley Pantuliano and Giambattista Parascandolo and Joel Parish and Emy Parparita and Alex Passos and Mikhail Pavlov and Andrew Peng and Adam Perelman and Filipe de Avila Belbute Peres and Michael Petrov and Henrique Ponde de Oliveira Pinto and Michael and Pokorny and Michelle Pokrass and Vitchyr H. Pong and Tolly Powell and Alethea Power and Boris Power and Elizabeth Proehl and Raul Puri and Alec Radford and Jack Rae and Aditya Ramesh and Cameron Raymond and Francis Real and Kendra Rimbach and Carl Ross and Bob Rotsted and Henri Roussez and Nick Ryder and Mario Saltarelli and Ted Sanders and Shibani Santurkar and Girish Sastry and Heather Schmidt and David Schnurr and John Schulman and Daniel Selsam and Kyla Sheppard and Toki Sherbakov and Jessica Shieh and Sarah Shoker and Pranav Shyam and Szymon Sidor and Eric Sigler and Maddie Simens and Jordan Sitkin and Katarina Slama and Ian Sohl and Benjamin Sokolowsky and Yang Song and Natalie Staudacher and Felipe Petroski Such and Natalie Summers and Ilya Sutskever and Jie Tang and Nikolas Tezak and Madeleine B. Thompson and Phil Tillet and Amin Tootoonchian and Elizabeth Tseng and Preston Tuggle and Nick Turley and Jerry Tworek and Juan Felipe Cerón Uribe and Andrea Vallone and Arun Vijayvergiya and Chelsea Voss and Carroll Wainwright and Justin Jay Wang and Alvin Wang and Ben Wang and Jonathan Ward and Jason Wei and CJ Weinmann and Akila Welihinda and Peter Welinder and Jiayi Weng and Lilian Weng and Matt Wiethoff and Dave Willner and Clemens Winter and Samuel Wolrich and Hannah Wong and Lauren Workman and Sherwin Wu and Jeff Wu and Michael Wu and Kai Xiao and Tao Xu and Sarah Yoo and Kevin Yu and Qiming Yuan and Wojciech Zaremba and Rowan Zellers and Chong Zhang and Marvin Zhang and Shengjia Zhao and Tianhao Zheng and Juntang Zhuang and William Zhuk and Barret Zoph},
  journal={arXiv preprint arXiv:2303.08774},
  year={2023}
}

@article{hendrycks2020measuring,
  title={Measuring massive multitask language understanding},
  author={Hendrycks, Dan and Burns, Collin and Basart, Steven and Zou, Andy and Mazeika, Mantas and Song, Dawn and Steinhardt, Jacob},
  journal={arXiv preprint arXiv:2009.03300},
  year={2020}
}

@article{geurts2006extremely,
  title={{Extremely Randomized Trees}},
  author={Geurts, Pierre and Ernst, Damien and Wehenkel, Louis},
  journal={Machine learning},
  volume={63},
  pages={3--42},
  year={2006},
  publisher={Springer}
}

@inproceedings{artetxe2017learning,
  title={Learning bilingual word embeddings with (almost) no bilingual data},
  author={Artetxe, Mikel and Labaka, Gorka and Agirre, Eneko},
  booktitle={Proceedings of the 55th Annual Meeting of the Association for Computational Linguistics (Volume 1: Long Papers)},
  pages={451--462},
  year={2017}
}

@inproceedings{zhang2021combining,
  title={Combining Static Word Embeddings and Contextual Representations for Bilingual Lexicon Induction},
  author={Zhang, Jinpeng and Ji, Baijun and Xiao, Nini and Duan, Xiangyu and Zhang, Min and Shi, Yangbin and Luo, Weihua},
  booktitle={Findings of the Association for Computational Linguistics: ACL-IJCNLP 2021},
  pages={2943--2955},
  year={2021}
}

@article{zampieri2019predicting,
  title={Predicting the type and target of offensive posts in social media},
  author={Zampieri, Marcos and Malmasi, Shervin and Nakov, Preslav and Rosenthal, Sara and Farra, Noura and Kumar, Ritesh},
  journal={arXiv preprint arXiv:1902.09666},
  year={2019}
}

@inproceedings{deselaers2009deep,
  title={A deep learning approach to machine transliteration},
  author={Deselaers, Thomas and Hasan, Sa{\v{s}}a and Bender, Oliver and Ney, Hermann},
  booktitle={Proceedings of the Fourth Workshop on Statistical Machine Translation},
  pages={233--241},
  year={2009}
}

@inproceedings{zhang2023twhin,
  title={Twhin-bert: A socially-enriched pre-trained language model for multilingual tweet representations at twitter},
  author={Zhang, Xinyang and Malkov, Yury and Florez, Omar and Park, Serim and McWilliams, Brian and Han, Jiawei and El-Kishky, Ahmed},
  booktitle={Proceedings of the 29th ACM SIGKDD conference on knowledge discovery and data mining},
  pages={5597--5607},
  year={2023}
}

@article{plaquet2023powerset,
  title={Powerset multi-class cross entropy loss for neural speaker diarization},
  author={Plaquet, Alexis and Bredin, Herv{\'e}},
  journal={arXiv preprint arXiv:2310.13025},
  year={2023}
}

@inproceedings{gouaillier2009mechatronic,
  title={{Mechatronic design of NAO humanoid}},
  author={Gouaillier, David and Hugel, Vincent and Blazevic, Pierre and Kilner, Chris and Monceaux, J{\'e}r{\^o}me and Lafourcade, Pascal and Marnier, Brice and Serre, Julien and Maisonnier, Bruno},
  booktitle={2009 IEEE international conference on robotics and automation},
  pages={769--774},
  year={2009},
  organization={IEEE}
}

@article{dubey2024llama,
  title={{The Llama 3 Herd of Models}},
  author={Dubey, Abhimanyu and Jauhri, Abhinav and Pandey, Abhinav and Kadian, Abhishek and Al-Dahle, Ahmad and Letman, Aiesha and Mathur, Akhil and Schelten, Alan and Yang, Amy and Fan, Angela and others},
  journal={arXiv preprint arXiv:2407.21783},
  year={2024}
}

@misc{ProoffReader2014,
author = {ProoffReader},
  title = {Methodology and analysis of letter distributions blog post \textasciitilde{} prooffreader plus},
  url = "https://web.archive.org/web/20210123135507/https://prooffreaderplus.blogspot.com/2014/05/methodology-and-analysis-of-letter.html",
month = {05},
year = {2014},
  note = "[Online; accessed 2025-02-07]"
}

@article{lundberg2017unified,
  title={A unified approach to interpreting model predictions},
  author={Lundberg, Scott M and Lee, Su-In},
  journal={Advances in neural information processing systems},
  volume={30},
  year={2017}
}

@inproceedings{ribeiro2016should,
  title={{"Why should I trust you?" Explaining the predictions of any classifier}},
  author={Ribeiro, Marco Tulio and Singh, Sameer and Guestrin, Carlos},
  booktitle={Proceedings of the 22nd ACM SIGKDD international conference on knowledge discovery and data mining},
  pages={1135--1144},
  year={2016},
  organization={ACM}
}

@misc{parliament2022constitution,
author = {{Parliament of Democratic Socialist Republic of Sri Lanka}},
  title = {The Constitution of The Democratic Socialist Republic of Sri Lanka},
  url = "https://www.parliament.lk/files/pdf/constitution.pdf",
month = {October},
year = {2022},
  note = "[Online; accessed 2025-08-26]"
}

@inproceedings{joulin2018loss,
  title={Loss in translation: Learning bilingual word mapping with a retrieval criterion},
  author={Joulin, Armand and Bojanowski, Piotr and Mikolov, Tom{\'a}{\v{s}} and J{\'e}gou, Herv{\'e} and Grave, Edouard},
  booktitle={Proceedings of the 2018 conference on empirical methods in natural language processing},
  pages={2979--2984},
  year={2018}
}

@inproceedings{feng2022language,
  title={{Language-agnostic BERT sentence embedding}},
  author={Feng, Fangxiaoyu and Yang, Yinfei and Cer, Daniel and Arivazhagan, Naveen and Wang, Wei},
  booktitle={Proceedings of the 60th Annual Meeting of the Association for Computational Linguistics (Volume 1: Long Papers)},
  pages={878--891},
  year={2022}
}

@inproceedings{nakamura2020fakeddit,
  title={{Fakeddit: A new multimodal benchmark dataset for fine-grained fake news detection}},
  author={Nakamura, Kai and Levy, Sharon and Wang, William Yang},
  booktitle={Proceedings of the twelfth language resources and evaluation conference},
  pages={6149--6157},
  year={2020}
}

@article{bocklisch2017rasa,
  title={{Rasa: Open Source Language Understanding and Dialogue Management}},
  author={Bocklisch, Tom and Faulkner, Joey and Pawlowski, Nick and Nichol, Alan},
  journal={arXiv preprint arXiv:1712.05181},
  year={2017}
}

@inproceedings{ronneberger2015u,
  title={{U-Net: Convolutional Networks for Biomedical Image Segmentation}},
  author={Ronneberger, Olaf and Fischer, Philipp and Brox, Thomas},
  booktitle={International Conference on Medical Image Computing and Computer-Assisted Intervention},
  pages={234--241},
  year={2015},
  organization={Springer}
}

@article{guo2024lightrag,
  title={{LightRAG: Simple and Fast Retrieval-Augmented Generation}},
  author={Guo, Zirui and Xia, Lianghao and Yu, Yanhua and Ao, Tu and Huang, Chao},
  journal={arXiv preprint arXiv:2410.05779},
  year={2024}
}

@article{lu2021vaenar,
  title={{VAENAR-TTS: Variational Auto-Encoder based Non-AutoRegressive Text-to-Speech Synthesis}},
  author={Lu, Hui and Wu, Zhiyong and Wu, Xixin and Li, Xu and Kang, Shiyin and Liu, Xunying and Meng, Helen},
  journal={arXiv preprint arXiv:2107.03298},
  year={2021}
}

@inproceedings{kim2021conditional,
  title={{Conditional Variational Autoencoder with Adversarial Learning for End-to-End Text-to-Speech}},
  author={Kim, Jaehyeon and Kong, Jungil and Son, Juhee},
  booktitle={International Conference on Machine Learning},
  pages={5530--5540},
  year={2021},
  organization={PMLR}
}

@article{russell1980circumplex,
  title={A circumplex model of affect.},
  author={Russell, James A},
  journal={Journal of personality and social psychology},
  volume={39},
  number={6},
  pages={1161},
  year={1980},
  publisher={American Psychological Association}
}

@inproceedings{bisk2020piqa,
  title={{PIQA: Reasoning about Physical Commonsense in Natural Language}},
  author={Bisk, Yonatan and Zellers, Rowan and Le Bras, Ronan and Gao, Jianfeng and Choi, Yejin},
  booktitle={Proceedings of the AAAI conference on artificial intelligence},
  volume={34},
  number={05},
  pages={7432--7439},
  year={2020}
}

@inproceedings{cavnar1994n,
  title={N-gram-based text categorization},
  author={Cavnar, William B and Trenkle, John M and others},
  booktitle={Proceedings of SDAIR-94, 3rd annual symposium on document analysis and information retrieval},
  volume={161175},
  pages={14},
  year={1994},
  organization={Las Vegas, NV}
}

@article{hussain2023efficient,
  title={{An Efficient and Robust Hand Gesture Recognition System of Sign Language Employing Finetuned Inception-V3 and Efficientnet-B0 Network.}},
  author={Hussain, Adnan and Ul Amin, Sareer and Fayaz, Muhammad and others},
  journal={Computer Systems Science \& Engineering},
  volume={46},
  number={3},
  year={2023}
}

@inproceedings{radford2021learning,
  title={{Learning Transferable Visual Models From Natural Language Supervision}},
  author={Radford, Alec and Kim, Jong Wook and Hallacy, Chris and Ramesh, Aditya and Goh, Gabriel and Agarwal, Sandhini and Sastry, Girish and Askell, Amanda and Mishkin, Pamela and Clark, Jack and others},
  booktitle={International conference on machine learning},
  pages={8748--8763},
  year={2021},
  organization={PmLR}
}

@inproceedings{jia2021scaling,
  title={{Scaling Up Visual and Vision-Language Representation Learning With Noisy Text Supervision}},
  author={Jia, Chao and Yang, Yinfei and Xia, Ye and Chen, Yi-Ting and Parekh, Zarana and Pham, Hieu and Le, Quoc and Sung, Yun-Hsuan and Li, Zhen and Duerig, Tom},
  booktitle={International conference on machine learning},
  pages={4904--4916},
  year={2021},
  organization={PMLR}
}

@article{barrault2023seamless,
  title={{Seamless: Multilingual Expressive and Streaming Speech Translation}},
  author={Barrault, Lo{\"\i}c and Chung, Yu-An and Meglioli, Mariano Coria and Dale, David and Dong, Ning and Duppenthaler, Mark and Duquenne, Paul-Ambroise and Ellis, Brian and Elsahar, Hady and Haaheim, Justin and others},
  journal={arXiv preprint arXiv:2312.05187},
  year={2023}
}

@inproceedings{logan2000mel,
  title={{Mel Frequency Cepstral Coefficients for Music Modeling}},
  author={Logan, Beth},
  booktitle={Ismir},
  volume={270},
  pages={1--11},
  year={2000},
  organization={Plymouth, MA}
}

@article{vasuthavaninfluence,
  title={{The Influence of Tamil Language on Other Languages}},
  author={Vasuthavan, Naveenkumar and Sivanadhan, Ilangkumaran},
  journal={International Journal of Academic Research in Business and Social Sciences},
  volume={15},
  number={8},
  year={2025}
}

@article{damodaram2026tamil,
  title={{Tamil words in early Prakrit inscriptions in Sri Lanka}},
  author={Damodaram Pillai, Karan},
  journal={Available at SSRN},
  year={2026}
}

@inproceedings{rei2023scaling,
  title={{Scaling up CometKiwi: Unbabel-IST 2023 submission for the quality estimation shared task}},
  author={Rei, Ricardo and Guerreiro, Nuno M and Pombal, Jos{\'e} and Van Stigt, Daan and Treviso, Marcos and Coheur, Luisa and de Souza, Jos{\'e} GC and Martins, Andr{\'e} FT},
  booktitle={Proceedings of the Eighth Conference on Machine Translation},
  pages={841--848},
  year={2023}
}

@article{dissanayake2025meh,
  title={{``Meh google and give'': Selected Sinhala Discourse Markers in Sri Lankan English}},
  author={Dissanayake, Rasudula and Perera, Kaushalya},
  journal={OUSL Journal},
  volume={20},
  number={2},
  year={2025}
}

@inproceedings{ao2022speecht5,
  title={{SpeechT5: Unified-Modal Encoder-Decoder Pre-Training for Spoken Language Processing}},
  author={Ao, Junyi and Wang, Rui and Zhou, Long and Wang, Chengyi and Ren, Shuo and Wu, Yu and Liu, Shujie and Ko, Tom and Li, Qing and Zhang, Yu and others},
  booktitle={Proceedings of the 60th Annual Meeting of the Association for Computational Linguistics (Volume 1: Long Papers)},
  pages={5723--5738},
  year={2022}
}

@inproceedings{snyder2018x,
  title={{X-Vectors: Robust DNN Embeddings for Speaker Recognition}},
  author={Snyder, David and Garcia-Romero, Daniel and Sell, Gregory and Povey, Daniel and Khudanpur, Sanjeev},
  booktitle={2018 IEEE international conference on acoustics, speech and signal processing (ICASSP)},
  pages={5329--5333},
  year={2018},
  organization={IEEE}
}

@inproceedings{heafield2011kenlm,
  title={{KenLM: Faster and smaller language model queries}},
  author={Heafield, Kenneth},
  booktitle={Proceedings of the sixth workshop on statistical machine translation},
  pages={187--197},
  year={2011}
}

@incollection{bezier1974mathematical,
  title={{Mathematical and practical possibilities of UNISURF}},
  author={Bezier, Pierre},
  booktitle={Computer aided geometric design},
  pages={127--152},
  year={1974},
  publisher={Elsevier}
}

@article{occhini2026artificial,
  title={{Artificial intelligence is creating a new global linguistic hierarchy}},
  author={Occhini, Giulia and Tanaka-Ishii, Kumiko and Barford, Anna and Tikochinski, Refael and Hu, Songbo and Reichart, Roi and Zhou, Yijie and Claus, Hannah and Petti, Ulla and Vuli{\'c}, Ivan and Debnath, Ramit and Korhonen, Anna},
  journal={arXiv preprint arXiv:2602.12018},
  year={2026}
}

@article{henrich2010weirdest,
  title={{The Weirdest People in the World?}},
  author={Henrich, Joseph and Heine, Steven J and Norenzayan, Ara},
  journal={Behavioral and brain sciences},
  volume={33},
  number={2-3},
  pages={61--83},
  year={2010},
  publisher={Cambridge University Press}
}

@inproceedings{kumar2026bhashasutra,
  title={{BhashaSutra: A Task-Centric Unified Survey of Indian NLP Datasets, Corpora, and Resources}},
  author={Kumar, Raghvendra and Raj, Devankar and Saha, Sriparna},
  booktitle={Proceedings of the 64th Annual Meeting of the Association for Computational Linguistics (Volume 1: Long Papers)},
  pages={11998--12064},
  year={2026}
}

@inproceedings{scaria2024instructabsa,
  title={{InstructABSA: Instruction Learning for Aspect Based Sentiment Analysis}},
  author={Scaria, Kevin and Gupta, Himanshu and Goyal, Siddharth and Sawant, Saurabh and Mishra, Swaroop and Baral, Chitta},
  booktitle={Proceedings of the 2024 Conference of the North American Chapter of the Association for Computational Linguistics: Human Language Technologies (Volume 2: Short Papers)},
  pages={720--736},
  year={2024}
}

@book{mackenzie1954world,
  title={{World Braille usage: a survey of efforts towards uniformity of Braille notation}},
  author={Mackenzie, Sir Clutha Nantes},
  year={1954},
  publisher={International Meeting on Braille Uniformity, Paris, 1950},
  url ={https://unesdoc.unesco.org/ark:/48223/pf0000071103}
}


%% file: sinhalaNLP.bib
@inproceedings{wijesiri2014building,
 title={{Building a wordnet for Sinhala}},
 author={Wijesiri, Indeewari and Gallage, Malaka and Gunathilaka, Buddhika and Lakjeewa, Madhuranga and Wimalasuriya, Daya and Dias, Gihan and Paranavithana, Rohini and de Silva, Nisansa},
 booktitle={Proceedings of the Seventh Global WordNet Conference},
 pages={100--108},
 year={2014}
}

@inproceedings{welgama2011towards,
 title={Towards a sinhala wordnet},
 author={Welgama, Viraj and Herath, Dulip Lakmal and Liyanage, Chamila and Udalamatta, Namal and Weerasinghe, Ruvan and Jayawardana, Tissa},
 booktitle={Proceedings of the Conference on Human Language Technology for Development},
 year={2011}
}

@Online {Sinhala2015,
 title = {Sinhala WordNet},
 date = {2018-12-17},
 url = {http://www.wordnet.lk/},
 urldate = {2015-05-29}
}

@inproceedings{dilshani2017corpus,
 title={A Corpus-Based Morphological Analysis of Sinhala Verbs.},
 author={Dilshani, W. S. N. and Dias, G},
 year={2017},
 organization={The Third International Conference on Linguistics in Sri Lanka, ICLSL 2017~…}
}

@inproceedings{fernando2013morphological,
 title={A Morphological Parser for Sinhala Verbs},
 author={Fernando, Niroshinie and Weerasinghe, Ruwan},
 booktitle={Proceedings of the International Conference on Advances in ICT for Emerging Regions},
 year={2013}
}

@inproceedings{hettige2006morphological,
 title={A Morphological analyzer to enable English to Sinhala Machine Translation},
 author={Hettige, Budditha and Karunananda, Asoka S},
 booktitle={Information and Automation, 2006. ICIA 2006. International Conference on},
 pages={21--26},
 year={2006},
 organization={IEEE}
}

@inproceedings{welgama2013evaluating,
 title={Evaluating a Machine Learning Approach to Sinhala Morphological Analysis},
 author={Welgama, Viraj and Weerasinghe, Ruvan and Niranjan, Mahesan},
 booktitle={Proceedings of the 10th International Conference on Natural Language Processing, Noida, India},
 year={2013}
}

@article{virajdefining,
 title={Defining the Gold Standard Definitions for the Morphology of Sinhala Words},
 author={Welgama, Viraj and Weerasinghe, Ruvan and Mahesan, Niranjan}
}

@article{herath1992analysis,
 title={Analysis system for Sinhalese unit structure},
 author={Herath, Susantha and Ikeda, Takashi and Ishizaki, Shun and Anzai, Y and Aiso, H},
 journal={Journal of Experimental \& Theoretical Artificial Intelligence},
 volume={4},
 number={1},
 pages={29--48},
 year={1992},
 publisher={Taylor \& Francis}
}

@inproceedings{herath1989sinhalese,
 title={Sinhalese morphological analysis: a step towards machine processing of Sinhalese},
 author={Herath, S and Ikeda, T and Yokoyama, S and Isahara, H and Ishizaki, S},
 booktitle={[Proceedings 1989] IEEE International Workshop on Tools for Artificial Intelligence},
 pages={100--107},
 year={1989},
 organization={IEEE}
}

@article{hettige2012multi,
 title={Multi-agent System Technology for Morphological Analysis},
 author={Hettige, B and Karunananda, A. S. and Rzevski, G},
 journal={Proceedings of the 9th Annual Sessions of Sri Lanka Association for Artificial Intelligence (SLAAI), Colombo},
 year={2012}
}

@inproceedings{nandathilaka2018rule,
 title={A Rule-based Lemmatizing Approach for Sinhala Language},
 author={Nandathilaka, Maheshi and Ahangama, Supunmali and Weerasuriya, G Thilini},
 booktitle={2018 3rd International Conference on Information Technology Research (ICITR)},
 pages={1--5},
 year={2018},
 organization={IEEE}
}

@inproceedings{herath1990formalization,
 title={Formalization of Sinhalese morphology},
 author={Herath, S and Ikeda, T and Ishizaki, S and Anzai, Y},
 booktitle={Proc. of the 40th National Congress of IPSJ. Jyouhou Syori Gakkai},
 volume={1},
 pages={327--328},
 year={1990}
}

@article{basnayakeplagiarism,
 title={Plagiarism detection in Sinhala language: A software approach},
 author={Basnayake, S and Wijekoon, H and Wijayasiriwardhane, T. K.}
}

@article{rajamanthrisinhala,
 title={Sinhala Language Plagiarism Tool with Internet Resources Using Natural Language Processing},
 author={Rajamanthri, L P and Thelijjagoda, S}
}

@inproceedings{rajamanthri2021plagiarism,
 title={Plagiarism Detection Tool for Sinhala Language with Internet Resources Using Natural Language Processing},
 author={Rajamanthri, Lochana and Thelijjagoda, Samantha},
 booktitle={2021 10th International Conference on Information and Automation for Sustainability (ICIAfS)},
 pages={156--160},
 year={2021},
 organization={IEEE}
}

@inproceedings{kumarasinghe2021sinmorphy,
 title={{SinMorphy: A Morphological Analyzer for the Sinhala Language}},
 author={Kumarasinghe, Kalindu and Dias, Gihan and Herath, Indu},
 booktitle={2021 Moratuwa Engineering Research Conference (MERCon)},
 pages={681--686},
 year={2021},
 organization={IEEE}
}

@inproceedings{kariyawasam2019rule,
 title={A rule based stemmer for Sinhala language},
 author={Kariyawasam, K T P M and Senanayake, S Y and Haddela, Prasanna S},
 booktitle={2019 14th Conference on Industrial and Information Systems (ICIIS)},
 pages={326--331},
 year={2019},
 organization={IEEE}
}

@inproceedings{senanayake2019enhanced,
 title={Enhanced Tokenizer for Sinhala Language},
 author={Senanayake, Shashini Y and Kariyawasam, K T P M and Haddela, Prasanna S},
 booktitle={2019 National Information Technology Conference (NITC)},
 pages={84--89},
 year={2019},
 organization={IEEE}
}

@inproceedings{kasthuriarachchi2019deep,
 title={Deep Learning Approach to Detect Plagiarism in Sinhala Text},
 author={KasthuriArachchi, Tharuka and Charles, E Y A},
 booktitle={2019 14th Conference on Industrial and Information Systems (ICIIS)},
 pages={314--319},
 year={2019},
 organization={IEEE}
}

@article{ekanayaka2023applying,
 title={{Applying Deep Learning for Morphological Analysis in the Sinhala Language}},
 author={Ekanayaka, Yasas and Pushpananda, Randil and Welgama, Viraj and Liyanage, Chamila},
 year={2023},
 pages={2-10},
 journal={The International Journal on Advances in ICT for Emerging Regions},
 doi={10.4038/icter.v16i2.7262},
 volume={16},
 issue={2},
 publisher={University of Colombo School of Computing}
}

@article{goonatillekestudy,
 title={{Study Morphological Complexity of Non-Related Languages to Build a Universal Morphological Model for Machine Translation}},
 author={Goonatilleke, M A S T and Hettige, B and Bandara, A M R R},
 tear={2024}
}

@inproceedings{dilshani2017comprehensive,
 title={A Comprehensive Part of Speech (POS) Tag Set for Sinhala Language.},
 author={Dilshani, N and Fernando, S and Ranathunga, S and Jayasena, S and Dias, G},
 year={2017},
 organization={The Third International Conference on Linguistics in Sri Lanka, ICLSL 2017~…}
}

@inproceedings{fernando2016comprehensive,
 title={Comprehensive Part-Of-Speech Tag Set and SVM Based POS Tagger for Sinhala},
 author={Fernando, Sandareka and Ranathunga, Surangika and Jayasena, Sanath and Dias, Gihan},
 booktitle={Proceedings of the 6th Workshop on South and Southeast Asian Natural Language Processing (WSSANLP2016)},
 pages={173--182},
 year={2016}
}

@inproceedings{fernando2018evaluation,
 title={Evaluation of Different Classifiers for Sinhala POS Tagging},
 author={Fernando, Sandareka and Ranathunga, Surangika},
 booktitle={2018 Moratuwa Engineering Research Conference (MERCon)},
 pages={96--101},
 year={2018},
 organization={IEEE}
}

@inproceedings{gunasekara2016hybrid,
 title={Hybrid Part of Speech Tagger for Sinhala Language},
 author={Gunasekara, Dilmi and Welgama, W. V. and Weerasinghe, A. R.},
 booktitle={Advances in ICT for Emerging Regions (ICTer), 2016 Sixteenth International Conference on},
 pages={41--48},
 year={2016},
 organization={IEEE}
}

@inproceedings{herath2004stochastic,
 title={A stochastic part of speech tagger for Sinhala},
 author={Herath, Dulip Lakmal and Weerasinghe, A. R.},
 booktitle={Proceedings of the 06th International Information Technology Conference},
 pages={27--28},
 year={2004}
}

@inproceedings{jayasuriya2013learning,
 title={Learning a stochastic part of speech tagger for sinhala},
 author={Jayasuriya, M and Weerasinghe, A. R.},
 booktitle={Advances in ICT for Emerging Regions (ICTer), 2013 International Conference on},
 pages={137--143},
 year={2013},
 organization={IEEE}
}

@article{jayaweera2011part,
 title={Part of Speech (POS) tagger for Sinhala language},
 author={Jayaweera, A. J. P. M. P. and Dias, N. G. J.},
 year={2011},
 publisher={University of Kelaniya}
}

@article{jayaweera2014hidden,
 title={Hidden Markov Model Based Part of Speech Tagger for Sinhala Language},
 author={Jayaweera, A. J. P. M. P. and Dias, N. G. J.},
 journal={arXiv preprint arXiv:1407.2989},
 year={2014}
}

@inproceedings{jayaweera2014unknown,
 title={Unknown words analysis in POS tagging of Sinhala language},
 author={Jayaweera, A. J. P. M. P. and Dias, N. G. J.},
 booktitle={Advances in ICT for Emerging Regions (ICTer), 2014 International Conference on},
 pages={270--270},
 year={2014},
 organization={IEEE}
}

@inproceedings{jayaweera2014handling,
 title={Handling issues with unknown words in POS Tagging},
 author={Jayaweera, A. J. P. M. P. and Dias, N. G. J.},
 year={2014},
 organization={Book of Abstracts, Annual Research Symposium 2014}
}

@article{jayaweera2016comparison,
 title={Comparison of Part of Speech taggers for Sinhala Language},
 author={Jayaweera, Manoj and Dias, N. G. J.},
 year={2016},
 publisher={Faculty of Graduate Studies, University of Kelaniya, Sri Lanka}
}

@article{jayaweera2012evaluation,
 title={Evaluation of Stochastic Based Tagging Approach for Sinhala Language},
 author={Jayaweera, A. J. P. M. P. and Dias, N. G. J.},
 year={2012},
 publisher={University of Kelaniya}
}

@inproceedings{kothalawala2019online,
 title={Online Learning for Solving Data Availability Problem in Natural Language Processing.},
 author={Kothalawala, Buddhi and Weerasinghe, Ruvan and Kumarasinghe, Prabhash},
 booktitle={NL4AI@ AI* IA},
 year={2019}
}

@inproceedings{withanage2020stochastic,
 title={A Stochastic Part of Speech Tagger for the Sinhala Language based on Social Media Data mining},
 author={Withanage, Shalki Ginthota and Silva, Thushari},
 booktitle={2020 20th International Conference on Advances in ICT for Emerging Regions (ICTer)},
 pages={137--142},
 year={2020},
 organization={IEEE}
}

@phdthesis{wijerathna2020svm,
 title={SVM Based Part of Speech Tagger For Sinhala Language},
 author={Wijerathna, Y A D S S},
 year={2020}
}

@inproceedings{sathsarani2022sinhala,
 title={{Sinhala Part of Speech Tagger using Deep Learning Techniques}},
 author={Sathsarani, M W A R and Thalawaththa, T P A B and Galappaththi, N K and Danthanarayana, J N and Gamage, Anjalie},
 booktitle={2022 6th International Conference on Computation System and Information Technology for Sustainable Solutions (CSITSS)},
 pages={1--6},
 year={2022},
 organization={IEEE}
}

@inproceedings{hettige2006parser,
 title={A Parser for Sinhala Language-First Step Towards English to Sinhala Machine Translation},
 author={Hettige, Budditha and Karunananda, Asoka S},
 booktitle={Industrial and Information Systems, First International Conference on},
 pages={583--587},
 year={2006},
 organization={IEEE}
}

@inproceedings{hettige2011computational,
 title={Computational model of grammar for English to Sinhala Machine Translation},
 author={Hettige, Budditha and Karunananda, A. S.},
 booktitle={Advances in ICT for Emerging Regions (ICTer), 2011 International Conference on},
 pages={26--31},
 year={2011},
 organization={IEEE}
}

@article{kanduboda2013usage,
 title={On the usage of Sinhalese differential object markers object marker \textit{/wa/} vs. object marker \textit{/ta/}},
 author={Kanduboda, A Prabath B},
 journal={Theory and Practice in Language Studies},
 volume={3},
 number={7},
 pages={1081},
 year={2013},
 publisher={Citeseer}
}

@inproceedings{liyanage2012computational,
 title={{A computational grammar of Sinhala}},
 author={Liyanage, Chamila and Pushpananda, Randil and Herath, Dulip Lakmal and Weerasinghe, Ruvan},
 booktitle={International Conference on Intelligent Text Processing and Computational Linguistics},
 pages={188--200},
 year={2012},
 organization={Springer}
}

@inproceedings{jayaweera2015restful,
 title={RESTFul POS tagging WEB service for Sinhala language},
 author={Jayaweera, A. J. P. M. P. and Dias, N. G. J.},
 booktitle={2015 Fifteenth International Conference on Advances in ICT for Emerging Regions (ICTer)},
 pages={50--57},
 year={2015},
 organization={IEEE}
}

@article{ilukkumbura2023sinhala,
 title={Sinhala Active Voice into Passive Voice Converter using Rule Based Approach with Grammar Error Correction},
 author={Ilukkumbura, Oshini and Rupasinghe, Sulochana},
 year={2023}
}

@inproceedings{jayasuriya2023grammar,
 title={Grammar Error Correction for Less Resourceful Languages: A Case Study of Sinhala},
 author={Jayasuriya, Pradeep and Wijesundara, Malitha and Thelijjagoda, Samantha and Kodagoda, Nuwan},
 booktitle={2023 IEEE 17th International Conference on Industrial and Information Systems (ICIIS)},
 pages={169--174},
 year={2023},
 organization={IEEE}
}

@inproceedings{stephen2024light,
 title={{Light Verb Constructions in Universal Dependencies for South Asian Languages}},
 author={Stephen, Abishek and Zeman, Daniel},
 booktitle={Proceedings of the Joint Workshop on Multiword Expressions and Universal Dependencies (MWE-UD)@ LREC-COLING 2024},
 pages={163--177},
 year={2024}
}

@inproceedings{dahanayaka2014named,
 title={Named entity recognition for Sinhala language},
 author={Dahanayaka, J. K. and Weerasinghe, A. R.},
 booktitle={Advances in ICT for Emerging Regions (ICTer), 2014 International Conference on},
 pages={215--220},
 year={2014},
 organization={IEEE}
}

@inproceedings{manamini2016ananya,
 title={Ananya-a Named-Entity-Recognition (NER) system for Sinhala language},
 author={Manamini, S. A. P. M. and Ahamed, A. F. and Rajapakshe, R. A. E. C. and Reemal, G. H. A. and Jayasena, S and Dias, G. V. and Ranathunga, S},
 booktitle={Moratuwa Engineering Research Conference (MERCon), 2016},
 pages={30--35},
 year={2016},
 organization={IEEE}
}

@inproceedings{senevirathne2015conditional,
 title={Conditional Random Fields based named entity recognition for Sinhala},
 author={Senevirathne, K. U. and Attanayake, N. S. and Dhananjanie, A. W. M. H. and Weragoda, W. A. S. U. and Nugaliyadde, A and Thelijjagoda, S},
 booktitle={2015 IEEE 10th International Conference on Industrial and Information Systems (ICIIS)},
 pages={302--307},
 year={2015},
 organization={IEEE}
}

@inproceedings{azeez2020fine,
 title={Fine-Grained Named Entity Recognition for Sinhala},
 author={Azeez, Rameela and Ranathunga, Surangika},
 booktitle={2020 Moratuwa Engineering Research Conference (MERCon)},
 pages={295--300},
 year={2020},
 organization={IEEE}
}

@phdthesis{anuruddha2021reinforcement,
 title={Reinforcement Learning for Sinhala Named Entity Recognition},
 author={Anuruddha, H M S},
 year={2021}
}

@inproceedings{wijesinghe2022sinhala,
 title={Sinhala Named Entity Recognition Model: Domain-Specific Classes in Sports},
 author={Wijesinghe, W M S K and Tissera, Muditha},
 booktitle={2022 4th International Conference on Advancements in Computing (ICAC)},
 pages={138--143},
 year={2022},
 organization={IEEE}
}

@article{mallikarachchi2021support,
 title={Support Vector Machine based Named Entity Recognition for Sinhala},
 author={Mallikarachchi, P S and Lorensuhewa, S A S and Kalyani, M A L},
 year={2021},
 publisher={Uva Wellassa University of Sri Lanka}
}

@article{ranathunga2024multi,
 title={{A Multi-way Parallel Named Entity Annotated Corpus for English, Tamil and Sinhala}},
 author={Ranathunga, Surangika and Ranasinghea, Asanka and Shamala, Janaka and Dandeniyaa, Ayodya and Galappaththia, Rashmi and Samaraweeraa, Malithi},
 journal={Natural Language Processing Journal},
  volume={11},
  pages={100160},
  year={2025},
  publisher={Elsevier}
}

@article{de2015Sinhala,
Author={de Silva, Nisansa},
title={{Sinhala Text Classification: Observations from the Perspective of a Resource Poor Language}},
Year={2015},
journal={},
misc={https://goo.gl/XmLJEq}

}

@inproceedings{kadupitiya2016sinhala,
 title={Sinhala Short Sentence Similarity Calculation using Corpus-Based and Knowledge-Based Similarity Measures},
 author={Kadupitiya, J. C. S. and Ranathunga, Surangika and Dias, Gihan},
 booktitle={Proceedings of the 6th Workshop on South and Southeast Asian Natural Language Processing (WSSANLP2016)},
 pages={44--53},
 year={2016}
}

@article{deepal2024siamese,
 title={Siamese Hybrid Network Approach for Sentence Similarity},
 author={Deepal, D A A and Bandara, A M R R and De Silva, P R S},
 journal={Vidyodaya Journal of Science},
 volume={27},
 number={02},
 year={2024}
}

@inproceedings{ranasinghe2025musts,
 title={{MUSTS: MUltilingual Semantic Textual Similarity Benchmark}},
 author={Ranasinghe, Tharindu and Hettiarachchi, Hansi and Orasan, Constantin and Mitkov, Ruslan},
 booktitle={Proceedings of the 63rd Annual Meeting of the Association for Computational Linguistics (Volume 2: Short Papers)},
 pages={331--353},
 year={2025}
}

@inproceedings{kadupitiya2017sinhala,
 title={Sinhala Short Sentence Similarity Measures using Corpus-Based Similarity for Short Answer Grading},
 author={Kadupitiya, J. C. S. and Ranathunga, Surangika and Dias, Gihan},
 booktitle={6th Workshop on South and Southeast Asian Natural Language Processing},
 pages={44--53},
 year={2017}
}

@inproceedings{nanayakkara2018clustering,
 title={Clustering Sinhala News Articles Using Corpus-Based Similarity Measures},
 author={Nanayakkara, Purnima and Ranathunga, Surangika},
 booktitle={2018 Moratuwa Engineering Research Conference (MERCon)},
 pages={437--442},
 year={2018},
 organization={IEEE}
}

@inproceedings{ekanayaka2018sinhala,
 title={{Sinhala news analysis using text mining and machine learning}},
 author={Ekanayaka, R K S K and Lorensuhewa, S A S and Kalyani, M A L},
 booktitle={Ruhuna International Science and Technology Conference},
 year={2018}
}

@inproceedings{arukgoda2014word,
 title={A Word Sense Disambiguation Technique for Sinhala},
 author={Arukgoda, Janindu and Bandara, Vidudaya and Bashani, Samiththa and Gamage, Vijayindu and Wimalasuriya, Daya},
 booktitle={2014 4th International Conference on Artificial Intelligence with Applications in Engineering and Technology},
 pages={207--211},
 year={2014},
 organization={IEEE}
}

@inproceedings{marasinghe2002word,
 title={Word sense disambiguation of Sinhala language with unsupervised learning},
 author={Marasinghe, C and Herath, Susantha and Herath, Ajantha},
 booktitle={Proc. International Conference on Information Technology and Applications},
 pages={25--29},
 year={2002}
}

@inproceedings{palihakkara2015dialogue,
 title={Dialogue Act Recognition for Text-based Sinhala},
 author={Palihakkara, Sudheera and Sahabandu, Dammina and Shamsudeen, Ahsan and Bandara, Chamika and Ranathunga, Surangika},
 booktitle={Proceedings of the 12th International Conference on Natural Language Processing},
 pages={367--375},
 year={2015}
}

@inproceedings{gunasekara2018context,
 title={Context aware stopwords for Sinhala Text classification},
 author={Gunasekara, S. V. S. and Haddela, Prasanna S},
 booktitle={2018 National Information Technology Conference (NITC)},
 pages={1--6},
 year={2018},
 organization={IEEE}
}

@inproceedings{gunasekara2018effective,
 title={Effective Domain Specific Stopwords Generation for Sinhala Text},
 author={Gunasekara, S V S and Haddela, Prasanna S},
 year={2018},
 organization={19th Conference on Postgraduate Research, International Postgraduate~…}
}

@inproceedings{lakmali2017effectiveness,
 title={Effectiveness of rule-based classifiers in Sinhala text categorization},
 author={Lakmali, K B N and Haddela, Prasanna S},
 booktitle={2017 National Information Technology Conference (NITC)},
 pages={153--158},
 year={2017},
 organization={IEEE}
}

@inproceedings{herath1990syntactic,
 title={Syntactic and semantic analysis of Sinhala: a step towards intelligence computing systems},
 author={Herath, S and Ishizaki, S and Ikeda, T and Anzai, Y and Aiso, H},
 booktitle={Proceedings. 5th IEEE International Symposium on Intelligent Control 1990},
 pages={316--324},
 year={1990},
 organization={IEEE}
}

@article{gallege2010analysis,
 title={Analysis of Sinhala Using Natural Language Processing Techniques},
 author={Gallege, Sajika},
 year={2010},
 publisher={CiteSeerX}
}

@article{welgama2012automatic,
 title={Automatic Text Summarization for Sinhala},
 author={Welgama, W V},
 year={2012}
}

@phdthesis{wimalasuriya2019automatic,
 title={Automatic Text Summarization for Sinhala},
 author={Wimalasuriya, O S},
 year={2019}
}

@inproceedings{chathuranga2019sinhala,
 title={Sinhala Sentiment Analysis using Corpus based Sentiment Lexicon},
 author={Chathuranga, P. D. T. and Lorensuhewa, S. A. S. and Kalyani, M. A. L.},
 booktitle={International Conference on Advances in ICT for Emerging Regions (ICTer)},
 volume={1},
 pages={7},
 year={2019}
}

@article{jayasinghe2019deep,
 title={Deep Learning Textual Entailment System For Sinhala Language},
 author={Jayasinghe, Sriyal Himesh and Sirts, Kairit},
 year={2019}
}

@inproceedings{kumari2019use,
 title={Use of LIME for Human interpretability in Sinhala document classification},
 author={Kumari, P. K. S. and Haddela, P. S.},
 booktitle={2019 International Research Conference on Smart Computing and Systems Engineering (SCSE)},
 pages={97--102},
 year={2019},
 organization={IEEE}
}

@inproceedings{demotte2020sentiment,
 title={{Sentiment Analysis of Sinhala News Comments using Sentence-State LSTM Networks}},
 author={Demotte, Piyumal and Senevirathne, Lahiru and Karunanayake, Binod and Munasinghe, Udyogi and Ranathunga, Surangika},
 booktitle={2020 Moratuwa Engineering Research Conference (MERCon)},
 pages={283--288},
 year={2020},
 organization={IEEE}
}
